\documentclass[10pt, oneside]{article}

\usepackage{fancyhdr}
\fancyheadoffset{ 0.5 cm}
\fancyfootoffset{ 0.5 cm}

\usepackage{titlesec}
\titlelabel{\thetitle.\quad}

\usepackage{graphicx}
\graphicspath{ {C:\Users\andre\Documents\Oxford Job (2020-2024)\Signature Methods in Machine Learning Survey Article} }

\usepackage{times}
\usepackage{geometry}
\usepackage{graphicx}
\usepackage{caption}
\usepackage{subcaption}
\usepackage{amssymb, amscd}
\usepackage{amsmath}
\makeatletter
\@addtoreset{equation}{section}
\makeatother
\numberwithin{equation}{section}
\usepackage{amsthm} 
\usepackage{mathrsfs}
\usepackage{epstopdf}
\usepackage{enumerate}
\usepackage{mathtools}
\usepackage{tikz}							
\usepackage{setspace}
\usepackage{amsfonts, amsmath, wasysym}

\usepackage[utf8]{inputenc} 
\usepackage{hyperref}       
\usepackage{url}            
\usepackage{booktabs}       
\usepackage{amsfonts}       
\usepackage{amsmath}
\usepackage{graphicx,wrapfig,lipsum}
\usepackage{amsfonts}
\usepackage{booktabs}
\usepackage{empheq}
\usepackage{amsthm}
\usepackage{caption}
\usepackage{subcaption}
\usepackage{mathrsfs}  
\usepackage{amssymb}
\usepackage{enumitem}

\usetikzlibrary{decorations.pathreplacing,calligraphy}
\usetikzlibrary{arrows.meta, decorations.text, decorations.markings, decorations.pathreplacing, angles, quotes, calligraphy}
\definecolor{bittersweet}{rgb}{1.0, 0.44, 0.37}
\usepackage{ifthen}
\usepackage{intcalc}
\usepackage{tikz}
\usepackage{multirow}

\usepackage{tikz}
\usetikzlibrary{calc,positioning}

\newcommand{\corners}{6pt}

\usepackage{shuffle}

\usepackage{float}
\restylefloat{table}

\usepackage{hyperref}
\hypersetup{
    colorlinks=true,
    linkcolor=red,
    urlcolor=purple,
    linktoc=all,
    citecolor=blue
           }
\usepackage{cite}

\usepackage{bbm}
\usepackage[draft]{optional}
\usepackage{color}

\usepackage{xcolor}
\newtagform{red}{\color{red}(}{)}

\usepackage{mathtools}

\usepackage{amsmath}
\makeatletter
\@addtoreset{equation}{section}
\makeatother
\numberwithin{equation}{section}

\usepackage{ifthen}
\usepackage{intcalc}
\usepackage{tikz}
\usepackage{multirow}
\usetikzlibrary{decorations.pathreplacing,calligraphy}
\usetikzlibrary{arrows.meta, decorations.text, decorations.markings, decorations.pathreplacing, angles, quotes, calligraphy}
\definecolor{bittersweet}{rgb}{1.0, 0.44, 0.37}

\theoremstyle{plain}
\newtheorem{theorem}{Theorem}[section]
\newtheorem{lemma}[theorem]{Lemma}

\theoremstyle{definition}
\newtheorem{definition}[theorem]{Definition}

\theoremstyle{remark}

\DeclareMathOperator{\DataSig}{DataS{\text{\i}}g}

\newcommand{\m}{\ensuremath{{\cal M}}}
\newcommand{\n}{\ensuremath{{\cal N}}}

\newcommand{\s}{\ensuremath{{\cal S}}}

\newcommand{\p}{\ensuremath{{\cal P}}}
\newcommand{\ca}{\ensuremath{{\cal A}}}
\newcommand{\cb}{\ensuremath{{\cal B}}}
\newcommand{\cc}{\ensuremath{{\cal C}}}
\newcommand{\cd}{\ensuremath{{\cal D}}}

\newcommand{\ck}{\ensuremath{{\cal K}}}
\newcommand{\ci}{\ensuremath{{\cal I}}}

\newcommand{\cl}{\ensuremath{{\cal L}}}

\newcommand{\cv}{\ensuremath{{\cal V}}}

\newcommand{\cH}{\ensuremath{{\cal H}}}
\newcommand{\cY}{\ensuremath{{\cal Y}}}

\newcommand{\al}{\alpha}
\newcommand{\be}{\beta}

\newcommand{\de}{\delta}

\renewcommand{\th}{\theta}

\newcommand{\vph}{\varphi}
\newcommand{\ep}{\varepsilon}

\newcommand{\R}{\ensuremath{{\mathbb R}}}
\newcommand{\Q}{\ensuremath{{\mathbb Q}}}
\newcommand{\N}{\ensuremath{{\mathbb N}}}
\newcommand{\B}{\ensuremath{{\mathbb B}}}

\newcommand{\Z}{\ensuremath{{\mathbb Z}}}
\newcommand{\C}{\ensuremath{{\mathbb C}}}

\newcommand{\bS}{\ensuremath{{\mathbb S}}}

\newcommand{\E}{\ensuremath{{\mathbb E}}}

\newcommand{\bbP}{\ensuremath{{\mathbb P}}}

\newcommand{\bx}{\textbf{ x}}
\newcommand{\by}{\textbf{ y}}
\newcommand{\ba}{\textbf{ a}}

\newcommand{\bl}{\textbf{ l}}
\newcommand{\bv}{\textbf{ v}}
\newcommand{\bolde}{\textbf{ e}}
\newcommand{\boldf}{\textbf{ f}}

\newcommand{\bA}{\textbf{ A }}
\newcommand{\bB}{\textbf{ B}}

\newcommand{\bKK}{\textbf{ K}}
\newcommand{\bLL}{\textbf{ L}}
\newcommand{\bJ}{\textbf{ J}}
\newcommand{\bW}{\textbf{ W}}
\newcommand{\bR}{\textbf{ R}}

\DeclareMathOperator{\inj}{inj}
\DeclareMathOperator{\proj}{proj}

\DeclareMathOperator{\sign}{sign}

\DeclareMathOperator{\LeadLag}{ Lead-Lag}
\DeclareMathOperator{\Time}{Time}
\DeclareMathOperator{\Timediff}{Time-diff}
\DeclareMathOperator{\inv}{inv}
\DeclareMathOperator{\Path}{Path}
\DeclareMathOperator{\Span}{Span}

\DeclareMathOperator{\Normal}{Normal}
\DeclareMathOperator{\Anomaly}{Anomaly}

\DeclareMathOperator{\dyadic}{dyadic}
\DeclareMathOperator{\overlap}{overlap}
\DeclareMathOperator{\var}{var}
\DeclareMathOperator{\RS}{RS}

\newcommand{\km}{{\rm km}}

\newcommand{\beq}{\begin{equation}}
\newcommand{\eeq}{\end{equation}}
\newcommand{\beqa}{\begin{equation}\begin{aligned}}
\newcommand{\eeqa}{\end{aligned}\end{equation}}
\newcommand{\brmk}{\begin{rmk}}
\newcommand{\ermk}{\end{rmk}}
\newcommand{\partref}[1]{\hbox{(\csname @roman\endcsname{\ref{#1}})}}

\newcommand{\K}{{\mathrm{K}}}

\newcommand{\Sig}{{\mathrm{Sig}}}
\newcommand{\LogSig}{{\mathrm{LogSig}}}
\newcommand{\LogODE}{{\mathrm{LogODE}}}

\newcommand{\Shuff}{{\mathrm{Shuff}}}
\newcommand{\bX}{{ \textbf{X} }}

\makeatletter
\def\dual#1{\expandafter\dual@aux#1\@nil}
\def\dual@aux#1/#2\@nil{\begin{tabular}{@{}c@{}}#1\\#2\end{tabular}}
\makeatother

\makeatletter
\def\three#1{\expandafter\three@aux#1\@nil}
\def\three@aux#1/#2/#3\@nil{\begin{tabular}{@{}c@{}c@{}}#1\\#2\\#3\end{tabular}}
\makeatother

\makeatletter
\def\four#1{\expandafter\four@aux#1\@nil}
\def\four@aux#1/#2/#3/#4\@nil{\begin{tabular}{@{}c@{}c@{}c@{}}#1\\#2\\#3\\#4\end{tabular}}
\makeatother

\title{Signature Methods in Machine Learning}
\author{Terry Lyons \& Andrew D. McLeod}
\date{ \today}

\begin{document}
\usetagform{red}
\maketitle
\begin{abstract}
Signature-based techniques give mathematical insight into the 
interactions between complex streams of evolving data. These 
insights can be quite naturally translated into numerical 
approaches to understanding streamed data, and perhaps because 
of their mathematical precision, have proved useful in analysing 
streamed data in situations where the data is irregular, and not 
stationary, and the dimension of the data and the sample sizes 
are both moderate.

Understanding streamed multi-modal data is exponential: a word 
in $n$ letters from an alphabet of size $d$ can be any one of 
$d^n$ messages. 
Signatures provide a ``lossy compression" of the information 
contained within such a stream by filtering out the 
parameterisation noise.
More concretely, suppose we have a time series with $3$ 
channels and $N$ samples.
There are $1 + 3N + \frac{3N(3N+1)}{2} = 
1 + \frac{9}{2}N(N+1)$ linearly independent quadratic 
polynomials defined on the time series.
But the signature of this time series truncated to depth
$2$ only consists of $1 + 3 + 3^2 = 13$ components
which is, in particular, independent of the number of samples
$N$.
However, whilst the independence of the number of samples $N$ 
has removed an exponential amount of noise, the dependence on 
the number of channels to the power of the depth ensures that 
an exponential amount of information remains.


This survey aims to stay in 
the domain where that exponential scaling can be managed directly. 
Scalability issues are an important challenge in many problems 
but would require another survey article and further ideas. 
This survey describes a range of contexts where the data sets 
are small and the existence of small sets of context free and 
principled features can be used effectively.

The mathematical nature of the tools can make their 
use intimidating to non-mathematicians. The examples 
presented in this article are intended to bridge this 
communication gap and provide tractable working examples 
drawn from the machine learning context. Notebooks are 
available online for several of these examples. 
This survey builds on the earlier paper of 
Ilya Chevryev and Andrey Kormilitzin which had 
broadly similar aims at an earlier point in the 
development of this machinery. 

This article illustrates how the theoretical insights 
offered by signatures are simply 
realised in the analysis of application data in a way 
that is largely agnostic to the data type. Larger and 
more complex problems would expect to address 
scalability issues and draw on a wider range of 
data science techniques.  

The article starts with a brief discussion of background 
material related to machine learning and signatures.
This discussion fixes notation and terminology whilst
simplifying the dependencies, but these background
sections are not a substitute for the extensive 
literature they draw from. 

Hopefully, by working some of the examples the reader 
will find access to useful and simple to deploy tools; 
tools that are moderately effective in analysing 
longitudinal data that is complex and irregular in 
contexts where massive machine learning is not a 
possibility.
\end{abstract}

\begin{center}
\textbf{93C15, 68Q32, 34F05} (Secondary).
\end{center}

\tableofcontents

\section{Introduction}
\label{intro}
Streams of sequential information can be found almost
everywhere 
in our society. They need to be understood, analysed, and 
interpreted. It might be a stream of spoken words, financial 
data, or ones hospital records. The simplest example of 
sequential information is probably a text. One sees quickly 
that the number of words is exponential in the number of 
letters. Most streams involve multiple channels, and 
understanding the consequences of the ``data" involve 
understanding how the channels interleave. Suppose one 
channel recorded arrivals of stock from deliveries at a 
supermarket. Another involved the demand from customers. 
To predict sales requires a detailed understanding of how 
these channels interact. The order of events matters. This 
can be very expensive using classical methodology. Smoothing 
the deliveries and the demand result in data that does not 
indicate the shortfall because the demand came before the 
arrival of deliveries. 

A stream of sequential information can be interpreted as 
a path. Reparameterisation then gives an infinite dimensional
group of symmetries. 
Typically symmetries are problematic for machine learning; 
different representations of the same object introduce an 
additional fact that the model must learn. 
One route around this issue for paths is to work with the 
\textit{unparameterised path}; that is, the equivalence class
with respect to the group of reparameterisations (sometimes
called a \textit{shape} or a \textit{contour}).
Only the data itself is important and not the particular 
choice of representation; we only want to extract information 
provided by the data that is \textit{independent} of the choice
of how the data is viewed. 
For example, we may be interested in which character a 
sequence of pen strokes results in, but it is of no importance
how quickly the character is drawn. A figure '3' is the same 
regardless of how quickly it is written.

The signature of a path is a forgetful summary that is 
intrinsically multidimensional. It captures the order in 
which things happen, the order in which the path visits 
locations, but completely ignores the parameterisation of 
the data. The parameterisation noise is filtered out.
It does not matter if the path was sampled 75 
times because there were 75 deliveries that week and 65 
times the week after. But it remembers the order of 
events perfectly. In a natural way, it produces a vector 
embedding of the path that is not sensitive to how often 
the path is sampled but is fully sensitive to the order 
of events on the different channels. This can allow the 
detection of such events far more efficiently than the 
customary time series approaches. The signature of a path 
is a mathematical transform that faithfully describes the 
curve, or the unparameterised path, but removes completely 
the infinite dimensional group of parameterisation based 
data symmetries. 

In this article, we survey a variety of contexts where 
signatures have been applied to machine learning. This 
survey is intended to act as a pseudo-sequel to the 
introduction provided in \cite{CK16}.
The applications selected for inclusion illustrate that the 
theoretical benefits, guaranteed by the underlying theory of 
path signatures, are being fully realised in practical 
problems. In some sense, our article may be thought of 
as a continuation of Section 2 in \cite{CK16} to cover 
more sophisticated uses of the signature transform in 
machine learning made after the appearance of \cite{CK16}. 

In the interest of readability we do not present this 
article as a proper sequel to \cite{CK16}. 
Instead, we cover some over-lapping material within 
our first few sections, hence the term pseudo-sequel. 
The inclusion of such material is two-fold.
Firstly, they provide a minimal coverage of the material 
required for the later surveys of the applications of
signature techniques for the reader unfamiliar with 
the basic ideas, making this work more self-contained.
Secondly, they allow us to fix notation and terminology 
that will be used later, forming a sort-of dictionary 
that can be referred back to during the latter sections.

The over-lapping material is approached from a 
different perspective to the presentation in \cite{CK16}. 
Subsection \ref{subsec:Illustrative_Example} provides a 
very basic introduction to regression via a simple example 
of monomial regression on the unit interval $[0,1] \subset \R$.
Subsection \ref{subsec:inc_stream_data} presents a sensible 
and informative way to convert a data stream of values to 
a data stream of increments. In particular, no information 
is lost under this transform.
Subsection \ref{subsec:math_framework} fixes the mathematical
framework for the regression problem that we will consider
throughout this article.

Subsection \ref{SRA} introduces some basic strategies to 
tackling the regression problem introduced in Subsection 
\ref{subsec:math_framework}.
The viewpoint adopted in Subsection \ref{SRA} is that of a 
beginner who has no prior experience with machine learning.
The machine learning 
material included in Section \ref{SRA} is only the 
essentials we later require. The reader seeking
a more comprehensive introduction to machine learning
is directed to the book \cite{ML_book}. 
The book \cite{ML_book} covers the fundamentals of 
\textit{machine learning}, \textit{classification}, 
\textit{regression}, \textit{neural networks} 
and \textit{deep learning} with exercises and coded examples. 
The more recent book \cite{LMR22} covers a broadly similar 
collection of topics whilst also providing coded examples. 
A major difference between \cite{ML_book} and \cite{LMR22}
is that the coded examples in \cite{ML_book} use
\href{https://www.tensorflow.org/}{\textit{TensorFlow}}, 
whilst the coded examples in \cite{LMR22} use
\href{https://pytorch.org/}{\textit{PyTorch}}.
In addition, both \cite{ML_book} and \cite{LMR22} include
introductions to more recent developments within the field 
of data science including, for example, 
\textit{large language models}, \textit{attention}, and 
\textit{transformers}.
A more theoretical perspective of the field can be found 
in \cite{FHT09}, for example.

Subsection \ref{dataset_to_path} explains the natural 
correspondence between a stream of increments and a 
continuous path. Additionally, Subsection 
\ref{dataset_to_path} presents some simple augmentations 
that can be applied to streamed data. Only augmentations 
used in later applications are covered. 

Subsection \ref{SI} informally introduces the signature 
transform as an attractive method of summarising a stream 
of increments via summarising the corresponding continuous 
path (cf. Subsection \ref{dataset_to_path}).
After informally introducing the signature as a method to 
summarise a path in Section \ref{SI} that captures all the 
relevant information, the mathematical framework required 
to rigorously talk about signatures is covered in Section 
\ref{Ten_Alg}. 
Keeping in mind our adopted machine learning perspective, 
signature properties and related theorems are all stated first
for continuous paths of bounded $p$-variation for
$1 \leq p < 2$ in Subsection \ref{sig_p<2_sec}. 
For the majority of this article we stay within the setting 
of considering signatures of continuous paths of bounded 
$1$-variation; and so the main theoretical requirements 
are the results presented in Subsection \ref{sig_p<2_sec}.

However, in places we necessarily require the more involved 
theory of \textit{rough paths}. 
Indeed, it is well known that the path $t \mapsto B_t$
for a standard Brownian motion $B_t$ only has finite 
$p$-variation for $p > 2$.
Thus the theory of rough paths is required to make sense
of the signature of a path arising from the evolution of 
Brownian motion. 
Consequently we provide a terse presentation of the aspects 
of rough path theory that we will require in Subsection 
\ref{rough_paths_sec}.

As a pseudo-sequel to \cite{CK16}, we assume the reader 
has had some exposure to signatures. Our discussion of 
signatures assumes a greater mathematical maturity than
\cite{CK16} and, in contrast to the example-heavy 
introduction given in \cite{CK16}, we favour a 
succinct presentation of the key theoretical aspects. 
Only the mathematical properties of signatures 
directly relevant to the later applications are included 
in Subsections \ref{Ten_Alg}, \ref{sig_p<2_sec} 
and \ref{rough_paths_sec}.
A wider range of properties of path signatures can be 
found in the survey article \cite{Lyo14}, whilst the 
reader seeking a comprehensive introduction to the theory 
is directed to the lecture notes \cite{CLL04}.

The entirety of Section \ref{background_material} is 
included with the aim of providing a dictionary for 
terminology and notation used throughout the later 
sections.
The presentation of this material is not designed to 
provide an in-depth coverage; only the specific elements 
required to understand the later applications are covered.
The reader already comfortable 
with \textit{Regression}, \textit{Feature Maps}, 
\textit{Kernels}, \textit{Rough Paths}, 
and the mathematical theory of \textit{Signatures}
may safely skim Subsections 
\ref{subsec:Illustrative_Example} -- \ref{rough_paths_sec}.

The material covered ceases over-lapping with the content 
of \cite{CK16} in Section \ref{good_props}. For the readers 
ease of navigation, we briefly
summarise the content included in the remaining sections.

\begin{itemize}
    \item Section \ref{good_props} provides both a basic 
    outline for the use of the signature transform as 
    a feature map, and the theoretical 
    results underpinning the suitability of the signature 
    transform as a feature map.
    \item Section \ref{CDE_log-ODE} introduces 
    \textit{Controlled Differential Equations} (CDEs) 
    and covers the \textit{log-ODE} method. The log-ODE method
    is a technique for numerically solving CDEs using 
    the \textit{log signature},
    which is itself introduced in Subsections 
    \ref{Ten_Alg}, \ref{sig_p<2_sec} 
    and \ref{rough_paths_sec}.
    \item Section \ref{coding_packages} provides an 
    introduction to the 
    \href{https://roughpy.org/}{\textit{RoughPy}}
    \cite{morley2024roughpy}
    python package for computing truncated signatures
    and log signatures.
    \item Section \ref{expect_sig} explains how signatures 
    can be extended to distributions \textit{without} any 
    restrictions on the distributions, 
    as in \cite{CO18} for example. We additionally 
    outline how this theory offers an approach to 
    distribution regression following \cite{BDLLS20}.
    \item Section \ref{Trunc_order} covers the work 
    \cite{BKLPS19} proposing how the signature transform 
    may be incorporated into a neural network. A particular 
    consequence is that the appropriate depth to 
    truncate the signature to should be treated as 
    a trainable parameter, and hence be determined by the data.
    \item Section \ref{full_sig_kernel_section} focuses 
    on the fact that the signature transform is a kernel. 
    It covers the work of \cite{CFLSY20} detailing 
    how the associated kernel function can be
    effectively approximated. This allows the 
    full signature transform to be used \textit{without} 
    truncation.
    \item Section \ref{LSDE} covers the introductions of
    \textit{Neural Controlled Differential Equations} 
    (Neural CDEs) made in \cite{FKLM20} and of 
    \textit{Neural Rough Differential Equations} 
    (Neural RDEs) made in \cite{FKLMS21}. The universality 
    guarantees associated with Neural CDEs are fundamentally
    reliant on the theory of path signatures, whilst 
    Neural RDEs are built around the theory of 
    log signatures.
    \item Section \ref{speech_emot_rec} covers the work 
    of \cite{LLNNSW19} on how signature transform methods 
    can be used in the extraction of emotion 
    from segments of speech.
    \item Section \ref{health_app} covers a range of 
    medical applications of signature transform methods.
    These include distinguishing between Bipolar Disorder 
    and Borderline Personality Disorder 
    \cite{AGGLS18,LLNSTWW20}, diagnosing 
    Alzheimer's disease \cite{GLM19}, 
    information extraction from medical prescriptions 
    \cite{BLNW20}, and early detection of Sepsis
    \cite{HKLMNS19,HKLMNS20}.
    \item Section \ref{LHAR_app} covers the work of 
    \cite{LNSY17} using signature transform methods to 
    recognise human actions from the motion
    of 20-40 markers located on the person. A python 
    notebook exists (see Subsection \ref{demo_notebook}) 
    allowing the reader to train their own classifier 
    (following a variant of the method proposed in 
    \cite{LNSY17}). 
    \item Section \ref{dist_reg_expect_sig} covers 
    the work of \cite{BDLLS20} realising the use of 
    a generalised signature transform for distribution 
    regression both with and without truncation of the 
    involved signatures.
    \item Section \ref{anom_detec} covers the use of  
    signatures for anomaly detection proposed in \cite{CCFLS20}.
    \item Section \ref{sec:randomised_signature} covers  
    \textit{randomised signatures} as developed in the works
    \cite{CGGOT21,AGTZ22}, and illustrates their use in 
    \cite{AGTZ22} for the detection of market manipulation 
    attempts from financial data.
\end{itemize}
\vskip 4pt
\noindent 
Whilst a wide range of applications are covered within this 
article, it remains a far from comprehensive presentation 
of the uses of path signature techniques within
machine learning.
Indeed the list of recent works \textit{not} discussed within 
this article includes uses of signature techniques in 
finance \cite{CGS22,NS22,BBFP23,CM23}, 
optimal stopping problems \cite{BHRS21,BPS23}, 
stochastic differential equations \cite{CST23a}, 
universal approximation theory \cite{CPS22,GRST22,CST23b},
neural network stability \cite{BFT22}, 
generalized Magnus expansion \cite{FHT21},
functional Taylor expansions \cite{DT22,DT25}, 
economic nowcasting \cite{CLMMNPRSSY23}, and 
longitudinal language modelling \cite{FLLTT23,BCKLLTT23}.
Moreover this list of works not included in this article 
is far from exhaustive.
Including every topic within a single article is infeasible.
But hopefully, after working their way through this article, 
the reader will be suitably equipped to read and understand 
other works considering the use of path signature techniques 
within a machine learning context.
\vskip 4pt
\noindent
\emph{Acknowledgements}: This work was supported by 
the $\DataSig$ Program under the EPSRC grant 
ES/S026347/1, the Alan Turing Institute under the
EPSRC grant EP/N510129/1, the Mathematical Foundations of
Intelligence: An ``Erlangen Programme" for AI under the
EPSRC Grant EP/Y028872/1,
the Data Centric Engineering 
Programme (under Lloyd's Register Foundation grant G0095),
the Defence and Security Programme (funded by the UK 
Government) and the Hong Kong Innovation and Technology 
Commission (InnoHK Project CIMDA). 
The example notebooks available via the $\DataSig$ website
were created by Peter Foster with various members of the 
$\DataSig$ team. This work was funded by the Defence
and Security
Programme (funded by the UK Government).
For the purpose of open access, the author has applied a
CC BY public copyright licence to any Author Accepted Manuscript
(AAM) version arising from this submission.
We thank the anonymous referee for both their careful reading 
of and valuable constructive feedback on an earlier version.

\section{Background Material}
\label{background_material}

\subsection{Illustrative Toy Regression Problem}
\label{subsec:Illustrative_Example}
Regression is a basic, yet extremely common, learning 
task in which real-valued functions are modelled using 
sets of data. 
The fundamental goal of regression is to learn a function
that predicts outcome values based on a specified 
collection of input features.
The following simple example of regression involving the 
consideration of the monomials on the unit interval 
$[0,1] \subset \R$ exhibits some of the theoretical 
considerations that are central to the use of signatures
within machine learning.

Suppose we want to learn
a continuous function $\rho : [0,1] \to \R$ using its 
values at a given collection of points 
$x_1,\ldots ,x_N \in [0,1]$, for some $N \in \N$. 
We assume that the points $x_1, \ldots ,x_N$ give a 
``good approximation" (in some sense) of $[0,1]$,
and that we want to learn the response of $\rho$ 
to any input in $[0,1]$.
An elementary rarely-used, yet still insightful, 
approach is to consider the functions 
$\phi_k : [0,1] \to \R$ for $k \in \N$, 
defined by $\phi_k(x) := x^k$, as the feature functions,
i.e. the monomials.
Then we fix a (large) $K_0 \in \N$ and try to express
\begin{equation}
	\label{lin_comb_poly}
		\rho \approx \sum_{k=0}^{K_0} a_k \phi_k.
\end{equation}
If \eqref{lin_comb_poly} is possible, then the 
coefficients $a_k$ are given by the linear equations
\begin{equation}
	\label{coeffs}
		y_i = \sum_{k=0}^{K_0} a_k \phi_k (x_i)
\end{equation}
for each $i \in \{1, \ldots , N \}$. We want to use 
the linear combination in \eqref{lin_comb_poly} to 
predict the value of $\rho$ at new instance in 
$[0,1]$. If the equations \eqref{coeffs} are 
non-degenerate, then the solution is unstable; 
any new instance will lead to a completely different 
set of coefficients $a_k$. Hence it is desirable that 
these equations should be degenerate, and solving 
\eqref{coeffs} often involves 
numerical techniques such as \textit{Singular Value 
Decomposition} (SVD) or \textit{gradient descent} 
algorithms involving suitable cost functions.
Degeneracy in the equations can also be handled via 
regularisation techniques including, for example,
\textit{Ridge regression}, \textit{LASSO regression}, and 
\textit{dropout}.
An in-depth discussion of such approaches may be found in 
\cite{ML_book} or \cite{LMR22}, for example.

The success of this approach to approximating $\rho$ 
is fundamentally reliant on the linear span 
of the set $\{ \phi_k \mid k \in \N \}$ being dense within 
the class of functions $C^0([0,1];\R)$.
In this case, since the monomials span an algebra and the 
unit interval is a compact subset of $\R$, the 
\textit{Stone--Weierstrass theorem} (see \cite{Sto48}, 
for example) provides the required density.

\subsection{Incremental Streamed Data}
\label{subsec:inc_stream_data}
A fundamental idea central to data science, 
illustrated by the simple monomial example in 
Subsection \ref{subsec:Illustrative_Example}, is 
that if we are able to represent data as vectors 
in some Euclidean space, 
then we can approximate continuous functions on compact 
domains by polynomials and learn the polynomial from the data.
From a theoretical perspective this is valid; the 
\textit{Stone--Weierstrass} theorem \cite{Sto48} tells us that,
on compact subsets, polynomials are a dense subset of the 
continuous functions. 

In practice this approach can work well when the data 
consists of time-series with moderate numbers of both 
channels and samples. 
Examples of tasks on which polynomial regression performs 
well include predicting the median house price for a 
district using 10 recorded attributes per district 
(see Chapter 2 in \cite{ML_book}), predicting the 
\textit{Traction Energy Consumption} of an urban car journey
based on knowledge of the time-of-day at which the journey 
is made \cite{GK19}, and predicting the number of deaths 
resulting from COVID-19 in US states based on four independent 
parameters \cite{BS22}.

But this approach runs into problems for tasks in which the 
order of events is important. In order to consider a toy problem
illustrating one particular issue, 
let $N \in \Z_{\geq 1}$ and suppose we have a
finite collection of time series, with each time series 
consisting of $2$ channels and $N$ samples.
Suppose that each time series $\bx$ has the following
structure. 
If $\bx = \{ (x_{1,i} , x_{2,i}) \}_{i=1}^N$ then the 
following properties are true.
\begin{itemize}
    \item For every $i \in \{1, \ldots , N \}$ we have 
    $x_{1,i} , x_{2,i} \in \{0,1\}$.
    \item For $j \in \{1,2\}$ if $i \in \{1, \ldots , N\}$ 
    and $x_{j,i} = 1$, then for every $k \in \{i , \ldots , N\}$
    we have $x_{j,k} = 1$.
    \item At least one of $x_{1,N}$ and $x_{2,N}$ is
    equal to $1$.
\end{itemize}
Consider the task of
determining which channel is the first to change from $0$ 
to $1$.

Each time series $\bx$ can be viewed
as an element in $\R^{2N}$. We could try to learn a 
quadratic polynomial that can determine which channel is 
the first to change from $0$ to $1$. However, ignoring any 
considerations regarding whether a quadratic polynomial 
capable of doing this even exists, we observe that there 
are $1 + 3N + 2N^2$ linearly independent quadratic 
polynomials determined on $\bx$.
Consequently we must consider linear combinations of 
$1 + 3N + 2N^2$ elements in order to learn our desired
quadratic polynomial. 
This quickly becomes intractable as the number of samples 
increases. 

A large number of samples is reasonable. The sampling 
frequency needs to be high enough to separate the increases
of the channels. 
If, for example, the time series correspond to a record of a 
patients heart rate and body temperature, with the change from 
$0$ to $1$ represents an increase in the heart rate or the 
body temperature respectively, then it may be the case that 
separating which one increases first could required taking a 
sample each second. 
This would result in $N=3600$ samples over each hour, leading
to the consideration of $25930801$ linearly independent 
quadratic polynomials. 

As we will see later (cf. Section \ref{coding_packages}), 
the signature provides 
a summary of time series that efficiently captures the 
order of events. In the toy problem above, 
the signature truncated to depth $2$ can 
determine which channel changes from $0$ to $1$ first. 
This involves only $7$ components rather than the 
$1 + 3N + 2N^2$ components of a linear combination of a 
basis of quadratic polynomials.
The key advantage is that the number of signature 
components required is \textit{independent} of the number 
of samples (though it does still grow exponentially with 
the number of channels).

Using the signature to summarise a time-series 
requires associating the time-series with a continuous path.  
A commonly used method is making a choice of rule for 
continuously interpolating between the entries of the 
time-series; examples include linear interpolation, 
rectilinear interpolation, or cubic spline interpolation.
A weakness of this approach is that the choice of interpolation 
is typically ad-hoc; there is no given natural best choice.
However, if the time-series is a stream of \textit{increments},
then the concatenation of the entries provides a sensible 
route to determining a path with which the time-series 
can be associated (cf. Section \ref{dataset_to_path}).
This natural association makes \textit{incremental} streamed 
data (i.e. streams whose entries are all incremental changes) 
more desirable for use with path signature techniques.

The incremental change in a channels values is often 
vital information. For example, suppose we have a stream 
containing the recorded temperatures of a hospital patient.
Observing the value $38^{\circ}C$ alone is insufficient 
to conclude whether the patient is doing well or not.
Knowledge of this values relation the the previous value 
is required. 
If the temperature was previously 
$39^{\circ}C$ then the new value of $38^{\circ}C$ suggests
a positive trajectory for the patient; if the previous 
value of the temperature however was $37^{\circ}C$, then 
the new value $38^{\circ}C$ suggests a more concerning 
trajectory for the patient, indicating that some form of
medical intervention may be required.

Moreover, there are situations in which \textit{only} the 
incremental changes are relevant, i.e. when the underlying 
actual values are not relevant. 
For example, doctors may be interested in knowing that
a patient has taken 2 paracetamol tablets since yesterday, 
but they will not be interested in knowing the precise 
number of tablets the patient has taken over their entire 
lifetime.
Similarly, in a financial context, the amount of volatility 
experienced over the past hour can be informative; the 
total amount of volatility experienced since the 
beginning-of-time is not.
When planning a journey, the number of buses from a 
particular stop in the next hour could be helpful; but 
knowing how many buses have ever departed from that stop 
is likely not helpful.
An interesting facet of these examples is that the irrelevancy 
of the actual value is, at least in part, due to its 
lack of availability. 

However, it is important to realise that their are situations 
in which the incremental changes alone are insufficient.
Indeed in the hospital patient temperature records example
above, observing the incremental change of $+0.5^{\circ}C$ 
is insufficient to conclude whether the patient is doing well.
If the last value before this change was 
$35^{\circ}C$ then this increase is a good sign. 
But if the last value before this change was $38^{\circ}C$ 
then this increase is not a good sign, indicating 
that some form of medical intervention may be required.

An important takeaway from this example is that knowledge of 
both a quantities actual values \textit{and} the incremental 
changes of the quantity are required to determine the context 
of the information provided by a stream.
Consequently, with the aim of using path signature techniques
to summarise streams, we seek a way to record/store \textit{any}
given stream as a stream of increments that encodes both 
the incremental changes of the values and the actual values 
themselves.
A definition of a stream can be found in Subsection 
\ref{subsec:math_framework} (cf. \eqref{eq:streams_def}).

Before turning our attention to achieving this, we first 
observe a third aspect of a data stream that is desirable to 
capture. Namely, whether each entry of the stream is the 
result of a new measurement being taken, or if it 
is the result of the previous value being carried across 
without a new measurement being taken.
The importance of this distinction can be again seen via 
the example of a hospital patients recorded temperatures.
Suppose the same temperature of $36.5^{\circ}C$ is recorded 
in successive entries. 
Then whether or not the second entry is a new measurement 
provides information. 
If it is not a new measurement, then it indicates that the 
patients condition is deemed stable enough to not require 
a new check of their temperature.
But if it is a new measurement, then it indicates that the 
patients condition was deemed concerning enough, for 
some unknown reason, to warrant a new check of their
temperature.

We seek a method of transforming any stream into a stream 
of increments that satisfies the following properties.

\begin{itemize}
    \item No information is lost; any information present
    in the original stream can be recovered from the 
    resulting stream of increments.
    \item Both the incremental changes and the actual 
    values of each channel of the original stream are 
    encoded in the resulting stream of increments.
    \item If available, whether or not each entry of 
    each channel of the original stream is the result of a
    new measurement or not is encoded in the resulting 
    stream of increments.
    \item The transform is injective; distinct streams 
    should be transformed to distinct streams of increments.
\end{itemize}
\noindent
We now informally discuss a transformation satisfying 
these properties in the simple setting of having a stream 
$\bx = (x_1 , \ldots , x_n)$ of real numbers; that is, for
every $i \in \{1, \ldots , n\}$ we have $x_i \in \R$. 
The first step is to mimic a ``pen-on, pen-off" style 
transformation and introduce a second channel recording a
$1$ for each entry that is the result of a new measurement, 
and a $0$ for each entry that is not the result of a 
new measurement. 
This results in a new stream $\bx' = \Big( (x_1 , a_1) , 
\ldots , (x_n , a_n) \Big)$ such that, for each 
$i \in \{1, \ldots , n\}$, $x_i \in \R$ and $a_i \in \{0,1\}$.
If it is unknown whether $x_i$ is the result of a new 
measurement or not we take $a_i := 1$; by default, we 
assume that each recorded value is the result of a new 
measurement unless explicitly told otherwise.

With an eventual aim of converting the stream $\bx'$ into 
a stream of incremental changes, we first encode 
information about the values $x_1 , \ldots , x_n$ in a 
manner that is invariant under translation.
We do this by explicitly encoding both the first value 
$x_1$ and the last value $x_n$ via ``invisibility-reset" type 
augmentations. 
To be precise, we replace the stream $\bx'$ by the new 
stream 
$\bx' = \Big( (0,0) , (x_1 , 0)  , (x_1 , a_1) , 
\ldots , (x_n , a_n) , (x_n, 0) , (0,0) \Big)$.
The ``invisibility-reset" aspect to encode the first value 
$x_1$ and the last value $x_n$ is done via a ``Lead-Lag" type 
augmentation in which one channel updates, and then there is 
a definite delay until the other channel updates. 
In this case, at the start we first update $0$ to $x_1$ and 
then update $0$ to $a_1$ after a delay; whilst at the end 
we update $a_n$ to $0$, and then update $x_n$ to $0$ after 
a delay.

This ``Lead-Lag" delayed updating (cf. Subsection 
\ref{dataset_to_path}) is a simple example of a sliding 
variable giving information at a point about what is 
happening on a neighbourhood of the point.
Unfortunately it accidentally fixes a parameterisation of 
the data set via the explicit choice of the number of delay 
steps. 
It would be possible to design time-invariant contextual 
variables to provide information at a point about what 
is happening on a neighbourhood of the point.
For example, taking the signature of a path over 
previous interval started at the time the path \textit{last} 
entered the closed unit ball centred at the current location
would give a more path intrinsic method.

Choosing to explicitly encode the first and last values 
is natural from a modelling perspective. 
It is these values that are particularly informative 
for a wide variety of tasks. 
For example, if our aim is to predict the subsequent 
values of a quantity (i.e. predict the next values of 
stock prices, patients body temperature etc) then the 
last known value of the quantity will be of particular 
significance.
But if the task is classify the outcome of a quantities 
evolution (i.e. medical diagnosis based on medical records,
whether an urban areas infrastructure will be sufficient to 
deal with demand etc) then the first value will be of 
particular significance. In these cases it is desirable 
to understand from the first recorded measurement what the 
likely outcome will be.

Moreover, the first and last entries of the stream are 
distinct from every other entry in that they only have 
one neighbouring entry. Every other entry has both a 
proceeding entry and a subsequent entry with which it 
can be compared. The first entry has no proceeding entry, 
and the last entry has no subsequent entry. 
Augmenting the stream by adding in additional entries 
to proceed the first entry and follow the last entry 
is a sensible way to make the first and last entries 
more uniform with respect to the other entries.

We choose to use the value $0$ to both proceed $x_1$ 
and succeed $x_n$. As mentioned previously, one benefit 
of this is that the value $x_1$ and $x_n$ will be explicitly
encoded in a translation invariant manner. In particular, 
they will remain encoded after the stream is replaced 
by its stream of incremental changes.
Another benefit is that comparing $x_1$ and $x_n$ with 
$0$ makes sense for all data types; it represents
comparison with the default setting of not having a 
measurement. 
Whilst in certain contexts it is arguably more sensible 
to compare these values with some notion of average (i.e. 
compare body temperature with average human body temperature 
value), there are settings in which such an average value 
does not make sense.
For example, if the stream represents a series of x-rays of 
a patient, it is unclear what meaning should be given to 
the average x-ray. But the notion of not having an x-ray 
does make sense.

Finally, the stream $\bx'$ is replaced by the stream of 
its incremental differences. That is, we set
\begin{equation}
    \label{eq:resulting_increment_stream}
        \bx' := \Big( (x_1,0) , (0,a_1) , 
        (x_2-x_1, a_2-a_1) , \ldots , 
        (x_n - x_{n-1}, a_n - a_{n-1}) , 
        (0 , -a_n) , (-x_n,0) \Big).
\end{equation}
This transformation takes a stream with one channel to 
a stream with two channels. 
For a general stream with $d$ channels, we will apply 
the transformation outlined above to each channel. Hence 
overall it will result in the stream with $d$ channels 
being transformed to a stream with $2d$ channels (cf. 
Subsection \ref{subsec:math_framework}).

We end this subsection with a brief discussion of some 
possible redundancies in this transformation.
The first is that if the value of the entries of the 
stream are known to be of no importance, then avoiding 
the invisibility-reset aspect of the transformation is 
a more cost-effective option. 
The second is that we only really require implementing 
the invisibility-reset aspect to one of $x_1$ or $x_n$.
After it has been implemented to one of these values, 
the other value is then recoverable from the resulting 
stream of increments. 
Our choice to include this aspect for both the first 
and last entries is made for two main reasons. 
Firstly, it emphasises the extra importance placed on 
the first and last values. 
Secondly, it makes our transformation more symmetrical, 
in particular ensuring that the concatenation of 
the resulting increments forms a closed path based at the
origin.

\subsection{Mathematical Framework}
\label{subsec:math_framework}
In this subsection we introduce the mathematical framework 
that we work in throughout this article.
Given an arbitrary set $\m$ we define the collection of 
\textit{streams} in $\m$ to be the set
\begin{equation}
    \label{eq:streams_def}
        \s(\m) := \bigcup_{k=1}^{\infty} \m^k
        = 
        \bigcup_{k=1}^{\infty} 
        \big\{ (x_1 , \ldots , x_k) \mid 
        x_1 , \ldots , x_k \in \m \big\}.
\end{equation}
Throughout this article we will consider streams in the 
setting that the set $\m$ is a real Banach space (complete 
normed linear vector space) that we denote by $W$. 
Moreover, we will assume that this Banach space $W$ is finite 
dimensional with dimension $d \in \Z_{\geq 1}$.

We consider the following regression problem.
Suppose that $M \in \Z_{\geq 1}$ is finite and we have
a collection of pairs $\big\{ (\bx_i , y_i) \big\}_{i=1}^M$
where, for each $i \in \{1, \ldots , M\}$,  $\bx_i \in \s(W)$  
and $y_i \in \R$.
The underlying idea is that for each $i \in \{1, \ldots , M\}$
the value $y_i \in \R$ represents the underlying systems 
response to the stream $\bx_i$.
Suppose we have a subset $\m \subset \s(W)$ such 
that $\{ \bx_1 , \ldots , \bx_M \} \subset \m$.
Then we can consider the task of 
transitioning from the collection of pairs 
$\big\{ (\bx_i , y_i) \big\}_{i=1}^M$ to a
continuous function 
$f : \m \to \R$ that accurately reflects the known responses
of the system to the inputs $\bx_1 , \ldots , \bx_M$ 
and is capable of predicting the response to new inputs.

Making sense of a continuous function $f : \m \to \R$ requires
a topology on $\m$. One particular way to equip $\m$ with a 
topology is to choose a topology for $\s(W)$ and 
then have $\m$ inherit this topology as a subset.
There are, of course, several different ways this could be 
done. As an example, we could 
define a norm on $\s(W)$ as follows. 
If $k \in \Z_{\geq 1}$ and 
$\bx = ( x_1 , \ldots , x_k) \in \s(W)$
then its norm is 
\begin{equation}
    \label{eq:norm_on_s(VxV)}
        \lvert \lvert \bx \rvert\rvert_{\s(W)}
        := 
        \sum_{j=1}^{k} \lvert\lvert x_j \rvert\rvert_{W}.
\end{equation}
This choice corresponds to using the $l^1(W)$ norm on 
the space of streams $\s(W)$, observing that any element
$\bx \in \s(W)$ can be viewed as an element in $l^1(W)$
whose entries are all eventually $0$.
Making an appropriate choice of norm for $\s(W)$
is a modelling problem. There is no choice that will always 
be the most sensible. Which choice of norm is most suitable 
will depend on the particular problem.

We approach this regression problem by first transforming 
each of the streams $\bx_1 , \dots , \bx_M \in \s(W)$ to
a stream of increments using the transformation informally 
introduced at the end of Subsection 
\ref{subsec:inc_stream_data}.
We first formalise this transformation. 
Consider a basis $e_1 , \ldots , e_d \in W$ for $W$.
Suppose $k \in \Z_{\geq 1}$ and 
let $\bx = (x_1 , \ldots , x_k) \in \s(W)$ be a stream 
such that, for each $j \in \{1, \ldots , k\}$, we have 
\begin{equation}
    \label{eq:stream_entries_wrt_basis}
        x_j = \sum_{l=1}^d x_{j,l}e_l
\end{equation}
for real coefficients $x_{j,1} , \ldots , x_{j,d} \in \R$.
If we let the $j^{\text{th}}$ row correspond to the 
coefficient of the basis element $e_j$, we can represent the
stream $\bx$ as 
\begin{equation}
    \label{eq:stream_as_col_vecs}
        \bx = \left\{
        \begin{pmatrix}
            x_{1,1} \\
            \vdots \\
            x_{1,d}
        \end{pmatrix} 
        ,
        \ldots 
        ,
        \begin{pmatrix}
            x_{k,1} \\
            \vdots \\
            x_{k,d}
        \end{pmatrix} 
        \right\}.
\end{equation}
Then we apply the transformation introduced at the end 
of Subsection \ref{subsec:inc_stream_data} to 
each channel in \eqref{eq:stream_as_col_vecs}.
That is, we transform the stream $\bx$ as represented 
in \eqref{eq:stream_as_col_vecs} to the new stream 
\begin{equation}
    \label{eq:incr_stream_as_col_vecs}
        \bx' := \left\{ 
        \begin{pmatrix}
            x_{1,1} \\
            0 \\
            \vdots \\
            x_{1,d} \\
            0
        \end{pmatrix} 
        ,
        \begin{pmatrix}
            0 \\
            -a_{1,1} \\
            \vdots \\
            0 \\
            -a_{1,d}
        \end{pmatrix}
        ,
        \begin{pmatrix}
            x_{2,1} - x_{1,1} \\
            a_{2,1} - a_{1,1} \\
            \vdots \\
            x_{2,d} - x_{1,d} \\
            a_{2,d} - a_{1,d}
        \end{pmatrix}
        ,
        \ldots 
        ,
        \begin{pmatrix}
            x_{k,1} - x_{k-1,1} \\
            a_{k,1} - a_{k-1,1} \\
            \vdots \\
            x_{k,d} - x_{k-1,d} \\
            a_{k,d} - a_{k-1,d}
        \end{pmatrix}
        ,
        \begin{pmatrix}
            0 \\
            -a_{k,1} \\
            \vdots \\
            0 \\
            -a_{k,d}
        \end{pmatrix}
        ,
        \begin{pmatrix}
            -x_{k,1} \\
            0 \\
            \vdots \\
            -x_{k,d} \\
            0
        \end{pmatrix} 
        \right\}.
\end{equation}
For every $i \in \{1, \ldots , k\}$ and 
$j \in \{1, \ldots , d\}$ the real number $a_{i,j} \in \{0,1\}$
records whether the recorded value $x_{i,j} \in \R$ is the 
result of a new measurement or not. 
We have $a_{i,j}=1$ if $x_{i,j}$ is the result of a new 
measurement, and $a_{i,j} = 0$ otherwise. 
If this information is unknown or not available, then 
the default value is to take $a_{i,j} = 1$.
That is, unless told otherwise we assume that each entry 
$x_{i,j}$ is the result of a new measurement.

The values of the first and last entries of the original 
stream $\bx$ are encoded in the first and last entries 
of the new stream $\bx'$ respectively.
The particular importance of the first and last values 
of a stream are emphasised by this explicit encoding.
In fact, the value of any entry of the original stream 
$\bx$ is recoverable from the new stream $\bx'$.
Moreover, if two streams $\bx_1$ and $\bx_2$ are distinct, 
then the new streams $\bx_1'$ and $\bx_2'$, resulting 
from the application of this transformation to $\bx_1$ 
and $\bx_2$ respectively, remain distinct.
No information is lost under this transformation.
Consequently this transformation of a stream to a 
stream of increments satisfies the wishlist of properties 
discussed in Subsection \ref{subsec:inc_stream_data}.

It is seen from \eqref{eq:incr_stream_as_col_vecs} 
that the stream $\bx'$ is an element in $\s(W \times \R^d)$.
Let $V := W \times \R^d$ and replace each of the streams 
$\bx_1 , \ldots , \bx_M \in \s(W)$ by the stream of
increments in $\s(V)$ resulting from applying the 
transformation determined by \eqref{eq:incr_stream_as_col_vecs}.
Making a slight abuse of notation, we continue to denote the 
resulting streams of increments in $\s(V)$ by 
$\bx_1 , \ldots , \bx_M \in \s(V)$.
Replace each stream $\bx$ in the subset $\m \subset \s(V)$ by 
the stream resulting from applying the transformation 
determined in \eqref{eq:incr_stream_as_col_vecs} to $\bx$.
Again we abuse notation and continue to denote the 
resulting subset of $\s(V)$ by $\m$. That is, we now have 
$\m \subset \s(V)$.

We now consider the following regression problem.
We have a finite collection of 
pairs $\Omega = \big\{ (\bx_i, y_i) \big\}_{i=1}^M$
where, for each $i \in \{1 , \ldots , M \}$, 
$\bx_i$ is a stream of increments in $V$ and $y_i \in \R$ 
represents the underlying 
systems response to the stream $\bx_i$.
We adopt the convention that $\Omega$ is a 
\textit{labelled dataset}.
We denote by $\Omega_{\s(V)}$ the projection of 
$\Omega$ to $\s(V)$, that 
is $\Omega_{\s(V)} := \{ \bx_i \mid i \in \{1 , \ldots , M\} \} 
\subset \s (V)$. 
We adopt the convention that $\Omega_{\s(V)}$ is a 
\textit{dataset}.
Observe that $\Omega_{\s(V)} \subset \m \subset \s(V)$.
We consider the task of transitioning from the collection 
of pairs $\Omega$ to a continuous function $f : \m \to \R$
that accurately reflects the known response of the system 
to the inputs $\bx_1 , \ldots , \bx_M$, and is capable 
of predicting the response to new inputs 
$\bx \in \s (V)$ that do \textit{not} 
correspond to an instance within $\Omega_{\s(V)}$.

Once again making sense of a continuous function 
$f : \m \to \R$ requires a choice of topology on $\m$.
This can again be achieved by equipping $\s(V)$ with a 
topology and letting $\m$ inherit the subspace topology.
There are, of course, numerous possible ways to equip $\s(V)$
with a topology. 
As an example, we could first choose a norm $N$ for $\R^d$.
Then we could define a norm
$\lvert\lvert \cdot \rvert\rvert_V$ on 
$V = W \times \R^d$ by defining 
$\lvert\lvert v \rvert\rvert_V :=
\lvert\lvert P_W(v)\rvert\rvert_W + N(P_d(v))$ 
for $v \in V$ where $P_W : V \to W$ and $P_d : V \to \R^d$ 
denote the respective projection maps.
Then we could subsequently define a norm on $\s(V)$ 
as done for $\s(W)$ in \eqref{eq:norm_on_s(VxV)}.
That is, if $k \in \Z_{\geq 1}$ and 
$\bx = ( x_1 , \ldots , x_k) \in \s(V)$
then its norm is 
\begin{equation}
    \label{eq:norm_on_s(V)}
        \lvert\lvert \bx \rvert\rvert_{\s(V)}
        := 
        \sum_{j=1}^{k} \lvert\lvert x_j \rvert\rvert_{V}.
\end{equation}
Making an appropriate choice of norm for $\s(V)$ is again 
a modelling problem. There is no choice that will always 
be the most sensible. Which choice of norm is most suitable 
will depend on the particular problem.

In this formulation, we implicitly assume that 
the dataset $\Omega_{\s(V)}$ gives a ``good approximation"
(in some sense) of the subset $\m$. That is, we
assume it is reasonable to try and learn the
systems response to inputs in $\m$ using only its 
responses to inputs in $\Omega_{\s(V)}$.  
This restriction is natural within learning tasks.

We would not expect a house value predictor learnt 
from rural house data to work well on houses in 
all locations. Nor would we expect a reaction time 
predictor learnt from professional racing drivers
to do a good job at predicting the reaction time
of members of the general public. And it is unlikely
that knowing the reaction of mice to a drug would 
allow the accurate prediction of the reaction of 
all animals to the drug.

The issue with all these examples is the presence
of elements in $\m$ baring little resemblance to 
\textit{any} element in $\Omega_{\s(V)}$. There are inner-city
houses with very little similarity to rural
houses, there are members of the general population
sharing few similar characteristics with professional
racing drivers, and there are animals whose biological
make-up is significantly different from that of mice. It is not
reasonable to expect to accurately predict a systems
response to inputs baring little resemblance to
the inputs for which the response is known.

It is easy to imagine several \textit{ad-hoc} methods
of ensuring that every element in $\m$ is somehow
similar to an element in $\Omega_{\s(V)}$.
But introducing a mathematical framework capable of
describing similarity is more challenging.
We will later discuss a rigorous notion of similarity
(cf. Section \ref{anom_detec}). 
But for now, we just assume that $\m \subset \s(V)$ is
a given subset and that our dataset $\Omega_{\s(V)}$ provides 
a "sufficiently good" approximation of $\m$, without
concerning ourselves with the precise details meant by
"sufficiently good".

\subsection{Simple Regression Approach}
\label{SRA}
We illustrate a basic approach to the regression problem 
outlined in Subsection \ref{subsec:math_framework}, and 
highlight some of the key theoretical 
properties required for this approach to be successful.
We will later present an approach 
to this regression task through the use of signatures, 
and validate that the required theoretical properties 
are satisfied (cf. Subsection 
\ref{SI} and Section \ref{good_props}).
For examples of working regression into machine learning
pipelines see \cite{ML_book} or \cite{LMR22}.

A core idea in many schemes for learning a function 
$f$ from a collection of pairs $\Omega$ is to 
mimic the use of the monomials in the basic example 
presented in Subsection \ref{subsec:Illustrative_Example}.
That it, first identify basic functions or features that 
are readily evaluated at each point in $\Omega_{\s(V)}$. 
This role is played by the monomials on the unit interval
$[0,1] \subset \R$ in Subsection 
\ref{subsec:Illustrative_Example}.
Once a collection of basic functions has been identified, 
we try to express the observed 
function as a linear combination of these basic functions. 

One approach to finding a suitable collection of feature 
functions is to identify a \textit{universal feature map} 
$\Psi$. That is, to find an embedding $\Psi : \m \to W$ 
for a Banach space $W$ for which
\begin{equation}
    \label{universal}
        \overline{\Psi (\m)^{\ast}} 
        =
        C^0 \big( \Psi (V) \big)
\end{equation} 
where the closure is taken with respect to the uniform 
$C^0$-norm on $W$.  
That is, we take the closure of $\Psi(\m)^{\ast}$ 
in $C^0(W;\R)$ where $C^0(W;\R)$ 
is equipped with the norm 
$\lvert\lvert A \rvert\rvert_{C^0(W;\R)} :=
\sup_{w \in W} \big\{ \lvert A(w) \rvert \big\}$.

A universal feature map $\Psi$ provides a 
representation of the dataset $\Omega_{\s(V)}$ as a subset 
$\Omega_W := \Psi \big( \Omega_{\s(V)} \big)$ of a 
Banach space $W$ in which continuous functions can be 
approximated by linear functions. 
That is, assuming \eqref{universal} is valid, 
if $F \in C^0 \big( \Psi (\m) \big)$ then given any 
$\ep > 0$ there exists a linear functional 
$\sigma : \Omega_W \to \R$ such that for every 
$w \in \Omega_W$ we have
$\lvert F(w) - \sigma(w) \rvert \leq \ep$.
This would allow us to use linear functions as 
our feature functions.
Before mimicking the approach of approximating a 
continuous function on the interval $[0,1]$ described in 
Subsection \ref{subsec:Illustrative_Example}, we need to 
understand how to evaluate $\phi(\bx_i)$ for elements 
$\phi \in \Psi(\m)^{\ast}$ and $i \in \ci$.

We seek a suitable universal embedding of the set 
$\m$ into some Banach space $W$. If we simultaneously 
found an embedding of $\m$ into the dual space $W^{\ast}$, 
then we would have many of the properties of what the 
machine learning literature refers to as a \textit{kernel}. 
To be more precise, a pair of embeddings
\begin{equation}
    \label{ker_embds}
        \Psi : \m \to W 
        \qquad \text{and} \qquad 
        \Phi : \m \to W^{\ast}
\end{equation}
would allow us to define a function 
$K : \m \times \m \to \R$ by setting
\begin{equation}
	\label{ker_def_gen}
		K(x,y) := \big< \Phi(x) , \Psi(y) \big>
\end{equation}
where $\big< \cdot , \cdot \big> : W^{\ast} \times W
\to \R$ denotes the natural dual-pairing map. 
That is, given $\phi \in W^{\ast}$ and $v \in W$ 
we define $\big< \phi , v \big> := \phi(v)$.

If $W$ is a Hilbert space, in which case its dual $W^{\ast}$
is isometric to itself and the natural dual-pairing map is 
simply the inner product on $W$, then the function $K$ defined 
in \eqref{ker_def_gen} is a kernel.
In this setting the kernel $K$ generates what is called a 
\textit{Reproducing Kernel Hilbert Space} (RKHS), denoted by 
$\cH_K$, that is a subset of the continuous functions 
$\m \to \R$, i.e. $\cH_K \subset C^0(\m;\R)$.
Originally developed by Aronszajn in \cite{Aro50}, numerous
recent works provide detailed exposition of this topic.
Two particular examples are the introductions provided in 
\cite{AM15} and \cite{PR16}.
We provide a brief outline of the construction of $\cH_K$, 
and direct the reader to either of \cite{AM15} or \cite{PR16}
for full details.
The RKHS $\cH_K$ can be constructed as follows.
\begin{itemize}
    \item Define $\cH_0 := \Span \Big( \big\{ K_z \mid z \in \m 
    \big\} \Big)$ where, for $z \in \m$, the function 
    $K_z : \m \to \R$ is defined by $K_z(v) := K(z,v)$ 
    for $v \in \m$.
    \item For $v,z \in \m$ define 
    $\big< K_z , K_v \big>_{\cH_0} := K(z,v)$ and 
    subsequently bilinearly extend 
    $\big< \cdot , \cdot \big>_{\cH_0}$ to the entirety of
    $\cH_0$.
    \item Prove that 
    $\Big( \cH_0 , \big< \cdot , \cdot \big>_{\cH_0} \Big)$
    is an inner product space and that if $f \in \cH_0$
    and $v \in \m$ then $f(v) = \big< f , K_v \big>_{\cH_0}$.
    \item Let $\cH_K$ be the collection of functions 
    $f : \m \to \R$ for which there exists a $\cH_0$-Cauchy 
    sequence $\big\{ f_n \big\}_{n=1}^{\infty} \subset \cH_0$
    such that $f_n \to f$ pointwise as $n \to \infty$.
    \item Show that $\big< f , g \big>_{\cH_K} := 
    \lim_{n \to \infty} \big< f_n , g_n \big>_{\cH_0}$
    for $\cH_0$-Cauchy sequences 
    $\big\{ f_n \big\}_{n=1}^{\infty} , 
    \big\{ g_n \big\}_{n=1}^{\infty} \subset \cH_0$
    converging pointwise to $f$ and $g$ respectively is a 
    well-defined inner product on $\cH_K$.
    \item Prove that $\cH_K$ is complete with respect to 
    this inner product, i.e. that 
    $\Big( \cH_K , \big< \cdot , \cdot \big>_{\cH_K} \Big)$
    is a Hilbert space.
    Moreover, it follows that if $f \in \cH_k$ and $v \in \m$ 
    then $f(v) = \big< f , K_v \big>_{\cH_k}$.
\end{itemize}
The property that for $f \in \cH_K$ and $v \in \m$ we have 
$f(v) = \big< f , K_v \big>_{\cH_K}$ is sometimes called
the \textit{reproducing property}.
It is worth mentioning that this outlined strategy 
works under weaker assumptions for the kernel 
$K : \m \times \m \to \R$. 
Indeed it is only required that the kernel be symmetric
in the sense that for every $v,u \in \m$ we have 
$K(u,v) = K(v,u)$, and positive definite in the sense 
that for every positive integer $n \in \Z_{\geq 1}$ and 
any $v_1 , \ldots , v_n \in \m$ and $a_1 , \ldots , a_n \in \R$
that $\sum_{i=1}^n \sum_{j=1}^n a_ia_jK(v_i,v_j) \geq 0$.
Both these properties are evidently satisfied when $K$ 
is given by a real inner product on a real vector space.

Returning to the setting in which we have embeddings
as outlined in \eqref{ker_embds} and the 
resulting function $K$ defined in \eqref{ker_def_gen},
we observe that we now have a concrete approach to the 
learning problem.
Recall we have a labelled dataset 
$\Omega = \big\{ (\bx_i , y_i) \mid i \in 
\{1 , \ldots , M\} \big\} \subset 
\s(V) \times \R$.
We could use the functions $K_{\bx_i} \in \cH_K$ 
for $i \in \{1 , \ldots , M\}$ as the feature functions. 
That is, first solve the system
\begin{equation}
    \label{ker_sys}
        y_i = \sum_{k=1}^M a_k K_{\bx_k} (\bx_i)
\end{equation}
for each $i \in \{1 , \ldots , M\}$, and then express the 
observed function $f$ as
\begin{equation}
    \label{ker_approx}
        f (\cdot):= \sum_{k=1}^M a_k K_{\bx_k} ( \cdot ).
\end{equation}
When $W$ is a Hilbert space the so-called 
\textit{Representer Theorem}, established in \cite{HSS01}, 
provides theoretical guarantees regarding the solvability
of the equations in \eqref{ker_sys} via gradient descent 
with respect to a suitable cost function, possibly involving 
a regularisation term, see \cite{HSS01} for full details.
However, the equations determined in \eqref{ker_sys} clearly
do not require $W$ to be a Hilbert space, and the outlined 
approach still makes sense under the weaker assumption that
$W$ is only a Banach space. 

Of course in this setting the function $K$ defined in 
\eqref{ker_def_gen} fails to be a kernel in the machine 
learning sense; indeed it need not be either symmetric 
of positive definite.
Consequently one cannot invoke the Representer Theorem 
established in \cite{HSS01} to provide theoretical 
guarantees regarding the use of cost function based 
gradient descent to solve the system of equations 
\eqref{ker_sys}.
However, it is in fact known 
that the Representer Theorem remains valid for certain types
of Banach spaces; see \cite{Sch21}. 
Hence there are settings in which just having the 
assumption of a pair of embeddings, one into a Banach space
$W$ and the other into its dual $W^{\ast}$, 
rather than an embedding into a Hilbert space remains 
sufficient to provide the theoretical guarantees of the 
Representer theorem.

A major benefit of the kernel approach outlined above is that
neither the dimension of $W$ or the explicit form 
of the embeddings $\Psi$ and $\Phi$ are involved.
The strategy only involves a \textit{finite} number 
of terms, namely the $M^2$ numbers $K(x_i,x_j)$, 
which is independent of the dimension of $W$. 
Assuming $K$ is known, computing these $M^2$ 
numbers does not require computing the embedded 
data points $\Phi(\bx_i)$ and $\Psi(\bx_i)$ themselves.
Consequently there is no restriction on the embeddings 
that may be considered, with embeddings into infinite 
dimensional Hilbert spaces $W$ being allowed both 
theoretically and in practice.
But the number of computations does grow quadratically 
with respect to the size of the dataset $\Omega_{\s(V)}$, which 
becomes problematic for large datasets $\Omega_{\s(V)}$.

Unlike the motivating example of using monomials as the 
feature functions 
(cf. Subsection \ref{subsec:Illustrative_Example}), 
there is no natural notion of 
ordering on the set $\big\{ K_{\bx_i} \mid i\in \ci \big\}$. 
For the monomials on $[0,1]$, namely the set 
$\big\{\phi_k :[0,1]\to \R ,\phi_k(x):= x^k \mid
k\in \Z_{\geq 0} \big\}$, 
then there is the grading given by the order of the monomials.
There is an intrinsic idea that the higher the order the 
more costly it is to use, and thus it is somehow preferable 
to use lower order monomials rather than ones of higher order;
i.e. using $\phi_2$ in place of $\phi_{10}$ would be less
costly.
A priori there is no such intrinsic grading on the set 
of functions 
$\big\{ K_{\bx_i} \mid i \in \{1 , \ldots , M\} \big\}$.
Given distinct $i,j \in \{1 , \ldots , M\}$, 
it is not clear whether 
there is any benefit between using $K_{\bx_i}$ or
$K_{\bx_j}$ to approximate a function.

\subsection{Streams of Increments as Paths}
\label{dataset_to_path}
A stream of increments determines a system 
whose state evolves over time. It may be viewed as a 
map from an ordered state space to the Banach space $V$,
and concatenating the increments provides an intrinsic 
way of embedding the stream within a continuous path 
$\gamma : I \to V$ for some compact interval $I \subset \R$.
Having obtained our streams of increments via the 
transformation detailed in Subsections 
\ref{subsec:inc_stream_data} and \ref{subsec:math_framework}, 
the continuous path resulting from the concatenation of 
the entries will be a closed loop based at the origin 
$0 \in V$.
This interpolation of a stream of increments is, 
in some sense, less \textit{ad-hoc} than the variety of 
interpolation schemes
that can be applied to a stream of values. 
Examples of several explicit choices of such interpolation
schemes, including linear interpolation, rectilinear 
interpolation, and cubic spline interpolation, can be 
found in \cite{CK16}.

The concatenation of the increments contained within a 
stream of increments determines an \textit{unparameterised} 
path in $V$; sometimes also referred to as a \textit{shape} 
or a \textit{contour}.
Each entry tells us how the system evolves; but the time 
over which this evolution happens is determined via a 
choice of how to view contour.
This choice should not influence the affect of the system. 
Our choice of how to view the contour should not affect 
the systems response to the contour.
We will summarise the effects of a stream of increments 
by using the signature to succinctly capture the effects 
of the contour resulting from the concatenation of the 
entries in the stream.

It is worth remarking that streams of increments can 
be \textit{augmented} in the same way that streams of values
often are. Typically, supposing that 
$\bx = ( x_{1} , \ldots , x_{k}) \in V^k$ 
for some $k \in \Z_{\geq 1}$, one considers a fixed map 
$\Theta : V \to W$ for some Banach space 
$W$, and subsequently uses $\Theta$ to transform $\bx$ 
to the augmented stream 
$\Theta ( \bx ) := 
\big( \Theta(x_{1}) , \ldots , \Theta (x_{k}) \big)  \in W^k$. 
The underlying idea is that the augmentation makes it 
easier to summarise the resulting stream, possibly 
by explicitly introducing a characteristic that is 
desirable for the particular model. Evidently the choice
of augmentation to use is a modelling problem; it will 
depend upon the specific task one is aiming to solve.

Once the dataset $\Omega_{\s(V)}$ is augmented by $\Theta$
to a dataset $\Omega_{\s(W)} := \big\{ \Theta(\bx) \mid 
\bx \in \Omega_{\s(V)} \big\} \subset \s(W)$, 
one can consider the regression problem of transitioning 
from the pairs $\big\{ (\Theta(\bx_i) , y_i) \mid i \in 
\{1 , \ldots , M\} \big\}$ to a continuous function 
$h : \Theta(\m) \to \R$.
Then the original desired continuous function $f : \m \to \R$
would be given by $f := h \circ \Theta$.

We end this section by explicitly introducing 
some of the more commonly used augmentations as maps 
$\s(V) \to \s(W)$ for a choice of Banach space $W$ that 
is specific to each augmentation.

The \textit{Lead-Lag} augmentation is the 
map $T_{\LeadLag} : \s(V) \to \s(V \times V)$ defined by
\begin{equation}
	\label{Lead-Lag}
		T_{\LeadLag} ( v_1 , \ldots , v_k ) := 
		\big( (v_1,v_1), (v_2, v_1), (v_2 , v_2), 
		\ldots ,  (v_j , v_j), (v_{j+1} , v_j), 
		\ldots , (v_k , v_k) \big).
\end{equation}
It takes a stream with $k$ entries to a stream with $2k$
entries, whilst splitting the original stream 
into two copies, labelled the future and the past. 
There is a delay between when the future is updated 
and when the past is subsequently updated.
The number of steps between the future being updated and 
the past being updated can be varied. 
Further, more than one past stream can be recorded, 
with each past stream having a different number of 
delay steps between when the future stream is updated 
and when it is itself updated. 

For the purpose of an explicit illustration of the 
Lead-Lag augmentation, take $V := \R$ and consider the stream 
$\Omega_L := \{ 1 , 6 , 3 , 9, 5\}$.
The basic Lead-Lag augmentation defined in 
\eqref{Lead-Lag} would map this stream to the stream
\[ \big\{ (1,1), (6,1), (6,6), (3,6), (3,3), (9,3), 
(9,9),(5,9),(5,5) \big\} \in \s \big( \R^2 \big).\]
If the delay is increased to $2$ steps, then we get 
the stream
\[ \big\{ (1,1),(6,1),(3,1), 
(3,6),(9,6),(5,6),(5,3),(0,3),(0,3),(0,9),
(0,9),(0,9),(0,5) \big\} \in \s \big( \R^2 \big).\]
We could record both the one-step delayed past and the 
two-step delayed path in a single augmentation. 
This would map $\Omega_L$ to the stream
\[ \left\{ 
\begin{pmatrix}1\\1\\1\end{pmatrix} ,
\begin{pmatrix}6\\1\\1\end{pmatrix} ,
\begin{pmatrix}6\\6\\1\end{pmatrix} ,
\begin{pmatrix}6\\6\\6\end{pmatrix} ,
\begin{pmatrix}3\\6\\6\end{pmatrix} ,
\begin{pmatrix}3\\3\\6\end{pmatrix} ,
\begin{pmatrix}3\\3\\3\end{pmatrix} ,
\begin{pmatrix}9\\3\\3\end{pmatrix} ,
\begin{pmatrix}9\\9\\3\end{pmatrix} ,
\begin{pmatrix}9\\9\\9\end{pmatrix} ,
\begin{pmatrix}5\\9\\9\end{pmatrix} ,
\begin{pmatrix}5\\5\\9\end{pmatrix} ,
\begin{pmatrix}5\\5\\5\end{pmatrix} 
\right\} \in \s \big( \R^3 \big).\]
The final variant of this transformation we mention 
involves not pausing the future stream whilst the past 
streams are updated. 
If $d$ past streams are being recorded, then the 
\textit{no future pause} augmentation maps a stream 
$(v_1, \ldots ,v_k) \in \s (V)$ to the stream
\begin{equation}
    \label{no_pause}
        \left\{ 
            \begin{pmatrix}v_1\\0\\0 \\ \vdots 
            \\ 0 \\ 0\end{pmatrix}, 
            \begin{pmatrix}v_2\\v_1\\0\\ \vdots 
            \\ 0 \\ 0 \end{pmatrix}, 
            \dots 
            \begin{pmatrix}v_{d+1}\\v_d\\v_{d-1}\\ \vdots 
            \\ v_2 \\v_1 \end{pmatrix}, 
            \begin{pmatrix}v_{d+2}\\v_{d+1}\\{v_d}\\ \vdots 
            \\ v_3 \\ v_2 \end{pmatrix},
            \dots
            \begin{pmatrix}v_k\\v_{k-1}\\v_{k-2}\\ \vdots 
            \\ v_{k-d+1} \\ v_{k-d} \end{pmatrix},
            \begin{pmatrix}0\\v_k\\v_{k-1}\\ \vdots 
            \\ v_{k-d+2} \\ v_{k-d+1} \end{pmatrix}, 
            \dots 
            \begin{pmatrix}0\\0\\0\\ \vdots 
            \\ v_k \\ v_{k-1} \end{pmatrix}, 
            \begin{pmatrix}0\\0\\0\\ \vdots \\0 
            \\ v_k \end{pmatrix}
        \right\} 
\end{equation}
in $\s \big( V^{d+1} \big)$. 
Returning to the dataset $\Omega_L$, applying the 
no future pause augmentation defined in 
\eqref{no_pause} with $d=2$ results in the new stream
\[ \left\{ 
	\begin{pmatrix}1\\0\\0\end{pmatrix} ,
	\begin{pmatrix}6\\1\\0\end{pmatrix} ,
	\begin{pmatrix}3\\6\\1\end{pmatrix} ,
	\begin{pmatrix}9\\3\\6\end{pmatrix} ,
	\begin{pmatrix}5\\9\\3\end{pmatrix} ,
	\begin{pmatrix}0\\5\\9\end{pmatrix} ,
	\begin{pmatrix}0\\0\\5\end{pmatrix} 
\right\} \in \s \big( \R^3 \big).\]
Lead-Lag type transformations are frequently 
used in financial markets modelling since they can 
encode the modelling assumption that no strategy 
should be able to use the present time stock value. 
Strategies are only allowed to use the past streams 
of the augmented dataset. 
Since the future stream is always updated first, 
with there being a definite delay before any of the 
past streams are subsequently updated, the strategies 
are prevented from making use of the present time stock 
values.

Another common augmentation is the \textit{Time} 
augmentation $T_{\Time} : \s(V) \to \s(V \times \R)$ 
defined by
\begin{equation}
    \label{Time}
        T_{\Time} \Big((v_1 , \ldots , v_k) \Big) :=
        \Big( (v_1 , t_0), \ldots , (v_j , t_j), 
        \ldots , (v_k , t_k) \Big)
\end{equation}
for a strictly increasing sequence of times 
$0 \leq t_1 < t_2 <\ldots < t_k$, ensuring the 
resulting path has a strictly monotonic component.	
A variant of this is the \textit{Time-difference} augmentation 
$T_{\Timediff} : \s(V) \to \s(V \times \R)$ defined by
\begin{equation}
	\label{Time-diff}
		T_{\Timediff} \Big( (v_1 , \ldots , v_k) \Big) :=
		\Big( (v_1 , t_0) , \ldots , 
		(v_j , t_j - t_{j-1}) , 
		\ldots ,(v_k , t_k - t_{k-1}) \Big)
\end{equation}
for a strictly increasing sequence of times 
$0 \leq t_1 < t_2 < \ldots < t_k$. Both the Time and 
Time-diff augmentations take streams consisting of $k$ 
entries to streams once again consisting of $k$ entries.

To illustrate these augmentations, consider the stream 
$\Omega_T := \{ 1 , 5 , 2, 9, 7, 6 \} \in \s (\R)$ and the 
strictly increasing sequence $0,  1 , 3, 6, 8 , 12 $. Then
\begin{equation}
	\label{time_aug}
		T_{\Time} \big( \Omega_T \big) = 
		\big\{ (1,0) , (5,1) , (2,3) , (9,6), (7,8), 
		(6,12) \big\} \in \s \big( \R^2 \big)
\end{equation}
and
\begin{equation}
	\label{time-diff_aug}
		T_{\Timediff} \big( \Omega_T \big) = 
		\big\{ (1,0), (5,1), (2,2), (9,3), (7,2), 
		(6,4) \big\} \in \s \big( \R^2 \big).
\end{equation}
The final augmentation we mention is the 
\textit{Invisibility Reset} augmentation 
$T_{\inv} : \s(V) \to \s(V \times \R)$ defined by
\begin{equation}
	\label{inv_reset}
		T_{\inv} ( v_1 , \ldots , v_k ) :=
		\Big( (v_1,1) , \ldots , (v_j,1) , 
		\ldots , (v_k,1) ,(v_k,0) ,(0,0) \Big)
\end{equation}
and taking streams with $k$ entries to streams with
$k+2$ entries. 
We illustrate this augmentation by applying it to both the
one-dimensional stream 
$\Omega_{ir1} := \{ 1,3,4,8,9 \}$ in $\s(\R)$ and the 
two-dimensional stream
$\Omega_{ir2} := \{(1,2),(3,4),(4,6),(5,9),(7,10)\}$ 
in $\s \big( \R^2 \big)$. This yields
\begin{equation}
	\label{T-inv-ir1}
		T_{\inv} \big( \Omega_{ir1} \big) =
		\big\{ (1,1),(3,1),(4,1),(8,1),(9,1),(9,0),
		(0,0) \big\} \in \s \big( \R^2 \big)
\end{equation}
and
\begin{equation}
	\label{T-inv-ir2}
		T_{\inv} \big( \Omega_{ir2} \big) = \left\{
		\begin{pmatrix}1\\2\\1\end{pmatrix} ,
		\begin{pmatrix}3\\4\\1\end{pmatrix} ,
		\begin{pmatrix}4\\6\\1\end{pmatrix} ,
		\begin{pmatrix}5\\9\\1\end{pmatrix} ,
		\begin{pmatrix}7\\10\\1\end{pmatrix} ,
		\begin{pmatrix}7\\10\\0\end{pmatrix} ,
		\begin{pmatrix}0\\0\\0\end{pmatrix} 
		\right\} \in \s \big( \R^3 \big).
\end{equation}
This augmentation is utilised in the transformation for 
converting a stream of values to a stream of increments 
that is detailed in Subsection \ref{subsec:math_framework}.
A particular consequence of this is that the signature of 
the resulting stream of increments will capture information 
related to the norm/size of the entries in the original 
stream of values (cf. Section \ref{good_props}).

\subsection{Signature Involvement}
\label{SI} 
The finite dataset $\Omega_{\s(V)}$ is a subset of 
the set of streams on $V$ denoted by $\s(V)$.
Thus given a stream $\bx \in \s(V)$, we seek to 
summarise $\bx$ (throw away irrelevant information) so 
as to capture its effect. 
Recalling our assumption that the streams we consider are
formed of incremental data, we observe that each stream 
can be associated with the contour/shape determined by 
the concatenation of the increments. 
Given a stream of increments $\bx \in \s(V)$ we denote the 
contour arising as the concatenation of the increments 
as $\Gamma_{\bx}$.
The approach followed in this article is to summarise 
a stream of increments $\bx \in \s(V)$ by summarising 
the contour $\Gamma_{\bx}$.

It is instructive to realise that the classical mathematical
description of paths is ill-suited to providing such a summary.
Assume that we have a parameterisation $\gamma : I \to V$
of the contour $\Gamma_{\bx}$ where $I \subset \R$ is a 
compact interval.
Then the classical description of the path $\gamma$ fails to 
adequately compress the information it contains.
The classical description requires precisely recording the 
position of the path at every instance. Even the 
probabilistic Kolmogorov approach (in which one considers 
fixed times $t_i$, open sets $O_i$, and considers the 
probability that, for all $i$, at time $t_i$ 
the path is within the set $O_i$) still places
emphasis on the location at a specified time. 

The limitations in using this approach to summarise paths 
become clear from considering some examples.
We would not try to summarise a movie by recording a 
second-by-second account, we would not describe a 
football match by providing the exact position of the
ball at each second and we would not describe a drawing 
by providing a second-by-second account of the location 
of the pencil tip.
The problem is that we are not throwing away any irrelevant 
information, making the extraction of the useful information 
very challenging. 

Recording only the major events of a path rather than all 
the events may seem like a solution that could be achieved 
via sampling; however this still run into problems. 
The issue is that the order in which the major events 
occur is frequently at least as important as the major 
events themselves.
For example, it could be critical from a diagnostic perspective
to know whether a patients heart rate or their breathing rate 
increased first.
The gap between the two events may be very 
small, meaning that a very high frequency sampling rate of 
both the pulse and the breathing rate is required to 
capture their order. But without a priori knowledge of 
when the first drop occurs we have to sample the entirety 
of both streams, resulting in a large amount of 
irrelevant information being recorded.
This problem is amplified as the number of events 
$n \geq 2$ whose order we are interested in increases.
We want a summary of the main events and the order in 
which they happen that is detailed enough to distinguish 
between different paths whilst still discarding irrelevant 
information.

We want to summarise the contour $\Gamma_{\bx}$ associated to 
the stream of increments $\bx$.
The contour $\Gamma_{\bx}$ is independent of the choice of 
parameterisation we make to view it.
This reflects the fact that the response of a system to a 
given stream $\bx$ is not affected by the speed at which 
the provided sequential increments happen. 
The features of the stream 
induce the same response regardless of how quickly 
we choose to experience them.
For example, consider writing the character `3'. The 
effect of this stream is a figure `3' drawn on our piece 
of paper, and obviously this effect remains 
the same irrespective of how quickly we draw it. Two 3's 
drawn with different speeds are still both a `3'; the 
change of speed with which we view the stream given by
`writing a figure 3' does \textit{not} change its effect 
in producing a `3' on our page. 

Consequently, if we want to summarise $\Gamma_{\bx}$ by 
summarising a parameterisation $\gamma : I \to V$ of 
$\Gamma_{\bx}$ defined on a compact interval $I \subset \R$,
we must summarise $\gamma$ in a manner that is invariant 
under reparameterisation.
That is, we must quotient out the symmetry induced by the 
group of reparameterisations so that the resulting summary 
of the contour $\Gamma_{\bx}$ depends only on the contour 
$\Gamma_{\bx}$ and \textit{not} on any particular choices 
made for the purposes of viewing it.
The information extracted from the contour $\Gamma_{\bx}$
should not depend on how we choose to access $\Gamma_{\bx}$.

The freedom of reparameterisation makes embedding 
a stream into a continuous path a reasonable idea. 
Each increment in a stream $\bx$ represents a change in
the state of the system.
Initially these changes appear to happen instantaneously, 
seemingly giving rise to a discontinuous path.
Hence it appears that the contour $\Gamma_{\bx}$ arising
from the concatenation of these changes in a continuous 
manner is not a good representation of the stream $\bx$.

However, under the assumption that reparameterisation 
does not affect the underlying system, 
we can imagine slowing the speed down and introducing 
new virtual time over which 
the next change to the system happens. By sufficiently 
slowing the speed, we can imagine that the seemingly 
instantaneous, with respect to the original time, 
affect of the next instance becomes a continuous 
change with respect to the new virtual time.

Recalling our machine learning motivations, after
associating each stream $\bx_i \in \Omega_{\s(V)}$ 
with its corresponding contour $\Gamma_{\bx_i}$, our
problem has been transformed to seeking
to learn a function that is an effect of these
unparameterised path. 
A consequence of \textit{rough path theory} is that the
\textit{signature} provides a summary of the paths affect
on systems that satisfies our list of requirements. 
The signature of a path determines this path in an
essentially unique way and
is invariant under reparameterisation of the path. 
This invariance allows the signature to readily record 
the order in which events occur
without recording precisely when they occur. 

The signature of a path even provides a natural feature set
of linear functionals capturing the aspects of the data
necessary to predict the effects of the path on systems.
The signature of a path $\gamma$ is the response of the
exponential nonlinear system to the path. In one 
dimension the response of the exponential linear system
captures the monomials, in the sense that the power
series of $\exp : \R \to \R$ is given by
\begin{equation}
    \label{exp_power}
		\exp(x) = \sum_{k=0}^{\infty} \frac{x^k}{k!}.
\end{equation}
The projection onto the $m^{\text{th}}$ degree term 
gives the $m^{\text{th}}$ degree monomial.
Thus, at least heuristically, we see why we expect the
signature to be a suitable feature map. 
In Section \ref{good_props}, we provide some of the details 
justifying these assertions. But first we must introduce
the space upon which the signature will live.

\subsection{Tensor Algebra and Signature}
\label{Ten_Alg}
Let $V$ be a $d$-dimensional Banach space.
The spaces of \textit{formal polynomials} and 
\textit{formal power series} over $V$ are given by
\begin{equation}
	\label{ten_alg}
        T(V) := \bigoplus_{k=0}^{\infty} V^{\otimes k}
        \qquad \text{and} \qquad 
		T((V)) := \prod_{k=0}^{\infty} V^{\otimes k}
\end{equation}
respectively where $V^{\otimes 0} := \R$. 
Both addition, scalar multiplication, and the
tensor product $\otimes$ 
extend to $T((V))$ as follows.
Suppose that $A = \prod_{k=0}^{\infty} a_k \in T((V))$, 
$ B = \prod_{k=0}^{\infty} b_k \in T((V))$, and 
$\lambda \in \R$. Then we define the operations 
\begin{equation}
    \label{add_and_mult_ops_def}
        A + B := \prod_{k=0}^{\infty} a_k + b_k 
        \qquad 
        \text{and}
        \qquad 
        A \otimes B := \prod_{n=0}^{\infty}
		\sum_{k=0}^{n} a_{k} \otimes b_{n-k}
\end{equation}
and recall the natural action of $\R$ on $T((V))$ given by
\begin{equation}
    \label{R_action_on_T((V))}
        \lambda A := \prod_{k=0}^{\infty} \lambda a_k.
\end{equation}
The operations defined in \eqref{add_and_mult_ops_def} 
and the action of $\R$ specified in \eqref{R_action_on_T((V))}
also make sense for elements in $T(V)$.

When equipped with the operations in
\eqref{add_and_mult_ops_def} 
and the action of $\R$ defined in \eqref{R_action_on_T((V))}
$T((V))$ becomes a real, non-commutative unital algebra, with 
unit $\textbf{1} = (1, 0 , 0 , \ldots)$, called the 
\textit{Tensor algebra} of $V$. It is within the tensor algebra
that the signature of a path $x : [a,b] \to V$ lives. 
Before detailing this, we first consider equipping $T((V))$ 
with a norm.

For convenience, for each $n \in \Z_{\geq 0}$, let 
$\pi_n : T((V)) \to V^{\otimes n}$ denote the projection map.
Moreover, for each $n \in \Z_{\geq 0}$, we may consider 
the \textit{truncated tensor algebra}
\begin{equation}
	\label{trun_ten_alg}
		T^{(n)}(V):= \prod_{k=0}^{n} V^{\otimes k}
\end{equation}
and denote the projection 
$T((V)) \to T^{(n)}(V)$ by $\Pi_n$.
One particular way to equip the tensor algebra $T((V))$
with a norm would be to choose \textit{admissible tensor norms},
in the sense originating in \cite{Sha50},
for all the tensor powers of $V$. 
For clarity, we give a precise definition below.

\begin{definition}[Admissible Tensor Norms]
\label{admissible_tensor_norm}
Let $V$ be a Banach space. The tensor powers of $V$ are
equipped with admissible norms if for each integer
$n \in \Z_{\geq 1}$ we have chosen a norm on $V^{\otimes n}$
such that the following conditions are satisfied.

\begin{enumerate}[label=(\Alph*)]
    \item\label{ten_norm_A} 
    For each $n \in \Z_{\geq 1}$ the norm 
    $\lvert\lvert \cdot \rvert\rvert_{V^{\otimes n}}$
    on $V^{\otimes n}$ is 
    invariant under the action of the symmetric group 
    $S_n$ on $V^{\otimes n}$. To elaborate, given 
    $\rho \in S_n$ and $a_1 \otimes \ldots \otimes a_n 
    \in V^{\otimes n}$, then 
    $\rho ( a_1 \otimes \ldots \otimes a_n )
    := a_{\rho(1)} \otimes \ldots \otimes a_{\rho(n)}$,
    and the action is extended to the entirety of
    $V^{\otimes n}$ by linearity.
    Then it is required that for 
    every $v \in V^{\otimes n}$ and every $\rho \in S_n$
    we have $\lvert\lvert \rho (v) \rvert\rvert_{V^{\otimes n}}
    = \lvert\lvert v \rvert\rvert_{V^{\otimes n}}$.
    \item\label{ten_norm_B} 
    For any $n,m \in \Z_{\geq 1}$ and any
    $v \in V^{\otimes n}$ and $w \in V^{\otimes m}$
    we have
    $\lvert\lvert v \otimes w \rvert\rvert_{V^{\otimes (n+m)}} 
    \leq 
    \lvert\lvert v \rvert\rvert_{V^{\otimes n}}
    \lvert\lvert w \rvert\rvert_{V^{\otimes m}}$.
    \item\label{ten_norm_C} 
    For any integers $n,m \in \Z_{\geq 1}$ and 
    for any dual elements
    $\phi \in \big(V^{\otimes n}\big)^{\ast}$
    and $\sigma \in \big(V^{\otimes m}\big)^{\ast}$
    we have
    $\lvert\lvert \phi \otimes \sigma
    \rvert\rvert_{\big(V^{\otimes (n+m)}\big)^{\ast}}
    \leq 
    \lvert\lvert \phi 
    \rvert\rvert_{\big(V^{\otimes n}\big)^{\ast}} 
    \lvert\lvert \sigma 
    \rvert\rvert_{\big(V^{\otimes m}\big)^{\ast}}$.
    Here, given any $k \in \Z_{\geq 1}$,
    the norm 
    $\lvert\lvert \cdot
    \rvert\rvert_{\left(V^{\otimes k}\right)^{\ast}}$
    denotes the dual-norm induced by
    $\lvert\lvert \cdot \rvert\rvert_{V^{\otimes k}}$.
\end{enumerate}
\end{definition}
\vskip 4pt
\noindent
It turns out (see \cite{Rya02}) that having 
\textit{both} the inequalities stated in 
Definition \ref{admissible_tensor_norm} \ref{ten_norm_B} 
and \ref{ten_norm_C} respectively ensures that
we in fact have equality in both estimates.
Hence if the tensor powers of $V$ are equipped
with admissible norms, we have equality 
in both the inequalities stated in Definition 
\ref{admissible_tensor_norm} \ref{ten_norm_B} and 
\ref{ten_norm_C} respectively.

Two common choices are the \textit{projective} and 
\textit{injective} tensor norms.
The projective tensor norm
is defined, for $n \geq 2$, on $V^{\otimes n}$ by
setting, for $v \in V^{\otimes n}$,
\begin{equation}
    \label{proj_ten_norm}  
        \lvert\lvert v \rvert\rvert_{\proj,n} 
        :=
        \inf \Bigg\{ 
        \sum_{i=1}^{\infty} \prod_{j=1}^n
        \lvert\lvert a_{j_i} \rvert\rvert_V 
        ~\Bigg\vert~ 
        v = 
        \sum_{i=1}^{\infty}
        a_{1_i} \otimes \ldots \otimes a_{n_i} 
        \text{ and }
        \sum_{i=1}^{\infty} \prod_{j=1}^n
        \lvert\lvert a_{j_i} \rvert\rvert_V 
        < \infty
        \Bigg\}.
\end{equation} 
The injective tensor norm
is defined, for $n \geq 2$, on $V^{\otimes n}$ by
setting, for $v \in V^{\otimes n}$,
\begin{equation}
    \label{inj_ten_norm}
        \lvert\lvert v \rvert\rvert_{\inj,n} := 
        \sup \Big\{ \lvert
        \vph_1 \otimes \ldots \otimes \vph_n (v)
        \rvert 
        \mid
        \vph_1 , \ldots , \vph_n \in V^{\ast} 
        \text{ and }
        \lvert\lvert \vph_1 \rvert\rvert_{V^{\ast}} =
        \ldots = 
        \lvert\lvert \vph_n \rvert\rvert_{V^{\ast}} = 1
        \Big\}.
\end{equation}
As observed in \cite{Sha50}. 
the injective and projective tensor norms are the 
prototype examples of admissible tensor norms in the following
sense.
Given any arbitrary choice of admissible tensor norms on 
the tensor powers of $V$ in the sense of 
Definition \ref{admissible_tensor_norm}, then for every 
$n \in \Z_{\geq 2}$, if
$\lvert\lvert \cdot \rvert\rvert_{V^{\otimes n}}$ denotes
the norm chosen for $V^{\otimes n}$, 
we have for every $v \in V^{\otimes n}$ that
(cf. Proposition 2.1 in \cite{Rya02})
\begin{equation}
    \label{inj<norm<proj}
        \lvert\lvert v \rvert\rvert_{\inj,n}
        \leq
        \lvert\lvert v \rvert\rvert_{V^{\otimes n}} 
        \leq
        \lvert\lvert v \rvert\rvert_{\proj , n}.
\end{equation}
Any choice of admissible tensor norms for the tensor powers
of $V$ could be used to define a norm on the tensor algebra
$T((V))$. For example, we could define
\begin{equation}
    \label{ten_alg_norm_example}
        \lvert\lvert v \rvert\rvert_{T((V))} 
        :=
        \sum_{n=0}^{\infty}
        \lvert\lvert \pi_n(v) \rvert\rvert_{V^{\otimes n}}
\end{equation}
for $v \in T((V))$ with the choice of the norm on 
$V^{\otimes 0} = \R$ as the usual absolute value.

For separable real Banach spaces $V$ it is known that one 
may define an inner product on $V$ so that the Hilbert space
resulting from the completion of $V$ with respect to this 
inner product is closely related to the original Banach 
space structure on $V$; see \cite{Kue70}.
In the case that $V$ is finite dimensional, as we are 
assuming, this can be achieved in the following way.
We begin by letting $\cb = \{ v_1 , \ldots , v_d \}$ 
be a basis for $V$, and subsequently denote the induced 
dual basis of $V^{\ast}$ by $\cb^{\ast} = 
\{ v_1^{\ast} , \ldots , v_d^{\ast} \}$. 
For each $n \in \Z_{\geq 2}$ the basis $\cb$ 
determines the basis
\begin{equation}
	\label{n_basis} 
		\cb^{\otimes n} := 
		\Big\{ v_{\bKK} = v_{k_1} \otimes 
		\ldots \otimes v_{k_n} \mid 
		\bKK = (k_1, \ldots , k_n) \in 
		\{ 1 , \ldots , d \}^n \Big\}		
\end{equation}
for $V^{\otimes n}$ and the corresponding dual basis
\begin{equation}
	\label{n_dual_basis}
		\big(\cb^{\ast}\big)^{\otimes n} := 
		\Big\{ v^{\ast}_{\bKK} = v^{\ast}_{k_1} \otimes 
		\ldots \otimes 
		v^{\ast}_{k_n} \mid 
		\bKK = (k_1, \ldots , k_n) \in 
		\{ 1 , \ldots , d \}^n \Big\}
\end{equation}
for $\big( V^{\ast} \big)^{\otimes n}$. 

We first use the basis $\cb$ to define an inner product on $V$. 
To do so we set
\begin{equation}
	\label{in_prod}
		\big< v_i , v_j \big>_V := \de_{ij} 
		= 
        \bigg\{ 
        \begin{array}{ccc}
            1 & \text{if} & i = j \\
            0 & \text{if} & i \neq j
        \end{array}
\end{equation}
and extending bilinearly to the whole of $V$.
Given any $n \in \Z_{\geq 2}$ we use the basis 
$\cb^{\otimes n}$ to define an inner product on
$V^{\otimes n}$. We set 
\begin{equation}
	\label{in_prod_extd}
		\big< v_{i_1} \otimes \ldots \otimes v_{i_n} ,
		v_{j_1} \otimes \ldots \otimes v_{j_n} 
		\big>_{V^{\otimes n}}
		:=
		\prod_{k=1}^{n} 
		\big< v_{i_k} , v_{j_k} \big>_V
		=
		\de_{i_1 j_1} \ldots \de_{i_n j_n}
\end{equation}
and again extending bilinearly to the whole of 
$V^{\otimes n}$. 
Finally, we define an inner product on the tensor 
algebra $T((V))$ by setting
\begin{equation}
	\label{ten_alg_inner_prod}
		\big< A , B \big>_{T((V))} 
		= 
		\sum_{n=0}^{\infty}
		\big< \pi_n (A) , \pi_n(B) \big>_{
        V^{\otimes n}
        }
\end{equation}
for $A,B \in T((V))$ where we choose the inner product 
on $V^{\otimes 0} = \R$ to be given by usual multiplication.
Let $\cH(V) := \overline{T((V))}$ denote the Hilbert space
given by the completion of the tensor algebra $T((V))$
with respect to this inner product.
Throughout this article we assume that the tensor algebra
$T((V))$ has been equipped with an inner product in this 
manner. 

The choices of inner products on the tensor powers of $V$
interact with the induced dual bases in the expected way.
To be more precise, first observe that  
for every $n \in \Z_{\geq 1}$ we have that
$\big(V^{\ast}\big)^{\otimes n} 
= \big( V^{\otimes n} \big)^{\ast}$. 
Consequently,
given $n \in \Z_{\geq 1}$, if 
$\bKK \in \{1, \ldots , d\}^n$ then the element 
$v^{\ast}_{\bKK} \in \cb^{\ast} \subset 
\big( V^{\otimes n} \big)^{\ast}$
is given by 
$v^{\ast}_{\bKK} (\cdot) = 
\big< \cdot , v_{\bKK} \big>_{V^{\otimes n}}$.

Momentarily ignoring regularity issues, 
the \textit{signature} of a continuous path $x : [a,b] \to V$
is the solution to the universal differential equation
\begin{equation}
    \label{sig_diff_eqn}
        d S_{a,t} (x)  = S_{a,t} (x) \otimes dx_t
        \qquad \text{with} \qquad
        S_{a,a}(x) = \textbf{1} = 
        (1, 0 , 0 , \dots ) \in T((V)),
\end{equation}
formalising that the signature $S(x)$ is the response 
of the exponential nonlinear system to the input $x$.
The subsequent Subsections \ref{sig_p<2_sec} and 
\ref{rough_paths_sec} are dedicated to examining the 
regularity issues related to the differential equation 
defined in \eqref{sig_diff_eqn}.

We end this subsection by defining certain subsets of 
the tensor algebra $T((V))$ and particular operations on 
elements within the tensor algebra $T((V))$ that will be 
useful in the later subsections.
First, we introduce the notation that $\tilde{T}((V))$ 
will denote the collection of elements $\ba \in T((V))$ 
for which $\pi_0(\ba) = 1 \in \R$, 
and $T_{>0}((V))$ will denote the collection of elements 
$\ba \in T((V))$ for which $\pi_0(\ba) > 0$.

By using the power series of the real-valued functions
$x \mapsto e^x$ and $x \mapsto \log(1+x)$, we define the
\textit{exponential} $\exp$ and \textit{logarithm} $\log$ of 
elements within the tensor algebra $T((V))$.

\begin{definition}[Exponential and Logarithm; Definition 2.20 
in \cite{CLL04}]
\label{def:ten_alg_exp_log} 
The \textit{exponential} map on the tensor algebra $T((V))$ is 
the map $\exp : T((V)) \to \tilde{T}((V))$ 
is defined for $\ba \in T((V))$ by 
\begin{equation}
    \label{eq:exp_def}
        \exp(\ba) := \sum_{n=0}^{\infty}
        \frac{\ba^{\otimes n}}{n!}.
\end{equation} 
The \textit{logarithm} map $\log : T_{> 0}((V)) \to T((V))$ is 
defined for $\ba \in T_{>0}((V))$ by
\begin{equation}
    \label{eq:log_def}
        \log(\ba) := \log(\pi_0(\ba)) + 
        \sum_{n=1}^{\infty} \frac{(-1)^n}{n}
        \bigg( \textbf{1} - \frac{\ba}{\pi_0(\ba)}
        \bigg)^{\otimes n}
\end{equation} 
where we recall that $\pi_0$ is the projection map 
$T((V)) \to \R =: V^{\otimes 0}$.
\end{definition}
\vskip 4pt
\noindent
The series defining the map $\exp$ in \ref{eq:exp_def} 
is convergent (see Lemma 2.19 in \cite{CLL04}).
Further, the series defining the map $\log$ in \ref{eq:log_def}
is purely algebraic involving only finitely many terms 
at each level. 
Consequently, it does not depend on the norms on the tensor 
powers of $V$.
Moreover, if we let $B_1$ denote the collection of elements 
$\ba \in T((V))$ with $\pi_0(\ba) = 0$, then 
$\exp : B_1 \to \tilde{T}((V))$ and 
$\log : \tilde{T}((V)) \to B_1$
are both one-to-one and the inverse of one and other; 
see Lemma 2.21 in \cite{CLL04}.

The tensor algebra $T((V))$ carries a Lie bracket 
$[\cdot,\cdot] : T((V)) \times T((V)) \to T((V))$ 
defined for $\ba , \bolde \in T((V))$ by 
\begin{equation}
    \label{eq:lie_bracket_def}
        [\ba,\bolde] := \ba \otimes \bolde - \bolde \otimes \ba.
\end{equation} 
Given linear subspaces $F_1 , F_2 \subset T((V))$ we
define 
\begin{equation}
    \label{eq:lie_bracket_lin_subsapces}
        [F_1,F_2] := \Span \Big( 
        \big\{ [\ba , \bolde] \mid \ba \in F_1 
        \text{ and } \bolde \in F_2 \big\}
        \Big).
\end{equation} 
Then we may define a sequence 
$\left\{ L_n \right\}_{n=0}^{\infty}$ of linear subspaces
of $T((V))$ recursively by 
\begin{equation}
    \label{eq:Lie_formal_series_level_n_def}
        L_0 := \{ 0 \} \subset \R =: V^{\otimes 0}, 
        \quad
        L_1 := V 
        \text{ and, for every } n \in \Z_{\geq 1}
        \quad
        L_{n+1} := [V , L_n] \subset V^{\otimes (n+1)}.
\end{equation} 
For each $n \in \Z_{\geq 0}$ the subspace 
$L_n \subset V^{\otimes n}$ defined in 
\eqref{eq:Lie_formal_series_level_n_def} is called the 
space of \textit{homogeneous Lie polynomials} 
of degree $n$.
The sequence $\{L_n\}_{n=0}^{\infty}$ enables us to define 
the space of \textit{Lie formal series} over $V$.

\begin{definition}[Lie formal series over $V$; Definition 2.22 
in \cite{CLL04}]
\label{def:Lie_formal_series}   
The space of \textit{Lie formal series} over $V$ is the subspace
$\cl((V)) \subset T((V))$ given by 
\begin{equation}
    \label{eq:Lie_formal_series_def}
        \cl((V)) := \Big\{ \bl \in T((V)) \mid 
        \text{for every } n \in \Z_{\geq 0}
        \text{ we have } \pi_n(\bl) \in L_n
        \Big\}.
\end{equation} 
Moreover, for each $n \in \Z_{\geq 0}$ we define the 
\textit{Lie polynomials} of degree $n$ to be the subset
$\cl^{(n)}(V) := \Pi_n \big( \cl((V)) \big)$.
\end{definition}
\vskip 4pt
\noindent
Finally we define the set of \textit{group-like} elements 
in $\tilde{T}((V))$. 

\begin{definition}[Group-like elements]
\label{def:group-like_elements}
The \textit{group-like} elements are those in the subset 
$G^{(\ast)} \subset \tilde{T}((V))$ defined by
\begin{equation}
    \label{eq:group-like_elements_def}
        G^{(\ast)} := \exp \Big( \cl((V)) \Big).
\end{equation} 
Further, for each integer $n \in \Z_{\geq 0}$, 
the \textit{free nilpotent group of order $n$} is 
defined as $G^{(n)} := \Pi_n \Big( G^{(\ast)} \Big)$.
\end{definition}
\vskip 4pt
\noindent
A particular consequence of \eqref{eq:group-like_elements_def}
is that $\cl((V)) = \log \Big( G^{(\ast)} \Big)$.
Moreover, as the name suggests, the group-like elements 
$G^{(\ast)}$ form a group under the tensor product operation
$\otimes$.
We will later see that the signature of a large class 
of paths lie within this group (cf. Subsections 
\ref{sig_p<2_sec} and \ref{rough_paths_sec}).

\subsection{Signature of Paths with Finite p-Variation}
\label{sig_p<2_sec}
In this subsection we cover the signature of a continuous
path $x : [a,b] \to V$ having finite $p$-variation for some 
$p \in [1,2)$.
We begin by introducing the notion of $p$-variation 
regularity.

If $p \in \R_{\geq 1}$ then the \textit{$p$-variation}
of a continuous path $x : [a,b] \to V$ is defined to be
\begin{equation}
    \label{p-variation_def}
        \lvert\lvert x \rvert\rvert_{p,[a,b]} :=
        \Bigg( 
        \sup_{a \leq t_0 < t_1 < \ldots < t_r \leq b} 
        \Bigg\{ 
        \sum_{j=0}^{r-1} 
        \big\lvert\big\lvert x_{t_{j+1}} - x_{t_j}
        \big\rvert\big\rvert_V^p 
        \Bigg\}
        \Bigg)^{\frac{1}{p}}.
\end{equation}
We denote the collection of continuous paths $[a,b] \to V$ 
with finite p-variation by $\cv^p([a,b],V)$. 
We can equip $\cv^p([a,b],V)$ with a norm called the 
\textit{$p$-variation norm} by defining, for 
$x \in \cv^p([a,b],V)$,
\begin{equation}
    \label{p-variation-norm_def}
        \lvert \lvert x \rvert \rvert_{\cv^p([a,b],V)} :=
        \lvert \lvert x \rvert \rvert_{p,[a,b]} +
        \sup_{t \in [a,b]} \Big\{ 
        \lvert \lvert x_t \rvert \rvert_V \Big\}.
\end{equation}
If $p,q \in \R_{\geq 1}$ with $p \leq q$ then 
$\cv^1([a,b],V) \subset \cv^p([a,b],V) \subset \cv^q([a,b],V)
\subset C^0([a,b],V)$ (see Proposition 1.7 in \cite{CLL04}, 
for example).
Further, if $\al \in (0,1]$ and $y \in C^0([a,b],V)$ is 
$\al$-H\"{o}lder continuous 
in the sense that there exists a constant $C > 0$ such that 
$\lvert \lvert y_t - y_s \rvert \rvert_V \leq
C \lvert t-s \rvert^\al$
whenever $s,t \in [a,b]$, 
then $y$ has finite $1/\al$-variation, i.e. 
$y \in \cv^{1/\al}([a,b],V)$.
Consequently, a typical Brownian path 
$B : [a,b] \to \R$ has finite $p$-variation for every 
$p > 2$, i.e. $B \in \cv^p([a,b],\R)$.
It is however known that the $2$-variation of $B$ is 
almost surely infinite \cite{Fre83}.
However, if we instead view the Brownian path as a path 
into $L^2$ rather than $\R$, then its $2$-variation on 
$[a,b]$ is finite; see \cite{CLL04}, for example.

If a continuous path $x:[a,b] \to V$ is in
$\cv^p([a,b],V)$ for a real number $1 \leq p < 2$ then 
its signature (i.e. the solution to the differential
equation \eqref{sig_diff_eqn}) can be constructed
explicitly using the \textit{Young integral}, which is a
continuous extension of the Stieltjes integral.
Given Banach spaces $V$ and $W$,  
the Young integral determines a way for a path 
$[a,b] \to \bLL(V,W)$ to be integrated along a path
$[a,b] \to V$. The notation $\bLL(V,W)$ denotes the space
of linear operators $V \to W$. To be more precise, let
$X : [a,b] \to V$ and $Y : [a,b] \to \bLL(V,W)$. 
Given a partition $\cd = (a=t_0 , t_1 , \ldots , t_r = b)$ 
of $[a,b]$ we let
$\lvert \cd \rvert := \min \big\{ t_{i+1} - t_i \mid
i \in \{0, \ldots , r-1\} \big\}$.
We first define 
\begin{equation}
    \label{Young_partition_integral_def}
        \int_{\cd} Y dX :=
        \sum_{i=0}^{r-1} Y_{t_r} \big[ 
        X_{t_{r+1}} - X_{t_r} \big].
\end{equation}    
Then, given $t \in [a,b]$, the Young integral of $Y$ along $X$ 
over the interval $[a,t]$ is defined as
\begin{equation}
    \label{Young_intgegral_def}
        \int_a^t Y_s dX_s := 
        \lim_{j \to \infty} \int_{\cd_j} Y dX
\end{equation}
where $\big\{ \cd_j \big\}_{j=1}^{\infty}$ is 
a sequence of partitions of $[a,t]$
with $\lvert \cd_j \rvert \to 0$ as $j \to \infty$.
When $p,q \in \R_{\geq 1}$ with
$\frac{1}{p} + \frac{1}{q} > 1$, 
$X \in \cv^p([a,b],V)$, and $Y \in \cv^q([a,b],\bLL(V,W))$
it can be established that the mapping
$t \mapsto \int_a^t Y_s dX_s$ defined in
\eqref{Young_intgegral_def} is a well-defined
path $[a,b] \to W$ 
with finite $p$-variation; see Theorem 1.16 in \cite{CLL04}.

Using the Young integral, if $p \in [1,2)$ and
$x \in \cv^p([a,b],V)$, then the signature of $x$ 
is given by the iterated integrals of $x$. 
That is, 
$S_{a,t}(x) = \prod_{n=0}^{\infty} S^n_{a,t}(x)$ where
\begin{equation}
    \label{sig_coeffs}
		S^0_{a,t}(x) \equiv 1
		\qquad \text{and, for } n \in \Z_{\geq 1}, \qquad 
        S^n_{a,t}(x) = 
		\idotsint\limits_{a < t_1 < \ldots < t_n < t} 
		d x_{t_1} \otimes \ldots \otimes d x_{t_n}.
\end{equation}
The path $t \mapsto S_{a,t}(x)$ is the unique solution 
to the differential equation stated in \eqref{sig_diff_eqn}
for the path $x$; see Lemma 2.10 in \cite{CLL04}, for example.

There are estimates known for the norms of the components
of the signature $S_{a,t}(x)$, i.e. there are known upper 
bounds for the quantities
$\lvert \lvert S^n_{a,t}(x) \rvert \rvert_{V^{\otimes n}}$.
In order to state these estimates we introduce the notion of 
a \textit{control} on the interval $[a,b]$. 
First, we define $\Delta_{[a,b]} := \big\{ (s,t) \mid
a \leq s \leq t \leq b \big\}$.
A control on $[a,b]$ is a continuous function 
$\omega : \Delta_{[a,b]} \to \R_{\geq 0}$ that is
super-additive,
in the sense that for every $s,t,u \in [a,b]$ with 
$s \leq u \leq t$ we have 
$\omega(s,u) + \omega(u,t) \leq \omega(s,t)$, 
and vanishes on the diagonal, in the sense that for every
$t \in [a,b]$ we have $\omega(t,t) = 0$.

Raising the $p$-variation of a path $x$ to the power of 
$p$ provides an example of a control. That is, the function 
$\omega_x : \Delta_{[a,b]} \to \R_{\geq 0}$ defined for 
$(s,t) \in \Delta_{[a,b]}$ by 
$\omega_x(s,t) := 
\lvert \lvert x \rvert \rvert_{p,[s,t]}^p$ is a control.
Given a general control
$\omega : \Delta_{[a,b]} \to \R_{\geq 0}$,
the $p$-variation of a path $x : [a,b] \to V$ is controlled 
by $\omega$ if for every $(s,t) \in \Delta_{[a,b]}$ we have 
$\lvert \lvert x \rvert \rvert_{p,[a,b]}
\leq \omega(s,t)^{1/p}$.

Now suppose that $p \in [1,2)$ and that $x \in \cv^p([a,b],V)$.
Then there exists a control
$\omega : \Delta_{[a,b]} \to \R_{\geq 0}$ such that for 
every $n \in \Z_{\geq 1}$ and every $(s,t) \in \Delta_{[a,b]}$
we have 
\begin{equation}
    \label{sig_term_fac_decay_p<2}
        \big\lvert\big\lvert S^n_{s,t}(x) 
        \big\rvert\big\rvert_{V^{\otimes n}}
        \leq 
        \frac{\omega(s,t)^n}{\Gamma 
        \Big( 1 + \frac{n}{p} \Big)}
\end{equation} 
where $\Gamma$ denotes the Gamma function. In particular, 
for positive real numbers $a > 0$ we have that 
$\Gamma(a) := \int_0^{\infty} r^{a-1} e^{-r} dr$, and 
for integers $m \in \Z_{\geq 1}$ we have 
$\Gamma (1 + m) = m!$.
This latter equality motivates the commonly used notation 
of writing $(n/p)!$ for $\Gamma(1 + n/p)$ in 
\eqref{sig_term_fac_decay_p<2}.

When $p=1$ one can take 
$\omega(s,t) := \omega_x(s,t) = \lvert\lvert x
\rvert\rvert_{1,[s,t]}$
and establish the estimate \eqref{sig_term_fac_decay_p<2} 
via a direct calculation; see Proposition 2.2 in \cite{CLL04}.
Consequently, in the case that $p=1$ we have the estimate 
that for every $n \in \Z_{\geq 1}$ and every 
$(s,t) \in \Delta_{[a,b]}$ that
\begin{equation}
    \label{sig_term_fac_decay_p=1}
        \big\lvert\big\lvert S^n_{s,t}(x)
        \big\rvert\big\rvert_{V^{\otimes n}}
        \leq 
        \frac{\lvert\lvert x \lvert\lvert_{1,[s,t]}^n}{n!}.
\end{equation}
One way to arrive at the estimates 
\eqref{sig_term_fac_decay_p<2} 
for $p \in (1,2)$ is to invoke
the \textit{Extension Theorem} for multiplicative functionals 
$\Delta_{[a,b]} \to T((V))$ appearing as Theorem 3.7 
in \cite{CLL04}.
A definition of what is meant by multiplicative functional
$\Delta_{[a,b]} \to T((V))$ is provided below 
(cf. \eqref{chen_id}) and can also be found in Definition
\ref{multiplicative_functional_def}, whilst a variant of the
\textit{Extension Theorem} appearing as Theorem 3.7
in \cite{CLL04}
is stated as Theorem \ref{extension_theorem} 
in Subsection \ref{rough_paths_sec}.
 
This Young integral approach does not work when $p \geq 2$;
however, the $p=2$ threshold is \textit{not} an artefact due to 
some particular limitation of the Young integral approach.
In fact no continuous extension of the 
Stieltjes integral can be rich enough to handle Brownian paths. 
Further, the seemingly simple function mapping a path to
the area enclosed by the path fails to be continuous in the
$2$-variation norm. A detailed discussion of these particular
issues can be found in Section 1.5 on \cite{CLL04}.

In the remainder of this subsection we
state some properties of signatures of paths with finite
$p$-variation for $p \in [1,2)$ that are
particularly useful from our machine learning perspective.
For this purpose we let $p \in [1,2)$ and fix a path
$x \in \cv^p([a,b],V)$.

The signature $S(x)$ is a \textit{multiplicative} functional 
$\Delta_{[a,b]} \to T((V))$. 
By multiplicative, we mean that Chen's identity holds; 
namely,
\begin{equation}
	\label{chen_id}
		S_{s,u}(x) \otimes S_{u,t}(x) = S_{s,t}(x)
\end{equation}
whenever $a \leq s \leq u \leq t \leq b$. A consequence 
of \eqref{chen_id} is that the signature of a concatenated 
path decomposes into the tensor product
of the signatures of the constituent parts. 
More precisely, if $x : [a,b] \to V$ and 
$y : [b,c] \to V$ are two continuous paths, then their 
concatenation $x \ast y$ is the path $[a,c] \to V$ 
defined by
\begin{equation}
	\label{concat}
		(x\ast y)_t := \bigg\{ 
        \begin{array}{ccc}
            x_t & \text{if} & a \leq t \leq b \\
            x_b - y_b + y_t & \text{if} & b < t \leq c.
        \end{array}
\end{equation} 
It then follows from \eqref{chen_id} that
\begin{equation}
	\label{chen_id_concat}
		S_{a,c}( x \ast y)
		=
		S_{a,b}(x) \otimes S_{b,c}(y).
\end{equation}
Given any positive integer $n \in \Z_{\geq 1}$
we define \textit{truncated signature}  
$S^{(n)}_{a,b}(x)$ by
\begin{equation}
    \label{trun_sig_def}
        S^{(n)}_{a,b}(x) :=
        \Pi_n \big( S_{a,b}(x) \big).
\end{equation}
It follows that $t \mapsto S^{(n)}_{a,t}(x)$ is the 
solution to a truncated version of the differential 
equation stated in \eqref{sig_diff_eqn} for the path $x$. 
To be precise, the mapping $[a,b] \to T^{(n)}(V)$ given 
by $t \mapsto S^{(n)}_{a,t}(x)$ solves the differential 
equation 
\begin{equation}
    \label{eq:n_truncated_sig_DE}
        d S^{(n)}_{a,t}(x) = S^{(n)}_{a,t}(x) \otimes dx_t 
        \qquad \text{with} \qquad
        S^{(n)}_{a,a} = \Pi_n( \textbf{1} ) 
        = (1 , 0 , \ldots , 0 ) \in T^{(n)}(V).
\end{equation} 
A detailed proof of this may be found in Chapter 7 of
\cite{FV10}, for example.

Given a word 
$\bKK = (k_1,\ldots ,k_n) \in \{1,\ldots ,d\}^n$
we define the \textit{coordinate iterated integral} 
$S_{a,b}(x)^{\bKK}$ by
\begin{equation}
	\label{c_i_i}
		S_{a,b}(x)^{\bKK}
		:=
		\big< S_{a,b}(x) , v_{\bKK} \big>_{T((V))}
		= 
		v^{\ast}_{\bKK} \big( S^n_{a,b}(x) \big).
\end{equation}
The name is justified by observing that in the case 
that the coefficients of the signature are given by 
\eqref{sig_coeffs}, we see via \eqref{in_prod_extd} that
\begin{equation}
	\label{c_i_i_smooth}
		S_{a,b} (x)^{\bKK} =
		\idotsint\limits_{a \leq t_1 \leq 
		\ldots \leq t_n \leq b}		
		\big< dx_{t_1} , v_{k_1} \big>_V \ldots 
		\big< dx_{t_n} , v_{k_n} \big>_V.
\end{equation}
The product of two coordinate iterated integrals
$S_{a,b}(x)^{\bKK}$ and $S_{a,b}(x)^{\bLL}$ 
yields a quadratic form. This quadratic form turns out to
coincides with a linear functional on elements within
the image of the signature in $T((V))$.

Before elaborating, we must introduce the
\textit{Shuffle product} $\shuffle$ defined by
\begin{equation}
	\label{shuff_prod_words}
		(i_1 , \ldots , i_n ) \shuffle
        (j_1 , \ldots , j_p )
		:=
		\sum_{\sigma \in \Shuff (n,p)} \Big(
		a_{\sigma^{-1}(1)} , \ldots , a_{\sigma^{-1}(n+p)}
		\Big)
\end{equation}
where 
$(a_1,\ldots,a_{n+p}):=(i_1,\ldots,i_n,j_1,\ldots,j_p )$ and 
\begin{equation}
	\label{shuff_perm}
		\Shuff(n,p) := \{ \sigma \in S_{n+p} \mid 
		\sigma(1) < \ldots < \sigma(n)
		\quad \text{and} \quad 
		\sigma(n+1) < \ldots < \sigma(n+p) \}.
\end{equation} 
For $n,p \in \N$, $\Shuff(n,p)$ lists all the ways that
two words, of length $n$ and $p$ respectively, can be
combined into a single word, of length $n+p$, whilst
preserving the order in which the letters of each
original word appear.

Returning to the coordinate iterated integrals
$S_{a,b}(x)^{\bKK}$ and $S_{a,b}(x)^{\bLL}$, 
on elements in the 
image of the signature in $T((V))$ their product
is given by the coordinate iterated integral 
associated to the shuffle product of the words
$\bKK$ and $\bLL$ (cf. Theorem 2.15 in \cite{CLL04}). 
That is,
\begin{equation}
	\label{shuff_prod_ten_alg}
		S_{a,b}(x)^{\bKK} 
		S_{a,b}(x)^{\bLL} = 
		S_{a,b}(x)^{\bKK \shuffle \bLL}.
\end{equation}
One consequence of \eqref{shuff_prod_ten_alg} is that 
the coordinate iterated integrals span an algebra
in $T((V))$.
Another consequence is that the signature is a 
homomorphism of paths with concatenation into the 
tensor algebra. Reversing a path produces the inverse
tensor, and \eqref{shuff_prod_ten_alg} establishes
that the range is closed under multiplication.
Hence the range of the signature forms a group in
the tensor algebra.

In fact, the group formed by the range of the signature 
is a subgroup of the group given by the 
\textit{group-like} elements $G^{(\ast)}$ introduced in 
Definition \ref{def:group-like_elements}.
In order to formalise this, we first extend the 
shuffle product to a binary operation on $T(V^{\ast})$.

Given elements $\bolde^{\ast} , \boldf^{\ast} \in T(V^{\ast})$
we define their \textit{shuffle product} 
$\bolde^{\ast} \shuffle \boldf^{\ast}$ as an element 
in $T(V^{\ast})$ as follows. 
If $\bolde^{\ast} = v^{\ast}_{\bKK}$ and 
$\boldf^{\ast} = v^{\ast}_{\bLL}$ for finite words 
$\bKK = (i_1 , \ldots , i_n) \in \{1, \ldots , d\}^n$ 
and $\bLL = (j_1 , \ldots , j_p) \in \{1, \ldots , d\}^p$, 
then we define 
\begin{equation}
    \label{eq:shuff_prod_lin_functionals}
        v^{\ast}_{\bKK} \shuffle v^{\ast}_{\bLL}
        := 
        \sum_{\sigma \in \Shuff(n,p)}
        v^{\ast}_{\Big(a_{\sigma^{-1}(1)} , \ldots , 
        a_{\sigma^{-1}(n+p)} \Big)}
\end{equation} 
where 
$(a_1,\ldots,a_{n+p}):=(i_1,\ldots,i_n,j_1,\ldots,j_p )$.
The definition in \eqref{eq:shuff_prod_lin_functionals} is
then extended by linearity to the entirety of $T(V^{\ast})$.

The result of Theorem 2.15 in \cite{CLL04} can be stated 
as follows. 
Let $S \big( \cv^p([a,b],V) \big) \subset \tilde{T}((V))$ 
denote the collection of elements in the tensor algebra 
$T((V))$ that arise as the signature of a path $[a,b] \to V$
with finite $p$-variation, i.e. the signature of an element 
in $\cv^p([a,b],V)$.
Then whenever 
$\bolde^{\ast} , \boldf^{\ast} \in T(V^{\ast})$
and $\ba \in S \big( \cv^p([a,b],V) \big) \subset 
\tilde{T}((V))$
we have that
\begin{equation}
    \label{eq:sig_image_group_prop}
        \bolde^{\ast} \shuffle \boldf^{\ast} (\ba) 
        =
        \bolde^{\ast}(\ba) \boldf^{\ast}(\ba)
\end{equation} 
It turns out that the subset of $\tilde{T}((V))$ where 
the identity \eqref{eq:sig_image_group_prop} holds for 
all $\bolde^{\ast} , \boldf^{\ast} \in T(V^{\ast})$ 
is a group (see Lemma 2.17 in \cite{CLL04}).
Moreover, one can then give the following equivalent 
definition of the group-like elements $G^{(\ast)}$.

\begin{lemma}[Equivalent definition of group-like; Variant of 
Theorem 2.23 in \cite{CLL04}]
\label{lemma:equiv_group-like_def}
Consider an element $\ba \in \tilde{T}((V))$ and 
consider $T(V^{\ast})$, 
the space of formal polynomials over $V^{\ast}$, 
to be equipped with the shuffle product.
Define the evaluation map 
$\text{ev}_{\ba} : T(V^{\ast}) \to \R$ 
by $\text{ev}_{\ba} \big( \bolde^{\ast} \big) 
:= \bolde^{\ast}(\ba)$.
Then
\begin{equation}
    \label{eq:equiv_group-like_def}
        \ba \in G^{(\ast)} 
        \qquad \text{if and only if} \qquad 
        \text{ev}_{\ba} : T(V^{\ast}) \to \R 
        \text{ is a morphism of algebras.}
\end{equation} 
\end{lemma}
\vskip 4pt
\noindent
The range of the signature is a proper subgroup of 
$G^{(\ast)}$, i.e. it is \textit{not} the entirety of 
$G^{\ast}$ (cf. Subsection \ref{rough_paths_sec}).
However, given any finite integer $n \in \Z_{\geq 0}$, 
the free nilpotent group $G^{(n)}$ of order $n$ 
(cf. Definition \ref{def:group-like_elements}) 
is the image of the truncated signature truncated to 
depth $n$ of the space $\cv^p([a,b],V)$. That is, 
$G^{(n)} = \Pi_n \circ S \big( \cv^p([a,b],V) \big)$
(see Proposition 2.27 in \cite{CLL04}).

Informally, it is helpful to think of the range of 
the signature map as a special curved space in the 
tensor algebra. As a result there is a lot of 
valuable structure. A particularly important map
is the logarithm; it is one to one on the group and 
provides a flat parameterisation of the group in 
terms of elements of the free Lie series (see Section 2
of \cite{CLL04} for details).
If $x : [a,b] \to V$ is a path with signature
$S_{a,b}(x) = \prod^{\infty}_{n = 0} S^n_{a,b}(x)$
then we know that $S^0_{a,b}(x) \equiv 1$.
Hence, if we define 
$\hat{S}_{a,b}(x) := S_{a,b}(x) - \textbf{1} \in T((V))$,
then we define the \textit{log signature} 
by taking the logarithm (defined in Definition 
\ref{def:ten_alg_exp_log}) of the element 
$S_{a,b}(x) = \textbf{1} + \hat{S}_{a,b}(x)$. 
This results in defining the Lie series 
$\LogSig_{a,b}(x) \in \cl((V))$ by 
(cf. \eqref{eq:log_def})
\begin{equation}
    \label{log_sig_series}
        \LogSig_{a,b}(x)
        := 
        \log S_{a,b}(x)
        \stackrel{\eqref{eq:log_def}}{=}
        \sum_{n=1}^{\infty}
        \frac{(-1)^{n-1}}{n} 
        \Big( \hat{S}_{a,b}(x) \Big)^{\otimes n}.
\end{equation}
Recall that the signature $S_{a,b}(x) \in \tilde{T}((V))$
since $\pi_0 \big( S_{a,b}(x) \big) = 1$.
Thus, via Lemma 2.21 in \cite{CLL04}, taking the 
exponential (defined in Definition \ref{def:ten_alg_exp_log}) 
of the log signature recovers the signature, so no 
information is lost.
The log signature 
extracts the same information as the signature, but
represents it in a more compact way.

Analogously to the definition of the truncated signature
in \eqref{trun_sig_def}, given any positive integer 
$n \in \Z_{\geq 1}$ we define the \textit{truncated log 
signature} (truncated to depth $n$) by
$\LogSig^{(n)}_{a,b}(x) := 
\Pi_n \big( \LogSig_{a,b}(x)\big)$.
That is, the truncated log signature (truncated to depth $n$)
is the projection of the full log signature to 
$\cl^{(n)}(V)$, the Lie polynomials of degree $n$.

\subsection{Rough Paths}
\label{rough_paths_sec}
In this subsection we outline how the theory
of \textit{rough paths} enables one to consider signatures
in the setting of $p$-variation regularity for $p \geq 2$. 
The core idea is to view the signature as the core object 
rather than the path itself.
In Subsection \ref{sig_p<2_sec} we observed that if 
$x \in \cv^p([a,b],V)$ for some $p \in [1,2)$, then 
its signature $S(x)$ is a multiplicative functional
$\Delta_{[a,b]} \to T((V))$ with finite $p$-variation 
in the sense that there exists a control 
$\omega : \Delta_{[a,b]} \to \R_{\geq 0}$ for which, 
given any $n \in \Z_{\geq 1}$ and any
$(s,t) \in \Delta_{[a,b]}$, we have 
\[ \big\lvert\big\lvert S^n_{s,t}(x)
\big\rvert\big\rvert_{V^{\otimes n}} 
\leq \frac{\omega(s,t)^n}{\Gamma\big(1 + n/p\big)}.\]
This motivates the following definitions.

\begin{definition}[Multiplicative Functional]
\label{multiplicative_functional_def}
Let $V$ be a Banach space, $[a,b] \subset \R$ an interval, 
and $n \in \Z_{\geq 1}$. Then a continuous function 
$X : \Delta_{[a,b]} \to T^{(n)}(V)$ is called a 
\textit{multiplicative functional of degree $n$} if it 
satisfies both of the following properties.
\begin{enumerate}[label=(\Alph*)]
    \item
    \label{mult_func_zero_level_term}
    For every $(s,t) \in \Delta_{[a,b]}$ we have that
    $X^0_{s,t} := \pi_0(X_{s,t}) = 1$.
    \item
    \label{chen_type_id}
    For every $s,u,t \in [a,b]$ with $s \leq u \leq t$ 
    we have that $X_{s,u} \otimes X_{u,t} = X_{s,t}.$
\end{enumerate}
A continuous function $\Delta_{[a,b]} \to T((V))$ satisfying
both properties \ref{mult_func_zero_level_term} and 
\ref{chen_type_id} is referred to as a 
\textit{multiplicative functional}.
\end{definition}

\begin{definition}[Finite $p$-Variation]
\label{finite_p_variation_gen_def}   
Let $V$ be a Banach space, $[a,b] \subset \R$ an interval,
$p \in \R_{\geq 1}$, 
and define $\be \in \R_{\geq 1}$ by
\begin{equation}
    \label{beta_def}
        \beta := p^2 \Bigg( 1 + \sum_{r=3}^{\infty} 
        \bigg( \frac{2}{r-2} \bigg)^{
        \frac{\lfloor p \rfloor + 1}{p}
        } \Bigg).
\end{equation}
Then a multiplicative functional $X : \Delta_{[a,b]} \to T((V))$
has \textit{finite $p$-variation} if there exists a control 
$\omega : \Delta_{[a,b]} \to \R_{\geq 0}$ such that 
for every integer $n \in \Z_{\geq 1}$ and every 
$(s,t) \in \Delta_{[a,b]}$ we have 
\begin{equation}
    \label{rough_path_finite_p_var}
        \big\lvert\big\lvert X^n_{s,t}
        \big\rvert\big\rvert_{V^{\otimes n}}
        \leq 
        \frac{\omega(s,t)^{\frac{n}{p}}}
        {\beta \Gamma \Big( 1 + \frac{n}{p} \Big)}
\end{equation}
where $\Gamma$ denotes the Gamma function.
\end{definition}
\vskip 4pt
\noindent
The constant $\beta \in \R$ in \eqref{beta_def} 
simplifies numerical constants in many of the fundamental 
results in the theory of rough paths. The precise value of
$\be$ is not of particular importance; one could change its
value without altering the class of 
multiplicative functionals determined by the requirement of 
finite $p$-variation.

We can now define a 
\textit{$p$-rough path} for general $p \in \R_{\geq 1}$.

\begin{definition}[$p$-Rough Path]
\label{p-rough_path_def}
Let $V$ be a Banach space, $[a,b] \subset \R$ be an interval, 
and $p \in \R_{\geq 1}$. Then a 
\textit{$p$-rough path} in $V$ is a multiplicative 
functional of degree $\lfloor p \rfloor$ 
(cf. Definition \ref{multiplicative_functional_def})
having finite $p$-variation 
(cf. Definition \ref{finite_p_variation_gen_def}).
The collection of $p$-rough paths in $V$ is denoted by 
$\Omega_p(V)$.
\end{definition}
\vskip 4pt
\noindent
The space of $p$-rough paths $\Omega_p(V)$ 
is \textit{not} closed under addition; having
$X,Y \in \Omega_p(V)$ does not mean that
$X + Y \in \Omega_p(V)$.
Consequently $\Omega_p(V)$ is not a vector space.

Given a $p$-rough path $X \in \Omega_p(V)$, for 
$i \in \Z_{\geq 1}$ we call $X^i := \pi_i(X) \in V^{\otimes i}$
the \textit{$i^{\text{th}}$ iterated integral} of $X$ 
despite there being no need for $X^i$ to be 
related to any kind of integral.

The following \textit{Extension Theorem} regarding 
$p$-rough paths is fundamental to the theory of rough paths.

\begin{theorem}[Extension Theorem; Variant of Theorem 3.7 
in \cite{CLL04}]
\label{extension_theorem} 
Let $V$ be a Banach space, $[a,b] \subset \R$ an interval, 
and $p \in \R_{\geq 1}$. Suppose that $X \in \Omega_p(V)$ 
is a $p$-rough path.
Then there exists a unique mutiplicative functional
$\tilde{X} : \Delta_{[a,b]} \to T((V))$
with finite $p$-variation that coincides with $X$ up to level 
$\lfloor p \rfloor$ in the sense that 
$\Pi_{\lfloor p \rfloor}\big( \tilde{X} \big) \equiv X$.
\end{theorem}
\vskip 4pt
\noindent
The Extension Theorem \ref{extension_theorem} allows
us to make the following definition of the \textit{signature} 
of a $p$-rough path $X \in \Omega_p(V)$.

\begin{definition}[Signature of a $p$-rough path]
\label{sig_of_p-rough_path_def}
Let $V$ be a Banach space, $[a,b] \subset \R$ an interval, 
$p \in \R_{\geq 1}$, and $X \in \Omega_p(V)$ be a $p$-rough 
path. The \textit{signature} of $X$ is the unique 
multiplicative functional $S(X) : [a,b] \to T((V))$ 
with finite $p$-variation such that 
$\Pi_{\lfloor p \rfloor} \big( S(X) \big) \equiv X$
in $T^{(\lfloor p \rfloor)}(V)$.
\end{definition}
\vskip 4pt
\noindent
Once again we define the \textit{log signature} of a $p$-rough 
path to be the Lie series in $\cl((V))$ resulting 
from taking the logarithm defined in Definition 
\ref{def:ten_alg_exp_log} of the signature.
If $X \in \Omega_p(V)$ is a $p$-rough path and 
$S(X) : [a,b] \to T((V))$ denote the signature of $X$,
the log signature of $X$ is the 
multiplicative functional 
$\LogSig(X) : \Delta_{[a,b]} \to T((V))$ 
defined for $(s,t) \in \Delta_{[a,b]}$ by
(cf. \eqref{eq:log_def})
\begin{equation}
    \label{log_sig_p-rough-path_formula}
        \LogSig_{s,t}(X) := 
        \sum_{n=1}^{\infty} 
        \frac{(-1)^{n-1}}{n}
        \big( S(X) - \textbf{1} \big)^{\otimes n}
\end{equation}
It is once again true that taking the exponential of the 
log signature recovers the signature. 
The log signature once again represents the same information
as the signature but in a more compact form.

We observe that Definition \ref{sig_of_p-rough_path_def} 
of the signature of a $p$-rough path is consistent 
with the definition of the signature given in Subsection 
\ref{sig_p<2_sec} when $p \in [1,2)$.
Indeed if $p \in [1,2)$ then a $p$-rough path 
$X \in \Omega_p(V)$ coincides with the signature of a 
path $x \in \cv^p([a,b],V)$ as defined in Subsection 
\ref{sig_p<2_sec}. 
To be more precise, first suppose that $x \in \cv^p([a,b],V)$.
Then the signature $S(x)$ is the unique $p$-rough path 
$X \in \Omega_p(V)$ satisfying, for every 
$(s,t) \in \Delta_{[a,b]}$, that
$\Pi_1(X_{s,t}) = ( 1 , x_t - x_s ) \in T^{(1)}(V)$.
Conversely, let $X \in \Omega_p(V)$ for $p \in [1,2)$.
Then if we define $x \in \cv^p([a,b],V)$ by defining 
$x_a := \pi_1(X_{a,a})$ and, for $t \in (a,b]$,
$x_t := \pi_1(X_{a,t}) + x_a$, then it turns out 
that $X$ coincides with the truncated signature 
$S^{(1)}(x)$ of this path $x$ \cite{CLL04}.
The \textit{Extension Theorem} \ref{extension_theorem} 
then ensures that $X \equiv S(x)$ as 
elements in the tensor algebra $T((V))$.
In this case, given an integer $i \in \Z_{\geq 1}$, 
the projection $X^i := \pi_i(X)$ of $X$ to $V^{\otimes i}$ is 
given by the $i^{\text{th}}$ iterated integral of the 
path $x$, providing some justification for choosing to 
call $X^i$ the $i^{\text{th}}$ iterated integral of $X$.

An illustrative example is the so-called Brownian rough path.
For this purpose we let $V = \R^d$ and consider an $\R^d$ 
valued Brownian motion $B : [0,T] \to \R^d$ for some $T > 0$.
For each $t \in [0,T]$ we will write 
$B_t = \big( B^1_t , \ldots , B^d_t \big)$ as the 
decomposition of $B_t$ with respect to the canonical basis
$e_1 , \ldots , e_d$ of $\R^d$.
Then we may define a multiplicative functional 
$I : \Delta_{[0,T]} \to T^{(2)}\big(\R^d\big)$ by setting, 
for $(s,t) \in \Delta_{[0,T]}$, 
\begin{equation}
    \label{Ito_mult_func}
        I_{s,t} = \Big( 1 , I^1_{s,t} , I^2_{s,t} 
        \Big) \in T^{(2)}\big(\R^d\big)
\end{equation}
where
\begin{equation}
    \label{Ito^1_def}
        I^1_{s,t} = \sum_{j=1}^d I^{1,j}_{s,t} e_j
        = \sum_{j=1} \Big( B^j_t - B^j_s \Big) e_j 
        = B_t - B_s
\end{equation} 
and 
\begin{equation}
    \label{Ito^2_def}
        I^2_{s,t} = \sum_{i=1}^d \sum_{j=1}^d 
        I^{2,ij}_{s,t} e_i \otimes e_j 
        = \sum_{i=1}^d \sum_{j=1}^d 	
        \int_{0}^T \int_0^{u_2} 
        dB^i_{u_1} dB^j_{u_2} e_i \otimes e_j.
\end{equation} 
The multiplicative functional $I$, which is sometimes called 
the \^{I}to $2$-multiplicative functional, is a $p$-rough path 
for any $2 < p < 3$ (see Corollary 3.17 in \cite{CLL04}).
The unique extension of $I$ to a multiplicative functional
$[0,T] \to T\big(\big(\R^d\big)\big)$
with finite $p$-variation then provides
an interpretation for the signature of a Brownian motion.

Given the important role played by the notion of finite
$p$-variation provided by Definition 
\ref{finite_p_variation_gen_def}, we would like a 
notion of the \textit{$p$-variation distance} between two 
$p$-rough paths.
To do so, we first observe that $\Omega_p(V)$ is a 
subset of the vector space 
$C_p^0 \big( \Delta_{[a,b]} ; T^{(\lfloor p \rfloor)}(V) \big)$
of continuous functions 
$\Delta_{[a,b]} \to T^{(\lfloor p \rfloor)}(V)$.
We can equip $\Omega_p(V)$ with a \textit{$p$-variation metric} 
$d_p$ by first defining it on the linear space 
$C_p^0 \big( 
\Delta_{[a,b]} ; T^{(\lfloor p \rfloor)}(V) \big)$.
That is, for $X,Y \in 
C_p^0 \big( \Delta_{[a,b]} ; 
T^{(\lfloor p \rfloor)}(V) \big)$
we define 
\begin{equation}
    \label{p_var_metric_def}
        d_p(X,Y) := \max_{i \in \{1, \ldots , 
        \lfloor p \rfloor \} }
        \sup_{a \leq t_0 \leq \ldots \leq t_r \leq b} 
        \Bigg( 
        \sum_{j=0}^{r-1} 
        \Big\lvert\Big\lvert 
        X^i_{t_j,t_{j+1}} - Y^i_{t_j,t_{j+1}}
        \Big\rvert\Big\rvert_{V^{\otimes i}}^{\frac{p}{i}}
        \Bigg)^{\frac{1}{p}}
\end{equation}
where for each $i \in \{1, \ldots , \lfloor p \rfloor \}$
and every $(s,t) \in \Delta_{[a,b]}$ we have 
$X^i_{s,t} := \pi_i (X_{s,t})$ and
$Y^i_{s,t} := \pi_i (Y_{s,t})$.
If the space of $p$-rough paths is equipped with the 
$p$-variation metric then $\big( \Omega_p(V), d_p \big)$
is a complete metric space \cite{LQ02}.
A more convenient characterisation of the notion of 
convergence induced by $d_p$ is provided in \cite{LQ02}
(see also Definition 3.12 in \cite{CLL04}).

Whilst the notion of $p$-rough path defined in 
Definition \ref{p-rough_path_def} extends the theory 
from Subsection \ref{sig_p<2_sec}, it does so in an abstract
theoretical manner. 
A more hands-on extension of the theory from Subsection 
\ref{sig_p<2_sec} can be made as follows.
Consider $q \in [1,2)$ and a path $x \in \cv^q([a,b],V)$. 
Then $x$ defines a $q$-rough path $X \in \Omega_q(V)$ by 
setting $X_{s,t} := ( 1 , x_t - x_s) \in T^{(1)}(V)$ 
for $(s,t) \in \Delta_{[a,b]}$.
Then given any $p \geq q$ the \textit{Extension Theorem} 
\ref{extension_theorem}
tells us that this functional $X$ can be extended 
to a multiplicative functional of degree $\lfloor p \rfloor$ 
with finite $q$-variation, and hence of finite $p$-variation.

In particular, a path with bounded variation (i.e. 
a path in $\cv^1([a,b],V)$) can be canonically viewed as a
$p$-rough path for any $p \geq 1$.
Thus under this identification of an element 
$x \in \cv^1([a,b],V)$ with a $p$-rough path 
$X \in \Omega_p(V)$, we can view $\cv^1([a,b],V)$ as a 
subset of $\Omega_p(V)$.
Hence we can consider the closure of $\cv^1([a,b],V)$ in 
$\Omega_p(V)$ with respect to the metric $d_p$ defined 
in \eqref{p_var_metric_def}.

\begin{definition}[Geometric $p$-Rough Path]
\label{geometric_p-rough_path_def}
Let $V$ be a Banach space, $[a,b] \subset \R$ an interval, 
and $p \in \R_{\geq 1}$. 
Then a \textit{geometric $p$-rough path} in $V$
is a p-rough path
$X \in \Omega_p(V)$ for which there exists
a sequence of $1$-rough paths 
$\{ X(n) \}_{n=1}^{\infty} \subset \Omega_1(V)$ with 
$d_p(X(n),X) \to 0$ as $n \to \infty$.
The space of geometric $p$-rough paths in $V$ is denoted by 
$G\Omega_p(V)$.
\end{definition}
\vskip 4pt
\noindent
It is evident that we have the inclusion that 
$G\Omega_p(V) \subset \Omega_p(V)$.
However, the inclusion is strict; not all $p$-rough paths 
are geometric $p$-rough paths. 
For example, the It\^{o} $2$-multiplicative functional 
$I$ defined 
in \eqref{Ito_mult_func} is a $p$-rough path for 
$p \in (2,3)$ but it is not geometric (see Section 3.3.1 in 
\cite{CLL04}).
But one can use an $\R^d$ valued Brownian motion to determine
a geometric $p$-rough path for $p \in (2,3)$.
This can be done by considering the dyadic piecewise linear 
approximation to the Brownian motion.

To be more precise, once again let $V = \R^d$ and consider 
an $\R^d$ valued Brownian motion $B : [0,T] \to \R^d$ for 
some $T > 0$. For each $n \in \Z_{\geq 0}$ let 
$x(n) : [0,T] \to \R^d$ be the continuous path that is 
linear off the dyadic points $k/2^n$ and satisfies that
$x(n)_{k/2^n} = B_{k/2^n}$. 
Further, for each $n \geq 0$ we can define a multiplicative
functional $S(n) : \Delta_{[0,T]} \to T^{(2)}(\R^d)$ 
of degree 2 by setting, for each $(s,t) \in \Delta_{[0,T]}$,
\begin{equation}
    \label{Strato_approx_func_def}
        S(n)_{s,t} := \Bigg( 1 , \int_s^t dx(n)_u ,
        \int_0^T \int_0^{u_2}
        dx(n)_{u_1} \otimes dx(n)_{u_2} \Bigg)
        \in T^{(2)}\big(\R^d\big).
\end{equation}
For any $p > 2$ the sequence $\{ S(n) \}_{n=1}^{\infty}$
is a Cauchy sequence in the $p$-variation topology 
\cite{Sip93,Fri05}.
Moreover its limit is given by 
\begin{equation}
    \label{Strato_func_def}
        S_{s,t} = \Bigg( 1 , B_t - B_s ,
        \frac{1}{2} \big( B_t - B_s \big)^{\otimes 2} 
        + A_{s,t} \Bigg) 
        \in T^{(2)}\big(\R^d\big)
\end{equation}
where $A_{s,t}$ is the L\'{e}vy area of the Brownian motion.
That is, for $i,j \in \{1, \ldots , d\}$ we have 
\begin{equation}
    \label{Levy_area_def}
        A_{s,t}^{ij} := \frac{1}{2} 
        \int_0^T \int_0^{u_2} dB^i_{u_1} dB^j_{u_2}
        - dB^j_{u_1}dB^i_{u_2}.
\end{equation}
The second iterated integral of the limit functional $S$
is nothing but the second iterated Stratonovich integral of
the Brownian motion; that is
\begin{equation}
    \label{strat_integral_form}
        S^2_{s,t} 
        \stackrel{\eqref{Strato_func_def}}{=}
        \frac{1}{2} \left( B_t - B_s \right)^{\otimes 2} 
        + A_{s,t}
        =
        \int_0^T \int_0^{u_2} 
        \circ dB_{u_1} \otimes \circ dB_{u_2}.
\end{equation}
Consequently, the signature of the functional $S$ is 
simply the Stratonovich signature of the Brownian motion, i.e.
the multiplicative functional 
$X : \Delta_{[0,T]} \to T\big(\big(\R^d\big)\big)$ with,
for $n \in \Z_{\geq 1}$,
its $n^{\text{th}}$ iterated integral 
$X^n : \Delta_{[0,T]} \to \big(\R^d\big)^{\otimes n}$ given by 
\begin{equation}
    \label{stratonovich_sig_nth_term}
    X^n_{s,t} := 
    \idotsint\limits_{s \leq u_1 \leq \dots \leq u_n \leq t}
    \circ dB_{u_1} \otimes \ldots \otimes \circ dB_{u_n}.
\end{equation}
This motivates the terminology that the multiplicative 
functional $S$ defined in \eqref{Strato_func_def}
is called the canonical Brownian rough path.
Proof that $S \in G\Omega_p\big(\R^d\big)$ for $p \in (2,3)$
may be found in \cite{Sip93}.

Recall from Subsection \ref{sig_p<2_sec} that if 
$x \in \cv^p([a,b],V)$ for a $p \in [1,2)$ then, 
as an element in the tensor algebra $T((V))$, 
the signature $S(x)$ is group-like, i.e. 
$S(x) \in G^{(\ast)}$ where the group-like elements 
are defined in Definition \ref{def:group-like_elements}.
In particular, given an integer $n \in \Z_{\geq 0}$, 
the truncated signature $S^{(n)}(x)$ takes its values 
in the free nilpotent group $G^{(n)}$ of order $n$ 
(cf. Definition \ref{def:group-like_elements}).
The algebraic properties related to the shuffle product
that define $G^{(n)}$ ensure that a geometric $p$-rough 
path takes its values in $G^{(\lfloor p \rfloor)}$.
Unfortunately this does \textit{not} characterise geometric 
rough paths amongst all rough paths. 

\begin{definition}[Weakly Geometric $p$-Rough Path]
\label{def:weakly_geometric_p_rough_path}   
Let $V$ be a Banach space, $[a,b] \subset \R$ an interval, 
and $p \in \R_{\geq 1}$. Then a 
\textit{weakly geometric $p$-rough path} in $V$ is a 
$p$-rough path $X \in G\Omega_p(V)$ which takes 
its values in $G^{(\lfloor p \rfloor )}$, the free nilpotent 
group of order $\lfloor p \rfloor$. 
We denote the space of weakly geometric $p$-rough paths 
by $WG\Omega_p(V)$.
\end{definition}
\vskip 4pt
\noindent 
The collection of weakly geometric $p$-rough paths 
$WG\Omega_p(V)$ is 
distinct from both the collection of geometric $p$-rough 
paths $G\Omega_p(V)$, and the collection of $p$-rough paths
$\Omega_p(V)$.
Indeed we have the following strict inclusions:
\begin{equation}
    \label{eq:strict_nesting_rough_path_spaces}
        G\Omega_p(V) \subset WG\Omega_p(V) \subset \Omega_p(V).
\end{equation}
It is not immediately obvious that the first inclusion is 
strict; the interested reader may find the details in 
\cite{FV06}.
The difference between geometric and weakly geometric rough 
paths is an annoying technicality that is, in spirit, similar 
to the distinction between $C^1$ and Lipschitz functions.
The difference between geometric rough paths and general rough 
paths is, however, substantial. 
The class of weakly geometric rough paths is the most 
important one to bear in mind for all the applications 
considered in this article.

We end this subsection with a remark about 
the interaction of the $p$-variation topology 
and the $q$-variation topology for real numbers 
$q \geq p \geq 1$.
Suppose that we have a $p$-rough path $X \in \Omega_p(V)$.
Then the \textit{Extension Theorem} \ref{extension_theorem}
tells us that $X$ can be canonically extended to a 
$q$-rough path, and thus we can slightly abuse notation 
and write $X \in \Omega_q(V)$.
In this way, any subset $\ck \subset \Omega_p(V)$ can 
be viewed as a subset $\ck \subset \Omega_q(V)$.

Now suppose that $\ck \subset \Omega_p(V)$ is a compact subset, 
which, in particular, means that $\ck$ is a compact subset
in the topology induced by the $p$-variation 
metric $d_p$ defined in \eqref{p_var_metric_def}.
Consider the map $E_{p,q} : \Omega_p(V) \to \Omega_q(V)$
mapping an element $Y \in \Omega_p(V)$
to its canonical extension in $\Omega_q(V)$.
That is, $E_{p,q}(Y)$ is the truncation to depth 
$\lfloor q \rfloor$ of the extension $\tilde{Y}$ of 
$Y$ provided by the \textit{Extension Theorem} 
\ref{extension_theorem}.
It follows from Theorem 3.10 in \cite{CLL04} that 
the map $E_{p,q}$ is continuous when the domain 
$\Omega_p(V)$ is equipped with the $p$-variation metric
$d_p$ defined in \eqref{p_var_metric_def} and the target 
$\Omega_q(V)$ is equipped with the $q$-variation metric
$d_q$ defined in \eqref{p_var_metric_def}.
Hence, being the continuous image of a compact set, 
the image $E_{p,q}(\ck)$ is a compact subset of 
$\big(\Omega_q(V),d_q\big)$.

Moreover, $E_{p,q}$ restricted to $\ck$ is an injective 
continuous mapping from a compact set $\ck$ into the 
Hausdorff space $\big(\Omega_q(V),d_q\big)$.
Thus the restriction of $E_{p,q}$ to $\ck$ is in fact
a homeomorphism onto its image, and so the 
$q$-variation topology on the image $E_{p,q}(\ck)$ coincides 
with the $p$-variation topology on $\ck$.

Therefore, for every $q \geq p$ the image $E_{p,q}(\ck)$
is a compact subset of $\big(\Omega_q(V) , d_q\big)$, and the 
$q$-variation topology induced on $E_{p,q}(\ck)$ by 
viewing it as a subset of $\Omega_q(V)$ coincides with 
the $p$-variation topology induced on $\ck$ as a subset of 
$\big(\Omega_p(V),d_p\big)$.

\section{Signature as a Feature Map}
\label{good_props}
We assume that $V$ is a finite $d$-dimensional Banach space, 
and that $\Omega \subset \s(V) \times \R$ is a labelled 
dataset. Following our conventions, this means that each 
element in $\Omega$ is a pair $(\bx,y) \in \s(V) \times \R$
where the stream $\bx$ is assumed to consist of 
incremental data. 
We assume that the streams of increments $\bx$ contained 
within the set $\Omega$ are of the form determined by the 
method of transforming a stream of values to a stream of 
increments detailed in Subsection \ref{subsec:math_framework}.
A particular consequence of this assumption is that the 
integer $d \in \Z_{\geq 1}$ must be even; see Subsection 
\ref{subsec:math_framework}.

Suppose that $M \in \Z_{\geq 1}$ is finite, and that
$\Omega = \big\{ (\bx_k , y_k) \in \s(V) \times \R \mid
k \in \{1 , \ldots , M \} \big\}$.
Recall that $\Omega_{\s(V)} \subset \s(V)$ denotes the 
projection of $\Omega$ onto $\s(V)$.
Moreover, we assume that
$\Omega_{\s(V)} \subset \m \subset \s(V)$, and our 
goal is to use the labelled dataset $\Omega$ to learn a
continuous function $f \in C^0(\m;\R)$ for the purpose of 
predicting the systems response to inputs 
$\bx \in \s(V)$ that are not within $\Omega_{\s(V)}$.
We reiterate that we are implicitly assuming that, in 
some sense, the subset $\m$ is sufficiently well 
approximated by $\Omega_{\s(V)}$ to make this task 
reasonable (cf. Subsection \ref{subsec:math_framework}).

We now outline a basic strategy for 
incorporating the signature as a feature map in this
regression task. 
\vskip 4pt
\noindent
\textbf{Simple Signature Regression Framework}
\begin{enumerate}[label=(\Alph*)]
    \item\label{Sig_Reg_Outline_A}
    Consider the set of contours $C = \big\{ \Gamma_{\bx_k} 
    \mid k \in \{1, \ldots , M\} \big\}$ where, for each 
    $k \in \{1, \ldots , M\}$, the contour $\Gamma_{\bx_k}$
    is the contour formed by the concatenation of the 
    entries within the stream of increments $\bx_k$.
    \item\label{Sig_Reg_Outline_B} 
    For each $k \in \{1, \ldots , M\}$ choose a parameterisation
    $\gamma_k : [0,1] \to V$ of the contour $\Gamma_{\bx_k}$.
    For each $k \in \{1, \ldots , M\}$, the stream $\bx_k$
    contains a finite number of increments, thus 
    each path $\gamma_k$ is of bounded variation. 
    That is, for each $k \in \{1, \ldots , M\}$ we have 
    $\gamma_k \in \cv^1([0,1],V)$, and so we can 
    consider its signature
    $S(\gamma_k) : \Delta_{[0,1]} \to T((V))$. 
    \item\label{Sig_Reg_Outline_C} 
    Choose $K_0 \in \Z_{\geq 1}$ and consider the 
    collection of truncated to depth $K_0$ signatures 
    $S^{(K_0)}(\gamma_k)$ for $k \in \{1 , \ldots , M\}$.
    Numerically solve the system of equations given by
    \begin{equation}
        \label{sig_learn_eqn}
            \sum_{i=0}^{K_0} 
            \sum_{\bJ \in \{1,\ldots,d\}^i}
            \al_i(\bJ) S_{0,1} 
            ( \gamma_k )^{\bJ}
            = y_k
    \end{equation}
    for each $k \in \{1 , \ldots , M\}$.
    This is a system of $M$ equations in 
    $1 + d + \ldots + d^{K_0} = \frac{d^{K_0 + 1} - 1}{d - 1}$
    variables.
    \item\label{Sig_Reg_Outline_D} 
    Take the learnt function to be $f : \m \to \R$ defined by
    \begin{equation}
        \label{sig_learnt_func}
            f(\bv) := 
            \sum_{i=0}^{K_0} 
            \sum_{\bJ \in \{1,\ldots,d\}^i}
            \al_i(\bJ) S_{0,1} 
            \big( \gamma(\bv) \big)^{\bJ}, 
    \end{equation}
    where $\gamma(\bv) : [0,1] \to V$ is a parameterisation 
    of the contour $\Gamma_{\bv}$ corresponding to the stream 
    $\bv \in \m$.
\end{enumerate}
\noindent
The sensibility of this approach and the attractiveness 
of the signatures as a feature map from a machine learning 
point of view, as claimed in Section \ref{SI}, 
rely on the following 
key theoretical properties of signatures. 
The first key property is that the 
infinite graded sequence of statistics provided by the 
signature identifies a contour in an essentially unique 
manner. 

\begin{theorem}[Uniqueness of signature, \cite{HL10,BGLY16}]
\label{tree_like_unique}
Let $V$ be a Banach space, $[a,b] \subset \R$ a compact 
interval, and $p \in \R_{\geq 1}$.
Then the signatures of two geometric $p$-rough paths 
$X,Y \in G\Omega_p(V)$ coincide if and only if the 
$X$ and $Y$ are tree-like equivalent.
\end{theorem}
\vskip 4pt
\noindent
The full technical details of \textit{tree-like} equivalence 
may be found in either of \cite{HL10} or \cite{BGLY16}, 
for example. Heuristically, the signature 
determines the path up to sections on which the path 
exactly retraces itself. 
In our setting such sections can be avoided by augmenting 
the initial streams to contain a monotonic channel.
Recalling our view that the data is in the form of streams 
of increments, this can be achieved, for example, by 
augmenting each stream to have a channel 
consisting of strictly positive entries.

One particularly important example of tree-like 
equivalence is when one path is a translation of the 
other. To be more precise, suppose that 
$\gamma : [a,b] \to V$ is a continuous path of bounded 
variation. 
Consider a fixed $v \in V$ and define a continuous 
path of bounded variation $\psi : [a,b] \to V$ by 
setting, for $t \in [a,b]$, $\psi_t := \gamma_t + v$.
Then $\gamma$ and $\psi$ are tree-like equivalent.
Consequently the signature is invariant under 
translation of a path.

This invariance motivates our use of the invisibility 
reset transformation (cf. Subsection \ref{dataset_to_path})
when detailing our method of transforming a stream of 
values into a stream of increments in Subsection 
\ref{subsec:math_framework}.
Indeed the invisibility reset transformation is used to 
explicitly encode both the first and last values of a 
stream into its resulting stream of increments in a manner
that is invariant under translation.
Consequently, the signature of the path resulting from 
the concatenation of these increments \textit{will} 
capture information related to the norm/size of the entries
in the original stream of values.
This information would \textit{not} be captured by the 
signature if we worked with the stream of incremental
changes obtained by only considering the differences between
successive entries of the original stream of values.

Another important example of tree-like equivalence 
is when one path is a reparameterisation of the other.
To be more precise, suppose that 
$\gamma : [a,b] \to V$ is a continuous path of bounded 
variation. Then if $\psi : [a,b] \to [a,b]$ is a 
continuously differentiable increasing surjection, 
then the paths $\gamma$ and $\gamma \circ \psi$ 
are tree-like equivalent. Consequently 
the signature is invariant under reparameterisation 
of a path; it essentially quotients out the 
infinite-dimensional group of time reparameterisation.
This formalises our earlier claims that the signature
filters out the parameterisation noise.

A consequence of this reparameterisation invariance
is that the signature can record the order of events
\textit{without} needing to record precisely when each 
event occurs. This is a distinct advantage compared to 
Fourier transform/wavelets techniques.

Consider, for example, monitoring financial markets for 
signs of insider trading. Three events that might be of 
interest are a phone call, a trade and price movement. 
The order of these three events is critical; the order 
phone call, trade and then price movement could indicate 
insider trading. To capture the order using Fourier series 
or wavelets we must sample each of the three channels 
representing the three events. 
The sampling frequency will likely need to be very high to 
resolve each channel accurately enough to capture the order,
leading to the consideration of large quantities of 
uninformative data. But the signature captures this 
information in the first two levels irrespective of how 
it is sampled, see example 3.8 in \cite{LLN16}.

Theorem \ref{tree_like_unique} ensures that, for each 
$k \in \{1, \ldots , M \}$, the choice of parameterisation 
$\gamma_k$ for the contour $\Gamma_k$ is irrelevant to 
the resulting signature $S(\gamma_k)$. 
Any two parameterisations of $\Gamma_k$ are reparameterisations
of one and other, and hence Theorem \ref{tree_like_unique}
means their signatures coincide.
In fact this provides a way to determining a topology 
on a space of unparameterised paths in $V$.

Indeed, given a fixed compact interval $[a,b] \subset \R$, 
we may consider the space of unparameterised paths in $V$ 
determined by taking the quotient of $\cv^1([a,b],V)$ 
with respect to the equivalence relation generated by the 
notion of tree-like equivalence \cite{HL10}.
We will denote the resulting space by $\Path_1([a,b],V)$, 
and each element within it is an equivalence class 
$[\gamma]$ for some $\gamma \in \cv^1([a,b],V)$.
By appealing to Theorem \ref{tree_like_unique}, we 
can define the signature of an element
$[\gamma] \in \Path_1([a,b],V)$ to be the signature 
of \textit{any} representative $\al \in \cv^1([a,b],V)$
of the equivalence class $[\gamma]$.

There is no canonical topology on $\Path_1([a,b],V)$ 
\cite{HL10}. A thorough investigation of various approaches 
to equipping $\Path_1([a,b],V)$ with a topology and the
subsequent properties of each resulting topology is 
provided in \cite{CT22}.
A concrete way to use the signature to determines a 
topology on $\Path_1([a,b],V)$ is to define a 
metric on $\Path_1([a,b],V)$ as follows. 
Given $[\gamma] , [\al] \in \Path_1([a,b],V)$ define 
\begin{equation}
    \label{metric_on_unparam_path_1}
        d_{\Path_1} \big( [\gamma] , [\al] \big) 
        := 
        d_1 \big( S(\gamma) , S(\al) \big)
\end{equation} 
where $S(\gamma) , S(\al) : \Delta_{[a,b]} \to T((V))$ 
denote the signatures
of $\gamma$ and $\al$ respectively, and $d_1$ is the 
$1$-variation metric defined in \eqref{p_var_metric_def}.
Theorem \ref{tree_like_unique} ensures that $d_{\Path_1}$
defined in \eqref{metric_on_unparam_path_1} is well-defined.

The next key property is that the terms of the signature 
decay in size factorially as the order increases. 

\begin{theorem}[Factorial Decay; Proposition 2.2 in \cite{CLL04}]
\label{fac_decay_thm}
Let $V$ be a Banach space, $[a,b] \subset \R$ a compact 
interval, and $\gamma \in \cv^1([a,b],V)$.
Let $S(\gamma) : \Delta_{[a,b]} \to T((V))$ 
denote the signature of $\gamma$.
Then for every $n \in \Z_{\geq 1}$ and every 
$(s,t) \in \Delta_{[a,b]}$, the $n^{\text{th}}$ component
of the signature
$S^n(\gamma) : \Delta_{[a,b]} \to V^{\otimes n}$
satisfies that
\begin{equation}
    \label{fac_decay}
        \big\lvert\big\lvert S^n_{s,t}(\gamma)
        \big\rvert\big\rvert_{V^{\otimes n}}
        \leq 
        \frac{\lvert\lvert \gamma \lvert\lvert_{1,[a,b]}^n}{n!}.
\end{equation}	
\end{theorem}
\vskip 4pt
\noindent
Consequently, in a norm based sense, the truncation of the 
signature to a finite depth manages to capture the terms 
with the most significant contribution.
Thus it is reasonable, again in a norm based sense, to 
imagine that by choosing a sufficiently large $K_0$ in 
\textbf{Simple Signature Regression Framework}
step \ref{Sig_Reg_Outline_C}, the resulting truncated 
signatures still manage to summarise the most 
significant affects of the paths.

Examining \textbf{Simple Signature Regression Framework}
steps \ref{Sig_Reg_Outline_C} and \ref{Sig_Reg_Outline_D},
the underlying idea is to approximate the systems 
response by a linear combination of signature components.
The sensibility of this relies upon which classes of 
functions are captured by such linear combinations.
Loosely, we need to know if linear combinations of 
signature components are capable of providing a good
approximation to an arbitrary continuous function.
The final key property is that this \textit{is} the case
since continuous functions of paths are approximately 
linear on signatures. Thus the signature is, in some sense, 
a \textit{universal nonlinearity} on paths.

\begin{theorem}[Universal Nonlinearity; Variant of Theorem 2.1 
in \cite{BDLLS20}]
\label{uni_nonlin}
Let $V$ be a Banach space, $[a,b] \subset \R$ a compact 
interval, $\ck \subset \Path_1([a,b],V)$ a 
compact subset, and $F : \ck \to \R$ a continuous function.
Then for any $\ep > 0$ there exists a truncation level 
$n \in \Z_{\geq 1}$ and a collection of real coefficients 
$\big\{ \al_i(\bJ) \mid i \in \{0 , \ldots , n\} \text{ and } 
\bJ \in \{1 , \ldots , d \}^i \big\}$
such that for every $[\th] \in \ck$ we have
\begin{equation}
    \label{universal_est}
        \Bigg\lvert F([\th]) - \sum_{i=0}^n 
        \sum_{\bJ \in \{1 ,\ldots , d\}^i }
        \al_i(\bJ) S_{a,b}(\th)^{\bJ}
        \Bigg\rvert \leq \ep. 
\end{equation}
\end{theorem}
\vskip 4pt
\noindent
In Theorem \ref{uni_nonlin} the 
topology on $\Path_1([a,b],V)$ is the topology 
determined by the metric 
$d_{\Path_1}$ defined in \eqref{metric_on_unparam_path_1}.
We sketch one approach to establishing this result.
The image $\big\{ S(\gamma) \mid [\gamma] \in \ck \big\}$
determines a compact subset of $\Omega_1(V)$.
Further, the shuffle product relation 
\eqref{shuff_prod_ten_alg} can be used to establish that 
the coordinate iterated integrals span an algebra.
Then the conclusion of Theorem \ref{uni_nonlin} is obtained 
by appealing to the \textit{Stone--Weierstrass Theorem} 
\cite{Sto48}.

Returning to \textbf{Simple Signature Regression Framework}
step \ref{Sig_Reg_Outline_A}, the 
subset $\big\{ \Gamma_k \mid k \in \{1, \ldots , M\} \big\}$
determines a finite subset of $\Path_1([0,1],V)$.
Hence Theorem \ref{uni_nonlin} is applicable,  
and we are guaranteed that, 
provided a sufficiently large truncation level is chosen, 
it will be possible to approximate \textit{any} continuous 
function on this finite subset by a linear combination 
of the coordinate iterated integrals of the truncated signature.

We have only examined the signatures of the contours
corresponding to the data set at the $1$-variation scale. 
An obvious motivation for doing so is that the contours 
naturally make sense at this scale; that is, as paths with 
bounded variation. 
But just because they make sense at this scale does not 
mean it is the most natural scale to work with. 
It might be more sensible to 
consider them at the $p$-variation scale for $p > 1$. 

As observed in Subsection \ref{rough_paths_sec},
if $\gamma \in \cv^1([a,b],V)$ and $p \geq 1$ 
then the truncation of the signature of $\gamma$ 
to depth $\lfloor p \rfloor$ give a canonical way to lift 
$\gamma$ to a $p$-rough path 
$X(p) = S^{(\lfloor p \rfloor)}(\gamma) \in \Omega_p(V)$.
Consequently one can canonically view $\cv^1([a,b],V)$
as a subset of $\Omega_p(V)$, and choose to work with 
the $p$-variation topology induced by the $p$-variation 
distance metric defined in \ref{p_var_metric_def}.
For each $j \in \{1 , \ldots , M\}$ and any $p \in \R_{\geq 1}$
let $X_j(p)$ denote the canonical lift of $\gamma_j$ to 
a $p$-rough path in $\Omega_p(V)$, i.e. 
$X_j(p) := S^{(\lfloor p \rfloor)}(\gamma_j)$.

Then as remarked at the end of Subsection 
\ref{rough_paths_sec}, the subset 
$\big\{ X_1(p) , \ldots , X_M(p) \big\}$ is both 
compact as a subset in $(\Omega_p(V),d_p)$, 
and additionally satisfies that
the $p$-variation topology induced on 
$\big\{ X_1(p) , \ldots , X_M(p) \big\}$ coincides with 
the $1$-variation topology induced on 
$\big\{ X_1(1) , \ldots , X_M(1) \big\}$ as a subset of 
$(\Omega_1(V),d_1)$.
Consequently, the theoretical guarantees of Theorem 
\ref{uni_nonlin} remain valid even after switching to 
viewing our collection of signatures at the 
$p$-variation scale for some $p > 1$.

The central idea has been to associate 
a particular $p$-rough path with each stream 
$\bx_k$ in the original data set $\Omega_{\s(V)}$.
Thus far we have considered doing so by first 
associating a $1$-rough path with the stream $\bx_k$, 
and, if deemed sensible, subsequently lifting this 
$1$-rough path to a $p$-rough path for $p > 1$.
The incremental information provided by the stream $\bx_k$ 
determines how the lift is defined.
However, if the stream $\bx_k$ contains more than just 
incremental data, there may be a more natural way of 
prescribing the higher order information in order to 
associate $\bx_k$ with a $p$-rough path. 

For example, suppose a stream provided increments 
corresponding to an $\R^d$ valued Brownian motion 
$B : [0,T] \to \R^d$ for some $T > 0$. 
Then both the It\^{o} $2$-multiplicative functional 
(cf. \eqref{Ito_mult_func})
and the canonical Brownian rough path 
(cf. \eqref{Strato_func_def})
defined in Subsection \ref{rough_paths_sec} 
determine distinct $p$-rough paths, for $p \in (2,3)$,
which could be associated to the stream.
However, if the stream additionally contained information
determining how, for example, the area evolves, then 
which of these two candidates is more appropriate to 
choose could be determined.

The point is that we could associate a $p$-rough path 
to a stream $\bx$ by appropriately prescribing the 
coordinate iterated integrals up to depth $\lfloor p \rfloor$.
The \textit{Extension Theorem} \ref{extension_theorem} 
would then tell us that all the coordinate iterated 
integrals at higher depths would be uniquely determined, 
and so we could follow the method outlined in 
\textbf{Simple Signature Regression Framework} 
using the signatures of these prescribed $p$-rough paths.
Consideration of Theorem \ref{tree_like_unique} 
suggests it is most sensible to associate each stream 
with a geometric $p$-rough path, since we then preserve the
same sense of uniqueness for the signature.

Indeed with this tree-like equivalence being preserved we 
could repeat our earlier approach. 
We can define $\Path_p([a,b],V)$ to be the quotient 
of $G\Omega_p(V)$ with respect to the equivalence relation 
generated by the notion of tree-like equivalence. 
Once again Theorem \ref{tree_like_unique} ensures we 
can define the signature of an element 
$[X] \in \Path_p([a,b],V)$ to be the signature of 
the geometric $p$-rough path $Y \in G\Omega_p(V)$ 
for \textit{any} representative $Y$ of the equivalence class
$[X]$.
The $p$-variation metric $d_p$ on $\Omega_p(V)$ can now be 
used to define a metric on $\Path_p([a,b],V)$.
Given $[X] , [Y] \in \Path_p([a,b],V)$ define 
\begin{equation}
    \label{metric_on_unparam_path_p}
        d_{\Path_p} \big( [X] , [Y] \big) 
        := 
        d_p \big( S(X) , S(Y) \big)
\end{equation} 
where $S(X) , S(Y) : \Delta_{[a,b]} \to T((V))$ 
denote the signatures
of $X$ and $Y$ respectively, and $d_p$ is the 
$p$-variation metric defined in \eqref{p_var_metric_def}.
Theorem \ref{tree_like_unique} ensures that $d_{\Path_p}$
defined in \eqref{metric_on_unparam_path_1} is well-defined.

The analogue of Theorem \ref{uni_nonlin} in this setting 
can then be established in exactly the same way.
Multiplicativity ensures the shuffle product remains valid, 
and so the coordinate iterated integrals (i.e. the components
of the signature) still span an algebra. Hence the 
\textit{Stone--Weierstrass} theorem \cite{Sto48}
may once again be 
invoked to determine that on compact subsets of 
$\Path_p([a,b],V)$ continuous functions can be arbitrarily 
well-approximated by linear combinations of coordinate 
iterated integrals (i.e. signature components).

The theoretical guarantees for the strategy outlined in 
\textbf{Simple Signature Regression Framework} make the 
signature an attractive choice of feature map.
The signature summarises a path in a way that is 
essentially unique and, in particular, that is invariant 
under reparameterisation of 
the path (cf. Theorem \ref{tree_like_unique}).
The coefficients of the signature (namely, the
coordinate iterated integrals) provide a set of 
feature functions that have a natural grading
(cf. Theorem \ref{fac_decay_thm}), and which are rich
enough to capture continuous functions on
path space (cf. Theorem \ref{uni_nonlin}).

The log signature is also an appealing candidate feature map.
As previously remarked, the log signature captures
the same information as the signature, and 
represents it in a more compact form. 
This seems ideal from our machine learning perspective.
However, whilst no information about the path is lost,
information regarding the nonlinearity is lost.

The coefficients of the signature are rich enough
to capture continuous functions on path space 
via linear combinations (cf. Theorem \ref{uni_nonlin}).
Thus the signature captures the universal nonlinearity
of the path, reducing all continuous nonlinearities
to linear combinations of its coefficients.
This property is not true for the log signature; 
whilst the coefficients of the log signature can 
be used to capture continuous functions, it is 
\textit{not} achieved via linear combinations.
Instead, more complicated nonlinear combinations of
the log signature coefficients are required.
Loosely, the signature captures both the information
\textit{and} the nonlinearity of a path, whilst the
log signature captures only the information.

\section{Controlled Differential Equations and the Log-ODE Method}
\label{CDE_log-ODE}
The \textit{log-ODE} method, developed within the area
of \textit{rough path theory} \cite{Lyo98,CLL04,FV10}, 
is a numerical method for solving \textit{Controlled 
Differential Equations} (CDEs) using the
log signature of the control path over short time intervals,
rather than relying on pointwise evaluations of the control
path. CDEs provide a framework under which one can give
a precise meaning for the response of a system to a path
according to some specified dynamics. 

We first precisely define the notion of a CDE driven by a path.
Let $t_0 , t_1 \in \R$ with $t_0 < t_1$ and $V,W$ both be a
Banach space. Suppose that $X : [t_0 , t_1] \to V$ is
a path. Let $\bLL(V,W)$ denote the set of continuous linear
mappings $V \to W$ and suppose that $f : W \to \bLL(V,W)$ is
continuous. Then a path $z : [t_0,t_1] \to W$ solves a
CDE controlled, or driven, by $X$ with initial condition 
$w \in W$ if 
\begin{equation}
    \label{CDE_eqn_gen}
        z_{t_0} = w
        \qquad \text{and} \qquad
        dz_t = f(z_t) dX_t
\end{equation}
for every $t \in (t_0,t_1]$.
The CDE framework provides a precise meaning for the
response of a system $z$ to the path $X$ according
to the dynamics $f$. In the case that $V := \R$ and
$X(t) := t$ for every $t \in [t_0,t_1]$ the CDE
\eqref{CDE_eqn_gen} reduces to an \textit{Ordinary Differential
Equation} (ODE).
An excellent introduction to CDEs
may be found in \cite{CLL04}. A comprehensive
exposition is provided in the textbook \cite{FV10}.
CDEs have been used within the field of
\textit{Deterministic Control Theory}; a comprehensive
introduction to deterministic control theory can be found
in \cite{Zab20}.

The link with signatures is apparent from our earlier
observation in Section \ref{Ten_Alg} 
(cf. \eqref{sig_diff_eqn}) that the
signature of a path $X$ solves the universal
non-commuting exponential CDE driven by $X$.
That is, the path $z : [t_0 , t_1] \to T((V))$
defined by $z_t := S_{t_0,t}(X)$ for $t \in [t_0,t_1]$
satisfies that
\begin{equation}
    \label{CDE_sig_eqn}
        z_{t_0} = \textbf{1} = 
		(1 , 0 , 0 , \dots ) \in T((V))
		\qquad \text{and} \qquad 
		d z_t = z_t \otimes dX_t
\end{equation}
for every $t \in (t_0,t_1]$. It turns out that
the solution of any linear CDE can be expressed in terms
of signature. For example, if the control path 
$X \in \cv^1([t_0,t_1],V)$ and 
$f(z_t)dX_t = B(dX_t)z_t$ for a bounded linear map
$B : V \to \bLL(W,W)$, then the solution to 
\eqref{CDE_eqn_gen} is given by 
\begin{equation}
    \label{lin_CDE_sol_sig_involve}
        z_t = \Bigg( \sum_{n=0}^{\infty} 
        B^{\otimes n} \Big( S^n_{t_0,t}(X) 
        \Big) \Bigg) [w]
\end{equation}
for any $t \in [t_0,t_1]$. This follows from the standard
Picard iteration argument coupled with the factorial 
decay rate for the signature components $S^n_{t_0,t}(X)$
(cf. Theorem \ref{fac_decay_thm} in this article); 
see \cite{CLL04,Lyo14}, for example.
Existence and uniqueness results for the general case
using \textit{rough path theory} may be found, for example, 
in \cite{Lyo98,CLL04,FV10}.

The map $f$ in \eqref{CDE_eqn_gen} can be thought of
as a linear map from $V$ to the space of vector fields 
on $W$. This object takes an element $v \in V$ and an
element $w \in W$ and produces a second element in $W$
representing the infinitesimal change to the state $z$
of the system that will occur if $X$ is changed
infinitesimally in the direction $v$ \cite{Lyo14}.
The link between truncated log signatures and 
vector fields (see Section 6 in \cite{Lyo14}, for example)
can be turned into a practical approach to understanding
CDEs. 

The naive approach of using a Taylor series based method
to approximate the solution using truncated log signatures
suffer stability issues. The log-ODE method avoids these
issues by returning the numeric issues back to
state-of-the-art ODE solvers. 
This is achieved by using the truncated log-signature to 
determine a path-dependent vector field $\tilde{f}$ such 
that the solution to the ODE 
$\frac{dx_t}{dt} = \tilde{f}(x_t)$ provides a good 
approximation to the solution of the original CDE.
Stability issues are now determined by the choice of ODE
solver. 

The log-ODE method has proven useful in
several rapidly developing areas. It is central to the
development of \textit{Neural Rough Differential Equations}
in \cite{FKLMS21} and their subsequent refinement in 
\cite{BEN24}. In \cite{FLO20} it is 
illustrated that the log-ODE method may be used for the 
numerical approximation of 
\textit{Stochastic Differential Equations} (SDEs). 
Neural Rough Differential Equations are covered in
Section \ref{LSDE} of this article.
The log-ODE method proposed in \cite{FLO20} exhibits 
high order convergence rates that are comparable with 
all contemporary high order methods for the numerical
approximation of SDEs, and can significantly outperform
lower order numerical methods for SDEs (such as the
Euler-Maruyama and Milstein methods).

We finish this section by following the presentation in 
Appendix A of \cite{FKLMS21} to illustrate the log-ODE method
under the assumption that both $V$ and $W$ are finite 
dimensional. Treatment of the infinite dimensional case 
may be found in \cite{BGLY14}, for example.
We assume $X \in \cv^1([0,T],V)$ has finite length 
and that $f : W \to \bLL( V , W)$ is either 
linear or bounded with $N$ bounded derivatives for some 
given $N \in \Z_{\geq 0}$. For $k \in \{0, \ldots , N\}$
we define 
$f^{\circ(k)} : W \to \bLL \big( V^{\otimes k} ,W\big)$ 
using the derivatives of $f$ 
as follows (see Definition A.6 in \cite{FKLMS21}).
Below we use the notation that 
$D f^{\circ(k-1)} : W \to 
\bLL( W , \bLL( V^{\otimes (k-1)} , W) )$ denotes the
Fr\'{e}chet derivative of $f^{\circ(k-1)}$. We additionally
use the convention that rounded brackets $(\cdot)$ are
used for evaluation at points in $W$, whilst square brackets
$[\cdot]$ have been used for evaluation at points in 
$V$ and its tensor powers.
\begin{itemize}
    \item For $k=0$ we define 
    $f^{\circ(0)} : W \to W$ 
    by setting 
    $f^{\circ(0)}(y) := y$ for every $y \in W$.
    \item For $k =1$ we define 
    $f^{\circ(1)} : W \to \bLL(V,W)$ 
    by setting 
    $f^{\circ(1)}(y) := f(y)$ for every $y \in W$.
    \item For the remaining $k \in \{2 , \ldots , N\}$ we 
    proceed inductively. Once
    $f^{\circ(k-1)} : W \to \bLL ( V^{\otimes(k-1)} , W)$ 
    has been defined, define 
    $f^{\circ(k)} : W \to \bLL ( V^{\otimes k} , W)$ 
    as follows. For each $y \in W$ we set $f^{\circ(k)}(y)$ 
    to be the unique $k$-linear map in 
    $\bLL ( V^{\otimes k} ,W)$ for which 
    \begin{equation}
        \label{circ(k)_def}
            f^{\circ(k)}(y)[ v \otimes v_{k-1} ] 
            = 
            Df^{\circ(k-1)}(y) \big( f(y)[v] \big) 
            \big[ v_{k-1} \big]
    \end{equation}
    whenever $v \in V$ and $v_{k-1} \in V^{\otimes (k-1)}$.
\end{itemize}
\noindent
As noted in \cite{FKLMS21}
the functions $f^{\circ(0)} , \ldots , f^{\circ(N)}$
naturally arise in the Taylor expansion associated to the 
CDE \eqref{CDE_eqn_gen}.
A combination of 
the functions $f^{\circ(0)} , \ldots , f^{\circ(N)}$ 
and the truncated log signature $\LogSig^N(X)$ is used to 
give an appropriate vector field $\tilde{f}$.

Suppose that $t_0 \leq s < t \leq t_1$ and that we know 
the solution at time $s$, i.e. that we are given $z_s$.
Further suppose that we have computed the truncated
log signature $\LogSig^{(N)}_{s,t}(X)$. Then we define 
a vector field $\tilde{f} : W \to \bLL(V,W)$ by 
\begin{equation}
    \label{log-ODE_vf_def}
        \tilde{f}(u) :=
        \sum_{k=0}^N f^{\circ(k)}(u)
        \bigg[
        \pi_k \Big( \LogSig^{(N)}_{s,t}(X) \Big)
        \bigg]
\end{equation}
for $u \in W$. Here $\pi_k : T^{(N)} (V) \to V^{\otimes k}$ 
denotes the projection map. We may then consider the ODE 
\begin{equation}
    \label{log-ODE_eqn}
        x(0) = z_s 
        \qquad \text{and} \qquad
        \frac{d x(r)}{dr} = \tilde{f}(x(r))  
\end{equation}
for all times $r \in (0,1]$. Then the log-ODE
approximation of $z_t$ (given $z_s$ and 
$\LogSig^{(N)}_{s,t}(X)$) is 
\begin{equation}
    \label{log-ODE_approx_def}
        \LogODE( z_s , f , 
        \LogSig^{(N)}_{s,t}(X)) 
        :=
        x(1).
\end{equation}
In fact the function
$\LogODE(z_s,f,\LogSig^{(N)}_{s,t}(X) , \cdot) : [s,t] 
\to W$ defined by
\begin{equation}
    \label{log-ODE_approx_func}
        \LogODE(z_s,f,\LogSig^{(N)}_{s,t}(X) , \tau)
        :=
        x \bigg( \frac{\tau - s}{t-s} \bigg)
\end{equation}
for $\tau \in [s,t]$ gives an approximation to the 
solution $z_{\tau}$ for $\tau \in [s,t]$.
Only the vector field $f$ 
and its (iterated) Lie brackets are required to construct
the log-ODE vector field $\tilde{f}$ defined in 
\eqref{log-ODE_vf_def} (see Remark A.10 in \cite{FKLMS21}). 
This is a consequence of the fact that the log signature 
of $X$ lies in a certain free Lie algebra; see 
Subsections \ref{Ten_Alg}, \ref{sig_p<2_sec}, and 
\ref{rough_paths_sec} in this article, or refer
Section 2.2.4 of \cite{CLL04} for the precise details.

Error estimates for the log-ODE method may be found in 
\cite{BGLY14} (including the full general case) where
the approximation error of the log-ODE method is 
quantified in terms of the regularity of the systems 
vector field $f$ and control path $X$ (also see Appendix
B in \cite{FKLMS21} for coverage of the theory in a 
simplified setting). These results make 
use of numerous technical details from 
\textit{rough path theory}, which one may find in either of
\cite{CLL04,FV10}, for example.
The simplified conclusion of the sophisticated theory is 
that log-ODEs can approximate CDEs.
Moreover, the convergence rate is controlled by the 
depth of log signature used and the step size considered
(i.e. the size of $t-s$ in our presentation above). 
Performance of the log-ODE method can be improved by 
appropriately selecting both the step size and the truncation
depth \cite{BGLY14}.

The more recent work \cite{BBL23} develops an adaptive 
algorithm for effective use of the log-ODE method for 
solving \textit{Rough Differential Equations} RDEs.
This method involves a problem-dependent way of determining
how to choose the two hyper-parameters governing the log-ODE 
method; namely, the length of the time intervals to consider 
the approximating ODE over, and the depth of log signature
to consider.
There is no universal right answer for choosing these 
parameters; the appropriate choice will depend on the 
differential equation through the regularity of the vector 
field $f$, the complexity of the paths, and the level 
of accuracy one desires. 

The error representation formula derived by the authors 
in \cite{BBL23} is the main tool for understanding the 
relationship between these three aspects for the log-ODE 
method. In particular, improved a posteriori uniform 
error estimates are achieved for a class of RDE 
approximation schemes in Section 3 of \cite{BBL23}, 
and it is further established that this class includes the 
log-ODE method in Section 4 of \cite{BBL23}.

In a very loose sense these estimates can be summarised 
as follows. Provided the vector field $f$ is smooth enough, 
that the time interval over which the approximation is 
sought is short enough, and that the truncation level 
for the log signature is deep enough, the error between the 
actual solution and the approximation resulting from the 
log-ODE method behaves like a power of the $p$-variation 
norm of the path $X$.
Quantifying this statement is a far from trivial task which 
we make no effort to do; the reader seeking the rigorous 
detailed statements is directed to \cite{BBL23}.

We end this section with the following remarks. 
It follows from the error estimates of \cite{BBL23} that 
one can always guarantee arbitrarily strong estimates 
for the log-ODE method by working over 
sufficiently small time intervals (i.e. by making $t-s$ 
sufficiently small).
But it is \textit{not} true that on a fixed time interval 
(i.e. a fixed value of $t-s$) that one can obtain 
arbitrarily strong estimates for the log-ODE method
by choosing a sufficiently 
high truncation order for the log signature.

\section{Computing Signatures}
\label{coding_packages}
Making use of rough path and  signature based techniques 
only becomes feasible if one is able to explicitly compute
signatures of paths. 
Software packages for managing such computations have 
evolved over time.
Initially there were python packages for computing path 
signatures directly from time series represented as
NumPy arrays 
of doubles, see either 
\href{https://esig.readthedocs.io/en/latest/#}{\textit{esig}} 
\cite{esig_ref} originating in 
2002, or 
\href{https://github.com/bottler/iisignature/tree/master}
{\textit{iisignature}} \cite{GB18}.
These packages could compute signatures and log signatures, 
but did not support efficient algorithms for computing 
signatures over multiple intervals, or any ability to manipulate
the signature or log signature tensors, e.g. exponentiate the 
log signature, or apply the Campbell--Baker--Hausdorff product.

Esig and iisignature are tools intended to be used without 
GPU based acceleration. Differentiable and GPU based tools 
with similar functionality, suitable for deep learning, 
include \textit{Signatory} \cite{KL20} and 
\textit{signax} \cite{signax_ref}; signatory depends upon 
\href{https://pytorch.org/}{PyTorch} \cite{PyT16} 
while signax depends upon
\href{https://jax.readthedocs.io/en/latest/index.html}{JAX}
\cite{JAX_ref}.

A C++ package \textit{libalgebra}, which has been in development
since 2002 and remains in very active development, 
is a powerful and comprehensive CPU based 
library with tools for manipulating signatures, Lie elements,
tensor algebra elements, and the corresponding dual objects 
(there are implementations of shuffle and half shuffle).
This templated code has been 
optimised for floating point coefficients, but 
importantly also supports rational coefficients, arbitrary fixed
precision coefficients, and even coefficients that are 
polynomials.
The former allows one to experiment and identify the precision 
required to execute high order calculations effectively.
The latter allows one to develop formulae for complex Lie 
operations, and is also extremely valuable for debugging 
when code does not produce expected outcomes.

Esig wrapped some of the most elementary functionality 
from an early version of libalgebra.

RoughPy \cite{morley2024roughpy}
is a major upgrade on the python interface to 
managing rough streams. It allows the construction of a 
stream object from raw data; one can then query this 
object over \textit{any} interval to get the signature or
log signature of the stream over the specified interval.
It is efficient and substantially streamlines the analysis 
of streamed data. RoughPy has all the functionality of 
the C++ package libalgebra mentioned above. 
In addition, it has automated GPU acceleration through 
\href{https://developer.nvidia.com/opencl#}{OpenCl} support.
It wraps a new C++ library 
\href{https://github.com/datasig-ac-uk/libalgebra-lite}
{libalgebra-lite} 
that is based upon libalgebra but 
with a reduced use of templates to allow better interaction
with the python wrapper.
In this survey we use RoughPy.
The documentation for RoughPy may be found at 
\href{https://roughpy.org/}{roughpy.org}.
RoughPy may be installed directly via \textbf{pip} 
(a standard package-management system used to install 
and manage software packages written in Python).

In the remainder of this section we give a brief
introduction to some of the functionality of RoughPy.
A more in-depth introduction to RoughPy can be found
in \cite{morley2024roughpy}, for example.

The main object in RoughPy is the 
\href{https://roughpy.org/user/basics.streams.html}{stream} 
object. The data is stored as a stream and, for example, 
signatures and log signatures are computed by querying 
the stream over an 
\href{https://roughpy.org/user/basics.intervals.html}{interval} 
chosen by the user.
The convention in RoughPy is that intervals are taken to 
be half-open in the sense that the start point is included 
but the end point is not. 
To be more precise, the python code 
`rp.RealInterval(0.0,2.0)' would generate the interval 
$[0,2) \subset \R$.

The convention of using half-open intervals is just 
one aspect of the careful way in which RoughPy works with 
intervals. The decision to take half-open intervals is done 
to ensure that RoughPy can handle jumps in the underlying 
signal that happen at \textit{any} particular time whilst 
still guaranteeing that combining signatures over 
adjacent intervals returns the signature over the entire
interval determined by their union.

Moreover, for efficiency,
an internal representation of a stream involving
caching the signature over dyadic intervals of different 
resolutions is used. Resolution refers to the
length of the finest granularity at which information about
the underlying data is stored.
Using the cache to recover the signature over any interval
has logarithmic complexity. 
If $n$ is the internal resolution of the stream, then
recovering the signature over any interval uses, at most, 
$2n$ tensor multiplications.
Any event in the stream occurs within one of these
finest granularity intervals.
If multiple events occur within the same interval, 
RoughPy resolves to a more complex log-signature which
correctly reflects the time sequence of the events within
the interval.

No query of the stream can see finer resolution 
than the internal resolution of the stream; a query can 
only access information over intervals that are a union of 
these finest resolution granular intervals.
Hence a query over any interval is replaced by a query 
over an interval whose endpoints are shifted to be 
consistent with the granular resolution. This is 
obtained by rounding each endpoint to the contained 
end-point of the unique half-open granular interval containing 
this point.
In particular, if both the left-hand and right-hand ends of 
the interval are contained in the half-open granular interval,
the interval is rounded to the empty interval. 
Specifying a resolution of 32 or 64 equates to using 
integer arithmetic.

We first illustrate the use of 'LieIncrementStream' 
for the purpose of computing a truncated
signature of a given stream of increments.
Suppose $N , d \in \Z_{\geq 1}$ and we have a stream 
of increments $\bx \in \s \big( \R^d\big)$ given by 
$\bx = ( I_1 , \ldots , I_N )$ for 
$I_1 , \ldots , I_N \in \R^d$.
Then the python code in Figure \ref{roughpy_simple_code}
computes the signature 
of $\bx$ truncated to depth $k \in \Z_{\geq 1}$.

\begin{figure}[ht]
	\centering
		\begin{tabular}{lc}
			$>>>$ import numpy as np \\
                $>>>$ import roughpy as rp \\
			$>>>$ x = np.array([ $I_1 , \ldots , I_N$ ]) \\
			$>>>$ stream = 
                rp.LieIncrementStream.from\_increments(x , 
                depth = k) \\
                $>>>$ interval = rp.RealInterval(0,N) \\
			$>>>$ sig = stream.signature(interval, depth=k)
		\end{tabular}
	\caption{Python code to compute the signature
        of the stream $\bx$ truncated to depth $k$
	using RoughPy}
	\label{roughpy_simple_code}
\end{figure}

In Figure \ref{roughpy_simple_code} `stream' is 
a stream whose (hidden) underlying data are the $N$ 
increments $I_1, \ldots , I_N \in \R^d$ and whose algebra
elements are truncated to a maximum depth $k$.
By default the increments are assumed to occur at 
parameter values equal to their row index in the 
provided data.
Thus here $I_1$ is assumed to occur at $0$, 
$I_2$ is assumed to occur at $1$, and so on until 
$I_N$ is assumed to occur at $N-1$.
Consequently, all parameters are contained within
the interval $[0,N)$ that is considered in Figure 
\ref{roughpy_simple_code} via the code
`interval = rp.RealInterval(0,N)'.

Instead of manually providing a depth when creating 
the stream 'stream', we could provide a 
\href{https://roughpy.org/user/basics.contexts.html}{context}.
A context is used to tell RoughPy the desired shape
that streams should be assumed to have. 
Thus the same output achieved in Figure
\ref{roughpy_simple_code}
could be obtained using the python code in
Figure \ref{roughpy_context_example}.

\begin{figure}[ht]
    \centering
    \begin{tabular}{lc}
			$>>>$ import numpy as np \\
                $>>>$ import roughpy as rp \\
			$>>>$ x = np.array([ $I_1 , \ldots , I_N$ ]) \\
                $>>>$ context = rp.get\_context(width = d, 
                depth = k, coeffs = rp.DPReal) \\
			$>>>$ stream = 
                rp.LieIncrementStream.from\_increments(x , 
                ctx = context) \\
                $>>>$ interval = rp.RealInterval(0,N) \\
			$>>>$ sig = stream.signature(interval, 
                ctx = context)
		\end{tabular}
    \caption{Python code to compute the signature of the
    stream $\bx$ truncated to depth $k$ using RoughPy 
    via a prescribed context.}
    \label{roughpy_context_example}
\end{figure}

The 'context' defined in Figure \ref{roughpy_context_example}
has width $d$ (since each entry in the stream $\bx$ is an 
element in $\R^d$), depth $k$ (since we want to truncate 
to depth $k$), and real coefficients (since we are working 
over the reals).

The output 'sig' in both Figures \ref{roughpy_simple_code} 
and \ref{roughpy_context_example} is a 
\href{https://roughpy.org/user/basics.free_tensors.html}
{FreeTensor} of width $d$, depth $k$, and ctype 'DPReal'.
Moreover, when printed, 'sig' returns both the values of
the coefficients of the truncated signature and the basis 
element to which each coefficient corresponds. 
We illustrate this via the following explicit example.
Fix $d = 2$, $k=3$, and consider the stream of increments
$\bx = ( (1,1) ) \in \s\big(\R^2\big)$.
Then we compute the signature of this stream $\bx$ 
truncated to depth $3$ using the python code in
Figure \ref{roughpy_signature_explicit_example}.

\begin{figure}[ht]
	\centering
		\begin{tabular}{lc}
			$>>>$ import numpy as np \\
			$>>>$ import roughpy as rp \\
                $>>>$ x = np.array([ [1.0,1.0] ]) \\
			$>>>$ context = rp.get\_context(width=2,depth=3,
                coeffs = rp.DPReal) \\
                $>>>$ stream = rp.LieIncrementStream.
                from\_increments(x, ctx = context) \\
			$>>>$ interval = rp.RealInterval(0,1) \\
			$>>>$ sig = stream.signature(interval,
                ctx = context) \\
                $>>>$ sig \\
                FreeTensor(width=2, depth=3, ctype=DPReal) \\
                $>>>$ print(sig) \\
                $\{$ 1() 1(1) 1(2) 0.5(1,1) 0.5(1,2) 0.5(2,1) 
                0.5(2,2) 0.166667(1,1,1) \\
                0.166667(1,1,2) 0.166667(1,2,1) 
                0.166667(1,2,2) 0.166667(2,1,1) \\ 
                0.166667(2,1,2) 0.166667(2,2,1) 
                0.166667(2,2,2) $\}$
		\end{tabular}
	\caption{Python code to compute signature of stream 
        $\bx \in \s\big(\R^2\big)$ truncated to depth $3$.
        Including the returned output of the commands 
        'sig' and 'print(sig)'.}
	\label{roughpy_signature_explicit_example}
\end{figure}

As expected, 'sig' is a FreeTensor of width $2$, depth $3$, 
and ctype DPReal.
The output returned by the command 'print(sig)' is the
collection of coefficients determining the signature 
truncated to depth $3$ along with the word (in brackets) 
determining which coordinate iterated integral each coefficient
represents.
For example, the entry $0.5(1,2)$ means that the value 
of the coordinate iterated integral corresponding to the 
word $(1,2)$ is $0.5$.
We recall from Subsection \ref{sig_p<2_sec} that in 
this simple setting this is just the value of the 
iterated integral $\int_0^1 \int_0^y dx dy$ where we 
have associated the canonical $x$ and $y$ coordinates of 
$\R^2$ with '1' and '2' respectively.

If one was instead interested in the log signature of the 
stream $\bx \in \s\big(\R^2\big)$, inserting the line of code
\begin{equation}
    \label{roughpy_lsigA}
        >>> \text{logsigA = 
        stream.log\_signature(interval,ctx=context)}
\end{equation}
into the code presented in Figure 
\ref{roughpy_signature_explicit_example}
would return the log signature truncated to depth $3$.
We could also compute the log signature by 
directly taking the logarithm of the already computed 
signature. That is, we could use the python code
\begin{equation}
    \label{roughpy_lsigB}
        >>> \text{logsigB = sig.log()}
\end{equation}
to compute the desired log signature.
However, there is a distinction between the outputs 
logsigA and logsigB of 
the approaches \eqref{roughpy_lsigA} 
and \eqref{roughpy_lsigB} respectively.
The respective outputs are
illustrated in Figure \ref{roughpy_log_sig_differences}.

\begin{figure}[ht]
    \centering
    \begin{tabular}{lc}
        $>>>$ logsigA \\
        Lie(width=2, depth=3, ctype=DPReal) \\
        $>>>$ print(logsigA) \\
        $\{$ 1(1) 1(2) $\}$ \\
        $>>>$ logsigB \\
        FreeTensor(width=2, depth=3, ctype=DPReal) \\
        $>>>$ print(logsigB) \\
        $\{$ 1(1) 1(2) -2.77556e-17(1,1,1) -2.77556e-17(1,1,2) \\
        -2.77556e-17(1,2,1) -2.77556e-17(1,2,2) 
        -2.77556e-17(2,1,1) \\ 
        -2.77556e-17(2,1,2) 
        -2.77556e-17(2,2,1) -2.77556e-17(2,2,2) $\}$
    \end{tabular}
    \caption{The outputs resulting from the commands 
    'logsigA', 'print(logsigA)', 'logsigB' and 'print(logsigB)' 
    where logsigA and logsigB are the results of the python 
    code in \eqref{roughpy_lsigA} and \eqref{roughpy_lsigB} 
    respectively.}
    \label{roughpy_log_sig_differences}
\end{figure}

The first difference is that logsigA is a
\href{https://roughpy.org/user/basics.lies.html}{Lie}
element whilst logsigB is a FreeTensor.
The essential difference is that a Lie element is in 
the Free Lie algebra generated by '1' and '2' and recorded
with respect to a \textit{Hall basis},
whilst a FreeTensor is in the Tensor algebra 
(cf. Subsection \ref{Ten_Alg}) recorded with respect to 
the basis generated by all string combinations of '1' and '2'.

The second difference is a result of numerical errors 
being introduced due to truncation. Instead of taking log 
of the full signature tensor, we are only taking log of 
the signature truncated to depth 3 to compute logsigB.
Consequently numerical errors enter the computations. 
In particular, the non-zero coefficients for the basis 
elements involving three entries from $\{1,2\}$ in 
Figure \ref{roughpy_log_sig_differences}, all 
of magnitude $10^{-17}$, should all be zero. 

We can also go from log signatures to signatures
by exponentiation of a log signature. However, in order to 
do so we must first convert the log signature from a 
Lie element to a FreeTensor, and then exponentiate.
As an example, consider a stream of increments 
$\bx = ( (1,0,0) , (0,1,0) , (0,0,1) ) \in \s\big(\R^3\big)$.
Then the following python code computes the signature of 
$\bx$ truncated to depth $2$ both directly and by 
taking the exponential of the log signature.

\begin{figure}[ht]
\centering
    \begin{tabular}{lc}
         $>>>$ import numpy as np \\
        $>>>$ import roughpy as rp \\
        $>>>$ x = np.array([ [1.0,0.0,0.0] , 
        [0.0,1.0,0.0] , [0.0,0.0,1.0] ]) \\
	$>>>$ context = rp.get\_context(width=3,depth=2,
        coeffs = rp.DPReal) \\
        $>>>$ stream = rp.LieIncrementStream.
        from\_increments(x, ctx = context) \\
	$>>>$ interval = rp.RealInterval(0,1) \\
	$>>>$ sigA = stream.signature(interval,
        ctx = context) \\
        $>>>$ sigA \\
        FreeTensor(width=3, depth=2, ctype=DPReal) \\
        $>>>$ print(sigA) \\
        $\{$ 1() 1(1) 1(2) 1(3) 0.5(1,1) 1(1,2) 1(1,3) 
        0.5(2,2) 1(2,3) 0.5(3,3) $\}$ \\
        $>>>$ logsig = stream.log\_signature(interval, 
        ctx = context) \\
        $>>>$ logsigT = context.lie\_to\_tensor(logsig) \\
        $>>>$ sigB = logsigT.exp() \\
        $>>>$ sigB \\
        FreeTensor(width=3, depth=2, ctype=DPReal) \\
        $>>>$ print(sigB) \\
        $\{$ 1() 1(1) 1(2) 1(3) 0.5(1,1) 1(1,2) 1(1,3) 
        0.5(2,2) 1(2,3) 0.5(3,3) $\}$
    \end{tabular}
\caption{Python code to compute signature of stream 
    $\bx \in \s(\R^3)$ truncated to depth $2$ both 
    directly and by computing the exponential of the 
    log signature.}
\label{roughpy_sig_from_logsig}
\end{figure}

As illustrated in Figure \ref{roughpy_sig_from_logsig},
both methods of computing the signature result in the same
output for this example.
It is worth remarking that the 'lie\_to\_tensor' function
requires a context to inform it about the shape of the 
input and output.

We can now provide the full illustration of the signatures
ability to record the order of events as commented on in 
Subsection \ref{subsec:inc_stream_data}.
For this purpose we recall the toy problem considered in 
Subsection \ref{subsec:inc_stream_data}.
Let $N \in \Z_{\geq 1}$ and suppose we have a
finite collection of time series, with each time series 
consisting of $2$ channels and $N$ samples.
Suppose that each time series $\bx$ has the following
structure. 
If $\bx = \big\{ (x_{1,i} , x_{2,i}) \big\}_{i=1}^N$ then the 
following properties are true.
\begin{itemize}
    \item For every $i \in \{1, \ldots , N \}$ we have 
    $x_{1,i} , x_{2,i} \in \{0,1\}$.
    \item For $j \in \{1,2\}$ if $i \in \{1, \ldots , N\}$ 
    and $x_{j,i} = 1$, then for every $k \in \{i , \ldots , N\}$
    we have $x_{j,k} = 1$.
    \item At least one of $x_{1,N}$ and $x_{2,N}$
    is equal to $1$.
\end{itemize}
Consider the task of
determining which channel is the first to change from $0$ 
to $1$.

Recall from Subsection \ref{dataset_to_path} that viewing 
each time series $\bx$ as an element in $\R^{2N}$ and 
trying to learn a quadratic polynomial capable of determining
which channel is the first to change from $0$ to $1$ involves
considering $1 + 3N + 2N^2$ linearly independent quadratic 
polynomials. If we consider $N=3600$ then we end up 
dealing with $25930801$ linearly independent 
quadratic polynomials. 

But computing the depth 2 signature will determine which 
channel changes first.
Observe that there are only two possible options. 
Either the stream values go in the order 
$(0,0) \to (1,0) \to (1,1)$ or the order
$(0,0) \to (0,1) \to (1,1)$.
Hence the channels of the increments are either 
equivalent to the stream 
$\bx_1 := ( (1,0) , (0,1) ) \in \s\big(\R^2\big)$
or to the stream 
$\bx_2 := ( (0,1) , (1,0) ) \in \s\big(\R^2\big)$.
The RoughPy output when the signature of $\bx_1$ truncated
to depth 2 is computed is
\begin{equation}
    \label{eq:roughpy_output_x1}
        \big\{ 1() \quad 1(1) \quad 1(2) \quad 0.5(1,1)
        \quad 1(1,2) \quad 0.5(2,2) \big\}.
\end{equation}
The RoughPy output when the signature of $\bx_1$ truncated
to depth 2 is computed is
\begin{equation}
    \label{eq:roughpy_output_x2}
        \big\{ 1() \quad 1(1) \quad 1(2) \quad 0.5(1,1)
        \quad 1(2,1) \quad 0.5(2,2) \big\}.
\end{equation}
We immediately see that the channel that is first to 
change from $0$ to $1$ is captured by the depth $2$ 
coordinate iterated integrals
corresponding to the words $(1,2)$ and $(2,1)$.
If the $x$-coordinate is the first to change from $0$ to 
$1$ then \eqref{eq:roughpy_output_x1} illustrates that 
the resulting signature has a coefficient of 
$1$ corresponding to the word $(1,2)$ and a coefficient of 
$0$ corresponding to the word $(2,1)$.
If the $y$-coordinate is the first to change from $0$ to 
$1$ then \eqref{eq:roughpy_output_x2} illustrates that
the resulting signature has a coefficient of 
$0$ corresponding to the word $(1,2)$ and a coefficient of 
$1$ corresponding to the word $(2,1)$.
Consequently the $7$ real numbers forming the coefficients 
of the truncation of the signature to depth $2$ are sufficient
to determine the order.
This is \textit{independent} of the number of samples $N$, 
and dealing with $7$ terms is simpler than dealing with
$1 + 3N + 2N^2$ quadratic polynomials.

It is possible to construct streams by providing the 
raw data of Lie increments with higher order terms by 
specifying the width.
For example, consider the stream of increments 
$\bx = \big( (1,1,1) , (1,-3,4), (0,1,2) \big)
\in \s\big(\R^3\big)$.
Consider taking width 2 and depth 2 so that the elements of a 
Lie element will have keys $(1,2,[1,2])$.
Then the python code in
Figure \ref{roughpy_prescribe_higher_order_terms_1}
will construct a stream from 
$\bx$ whose underlying Lie increments are width 2, depth 2.

\begin{figure}[ht]
	\centering
		\begin{tabular}{lc}
			$>>>$ import numpy as np \\
			$>>>$ import roughpy as rp \\
                $>>>$ x = np.array([ [1.0,1.0,1.0] , 
                [1.0,-3.0,4.0] , [0.0,1.0,2.0] ]) \\
			$>>>$ context = rp.get\_context(width=2,depth=2,
                coeffs = rp.DPReal) \\
                $>>>$ stream = rp.LieIncrementStream.
                from\_increments(x, ctx = context) \\
			$>>>$ interval1 = rp.RealInterval(0,1) \\
                $>>>$ interval2 = rp.RealInterval(0,2) \\
                $>>>$ interval3 = rp.RealInterval(0,3) \\
			$>>>$ logsig1 = stream.log\_signature(interval1,
                ctx = context) \\
                $>>>$ logsig2 = stream.log\_signature(interval2,
                ctx = context) \\
                $>>>$ logsig3 = stream.log\_signature(interval3,
                ctx = context) \\
                $>>>$ print(logsig1) \\
                $\{$ 1(1) 1(2) 1([1,2]) $\}$ \\
                $>>>$ print(logsig2) \\
                $\{$ 2(1) -2(2) 3([1,2]) $\}$ \\
                $>>>$ print(logsig3) \\
                $\{$ 2(1) -1(2) 6([1,2]) $\}$
		\end{tabular}
	\caption{Python code to use the stream 
        $\bx \in \s\big(\R^3\big)$ to prescribe the depth 2 
        information. Includes outputs 
        resulting from computing and printing the 
        log signature over a range of intervals.}
	\label{roughpy_prescribe_higher_order_terms_1}
\end{figure}

As a final example, consider the stream
$\by = \big\{ (1,1,2.5,0.75,-1.5) , 
(1.2,3.4,-2.5,0.98,0.4) \big\} \in \s\big(\R^5\big)$.
Suppose we want to determine a Lie element of width 2 and 
depth 3 from the stream $\by$, which will have keys 
$(1,2,[1,2],[1,[1,2]],[2,[1,2]])$.
Then the python code in
Figure \ref{roughpy_prescribe_higher_order_terms_2}
will construct a stream from 
$\by$ whose underlying Lie increments are width 2, depth 3.

\begin{figure}[ht]
	\centering
		\begin{tabular}{lc}
			$>>>$ import numpy as np \\
			$>>>$ import roughpy as rp \\
                $>>>$ y = np.array([ [1.0,1.0,2.5,0.75,-1.5] , 
                [1.2,3.4,-2.5,0.98,0.4] ]) \\
			$>>>$ context = rp.get\_context(width=2,depth=3,
                coeffs = rp.DPReal) \\
                $>>>$ stream = rp.LieIncrementStream.
                from\_increments(x, ctx = context) \\
			$>>>$ interval1 = rp.RealInterval(0,1) \\
                $>>>$ interval2 = rp.RealInterval(0,2) \\
			$>>>$ logsig1 = stream.log\_signature(interval1,
                ctx = context) \\
                $>>>$ logsig2 = stream.log\_signature(interval2,
                ctx = context) \\
                $>>>$ print(logsig1) \\
                $\{$ 1(1) 1(2) 2.5([1,2]) -0.75([1,[1,2]]) 
                -1.5([2,[1,2]])$\}$ \\
                $>>>$ print(logsig2) \\
                $\{$ 2.2(1) 4.4(2) 1.1([1,2]) 
                -2.55667([1,[1,2]]) -7.04([2,[1,2]]) $\}$ 
		\end{tabular}
	\caption{Python code to use the stream 
        $\by \in \s\big(\R^5\big)$ to prescribe the depth 3
        information. Includes outputs 
        resulting from computing and printing the 
        log signature over a range of intervals.}
	\label{roughpy_prescribe_higher_order_terms_2}
\end{figure}

It is both informative and instructive to work through
some examples of using signatures within a relatively 
simple machine learning context. There are a number of 
demonstration notebooks available via the 
\href{https://www.datasig.ac.uk}{DataSig} website. The
\href{https://github.com/pafoster/path_signatures_introduction}
{\textit{Introduction to Path Signatures}} python notebook
is designed to familiarise the user with the key concepts. 
Primarily aimed at data scientists and machine learning 
practitioners, it uses simple illustrative examples to 
explore the behaviour of path signatures and of the affect 
that various stream transformations (including those covered
in Section \ref{dataset_to_path}) have on the signature of 
a stream.

The 
\href{https://github.com/datasig-ac-uk/signature_applications/tree/master/mnist_classification}
{\textit{Handwritten Digit Classification}} python notebook 
demonstrates the use of path signatures (via esig) 
for handwritten digit classification. 
Given sequences of pen strokes contained in the 
\href{https://edwin-de-jong.github.io/blog/mnist-sequence-data/}
{MNIST sequence dataset}, 
corresponding to the MNIST handwritten digit data set by 
Edwin D. De Jong, path signatures of transformations of 
these streams are computed and incorporated as features 
into a linear classifier. The notebook additionally 
explores combining the features with unsupervised 
classification techniques.

\section{Expected Signature}
\label{expect_sig}
The signature provides a summary of a path that is adequate 
for predicting its effect on a broad range of different 
systems whilst managing to discard irrelevant information.
Frequently we are interested in understanding more than a 
single path at a time; that is, we want to adequately 
summarise a collection of paths in order to predict their
collective effect on systems. The paths traced out by 
multiple points on an object will give more information 
about the objects trajectory and motion than a single point. 
The effect of Covid-19 restrictions on the reproduction (R) 
number of the virus are more accurately predicted knowing 
the levels of compliance of numerous people across
several locations, as opposed to relying on the level of 
compliance of a single individual.

A distribution can summarise a collection of paths; 
for example, if $M \in \Z_{\geq 1}$ then a finite collection 
$\{ \gamma_1 , \ldots , \gamma_M \} 
\subset \cv^1 ([a,b] , V)$ can be 
summarised by the empirical measure
$\de := \frac{1}{M} \sum_{j=1}^M \de_{\gamma_j}$.
We would like to summarise distributions in an analogous 
manner to how the signature summarises paths.
That is, loosely speaking, compress the information 
whilst retaining enough detail to distinguish between different 
distributions.
For paths the signature achieves this by filtering out the 
parameterisation noise (cf. Section \ref{good_props}).
Typically, summarising distributions requires 
requires restricting the class of distribution 
considered by imposing additional constraints (cf. Brownian 
Motion, Markov processes etc). 
It turns out that the signature allows one to summarise 
\textit{all} distributions without restriction.

Let $p \in \R_{\geq 1}$ and let 
$\p \Omega_p(V)$ denote the space of (Borel) probability
measures on the space of $p$-rough paths $\Omega_p(V)$.
Then the \textit{Expected Signature} is the map 
$\bS_p : \p \Omega_p(V) \to T((V))$ defined by
\begin{equation}
    \label{expect_sig_p_def}
        \bS_p(\mu) :=
        \E_{\mu} \big[ S(X) \big]
        =
        \prod_{n=0}^{\infty} 
        \E_{\mu} \big[ S^n(X) \big].
\end{equation}
The expected signature defined in \eqref{expect_sig_p_def}
allows us to define the expected signature of a stochastic 
process.

For this purpose, suppose that $X_t$ is a stochastic process 
on $V$ for $t \in [0,1]$ under a probability space $(\n,\bbP)$.
First suppose that $p \in [1,2)$ and that
for $\bbP$-almost every $\omega \in \n$ 
the path $[0,1] \to V$ defined by $s \mapsto X_s(\omega)$
is in $\cv^p([0,1],V)$.
Let $Q : \n \to \cv^p ( [0,1] , V)$ be the map taking 
$\omega \in \n$ to the path 
$\big(s \mapsto X_s(\omega) \big)_{s \in [0,1]}$.
The push-forward $Q_{\ast} \bbP$ of the measure 
$\bbP$ is an element in $\p \cv^p([0,1],V)$.
Recalling from Subsection \ref{rough_paths_sec} 
that $p \in [1,2)$ means that $p$-rough paths in 
$\Omega_p(V)$ are the signatures of paths in 
$\cv^p([0,1],V)$, we may view $Q_{\ast} \bbP$ as an  
element in $\p \Omega_p(V)$.
Thus we can make sense of the 
\textit{expected signature of $X_t$} as
$\bS_p \big( Q_{\ast} \bbP \big)$, with it being common to 
abuse notation and write $\bS_p(X)$.
When the value of $p \in [1,2)$ is either clear from the 
context or not of particular importance, it is common 
to drop it from the notation and write $\bS(X)$ 
for the expected signature of the process $X_t$.

If the sample paths of the stochastic process are only 
of finite $p$-variation for $p \geq 2$ then making sense of
the expected signature of the process is more involved.
Suppose that this is now the case, and assume that for 
$\bbP$-almost every $\omega \in \n$ there are, 
for $i \in \{2 , \ldots , \lfloor p \rfloor \}$,
functions
$A[i](\omega) : \Delta_{[a,b]} \to V^{\otimes i}$ such that 
the functional 
$\bX(\omega) : \Delta_{[a,b]} \to T^{(\lfloor p \rfloor)}(V)$
defined for $(s,t) \in \Delta_{[a,b]}$ by
\begin{equation}
    \label{process_p_rough_path_func}
        \bX_{s,t}(\omega) := 
        \Big( 1 , X_t(\omega) - X_s(\omega) , 
        A_{s,t}[2](\omega) , \ldots , 
        A_{s,t}[\lfloor p \rfloor] (\omega) \Big)
        \in
        T^{(\lfloor p \rfloor)}(V)
\end{equation}
satisfies that $\bX(\omega) \in \Omega_p(V)$.
Let $Q : \n \to \Omega_p(V)$ be the map taking 
$\omega \in \n$ to the $p$-rough path 
$\bX(\omega) \in \Omega_p(V)$.
The push-forward $Q_{\ast}\bbP$ of the measure $\bbP$ 
is an element in $\p \Omega_p(V)$.
Consequently we can make sense of the 
\textit{expected signature of $X_t$} as 
$\bS_p(Q_{\ast}\bbP)$ with it being common to abuse 
notation and write $\bS_p(X)$.
When the value of $p \geq 2$ is either clear from the 
context or not of particular importance, it is common 
to drop it from the notation and write $\bS(X)$ 
for the expected signature of the process $X_t$.

In this latter case we observe that the increments of 
the process $X_t$ are no longer sufficient to determine 
its expected signature. Instead, we are required to 
additionally prescribe its higher order properties such
as its area.
A standard Brownian motion $B : [0,T] \to \R^d$ 
for some $T > 0$ provides an instructive example.
Given a $p \in (2,3)$ it is known that the process
$t \mapsto B_t$ possesses finite $p$-variation.
Consequently, in order to define its expected signature 
we must first make a choice of depth $2$ term; that is, 
we need to prescribe a suitable function 
$\Delta_{[0,T]} \to T^{(2)}(\R^d)$.

Both the It\^{o} $2$-multiplicative functional 
(cf. \eqref{Ito_mult_func})
and the canonical Brownian rough path 
(cf. \eqref{Strato_func_def})
defined in Subsection \ref{rough_paths_sec} 
provide choices of depth $2$ terms that result in 
distinct $p$-rough paths whose depth $1$ terms are 
given by the increments $B_t - B_s$ of the Brownian motion.
The resulting expected signatures are both candidates 
for the expected signature of the Brownian motion. 
Without making a choice of depth $2$ term there is no 
canonical way to choose which one should be taken to be 
the expected signature of the Brownian motion $B_t$.

The expected signature of a process $X_t$ is 
the natural generalisation of the moments of the 
process $X_t$. 
Moreover, in \cite{BO19}, it is shown 
that the \textit{Log Expected Signature} $\log \bS (X)$ 
provides the natural generalisation of the cumulants of 
$X_t$.

Beginning with the work of Fawcett in his thesis 
\cite{Faw03}, there has been a large number of works 
investigating the relationship between the expected 
signature $\bS(X)$ and the law of $S(X)$.
Fawcett originally proved that if the measure 
$Q_{\ast} \bbP$ is compactly supported, then the law of 
$S(X)$ is uniquely determined by $\bS(X)$.
Subsequently, it has been established that if the radius 
of convergence of the power series 
$\sum_{n=0}^{\infty} z^n \E \lvert\lvert S^n(X) 
\rvert\rvert$
is infinite, then the expected signature uniquely 
determines the law of $S(X)$ \cite{Che13}.
Further refinements to the extent to which the expected 
signature of a stochastic process $X_t$ determines the 
law of $X_t$ may be found in \cite{CL13,CO18}.

For Brownian motion $B_t$ with L\'{e}vy area 
(i.e. the canonical Brownian rough path defined in 
Subsection \ref{rough_paths_sec}; cf. \eqref{Strato_func_def}) 
on a bounded $C^1$ domain $\Omega \subset \R^d$, the 
investigation of the finiteness of the radius of 
convergence for the power series 
$\sum_{n=0}^{\infty} z^n \E \lvert\lvert S^n(B_t) 
\rvert\rvert$
was initiated in \cite{LN11}, where the authors 
established that the radius of convergence was 
strictly positive. In \cite{CL13} it is proven that 
this radius of convergence is infinite, provided $B_t$ 
is considered up to a fixed finite time $T > 0$. 
Consequently, for such Brownian motions, the expected 
signature determines the law. But recently 
it has been established that this is \textit{not} true 
for the exit time of Brownian motion in two-dimensions.
More concretely, if $B^z_t$ is the standard Brownian motion 
in $\R^2$ started at $z \in \B^2(0,1)$ and stopped at the 
first time it hits the boundary $\partial \B^2(0,1)$, then 
radius of convergence of the corresponding power series is 
finite; see \cite{BDMN19} for full details. 

Of greater interest from the machine learning perspective 
is that the expected signature provides a systematic way 
to describe probability measures on paths in 
terms of their effects. An early breakthrough was the 
realisation that complex path measures (such as the Wiener 
measure) can be effectively approximated 
by a measure supported on finitely many paths having 
the same expected signature on a truncated tensor algebra 
$T^{(n)}(V)$ \cite{LV04, LL11, LY13}. Unlike the 
\textit{Martingale Problem} approach originating in 
\cite{SV79}, there is no restriction on the probability 
measures considered. 

A result from \cite{CO18} establishes that the expected 
signature essentially uniquely determines distributions on 
compact subsets of $\p \cv^1 ([a,b],V)$. It is proven 
in Theorem 5.6 of \cite{CO18} that when restricted to 
compact subsets, the expected signature is
injective up-to tree-like equivalence (see Appendix B of 
\cite{CO18} for full details). The important point from 
our perspective is that if the set of
tree-like paths is removed from $\cv^1 ([a,b],V)$, 
then the expected signature is injective on compact 
subsets of $\p \cv^1 ([a,b],V)$.
Let $\mathcal{C} ([a,b],V) \subset \cv^1 ([a,b],V)$ 
denote the subset resulting from removing all tree-like 
paths.

The work \cite{BDLLS20} establishes a universality property
for the expected signature analogous to the universality 
property of the signature itself 
(cf. Theorem \ref{uni_nonlin}). 
For the signature, this universality is that, on compact 
subsets of path-space, the components of the signature 
(i.e. the coordinate iterated integrals) span the 
set of continuous functions.
An almost identical result turns out to be true for the 
space $C^0 \big( \p \ck \big)$, where 
$\ck \subset \mathcal{C} ([a,b],V)$ is compact and
$\p \ck$ denotes the space of Borel probability measures
on $\ck$,
provided we first lift the expected signature map so that
it maps a probability measure $\mu$ to a path
$[a,b] \to T((V))$, rather than a single element of $T((V))$.

Before introducing the lift of the expected signature map 
to the so-called \textit{Pathwise Expected Signature}
$\Phi_{\Path} : \p \mathcal{C} ([a,b],V) \to 
\cv^1 \big( [a,b] , T((V)) \big)$, we first fix a choice
of topology on the space of Borel probability measures
$\p \ck$ on a compact subset
$\ck \subset \mathcal{C} ([a,b],V)$.
For this purpose we consider the following notion of 
weak convergence in $\p \ck$. 

Let $C^0(\ck)$ denote the space of real-valued 
continuous functions $\ck \to \R$.
Then any $\nu \in \p \ck$ can be viewed as a bounded
linear functional $C^0(\ck) \to \R$ by defining 
$\nu[\vph] := \int_{\ck} \vph(z) d\nu(z)$
for $\vph \in C^0(\ck)$. 
Then a sequence $\{ \mu_n \}_{n=1}^{\infty} \subset \p \ck$
is said to weakly converge to $\mu \in \p \ck$ if 
for every $f \in C^0(\ck)$ we have that 
$\mu_n[f] \to \mu[f]$ as $n \to \infty$.
It is established in \cite{BDLLS20} (cf. Theorem A.1)
that the expected signature determines a weakly continuous
map $\p \ck \to T((V))$ where $T((V))$ is equipped 
with the topology induced by 
$\big< \cdot , \cdot \big>_{T((V))}$
introduced in Subsection \ref{Ten_Alg}.

The \textit{Pathwise Expected Signature} is the map 
$\Phi_{\Path} : \p \mathcal{C} ([a,b],V) \to 
\cv^1 \big( [a,b] , T((V)) \big)$
defined by
\begin{equation}
	\label{pathwise_expected_sig}
		\Phi_{\Path} (\mu)_t := 
		\E_{\mu} \big[ S_{a,t}(X) \big] 
		\qquad \text{so that} \qquad
		\Phi_{\Path} (\mu)_b = \bS (\mu).
\end{equation}
It is established in \cite{BDLLS20} (cf. Theorem A.4) 
that if $\ck \subset \mathcal{C} ([a,b],V)$ is compact, then 
the pathwise expected signature determines a 
weakly continuous map $\p\ck \to \cv^1 ( [a,b] , T((V)))$.

The coordinate iterated integrals forming the components of 
$S_{a,b}\big( \Phi_{\Path} (\mu ) \big)$ then span 
$C^0 \big( \p \ck \big)$ 
for compact subsets $\ck \subset \mathcal{C} ([a,b],V)$.
That is, any weakly continuous function on 
$\p \ck$ can be uniformly well-approximated by a linear
combination of the terms in the signature of the 
pathwise expected signature.

\begin{theorem}[Expected Signature Universality; 
Theorem 3.2 in \cite{BDLLS20}]
\label{dist_uni_nonlin}
Let $ \mathcal{C} ([a,b],V) \subset \cv^1 ([a,b],V)$ 
denote the subset resulting from removing all tree-like 
paths.
Suppose $\ck \subset \mathcal{C} ([a,b],V)$ 
is compact and consider a weakly continuous function 
$F : \p \ck \to \R$.
Then for any $\ep > 0$ there exists a truncation level 
$m \in \N$ and real coefficients 
$\big\{ \al_k(\bJ) \mid k \in \{0, \ldots , m\} 
\text{ and } \bJ \in \{1, \ldots , d \}^k \big\} \subset \R$
such that for any $\mu \in \p \ck$ we have
\begin{equation}
	\label{dist_uni_approx} 
		\Bigg| F ( \mu) - \sum_{k=0}^m
		\sum_{\bJ \in \{ 1 , \ldots , d\}^k }
		\al_k(\bJ ) S_{a,b}
		\left(\Phi_{\Path}(\mu)\right)^{\bJ} 
		\Bigg| < \ep.
\end{equation}
\end{theorem}
\vskip 4pt
\noindent
The components $S^n_{a,b}\big(\Phi_{\Path} (\mu)\big)$ 
have the factorial decay associated with all signature 
components (cf. Theorem \ref{fac_decay_thm}).
Consequently, via the same reasoning used for the signature 
of a path in Section \ref{good_props}, the pathwise expected 
signature gives a universal feature map
providing a graded set of feature functions 
(the coordinate iterated integrals of 
$S \big( \Phi_{\Path} (\mu) \big)$) that are detailed 
enough to essentially uniquely determine the distribution 
$\mu$. Again the decay of the coordinate iterated 
integrals tells us that the more easily computed lower order 
terms will typically be more informative than the higher 
order ones. The pathwise expected signature provides a 
route to tackling distribution regression. This is elaborated
on in Section \ref{dist_reg_expect_sig} of this article.

The
\href{https://github.com/datasig-ac-uk/signature_applications/blob/master/drone_identification/drone_identification.ipynb}
{\textit{Drone Identification}} python notebook
is an instructive example of using expected path signatures 
in a classification task that may be worked through by the 
reader. In the notebook a classifier for 
identifying drones is constructed based on the following 
assumption. When a radio pulse is reflected off a drone, 
the reflected signal received back by the observer is a 
combination of the reflection caused by the drone's body 
and the reflection caused by the drone's propeller.
The expected path signature is used to characterise the 
random behaviour in reflected signals. Estimates of expected
path signatures are used as feature vectors for the task of
distinguishing between drone and non-drone objects. 
The task of predicting the number of rotations
per minute (rpm) of a drone's propeller is additionally 
considered.

\section{Truncation Order Selection}
\label{Trunc_order}
Path signatures live in the 
infinite-dimensional tensor algebra $T((V))$. 
Consequently, applications seemingly require the 
selection of some finite sub-collection of terms.
The factorial decay of the components of the signature 
(cf. Theorem \ref{fac_decay_thm}) means it is common to 
choose the first $N$ terms since these will typically 
be the largest. Recall that the first $N$ terms of a 
signature $S(x)$ are called the 
\textit{truncated signature of depth $N$} and denoted by 
$S^{(N)}(x) := \Pi_N ( S(x))$. 

From a theoretical point of view this is not an issue.
Indeed, on compact subsets, we know that provided 
a sufficiently large truncation depth $N$ is chosen, 
we are able to approximate continuous functions by 
linear combinations of the coordinate iterated integrals 
of the truncated to depth $N$ signatures (cf. Theorem 
\ref{uni_nonlin}).
However, from a practical perspective there is a potential
problem.

The theoretical guarantees provide no upper bound on 
the truncation depth required. 
In particular, a large value of $N$ may be required to 
adequately capture continuous functions with 
non-trivial dependence on higher order signature terms.
This can become problematic due to the exponential growth 
in the number of components in each term of a signature 
with respect to the depth.
For example, suppose $V$ is $d$-dimensional for some integer 
$d \in \Z_{\geq 2}$. Then given 
an integer $k \in \Z_{\geq 1}$, the depth $k$ term of a 
signature of a path in $V$ consists of $d^k$ components.
Consequently, the truncation of the signature to depth 
$N$ involves 
$$1 + d + \ldots + d^N = \frac{d^{N+1} - 1}{d - 1}$$
which quickly becomes intractable for moderately sized 
values of $d$ and $N$.
Despite signatures compressing the information of a path 
down to a countable graded collection of statistics, 
further compression is required to enable directly 
learning a suitable linear combination. 

Introducing a suitable augmentation of the dataset 
\textit{before} taking the signature offers a solution to 
this problem. Recall (cf. Section \ref{dataset_to_path}) 
that by augmenting the dataset $\Omega_{\s(V)}$, we mean 
considering a map $\Theta : V \to W$ for some 
Banach space $W$. 
Any stream $\bx \in \s(V)$ can be transformed to be a
stream in $\s(W)$ via the element-wise application of 
the map $\Theta$.
Hence we can work with the augmented dataset 
$\Omega_{\s(W)} \subset \s(W)$ resulting from using $\Theta$
to transform each stream in $\Omega_{\s(V)}$ to a stream 
in $\s(W)$.

If we take 
$W := V \times U$ and $\Theta (v) := ( v , \varphi(v))$ 
for a suitable $\varphi : V \to U$, 
the information from the higher order terms of the signature 
of the stream $\Omega_V$ is contained within the 
signature of the stream $\Theta \left( \Omega_V \right)$ 
truncated to a manageable depth.
A common situation is that the Banach space $V$ is 
$\R^d$ for some $d \in \Z_{\geq 1}$, and that the Banach
space $U$ is chosen to be $\R^e$ for some $e \in \Z_{\geq 1}$ 
so that $W = \R^{d+e}$.

We have already seen several explicit examples of possible 
augmentations in Section \ref{dataset_to_path}, and each 
augmentation offers different benefits.
A Lead-Lag augmentation aides the study of a streams 
variance via its signature (see \cite{CK16}), the 
invisibility-reset augmentation (cf. \eqref{inv_reset}) 
ensures that the signature of the transformed stream 
will contain information on the initial position which 
is otherwise lost due to the translation-invariance of 
the signature, and the time augmentation 
(cf. \eqref{Time}) introduces a monotonic coordinate 
that can ensure the resulting path is uniquely 
determined by its signature 
(cf. Theorem \ref{tree_like_unique}).

Many works choose the augmentation map $\Theta$ via 
experimentation with several explicit options, see 
\cite{CO18,KO19} for example. It is 
proposed in \cite{BKLPS19} that the augmentation map 
should be data dependent and be learnt, i.e to 
consider $\Theta = \Theta^{\th}$ for a trainable 
parameter $\th$. This removes any limitations on the 
form of $\Theta$, allowing complicated nonlinear 
influences from the higher order signature terms to be 
accurately captured. In particular, the augmentation 
may form part of a \textit{Neural Network}.

Initially inspired by the biological workings of a human 
brain, a neural network is a collection of functions 
(called nodes) that are arranged in layers. 
Each layer connects to its neighbours, and the output of 
a layer is used as the input data for the next layer.
Typically, each node has an associated 
\textit{activation function} controlling whether or not 
the nodes function is run depending on the received input.
Empirically, simple initial inputs and a large number of 
layers works better than more complex initial specifications 
teamed with fewer layers.
Mathematically, neural networks may be thought of as linear 
functions composed with nonlinearity.
Whilst complicated nonlinear transformations are allowed 
between layers, all optimisation happens at the layers 
themselves, where we are always seeking
a linear combination of a specified collection of functions.
Detailed and thorough introductions to neural networks 
may be found in \cite{ML_book}, \cite{Hea12}, \cite{Gur97}, 
for example. 

Recall that, on compact subsets, continuous functions of 
paths can be arbitrarily well-approximated by linear 
combinations of coordinate iterated integrals
(cf. Theorem \ref{uni_nonlin}). In this sense, the 
signature is a universal nonlinearity of streams, 
which makes it a candidate for the nonlinear
transformation between layers within a neural network. 
Using the signature in this manner requires adapting the 
signature to map a stream to a stream, rather 
than mapping a stream to statistics with no obvious 
stream-like properties. 
In \cite{BKLPS19} this is achieved by lifting the input 
stream $\Omega_V$ to a stream of streams, making use of 
the observation that a path $X : [a,b] \to V$ induces a 
path-of-paths in path space by mapping 
$t \mapsto X \rvert_{[a,t]}$ for $t \in [a,b]$.

This idea makes it easy to see that a stream 
$(x_1, \ldots , x_k)$ can induce the stream-of-streams 
given by
$\Big( (x_1 ,x_2) , (x_1,x_2,x_3) , \ldots, 
(x_1 , \ldots , x_k) \Big)$, for example.
But there is no unique way to induce a stream-of-streams 
from a stream. A generic \textit{lift} 
$l : \s (V) \to \s ( \s(V))$ can be considered,
with the signature of $l(X)$ for $X \in \s(V)$ to be the 
stream of signatures of the components. More precisely, 
if $X \in \s(V)$ then 
$l(X) = ( l(X)_1 , \ldots  , l(X)_m )$ for some 
$m \in \N$, where $l(X)_j \in \s(V)$ for each 
$j \in \{1, \ldots ,m\}$.
The signature of $l(X)$ is then taken to be the stream 
$\Big( S \big( l(X)_1 \big) , \ldots , 
S \big( l(x)_m \big) \Big) \in 
\s \big( T((V)) \big)$.
The particular choice of lift will be determined by the 
modelling assumptions. Several example lifts $l$ may 
be found in \cite{BKLPS19}.

Having modified the signature to transform a stream to a 
stream, the authors define a \textit{deep signature model} 
in which several learnt augmentation steps are implemented.
The output of each layer is a linear function of the 
signature, and the signature of this forms the input 
functions for the next layer. The shuffle product plays a 
vital role in understanding how to compute $S(L(S(X)))$ 
for a linear function $L$ \cite{CLL04}. 
Two simple architectures involving signature 
layers are shown in Figure \ref{possible_sig_layers}, 
and full details of the proposed models can be 
found in \cite{BKLPS19}.

\begin{figure}[ht]
\centering
\resizebox{\linewidth}{!}{
\begin{tikzpicture}[shorten >=1pt,
        draw=black!50, node distance=\layersep]	
		\draw[rounded corners=\corners, black, 
		fill=green!30, thick]
		(0,0) rectangle ++(1,1);
		\node[text width=4em, text centered] at 
		(0.5,0.5) {$\mathbf x$};		
		\node[text width=4em, text centered] at 
		(0.5,1.5) {Input stream};		
		\draw[black, thick, ->] (1,0.5) -- (1.5,0.5);		
		\draw[rounded corners=\corners, black, 
		fill=blue!30, thick]
		(1.5,0) rectangle ++(1,1);		
		\node[text width=4em, text centered] at 
		(2,1.5) {Signature transform};
		\node[text width=4em, text centered] at 
		(2,0.5) {  $S^{(N)}$  };
		\draw[black, thick, ->] (2.5,0.5) -- (3,0.5);		
		\draw[rounded corners=\corners, black, 
		fill=blue!30, thick]
		(3,0) rectangle ++(1,1);		
		\node[text width=4em, text centered] at 
		(3.5,1.5) {Neural network};
		\node[text width=6em, text centered] at 
		(3.5,0.5) { $f^\theta$ };
		\draw[black, thick, ->] (4,0.5) -- (4.5,0.5);		
		\draw[rounded corners=\corners, black, 
		fill=red!30, thick]
		(4.5,0) rectangle ++(1,1);
		\node[text width=4em, text centered] at 
		(5,1.5) {Output};
		\node[text width=6em, text centered] at 
		(5,0.5) {$\sigma$};	
        \node[text width = 20em, text centered] at
        (3.0,-0.5) {Neural-signature model.};
        \node[text width = 20em, text centered] at
        (3.0,-1.0) {Trainable parameters: $\th$.};
		\draw[rounded corners=\corners, black, 
		fill=green!30, thick]
		(6.5,0) rectangle ++(1,1);		
		\node[text width=4em, text centered] at 
		(7.0,0.5) {$\mathbf x$};		
		\node[text width=4em, text centered] at 
		(7.0,1.5) {Input stream};
		\draw[black, thick, ->] (7.5,0.5) -- (8.0,0.5);		
		\draw[rounded corners=\corners, black, 
		fill=blue!30, thick]
		(8.0,0) rectangle ++(1,1);		
		\node[text width=4em, text centered] at 
		(8.5,1.5) {Feature map};
		\node[text width=4em, text centered] at 
		(8.5,0.5) { $\Phi$ };
		\draw[black, thick, ->] (9.0,0.5) -- (9.5,0.5);		
		\draw[rounded corners=\corners, black, 
		fill=blue!30, thick]
		(9.5,0) rectangle ++(1,1);		
		\node[text width=4em, text centered] at 
		(10.0,1.5) {Signature transform};
		\node[text width=4em, text centered] at 
		(10.0,0.5) {$S^{(N)}$};
		\draw[black, thick, ->] (10.5,0.5) -- (11.0,0.5);		
		\draw[rounded corners=\corners, black, 
		fill=blue!30, thick]
		(11.0,0) rectangle ++(1,1);		
		\node[text width=4em, text centered] at 
		(11.5,1.5) {Neural network};
		\node[text width=6em, text centered] at 
		(11.5,0.5) { $ f^{\theta}$  };	
		\draw[black, thick, ->] (12.0,0.5) -- (12.5,0.5);
		\draw[rounded corners=\corners, black, 
		fill=red!30, thick]
		(12.5,0) rectangle ++(1,1);	
		\node[text width=4em, text centered] at 
		(13.0,1.5) {Output};
		\node[text width=6em, text centered] at 
		(13.0,0.5) {$\sigma$};	
        \node[text width=20em, text centered] at
        (10.0,-0.5) {Neural-signature-augment model.};
        \node[text width=20em, text centered] at
        (10.0,-1.0) {Trainable parameters: $\th$.};
\end{tikzpicture}
}
\centering
\caption{Two simple architectures illustrating 
the possible inclusion of a signature layer -
Figure 1 in \cite{BKLPS19}.
Reproduced with permission.
Copyright held by authors of \cite{BKLPS19}.}
\label{possible_sig_layers}
\end{figure}

The required computation and back-propagation of 
the signature transform is efficiently managed by 
the \textit{Signatory} package \cite{KL20}. 
The resulting model obtained in \cite{BKLPS19} is 
compared with several other techniques via the 
task of learning the \textit{Hurst parameter} $H$ of
a fractional Brownian motion
$B^H$. That is, to learn the map $\bx^H \mapsto H$ 
for an implementation
\begin{equation}
	\label{hurst_learn}
		\bx^H = \Big( \big( t_0 , B^H_{t_0} \big) , 
		\ldots , \big( t_n , B^H_{t_n} \big) \Big)
		\in \s \big( \R^2 \big)
\end{equation}
for some realisation of $B^H$. Empirical evidence suggests 
fractional Brownian motions are closely linked with 
financial market data \cite{GJR18}, and estimating the 
Hurst parameter is a non-trivial task \cite{LLLN09}.

A total of 700 samples was considered, with 600 being 
used as the training set and the remaining 100 forming 
the test set. Each individual sample was an instance of 
fractional Brownian motion, sampled at 300 time steps 
of the interval $[0,1]$, with Hurst parameters
in the range $[ 0.2 , 0.8 ]$.
Every model was trained for 100 epochs with the loss 
function take to be the mean squared error (MSE).
The results are summarised in the Table 
\ref{deep_sig_model_results}. 

\begin{table}[ht]
	\centering
	\begin{tabular}{|c|c|c|}
        \hline 
		\textbf{Model}& \textbf{Mean} & \textbf{Variance} \\ 
		\hline		
		Rescaled Range & $7.2 \times 10^{-2}$ & 
		$3.7 \times 10^{-3}$ \\ 
		LSTM & $4.3 \times 10^{-2}$ &
		$8.0 \times 10^{-3}$\\
		Feedforward &$2.8 \times 10^{-2}$ & 
		$3.0 \times 10^{-3}$ \\
		Neural-Sig &$1.1 \times 10^{-2}$ & 
		$8.2 \times 10^{-4}$ \\
		GRU &$3.3 \times 10^{-3}$ & 
		$1.3 \times 10^{-3}$ \\
		RNN &$1.7 \times 10^{-3}$ & 
		$4.9 \times 10^{-4}$ \\
		\textbf{DeepSigNet} & $\mathbf{2.1 \times 10^{-4}}$ & 
		$\mathbf{8.7 \times 10^{-5}}$ \\
		\textbf{DeeperSigNet} & $\mathbf{1.6 \times 10^{-4}}$ & 
		$\mathbf{2.1 \times 10^{-5}}$ \\
        \hline 
	\end{tabular}
	\caption{Final test MSE for the different models - 
	Variant of Table 1 in \cite{BKLPS19}}
	\label{deep_sig_model_results}
\end{table}

The \textit{Rescaled Range} method is a mathematically 
derived method involving no learning, see \cite{Hur51}.
The \textit{Long Short-Term Memory} (LSTM), GRU and 
\textit{Recurrent Neural Network} (RNN) models provide 
baselines in the context of recurrent 
neural networks. The \textit{Neural-Sig} model provides 
a baseline in the context of signatures, with only a 
single layer of augmentation being 
used. The \textit{DeepSigNet} and \textit{DeeperSigNet} are 
both deep signature models featuring multiple layers of 
augmentation. Whilst the traditional signature based 
methods perform worse than the recurrent models, the 
deep signature models out perform all other models by 
at least an order of magnitude. Full details of this 
comparison and the baseline models may be found in 
section 4.2 and appendix B.2 of \cite{BKLPS19}.

\section{Signature Kernel} 
\label{full_sig_kernel_section}
The signature provides a universal feature map embedding 
the dataset $\Omega_{\s(V)}$ into the tensor algebra $T((V))$.
By considering the inner product we equipped the tensor 
algebra $T((V))$ with in Section \ref{Ten_Alg}, 
we can define the \textit{Signature Kernel}  
\begin{equation}
    \label{sig_kernel}
        K_{x,y} (s,t) := 
        \big< S_{a,s}(x) , S_{c,t}(y) \big>_{T((V))}
\end{equation}
for $x \in \cv^1 ([a,b],V)$, $y \in \cv^1 ([c,d],V)$, 
$s \in [a,b]$ and $t \in [c,d]$.

The completion $\cH(V) := \overline{T((V))}$
is a Hilbert space, 
meaning that the signature provides us with a kernel. 
Recall that we consider a labelled data set 
$\Omega \subset \s(V) \times \R$ given, for some 
$M \in \Z_{\geq 1}$, by 
$\Omega = \big\{ (\bx_k , y_k) \in \s(V) \times \R \mid 
k \in \{1 , \ldots , M \} \big\}$, and we take 
$\Omega_{\s(V)} \subset \s(V)$ to be the
projection of $\Omega$ onto $\s(V)$.
Moreover we assume that each stream $\bx \in \Omega_{\s(V)}$
consists of incremental data.
For each $k \in \{1 , \ldots , M\}$ let $\Gamma_k$ denote
the contour resulting from the concatenation of the entries
in the stream $\bx_k$, and 
choose a parameterisation $\gamma_k \in \cv^1([0,1],V)$
of $\Gamma_k$.

Then we can consider the function
$\K_{\Omega_{\s(V)}} : \Omega_{\s(V)} \times \Omega_{\s(V)} 
\to \R_{\geq 0}$ given by
\begin{equation}
    \label{sig_kernel_1}
        \K_{\Omega_{\s(V)}} (\bx_i,\bx_j) 
        := 
        K_{\gamma_i , \gamma_j} ( 1 , 1 )  
        = 
        \big< S_{0,1}(\gamma_i),
        S_{0,1}(\gamma_j)\big>_{T((V))}
\end{equation}
for $i,j \in \{1, \ldots , M\}$. 
In \cite{CFLSY20} the
authors consider the use of the kernel function 
$\K_{\Omega_{\s(V)}}$, defined in \eqref{sig_kernel_1},
for regression tasks. Via their observation that 
the function $(s,t) \mapsto K_{x,y}(s,t)$ defined 
in \eqref{sig_kernel} satisfies a linear second-order
hyperbolic PDE (a \textit{Goursat problem} \cite{Lee60}),
the authors introduce a method to use the full signature
transform for regression tasks \textit{without} any 
truncation via use of the kernel function $\K_{\Omega_{\s(V)}}$.
The primary aim for the remainder of this section is 
to summarise this approach developed in \cite{CFLSY20}. 

Kernels on truncated signatures are considered in \cite{KO19}
via the development of a technique for transforming a kernel
on $V$ to a kernel on $\s(V)$.
The approach followed in \cite{KO19} is to use truncated 
kernels (in which the full signature is replaced by 
its truncation to some finite depth)
to approximate the full untruncated signature kernel.
For paths of bounded variation, there are 
efficient algorithms to compute this truncated kernel, 
with the practical consequences explored in \cite{OT20}.

In \cite{CFLSY20} the full untruncated signature kernel is 
studied.  It is established that for two continuously 
differentiable paths $x$ and $y$, the full untruncated 
signature kernel $K_{x,y} (\cdot,\cdot)$ solves the 
following PDE.

\begin{theorem}[Signature Kernel Solves PDE; 
Theorem 2.5 in \cite{CFLSY20}]
\label{sig_ker_pde_smooth_case}
Let $I = [a,b]$ and $J = [c,d]$ be compact 
intervals and suppose $x \in C^{1} ( I ; V)$ and 
$y \in C^1 ( J ; V)$ for some Banach space $V$. Consider 
the bilinear form $K : I \times J \to \R$ defined by
\begin{equation}
	\label{ker_def}
		K_{x,y} (s,t) := 
		\big< S_{a,s}(x) , S_{c,t}(y) 
		\big>_{T((V))}.
\end{equation}
Then $K_{x,y}(\cdot , \cdot)$ solves the following 
linear second-order hyperbolic PDE
\begin{equation}
	\label{goursat_eqn}
		\frac{\partial^2 K_{x,y}}{\partial s \partial t} 
		= \big< \dot{x}_s , \dot{y}_t \big> K_{x,y}
\end{equation}
with initial conditions 
$K_{x,y} (a, \cdot ) = K_{x,y} (\cdot , c) = 1$ 
and where $\dot{x}_s := \frac{d x_p}{dp} \Big\rvert_{p=s}$
and $\dot{y}_t := \frac{d y_q}{dq} \Big\rvert_{q=t}$.
\end{theorem}
\vskip 4pt
\noindent
The PDE in \eqref{goursat_eqn} is an example of a 
\textit{Goursat problem}, for which existence and uniqueness 
results originate in \cite{Lee60}. 
Consequently, for sufficiently regular paths, the existence 
and uniqueness of $K$ solving \eqref{ker_def} is guaranteed 
(cf. Theorem 3.1 in \cite{CFLSY20}).
The content of Theorem \ref{sig_ker_pde_smooth_case} makes 
sense for rougher paths. That is, the PDE 
\eqref{goursat_eqn} can be made sense of for paths
$x$ and $y$ of lower regularity by considering its
integral form
\begin{equation}
	\label{ker_eqn}
		K_{x,y}(s,t) = 1 + 
		\int_{a}^s \int_c^t 
		\big< S_{a,p}(x) , S_{c,q}(y) \big>_{T((V))}
		\big< dx_p , dy_q \big>_V
\end{equation}
and making an appropriate definition for the 
quantity $\big< dx_p , dy_q \big>_V$.
This is done in section 4 in \cite{CFLSY20} 
for the setting that the paths $x$ and $y$ are 
geometric rough paths 
(cf. Definition \ref{geometric_p-rough_path_def} in 
this article).
The main technicality is using the theory of integrating 
one-forms along rough paths to make sense of the 
double integral in \eqref{ker_eqn} in this rougher 
setting. 
A review of the theory of integrating one-forms along 
rough paths can be found in Appendix B of \cite{CFLSY20};
a more thorough presentation of this theory can be found 
in section 4 of \cite{CLL04}.
The full details of how to make sense of the 
\textit{rough double integral} can be found in section 4 of 
\cite{CFLSY20}.

Returning to the setting of paths of bounded variation,
numerical schemes to approximate $K_{x,y}$ are developed 
in \cite{CFLSY20}, which we now illustrate in the case 
that $V = \R^d$ for some $d \in \N$. 
Assume that $I,J \subset \R$ are compact intervals and 
that $x : I \to \R^d$ and $y : J \to \R^d$ are both 
piecewise linear. 
Then the PDE \eqref{goursat_eqn} becomes
\begin{equation}
	\label{pl_goursat}
		\frac{\partial^2 K_{x,y}}{\partial s \partial t} 
		= C K_{x,y}
\end{equation}
on each domain 
$\cd_{ij} := \big[u_i , u_{i+1} \big] \times 
\big[ v_j , v_{j+1} \big]$ that 
$C := \big< \dot{x}_s ,\dot{y}_t \big>$ is constant on.
In integral form, \eqref{pl_goursat} reads
\begin{equation}
	\label{integral_pl_goursat}
		K_{x,y} (s,t) = K_{x,y} (s,v) + 
		K_{x,y} (u,t) - K_{x,y} (u,v) + 
		C \int_{u}^s \int_{v}^t K_{x,y} (r,w) dr dw
\end{equation}
for $(s,t),(u,v) \in \cd_{ij}$ with $u\leq s$ and $v\leq t$.
Approximations of the double-integral in 
\eqref{integral_pl_goursat} can be used to develop
finite difference schemes to approximate $K_{x,y}$.
The relatively simple approximation
\begin{equation}
	\label{explicit_est}
		\int_{u}^s \int_{v}^t K_{x,y} (r,w) dr dw 
		\approx \frac{1}{2} \big( K_{x,y}(s,v) + 
		K_{x,y}(u,t) \big) (s-u)(t-v),
\end{equation}
is used in \cite{CFLSY20} to develop such a scheme.

In order to more precisely discuss this finite 
difference scheme, let $I, J$ be compact intervals and
$\cd_1 := \{u_0 < u_1 < \ldots < u_{m-1} < u_m\}$ 
and $\cd_2 := \{v_0 < v_1 < \ldots < v_{n-1} < v_n\}$ 
be partitions of $I$ and $J$ respectively for which $x$ 
is piecewise linear with respect to $\cd_1$ and $y$ is 
piecewise linear with respect to $\cd_2$.
Then \eqref{explicit_est} can be 
used to define numerical schemes for \eqref{goursat_eqn} 
on the grid $P_0 := \cd_1 \times \cd_2$, and its dyadic 
refinements.
For $\lambda \in \N_0$ we define $P_{\lambda}$ as 
the dyadic refinement of $P_0$ such that
\begin{equation}
	\label{dyadic_refine}
		P_{\lambda} \cap \Big( [ u_i , u_{i+1}] 
		\times [ v_j , v_{j+1}] \Big)
		=
		\bigcup_{k,l =0}^{2^{\lambda}}
		\bigg\{ \bigg( u_i + \frac{k}{2^{\lambda}} 
		\big(u_{i+1} - u_i\big) , v_j + \frac{l}{2^{\lambda}} 
		\big(v_{j+1} - v_j\big)  \bigg) \bigg\}.
\end{equation}
In \cite{CFLSY20} the authors define the 
finite difference scheme 
\begin{align}
	\label{explicit_scheme}
		\hat{K} \big( s_{i+1} , t_{j+1} \big) &= 
		\hat{K} \big( s_{i+1} , t_j \big) + 
		\hat{K} \big( s_i , t_{j+1} \big) - 
		\hat{K} \big( s_i , t_j \big) + \notag \\
		&\quad \quad \frac{1}{2} 
		\big< x_{s_{i+1}} - x_{s_i} , 
		y_{t_{j+1}} - y_{t_j} \big> 
		\Big( \hat{K} \big( s_{i+1} ,t_j \big) 
		+ \hat{K} \big( s_i ,t_{j+1} \big) \Big),
\end{align}
with $\hat{K} (s_0 , \cdot ) = \hat{K} ( \cdot , t_0) = 1$, 
on the grid 
\begin{equation}
	\label{plambda}
		P_{\lambda} =
		\bigcup_{i=0}^{m-1} \bigcup_{j=0}^{n-1} 
		~\bigcup_{k,l =0}^{2^{\lambda} - 1}
		\bigg\{ \bigg( u_i + \frac{k}{2^{\lambda} } 
		\big( u_{i+1} - u_i\big) , v_j + 
		\frac{l}{2^{\lambda}} \big(v_{j+1} - v_j\big)  
		\bigg) \bigg\}
		=: \bigcup_{i = 0}^{2^{\lambda } n} 
		~\bigcup_{j=0}^{2^{\lambda}m} 
		\big\{ ( s_i , t_j) \big\}.
\end{equation}
They further establish that 
refining the discretisation of the grid used to 
approximate the PDE yields convergence to the true value.

\begin{figure}[ht]
\begin{center}
\includegraphics[width=\textwidth]{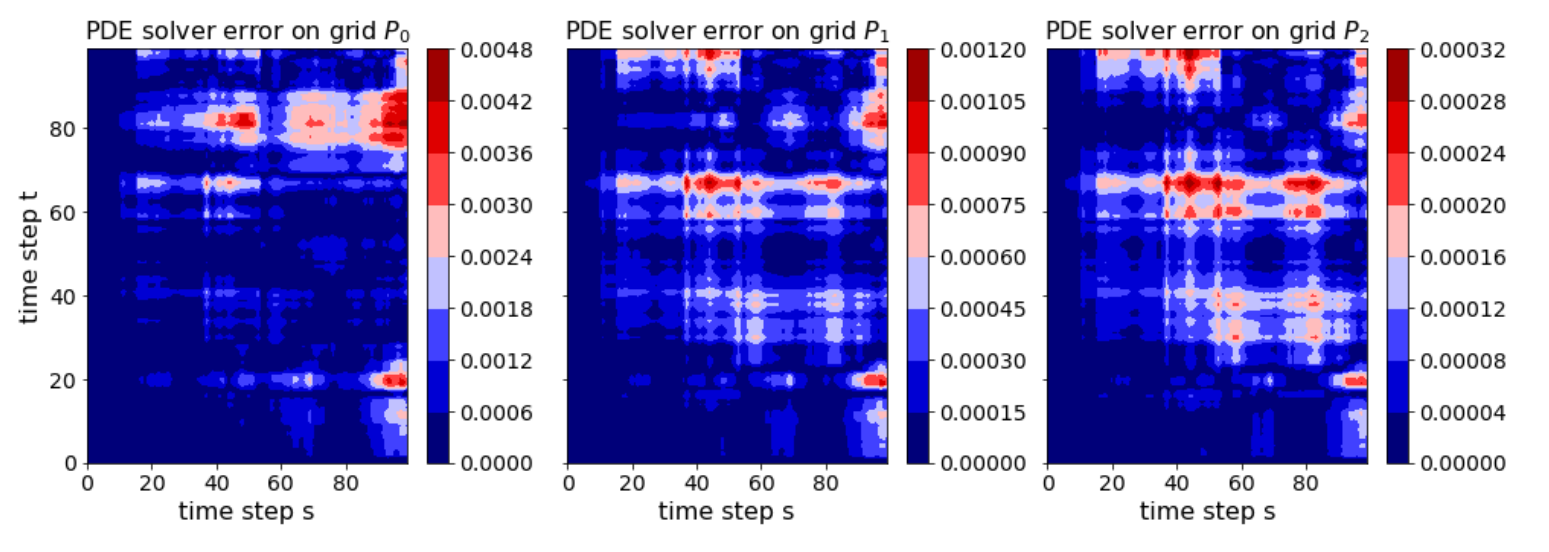}
\caption{Example of error distribution of $k_{x,y}(s,t)$ 
on the grids $P_0$, $P_1$ and $P_2$---Figure 2 in 
\cite{CFLSY20}. 
Copyright \copyright 2021 Society for Industrial and 
Applied Mathematics. Reprinted with permission. All 
rights reserved.}
\label{figure1_CFLSY20}
\end{center}
\end{figure}

\begin{theorem}[Global Convergence; Theorem 2.8 
in \cite{CFLSY20}]
\label{global_conv}
Assume that $x$ and $y$ are piecewise linear paths as 
above satisfying that
\begin{equation}
	\label{upper_bound}
		\sup_{I \times J} \big\lvert 
		\big< \dot{x}_s , \dot{y}_t \big> \big\rvert 
		\leq M
\end{equation}
for some $M > 0$. Then there exists a constant $C_1 > 0$, 
depending on $M$ and $K_{x,y}$, such that for any 
$\lambda \geq 0$ the numerical solution 
$\hat{K}$ obtained by applying the scheme define by
\eqref{explicit_scheme} on $P_{\lambda}$ satisfies
\begin{equation}
	\label{conv_rate}
		\sup_{I \times J} 
		\Big\lvert K_{x,y}(s,t) - \hat{K}(s,t) \Big\rvert
		\leq \frac{C_1}{2^{2\lambda}}.
\end{equation}
\end{theorem}
\noindent
In \cite{CFLSY20} it was found that rescaling the 
paths $x$ and $y$ ensures 
that coarse partition choices of $\lambda = 0,1$ 
were sufficient to provide good approximations, as 
illustrated in Figure \ref{figure1_CFLSY20}.

\begin{figure}[ht]
\begin{center}
\includegraphics[width=\textwidth]{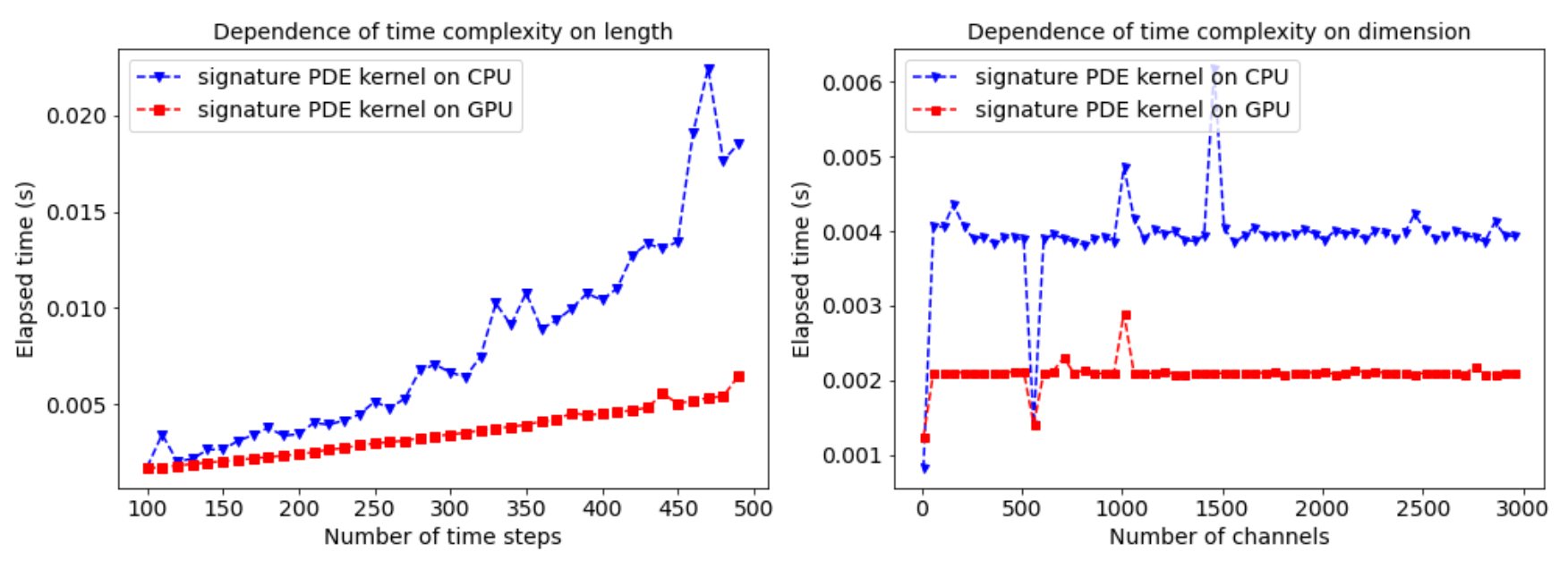}
\caption{Comparison of the elapsed time(s) to reach 
an accuracy of $10^{-3}$ from a target value 
obtained by solving the signature kernel PDE 
on a fine discretization grid ($P_5$). Each run 
simulates 5 (piecewise linear interpolations of)
Brownian paths. The left plot shows the 
dependency on the length of two paths in dimension
2. The right plot shows the dependency on the 
dimension of two paths of a fixed length. 
Reproduction of Figure 3 in \cite{CFLSY20}. 
Copyright \copyright 2021 Society for Industrial and 
Applied Mathematics. Reprinted with permission. All 
rights reserved.}
\label{figure2_CFLSY20}
\end{center}
\end{figure}

It is noted in \cite{CFLSY20} that the 
``PDE structure" allows for efficient GPU implementation
of the scheme via parallelization. 
Figure \ref{figure2_CFLSY20}, which is Figure 3 in 
\cite{CFLSY20}, illustrates the complexity reduction 
from quadratic on CPU to linear on GPU.

In \cite{CFLSY20}, the untruncated signature 
kernel is compared against 
other kernels via the task of classification with 
Support Vector Machine (SVM) classifiers \cite{Vap98}.
SVM classifiers have been successfully used in medicine 
\cite{BCDFHS00}, image retrieval \cite{CT01} and text 
classification \cite{KT01}, for example.
Given $X = \{ x_1 , \ldots , x_n\}$ and a reproducing 
kernel $k$ on $X$ with associated reproducing kernel 
Hilbert space (RKHS) $H_k$, consider the pairs 
$\big\{ (x_i , y_i) \mid i \in \{1,\ldots ,n\} \big\}$.
For binary classification we have $y_i \in \{-1,1\}$ 
and the binary SVM classification algorithm aims at 
solving the minimisation problem
\begin{equation}
	\label{SVM_min}
		\min_{f \in H_k} \sum_{i=1}^n 
		L ( y_i , f(x_i)) + 
		\lambda \lvert\lvert f \rvert\rvert_{H_k}
\end{equation}
where $L ( y_i , f (x_i) ) 
:= \max \big\{ 0 , 1 - y_i f (x_i) \big\}$ and 
$\lambda$ is a penalty hyper-parameter. 
The optimal solution $f_{\ast}$ is given by
\begin{equation}
	\label{SVM_soln}
		f_{\ast} (x) := \sign \bigg( \al_0 + 
		\sum_{i=1}^n \al_i y_i k ( x , x_i )  \bigg)
\end{equation}
where $\al_0 , \ldots , \al_n$ are scalar coefficients 
solving a certain quadratic programming problem, see 
\cite{SS18} for full details.

Selecting an appropriate kernel when $X$ consists
of multivariate time-series is difficult \cite{SS09}.
When all time-series in $X$ have the same length, 
the standard kernels (i.e. linear, polynomial,
Gaussian etc) work well. But outside this setting,
kernels specifically designed for sequential data 
are required. The performances of an SVM classifier 
equipped with a range of different kernels on various 
multivariate time-series UEA datatsets \cite{BBKLL17} 
are presented in Table \ref{SVM_results}.
The \textit{global alignment kernel} (GAK) \cite{Cut11} 
is the only other kernel proposed for sequential data, 
and depends on a hyperparamter $\beta \in (0,1]$.
The full untruncated signature kernel (Sig-PDE) SVM is 
among the top 2 classifiers across all but two of
the datasets. Sig-PDE systematically outperforms 
its truncated counterpart Sig(n).
Further numerical applications of the signature kernel
can be found in section 5 of \cite{CFLSY20}.

\begin{table}[ht]	
	\centering
	\begin{tabular}{|c|c|c|c|c|c|}
        \hline
		\textbf{Datasets/Kernel} & Linear & 
		RBF & GAK & Sig(n) & Sig-PDE \\
		\hline
		ArticularyWordRecognition & 98.0 & 98.0 &
		98.0 & 92.3 & \textbf{98.3} \\
		BasicMotions & 87.5 & 97.5 &
		97.5 & 97.5 & \textbf{100.0} \\
		Cricket & 91.7 & 91.7 & 
		\textbf{97.2} & 86.1 & \textbf{97.2} \\
        ERing & 92.2 & 92.2 & \textbf{93.7} & 84.1 & 
        93.3 \\
        Libras & 73.9 & 77.2 & 79.0 & 
        \textbf{81.7} &  \textbf{81.7} \\
        NATOPS & 90.0 & 92.2 & 90.6 & 88.3 &
        \textbf{93.3} \\
        RacketSports & 76.9 & 78.3 & 84.2 & 80.2 & 
        \textbf{84.9} \\
        FingerMovements & 57.0 & 60.0 & 
        \textbf{61.0} & 51.0 & 58.0 \\
        Heartbeat & 70.2 & 73.2 &  70.2 &  72.2 & 
        \textbf{73.6} \\
        SelfRegulationSCP1 & 86.7 & 87.3 & 
        \textbf{92.4} & 75.4 & 88.7 \\
        UWaveGestureLibrary & 80.0 & \textbf{87.5} & 
        \textbf{87.5} & 83.4 & 87.0 \\
        \hline 
	\end{tabular}
	\caption{Test set classification accuracy 
	(in $\%$) on UEA multivariate time series 
	datasets - Table 1 in \cite{CFLSY20}}
	\label{SVM_results}
\end{table}

Subsequent to the appearance of \cite{CFLSY20}, numerous
works have considered the development and application of 
signature kernel techniques within a wide range of settings.
We end this section with a far-from-exhaustive summary
of some of the further developments of signature kernel
techniques.

\textit{General Signature Kernels} (GSKs) have been 
introduced in \cite{CLX21}. 
Recall that
for $A,B \in T((V))$ we have that
$\big< A , B \big>_{T((V))} = \sum_{n=0}^{\infty}
\big< \pi_n (A) , \pi_n(B) \big>_{V^{\otimes n}}$
(cf. \eqref{ten_alg_inner_prod}).
Loosely, the GSKs proposed in \cite{CLX21} 
are obtained by choosing coefficients
$\{ \phi(n) \}_{n=0}^{\infty} \subset \C$ that are
\textit{not} required to be $1$, and subsequently
replacing the inner product
$\big< \cdot , \cdot \big>_{T((V))}$ by
$\big< \cdot , \cdot \big>_{\phi}$ which is defined
for $A,B \in T((V))$ by
$\big< A , B \big>_{\phi} := 
\sum_{n=0}^{\infty} \phi(n)
\big< \pi_n (A) , \pi_n(B) \big>_{V^{\otimes n}}$.
Of course $\big< \cdot , \cdot \big>_{\phi}$ is not
necessarily an inner product on $T((V))$, but one may 
still define an associated GSK by  
\begin{equation}
    \label{gen_sig_kernel}
        K_{x,y}^{\phi}(s,t) :=
        \big< S_{a,s}(x) , S_{c,t}(y) \big>_{\phi}
        =
        \sum_{n=0}^{\infty} \phi(n)
        \big< S^n_{a,s}(x) , S^n_{c,t}(y)
        \big>_{V^{\otimes n}}
\end{equation}
for $x \in \cv^1 ([a,b],V)$, 
$y \in \cv^1 ([c,d],V)$, $s \in [a,b]$ and $t \in [c,d]$.
In many situations GSKs can be 
interpreted as an average of PDE solutions, and may 
consequently be computed via suitable quadrature rules
\cite{CLX21}. The extension of this analysis allows the 
authors of \cite{CLX21} to obtain closed-form formulae
for expressions involving the expected (Stratonovich)
signature of Brownian motion and to articulate a connection
between signature kernels and the hyperbolic development 
map.

In a different direction the signature kernel techniques
developed in \cite{CFLSY20} are central to the work on
Optimal Stopping Problems in \cite{HLLLS23}.
A major achievement of \cite{HLLLS23} is the development of a
rigorous mathematical framework in which an
Optimal Stopping Problem can be recast as a higher order
kernel mean embedding regression based on the notions of higher
rank signatures of measure-valued paths and adapted topologies.
This overcomes the issue in Optimal Stopping Problems
arising from mathematical finance that the corresponding
value functions are, in general, discontinuous with respect
to the weak topology. Full details may be found in
\cite{HLLLS23}.

The final use of the signature kernel techniques developed
in \cite{CFLSY20} that we mention here is in the development
of an efficient method for the computation of a sparse
collection of iterated integrals \cite{SS24}.
The authors of \cite{SS24} introduce a technique to effectively
isolate specific groups of signature coefficients including,
in particular, a singular coefficient at any depth.
The empirical effectiveness of this approach is demonstrated
via the task of the construction of $N$-step Euler schemes for
sparse CDEs \cite{SS24}.

\section{Neural CDEs and Neural RDEs}
\label{LSDE}
In this section we discuss 
\textit{Neural Controlled Differential Equations}
(Neural CDEs) \cite{FKLM20} and their subsequent
development to \textit{Neural Rough Differential Equations}
(Neural RDEs) \cite{FKLMS21}.
Loosely speaking, both are techniques in which Neural
networks are utilised to learn solutions to differential 
Equations.
We begin with a brief discussion of 
\textit{Neural Ordinary Differential Equations} (Neural ODEs). 

Neural ODEs are a popular and successful ODE 
based approach to modelling temporal dynamics.
They aim to approximate a mapping $F : \R^q \to \R^l$
by learning linear maps $l_{\th}^1 : \R^q \to \R^d$,
$l^2_{\th} : \R^d \to \R^l$, and a function
$f_{\th} : \R^d \to \R^d$, all depending on 
parameter $\th$, for which the following is true.
Given $x \in \R^q$, let $z : [0,T] \to \R^d$ denote the
solution to the ODE  
\begin{equation}
    \label{Neural_ODE_eqn}
        z(0) = l^1_{\th}(x) 
        \qquad \text{and} \qquad
        z(t) = z(0) + \int_0^T f_{\th}(z(s)) ds
\end{equation}
for every $t \in (0,T]$. Then the value 
$y := l^2_{\th}(z(T)) \in \R^l$ gives a good approximation
for $F(x) \in \R^l$, i.e. $y \approx F(x)$.
Explicit dependence of $f_{\th}$ on $s$ is only obtained
by including $s$ as an extra dimension in $z_s$;
see Appendix B in \cite{BCDR18}. 
The use of ODEs leads to the following fundamental issue.
Once the parameter $\th$ has been learnt, the solution
to the ODE is determined by the initial condition. 
There is no direct mechanism for adjusting the
trajectory to incorporate data that arrives later
\cite{FKLM20}.

In \cite{FKLM20} the authors resolve this issue by using
CDEs, resulting in their development of Neural CDEs.
We detail their approach for a fully-observed yet
potentially irregularly sampled time-series
$\bx = \big( (t_0 , x_0) , (t_1 , x_1) , \ldots , 
(t_n , x_n) \big)$ for timestamps
$t_0 < t_1 < \ldots < t_n$ and observations
$x_0 , x_1 , \ldots , x_n \in \R^p$.
The natural cubic spline $X : [t_0,t_n] \to \R^{p+1}$
with knots at $t_0 , \ldots , t_n$ such that
$X_{t_i} = (x_i,t_i)$ is chosen to approximate the
underlying process observed via $\bx$. The resulting 
differentiability of the control path $X$ allows one 
to treat the $dX_s$ term as $\frac{d X_s}{ds} ds$.
In Appendix A of \cite{FKLM20} it is justified that the 
choice of natural cubic spline for $X$ has the minimum 
regularity required.

Let $f_{\th} : \R^{d} \to \R^{d \times (p+1)}$ be any
neural network model depending on parameters $\th$ and 
having a hidden state of size $d$.
Let $\zeta_{\th} : \R^{p+1} \to \R^d$ be any neural 
network depending on parameters $\th$.
Then the Neural CDE model is defined to be the 
solution of the CDE
\begin{equation}
    \label{NCDE_model_eqn}
        z_{t_0} = \zeta_{\th} (x_0,t_0)
        \qquad \text{and} \qquad 
        z_t = z_{t_0} + \int_{t_0}^t f_{\th}(z_s) dX_s
\end{equation}
for all $t \in (t_0,t_n]$. 
The integral in \eqref{NCDE_model_eqn} is the
\textit{Riemann--Stieltjes} integral.
This initial condition is 
used to avoid translational invariance.
The model output can be taken 
as either the evolving process $z$ or its terminal value
$z_{t_n}$, and the final prediction should typically be
given by a linear map applied to this output.

The essential difference between Neural CDEs and Neural ODEs 
is that the equation \eqref{NCDE_model_eqn} for Neural CDEs
is driven by the data process $X$,
whilst the corresponding equation for Neural ODEs is driven 
only by the 
identity map $\R \to \R$. The Neural CDE is naturally 
adapting to incoming data; changes in $X$ change the 
local dynamics of the system. This difference is illustrated 
in Figure \ref{figure1_FKLM20}, which is Figure 1 in 
\cite{FKLM20}.

\newcommand{\mainpicturedata}{
	\draw[thick, ->] (-1, 0) -- (5, 0);
	\node at (0, 0)[circle,fill, inner sep=1.5pt] {};
	\node at (0.6, 0)[circle,fill, inner sep=1.5pt] {};
	\node at (2, 0)[circle,fill, inner sep=1.5pt] {};
	\node at (4, 0)[circle,fill, inner sep=1.5pt] {};
	\node at (0, -0.3) {$t_1$};
	\node at (0.6, -0.3) {$t_2$};
	\node at (2, -0.3) {$t_3$};
	\node at (3, -0.3) {$\cdots$};
	\node at (4, -0.3) {$t_n$};
	\node at (5.6, 0) {Time};
	
	\draw[blue!80!white, thick, cap=round] (-1, 0.3) .. controls ++(45:0.5) and ++(185:0.2) .. (0, 1);
	\draw[blue!80!white, thick, cap=round] (0, 1) .. controls ++(185+180:0.2) and ++(135:0.4) .. (0.6, 0.7);
	\draw[blue!80!white, thick, cap=round] (0.6, 0.7) .. controls ++(135+180:0.4) and ++ (190:0.3) .. (2, 0.6);
	\draw[blue!80!white, thick, cap=round] (2, 0.6) .. controls ++(190+180:0.3) and ++ (135:0.7) .. (4, 0.8);
	\draw[blue!80!white, thick, cap=round] (4, 0.8) .. controls ++(135+180:0.7) and ++(200:0.2) .. (5, 0.6);
	\node at (0, 1)[circle,draw=black, fill=cyan, inner sep=1.5pt] {};
	\node at (0.6, 0.7)[circle,draw=black, fill=cyan, inner sep=1.5pt] {};
	\node at (2, 0.6)[circle,draw=black, fill=cyan, inner sep=1.5pt] {};
	\node at (4, 0.8)[circle,draw=black, fill=cyan, inner sep=1.5pt] {};
	\node at (0, 0.7) {$x_1$};
	\node at (0.6, 0.4) {$x_2$};
	\node at (2, 0.3) {$x_3$};
	\node at (4, 0.5) {$x_n$};
	\node at (5.6, 0.6) {Data $\mathbf{x}$};
	}

\newcommand{\mainpicturedataB}{
	\draw[thick, ->] (8, 0) -- (14, 0);
	\node at (9, 0)[circle,fill, inner sep=1.5pt] {};
	\node at (9.6, 0)[circle,fill, inner sep=1.5pt] {};
	\node at (11, 0)[circle,fill, inner sep=1.5pt] {};
	\node at (13, 0)[circle,fill, inner sep=1.5pt] {};
	\node at (9, -0.3) {$t_1$};
	\node at (9.6, -0.3) {$t_2$};
	\node at (11, -0.3) {$t_3$};
	\node at (12, -0.3) {$\cdots$};
	\node at (13, -0.3) {$t_n$};
	\node at (14.6, 0) {Time};
	
	\draw[blue!80!white, thick, cap=round] (8, 0.3)
    .. controls ++(45:0.5) and ++(185:0.2) .. (9, 1);
	\draw[blue!80!white, thick, cap=round] (9, 1)
    .. controls ++(185+180:0.2) and ++(135:0.4) .. (9.6, 0.7);
	\draw[blue!80!white, thick, cap=round] (9.6, 0.7)
    .. controls ++(135+180:0.4) and ++ (190:0.3) .. (11, 0.6);
	\draw[blue!80!white, thick, cap=round] (11, 0.6)
    .. controls ++(190+180:0.3) and ++ (135:0.7) .. (13, 0.8);
	\draw[blue!80!white, thick, cap=round] (13, 0.8)
    .. controls ++(135+180:0.7) and ++(200:0.2) .. (14, 0.6);
	\node at (9, 1)[circle,draw=black, fill=cyan,
    inner sep=1.5pt] {};
	\node at (9.6, 0.7)[circle,draw=black, fill=cyan,
    inner sep=1.5pt] {};
	\node at (11, 0.6)[circle,draw=black, fill=cyan,
    inner sep=1.5pt] {};
	\node at (13, 0.8)[circle,draw=black, fill=cyan,
    inner sep=1.5pt] {};
	\node at (9, 0.7) {$x_1$};
	\node at (9.6, 0.4) {$x_2$};
	\node at (11, 0.3) {$x_3$};
	\node at (13, 0.5) {$x_n$};
	\node at (14.6, 0.6) {Data $\mathbf{x}$};
	}

\begin{figure}[ht]
\begin{center}
	\begin{tikzpicture}[scale=0.85]
	\mainpicturedata
	
	\draw[olive, thick, cap=round, ->] (0, 2) -- (0, 1.8);
	\draw[green!80!black, thick, cap=round] (0, 1.8)
    .. controls (0.3, 1.4) and (0.7, 2.4) .. (0.6,2.2);
	\draw[olive, thick, cap=round, ->] (0.6,2.2)
    -- (0.6,1.7);
	\draw[green!80!black, thick, cap=round] (0.6,1.7)
    .. controls ++(10:0.7) and ++(160:0.6) .. (2,1.9);
	\draw[olive, thick, cap=round, ->] (2,1.9) -- (2,2.5);
	\draw[green!80!black, thick, cap=round] (2,2.5)
    .. controls ++(0:0.6) and ++(210:0.9) .. (4,2);
	\draw[olive, thick, cap=round, ->] (4,2) -- (4, 2.2);
	
	\draw[dashed, ->] (0, 1.1) -- (0, 1.7);
	\draw[dashed, ->] (0.6, 0.8) -- (0.6, 1.65);
	\draw[dashed, ->] (2, 0.7) -- (2, 1.85);
	\draw[dashed, ->] (4, 0.9) -- (4, 1.95);
	
	\node at (5.6, 2.2) {Hidden state $z$};

	\mainpicturedataB
	
	\draw[orange!95!black, thick, cap=round] (9, 2)
    .. controls ++(0:0.2) and ++(150:0.4) .. (10, 1.7);
	\draw[orange!95!black, thick, cap=round] (10, 1.7)
    .. controls ++(150+180:0.4) and ++ (185:0.3) .. (11, 1.6);
	\draw[orange!95!black, thick, cap=round] (11, 1.6)
    .. controls ++(185+180:0.3) and ++ (165:0.7) .. (13, 1.8);
	\node at (14.6, 1.8) {Path $X$};
	
	\draw[dashed, ->] (9, 1.1) -- (9, 1.95);
	\draw[dashed, ->] (9.6, 0.8) -- (9.6, 1.8);
	\draw[dashed, ->] (11, 0.7) -- (11, 1.55);
	\draw[dashed, ->] (13, 0.9) -- (13, 1.75);
	
	\draw[dashed, ->] (9, 1.1) -- (9.4, 1.85);
	\draw[dashed, ->] (9.6, 0.8) -- (9.2, 1.9);
	\draw[dashed, ->] (9.6, 0.8) -- (10, 1.65);
	\draw[dashed, ->] (9.6, 0.8) -- (10.4, 1.5);
	\draw[dashed, ->] (9.6, 0.8) -- (10.8, 1.5);
	\draw[dashed, ->] (11, 0.7) -- (10.2, 1.55);
	\draw[dashed, ->] (11, 0.7) -- (10.6, 1.5);
	\draw[dashed, ->] (11, 0.7) -- (11.4, 1.6);
	\draw[dashed, ->] (13, 0.9) -- (12.6, 1.8);
	
	\draw[green!80!black, thick, cap=round] (9, 3)
    .. controls ++(-30:0.4) and ++(210:0.5) .. (10, 3);
	\draw[green!80!black, thick, cap=round] (10, 3)
    .. controls ++(210+180:0.5) and ++(180:0.7) .. (11.5, 3.5);
	\draw[green!80!black, thick, cap=round] (11.5, 3.5)
    .. controls ++(0:0.7) and ++(150:0.5) .. (13, 3);
	
	\draw[dashed, ->] (9, 2.05) -- (9, 2.95);
	\draw[dashed, ->] (9.2,2) -- (9.2,2.85);
	\draw[dashed, ->] (9.4,1.95) -- (9.4,2.8);
	\draw[dashed, ->] (9.6,1.9) -- (9.6,2.8);
	\draw[dashed, ->] (9.8,1.85) -- (9.8,2.85);
	\draw[dashed, ->] (10,1.75) -- (10,2.95);
	\draw[dashed, ->] (10.2, 1.7) -- (10.2, 3.05);
	\draw[dashed, ->] (10.4, 1.6) -- (10.4, 3.15);
	\draw[dashed, ->] (10.6, 1.6) -- (10.6, 3.25);
	\draw[dashed, ->] (10.8, 1.6) -- (10.8, 3.35);
	\draw[dashed, ->] (11, 1.65) -- (11, 3.4);
	\draw[dashed, ->] (11.2, 1.7) -- (11.2, 3.45);
	\draw[dashed, ->] (11.4, 1.75) -- (11.4, 3.45);
	\draw[dashed, ->] (12.6, 1.9) -- (12.6, 3.2);
	\draw[dashed, ->] (12.8, 1.9) -- (12.8, 3.05);
	\draw[dashed, ->] (13, 1.85) -- (13, 2.95);
	\node at (14.6,3) {Hidden state $z$};
	\end{tikzpicture}
\caption{ \textbf{Left:} The hidden state is modified 
at each observation, and potentially continuously evolved
between observations. 
\textbf{Right:} The hidden state in the Neural CDE model 
has continuous dependence on the observed data.
Figure 1 in \cite{FKLM20}. Reproduced with permission.
Copyright held by authors of \cite{FKLM20}.}
\label{figure1_FKLM20}
\end{center}
\end{figure}

In Appendix B in \cite{FKLM20} the authors establish that
Neural CDEs are universal approximators. The precise 
statement may be found as Theorem B.14 in \cite{FKLM20}; 
informally it guarantees that a continuous function from 
the space of time-series can be arbitrarily well-approximated
locally by a linear map applied to the terminal value of 
a Neural CDE. The essential idea is that CDEs can be used
to approximate bases of continuous functions on path space.
This is itself a consequence of the
\textit{universal nonlinearity} property of the path signature
(cf. Theorem \ref{uni_nonlin} in this article) 
and that the path signature solves a CDE 
(cf. \eqref{CDE_sig_eqn}).
Full details may be found in Appendix B of \cite{FKLM20}.

The efficacy of Neural CDEs is demonstrated in \cite{FKLM20}
via the task of written character classification. The 
CharacterTrajectories dataset from the UEA time series 
classification archive \cite{BBDFKLLS18} is considered.
This is a dataset of 2858 time series, each of length 182, 
consisting of the $x$ and $y$ coordinate positions and the
pen tip force whilst a Latin alphabet character is written
in a single stroke. The task is to classify which of the 
20 different characters are written.

Three experiments are run in \cite{FKLM20} in which
30\%, 50\% and 70\% of the observations are selected to
be dropped. The selections are made uniformly at random
and independently for each time series. This results in a
dataset of irregularly sampled, completely observed
time series. The Neural CDE model is compared with several
ODE and RNN based models, details of which may be found in
\cite{FKLM20}. The results are summarised in
Table \ref{FKLM20_Table1} below, 
which is a variant of Table 1 in \cite{FKLM20}.
The randomly removed data is the same for every model 
and every repeat.
The Neural CDE outperforms every other model considered.
Moreover, the Neural CDE performance remains roughly
constant as the percentage of dropped data increases, 
whereas the accuracy of the other models start to decrease.

\begin{table}[ht]	
	\centering
	\begin{tabular}{|c|c|c|c|}
        \hline
		Model & 30\% dropped & 50\% dropped 
		& 70\% dropped \\
		\hline
		GRU-ODE & $92.6 \pm 1.6$ & 
		$86.7 \pm 3.9$& $89.9 \pm 3.7$ \\
		GRU-$\Delta$t & $93.6 \pm 2.0$ &
		$91.3 \pm 2.1$ & $90.4 \pm 0.8$ \\
		GRU-D & $94.2 \pm 2.1$ & 
		$90.2 \pm 4.8$ & $91.9 \pm 1.7$ \\
		ODE-RNN & $95.4 \pm 0.6$ & 
		$96.0 \pm 0.3$ & $95.3 \pm 0.6$ \\
		Neural CDE & $\mathbf{98.7 \pm 0.8}$ & 
		$\mathbf{98.8 \pm 0.2}$ & 
		$\mathbf{98.6 \pm 0.4}$ \\
        \hline
	\end{tabular}
	\caption{Test accuracy 
	(mean\% $\pm$ std\%, computed across five runs) on
    CharacterTrajectories - Table 1 in \cite{FKLM20}}
	\label{FKLM20_Table1}
\end{table}

The performance of Neural CDEs is investigated through
further numerical experiments in \cite{FKLM20}.
In addition to performance, the ability to utilise 
memory-efficient adjoint-based backpropagation in Neural 
CDEs ensures that they use significantly less
memory than other approaches \cite{FKLM20}. It is also 
explained in \cite{FKLM20} how the Neural CDE approach 
can be applied to partially observed data.
Consequently Neural CDEs overcome some of the issues 
that can arise in the use of \textit{Recurren Neural 
Networks} (RNNs) for irregularly sampled or partially 
observed time series \cite{FKLM20}.
The subsequent work \cite{KLMY21} develops
Neural CDEs for \textit{online use}; that is, to learn and 
predict in real-time where new data arrives during 
inference. The key component is to replace the use
of a natural cubic spline for the control path $X$ by 
a new control signal satisfying four ideal requirements;
see \cite{KLMY21} for full details.

The Neural CDE approach developed in \cite{FKLM20} 
numerically solves the resulting CDE via evaluations of
the control path $X$. Consequently, Neural CDEs performance
drops for long time series, and the large number of 
forward operations within each training epoch result in 
prohibitive training time \cite{FKLMS21}.
In \cite{FKLMS21} these issues are overcome by using the
log-ODE method (cf. Section \ref{CDE_log-ODE} in this article)
to numerically solve the resulting CDE.
The main idea is that the log-ODE method offers a way to 
update the hidden state of a Neural CDE over large intervals,
dramatically reducing the effective length of the time series
\cite{FKLMS21}.
An additional benefit of this approach is its removal of the
requirement for the control path $X$ to be differentiable.
The resulting approach is termed 
\textit{Neural Rough Differential Equations} (Neural RDEs).
Neural RDEs still make use of the memory-efficient 
continuous-time adjoint backpropagation; the resulting
decrease in memory requirement is increasingly relevant 
as the time series length increases.

We detail the Neural RDE approach for a fully-observed 
yet potentially irregularly sampled time-series 
$\bx = \big( (t_0 , x_0) , (t_1 , x_1) , \ldots , 
(t_n , x_n) \big)$ for timestamps 
$t_0 < t_1 < \ldots < t_n$ and observations
$x_0 , x_1 , \ldots , x_n \in \R^p$.
Construct a piecewise linear interpolation 
$X :[t_0,t_n] \to \R^{p+1}$ such that, for each 
$i \in \{0, \ldots , n\}$, we have $X_{t_i} = (t_i,x_i)$.

For a choice of positive integer $m$ much smaller than $n$,
pick points $t_0 = r_0 < r_1 < \ldots < r_m = t_n$. 
Whilst it is not a requirement, it is typical to choose 
these points to be spaced equally apart. The choice and 
spacing are hyper-parameters of the model.
For a chosen depth hyper-parameter $N \in \Z_{\geq 1}$ 
we compute the truncated log signature 
$\LogSig^{(N)}_{r_i,r_{i+1}}(X)$ for each 
$i \in \{0, \ldots , m-1\}$.
The log-ODE method (see Section \ref{CDE_log-ODE} in 
this article) is then used to numerically solve
the CDE \eqref{NCDE_model_eqn} on the interval
$[t_0,t_n]$ by inductively
solving it over the interval $[r_i,r_{i+1}]$. That is,  
let $i \in \{0, \ldots , m-1\}$ and assume that the 
solution $z_t$ has already been defined for $t \in [0,r_i]$.
Consider the equation specified in 
\eqref{NCDE_model_eqn} over the
interval $[r_i,r_{i+1}]$ with initial condition $z_{r_i}$.
Then the solution is extended to interval $[t_0,r_{i+1}]$
by defining $z_{\tau} := \LogODE \Big(z_{r_i}, f_{\th}, 
\LogSig^{(N)}_{r_i,r_{i+1}}(X) , \tau \Big)$
(cf. \eqref{log-ODE_approx_func})
for $\tau \in (r_i,r_{i+1}]$.
Recall from Section \ref{CDE_log-ODE} that this only 
involves solving an ODE over $[r_i,r_{i+1}]$, which 
may be done using standard ODE solvers.
Figure \ref{figure1_FKLMS21} below gives a high level 
comparison of the CDE and RDE based approaches; it appears 
as Figure 1 in \cite{FKLMS21}

\definecolor{bittersweet}{rgb}{1.0, 0.44, 0.37}

\pgfmathsetmacro{\pathyshift}{0.8}
\global\def\rs{{0, 1, 2.1, 3.3, 4.5, 5.5, 6.5}}
\global\def\lsone{{2.15, 2.3, 2.1, 2.4, 2.1, 2.4}}
\global\def\lstwo{{2.1, 2.4, 2.2, 2.5, 2.2, 2.1}}
\global\def\lsthree{{2.3, 2.1, 2.15, 2.35, 2.5, 2.2}}
\global\def\lspathraise{0.7}
\global\def\data{{0.4, 0.7, 0.9, 0.5, 0.6, 0.9, 0.8, 0.5, 0.3, 0.3, 0.8, 1.0, 0.2, 0.4, 0.7, 0.5, 1.0, 1.0, 0.8, 0.4, 0.2, 0.6, 0.8, 0.5, 0.3, 0.5, 0.2, 0.1, 0.5, 0.2, 0.4, 0.6, 0.3, 0.2, 0.1, 0.7, 0.6, 0.2, 0.5, 0.4, 0.4, 0.1}}

\definecolor{colorls1}{HTML}{66C2A5}
\definecolor{colorls2}{HTML}{8DA0CB}
\definecolor{colorls3}{HTML}{FC8D62}

\def\arrnocomma {
    (0.0, 0.5903005787234183)
    (0.15853658536585366, 0.31548220294099943)
    (0.3170731707317073, 0.5723798869840881)
    (0.47560975609756095, 0.7592559931584444)
    (0.6341463414634146, 0.46000436695787333)
    (0.7926829268292683, 0.8268714937427233)
    (0.9512195121951219, 0.8165360374858748)
    (1.1097560975609757, 0.40954054026875836)
    (1.2682926829268293, 0.7450130964733412)
    (1.4268292682926829, 0.3631221102604148)
    (1.5853658536585367, 0.7482435510764813)
    (1.7439024390243902, 0.917384718293228)
    (1.9024390243902438, 0.5573926930403175)
    (2.0609756097560976, 0.7663748137356283)
    (2.2195121951219514, 0.6694620736634)
    (2.3780487804878048, 1.1176346402925919)
    (2.5365853658536586, 0.949415028614423)
    (2.6951219512195124, 0.3177597410442012)
    (2.8536585365853657, 0.6693951873464801)
    (3.0121951219512195, 0.9294900828561168)
    (3.1707317073170733, 0.9328267954184619)
    (3.3292682926829267, 0.663301293119932)
    (3.4878048780487805, 1.256064569672367)
    (3.6463414634146343, 0.472180216368223)
    (3.8048780487804876, 1.2501090850305931)
    (3.9634146341463414, 0.6636785039153019)
    (4.121951219512195, 0.650984855539094)
    (4.280487804878049, 0.7248161591552023)
    (4.439024390243903, 0.5481046159182468)
    (4.597560975609756, 1.0235344968993427)
    (4.7560975609756095, 1.0369378751495655)
    (4.914634146341464, 1.1262114265391723)
    (5.073170731707317, 0.3962487903451824)
    (5.2317073170731705, 0.9389413773558675)
    (5.390243902439025, 1.2096540128425803)
    (5.548780487804878, 1.079908936992647)
    (5.7073170731707314, 1.2109400195109703)
    (5.865853658536586, 0.37773305265847396)
    (6.024390243902439, 0.5808720577974682)
    (6.182926829268292, 0.5390596533228442)
    (6.341463414634147, 0.6221988967136689)
    (6.5, 0.7650561612631395)
}

\newcommand{\arrcomma}{
    (0.0, 0.5903005787234183), (0.15853658536585366, 0.31548220294099943), (0.3170731707317073, 0.5723798869840881), (0.47560975609756095, 0.7592559931584444), (0.6341463414634146, 0.46000436695787333), (0.7926829268292683, 0.8268714937427233), (0.9512195121951219, 0.8165360374858748), (1.1097560975609757, 0.40954054026875836), (1.2682926829268293, 0.7450130964733412), (1.4268292682926829, 0.3631221102604148), (1.5853658536585367, 0.7482435510764813), (1.7439024390243902, 0.917384718293228), (1.9024390243902438, 0.5573926930403175), (2.0609756097560976, 0.7663748137356283), (2.2195121951219514, 0.6694620736634), (2.3780487804878048, 1.1176346402925919), (2.5365853658536586, 0.949415028614423), (2.6951219512195124, 0.3177597410442012), (2.8536585365853657, 0.6693951873464801), (3.0121951219512195, 0.9294900828561168), (3.1707317073170733, 0.9328267954184619), (3.3292682926829267, 0.663301293119932), (3.4878048780487805, 1.256064569672367), (3.6463414634146343, 0.472180216368223), (3.8048780487804876, 1.2501090850305931), (3.9634146341463414, 0.6636785039153019), (4.121951219512195, 0.650984855539094), (4.280487804878049, 0.7248161591552023), (4.439024390243903, 0.5481046159182468), (4.597560975609756, 1.0235344968993427), (4.7560975609756095, 1.0369378751495655), (4.914634146341464, 1.1262114265391723), (5.073170731707317, 0.3962487903451824), (5.2317073170731705, 0.9389413773558675), (5.390243902439025, 1.2096540128425803), (5.548780487804878, 1.079908936992647), (5.7073170731707314, 1.2109400195109703), (5.865853658536586, 0.37773305265847396), (6.024390243902439, 0.5808720577974682), (6.182926829268292, 0.5390596533228442), (6.341463414634147, 0.6221988967136689), (6.5, 0.7650561612631395)
}

\newcommand{\arrarr}{
    {0.0, 0.5903005787234183}, {0.15853658536585366, 0.31548220294099943}
}

\global\def\xarrm{{0.0, 0.15853658536585366, 0.3170731707317073, 0.47560975609756095, 0.6341463414634146, 0.7926829268292683, 0.9512195121951219, 1.1097560975609757, 1.2682926829268293, 1.4268292682926829, 1.5853658536585367, 1.7439024390243902, 1.9024390243902438, 2.0609756097560976, 2.2195121951219514, 2.3780487804878048, 2.5365853658536586, 2.6951219512195124, 2.8536585365853657, 3.0121951219512195, 3.1707317073170733, 3.3292682926829267, 3.4878048780487805, 3.6463414634146343, 3.8048780487804876, 3.9634146341463414, 4.121951219512195, 4.280487804878049, 4.439024390243903, 4.597560975609756, 4.7560975609756095, 4.914634146341464, 5.073170731707317, 5.2317073170731705, 5.390243902439025, 5.548780487804878, 5.7073170731707314, 5.865853658536586, 6.024390243902439, 6.182926829268292, 6.341463414634147, 6.5}}

\global\def\yarrm{{0.5903005787234183, 0.31548220294099943, 0.5723798869840881, 0.7592559931584444, 0.46000436695787333, 0.8268714937427233, 0.8165360374858748, 0.40954054026875836, 0.7450130964733412, 0.3631221102604148, 0.7482435510764813, 0.917384718293228, 0.5573926930403175, 0.7663748137356283, 0.6694620736634, 1.1176346402925919, 0.949415028614423, 0.3177597410442012, 0.6693951873464801, 0.9294900828561168, 0.9328267954184619, 0.663301293119932, 1.256064569672367, 0.472180216368223, 1.2501090850305931, 0.6636785039153019, 0.650984855539094, 0.7248161591552023, 0.5481046159182468, 1.0235344968993427, 1.0369378751495655, 1.1262114265391723, 0.3962487903451824, 0.9389413773558675, 1.2096540128425803, 1.079908936992647, 1.2109400195109703, 0.37773305265847396, 0.5808720577974682, 0.5390596533228442, 0.6221988967136689, 0.7650561612631395}}

\global\def\xarrlen{0, 1, 2, 3, 4, 5, 6, 7, 8, 9, 10, 11, 12, 13, 14, 15, 16, 17, 18, 19, 20, 21, 22, 23, 24, 25, 26, 27, 28, 29, 30, 31, 32, 33, 34, 35, 36, 37, 38, 39, 40, 41}

\newcommand\mypath[1]{
    \draw [green!50, line width=0.5mm] plot [#1] coordinates {
        \arrnocomma
    };
}


\newcommand\ncdenrdediagram[1]{
    \resizebox{0.45\linewidth}{!}{
    \begin{tikzpicture}
        \pgfmathsetmacro{\logsigversion}{#1}
    
        \draw[black, thick, ->] (0, 0) -- (7, 0);
        
        \pgfmathsetmacro{\summaryh}{1.7}
        \pgfmathsetmacro{\summaryeps}{0.01}
        \pgfmathsetmacro{\summaryd}{0.5}
        \pgfmathsetmacro{\cdeh}{2.5}
        \pgfmathsetmacro{\cded}{1}
        
        \draw (\rs[0], -0.05) -- (\rs[0], 0.05);
        \draw (\rs[6], -0.05) -- (\rs[6], 0.05);
        \node at (\rs[0], -0.25) {\small $t_0$};
        \node at (\rs[6], -0.25) {\small $T$};

        \ifthenelse{\logsigversion=1}
            {
                \foreach \d in {1, 2, 3, 4, 5, 6} {
                    \draw [fill=red!10, draw=red!50, rounded corners, dashed] (\rs[\d - 1] + \summaryeps, \summaryh) rectangle (\rs[\d] - \summaryeps, \summaryh + \summaryd) node[pos=.5] {};
                    
                    \draw [black!50, ->] (0.5 * \rs[\d - 1] + 0.5 * \rs[\d], \summaryh + \summaryd) -- (0.5 * \rs[\d - 1] + 0.5 * \rs[\d], \cdeh);
                    
                    \draw [black!50, ->] (0.5 * \rs[\d - 1] + 0.5 * \rs[\d], \cdeh + \cded) -- (0.5 * \rs[\d - 1] + 0.5 * \rs[\d], \cded + \cdeh + 0.5);
                }
            }
            {
            }
            
        \pgfmathsetmacro{\cdeeps}{0.}
        \ifthenelse{\logsigversion=0}
            {
                \draw [fill=orange!10, draw=orange!50, rounded corners, dashed] (\rs[0], \cdeh - \cdeeps) rectangle (\rs[6], \cdeh + \cded - \cdeeps) node[pos=.5] {Neural CDE};
            }
            {
                \draw [fill=orange!10, draw=orange!50, rounded corners, dashed] (\rs[0], \cdeh) rectangle (\rs[6], \cdeh + \cded) node[pos=.5] {Neural RDE};
            }
        
        \ifthenelse{\logsigversion=1}{
                \mypath{}
            }
            {
                \mypath{smooth}
            }
        \foreach \coord [count=\i] in \arrcomma {
            \node at \coord [circle, draw=black, fill=green!50, inner sep=0.5pt] {};
        }
        \foreach \i in \xarrlen {
            \ifthenelse{\logsigversion=1}{
                \draw [black!50, ->] (\xarrm[\i], \yarrm[\i]) -- (\xarrm[\i], \summaryh);
            }
            {
                \draw [black!50, ->] (\xarrm[\i], \yarrm[\i]) -- (\xarrm[\i], \cdeh);
                \draw [black!50, ->] (\xarrm[\i], \cdeh + \cded) -- (\xarrm[\i], \cdeh + \cded + 0.5);
            }
        }
        
        \pgfmathsetmacro{\hiddenh}{1.5}
        \draw[blue!50, line width=0.5mm, cap=round] (\rs[0], 3 + \hiddenh) .. controls (1.5, 1.6 + \hiddenh) and (\rs[4], 4 + \hiddenh) .. (\rs[6], 2.5 + \hiddenh);
        
        \node at (8, 0) {\footnotesize Time};
        \ifthenelse{\logsigversion=0}{
                \node (cde) at (8, 3) {CDE}; 
                \node (path) at (8, 0.8) {\footnotesize \begin{tabular}{c}Path, $X_t$\\(smoothed)
                \end{tabular}};
            }
            {
                \node (path) at (8, 0.8) {\footnotesize \begin{tabular}{c}Path, $X_t$\\ \end{tabular}};
            }
        \ifthenelse{\logsigversion=1}{
            \node (summarisations) at (8, 1.95) {\footnotesize Summaries}; 
            \node (rde) at (8, 3) {RDE}; 
        }{}
        \node (hidden) at (8, 4.2) {\footnotesize Response};
    
        \ifthenelse{\logsigversion=1}
            {
                \draw[->] (path) -- (summarisations);
                \draw[->] (summarisations) -- (rde);
                \draw[->] (rde) -- (hidden);
            }
            {
                \draw[->] (path) -- (cde);
                \draw[->] (cde) -- (hidden);
            }

    \end{tikzpicture}
    }
}


\begin{figure}[ht]
    \centering
    \subfloat[CDE approach proposed in \cite{FKLM20}
    in which data is smoothly interpolated and pointwise 
    derivative information is used to drive the CDE.]
    {\ncdenrdediagram{0}}
    \qquad
    \subfloat[RDE approach proposed in \cite{FKLMS21}
    in which local interval summarisations of the data 
    given by truncated log signatures are computed and 
    used to drive the response over the interval.]
    {\ncdenrdediagram{1}}
    \caption{
    Figure 1 in \cite{FKLMS21}.
    Reproduced with permission. 
    Copyright held by authors of \cite{FKLMS21}.}
    \label{figure1_FKLMS21}
\end{figure}

Numerical experiments are conducted in \cite{FKLMS21}
to illustrate the benefits of the Nerual RDE approach 
in the setting of long time series. 
One task considers the EigenWorm 
dataset from the UEA archive \cite{BBKLL17}.
This consists of time series of length 17 984 and 
6 channels (including time) corresponding to the movement
of a roundworm. 
The task is to classify each worm as either wild-type
or one of four mutant-type classes.

The regular sampling allows the choice of $t_i := i$. 
Neural RDEs are run for all depths $N \in \{2,3\}$ and 
all step sizes in $\{ 2^j \mid j \in \{1 , \ldots , 10\} \}$.
A Neural CDE model and an ODE-RNN model (introduced in 
\cite{CDR19}) are used for baseline comparison.
Precise details of hyper-parameter selection, optimisers,
and other model parameter choices may be found in
Appendix C of \cite{FKLMS21}.

The experiment is run three times for each model and
each hyper-parameter combination, with the mean and 
standard deviation of the accuracy recorded. The results
are presented in Table \ref{FKLMS21_Table1} below, which is 
a version of Table 1 in \cite{FKLMS21}.
For brevity, the results are only displayed for 
a sub-selection of step sizes considered; the full results
can be found in Table 7 in Appendix D of \cite{FKLMS21}.
Amongst the results included in Table \ref{FKLMS21_Table1}
we observe that the best accuracy results are returned by
Neural RDE models, and that the accuracy achieved is a
noticeable improvement over that resulting from either of
ODE-RNN or Neural CDE models.
This improvement over Neural CDE performance comes at
the cost of Neural RDE models being more memory intensive.
However, Neural RDE models remain significantly less
memory intensive than ODE-RNN models.

\begin{table}[ht]	
	\centering
	\begin{tabular}{|c|c|c|c|}
        \hline
		\textbf{Model} & \textbf{Step} & 
		\textbf{Test Accuracy (\%)} &
		\textbf{Memory (Mb)} \\
		\hline
		  & 4 & $35.0 \pm 1.5$ & 3629.3\\
		 \textbf{ODE-RNN} & 32 & $32.5 \pm 1.5$ & 532.2\\
		  & 128 & $47.9 \pm 5.3$ & 200.8 \\
		  \hline
		  & 4 & $66.7 \pm 11.8$ & 46.6\\
		 \textbf{Neural CDE} & 32 & $64.1 \pm 14.3$ & 8.0 \\
		  & 128 & $48.7 \pm 2.6$ & 3.9  \\
		  \hline
		  & 4 & $\mathbf{83.8 \pm 3.0}$ & 180.0 \\
		 \textbf{Neural RDE (depth 2)} & 32 & $67.5 \pm 12.1$ 
		 & 28.1 \\
		  & 128 & $\mathbf{76.1 \pm 5.9}$ & 7.8 \\
		  \hline
		  & 4 & $76.9 \pm 9.2$ & 856.8 \\
		 \textbf{Neural RDE (depth 3)} & 32 & 
		 $\mathbf{75.2 \pm 3.0}$ & 134.7 \\
		  & 128 & $68.4 \pm 8.2$ & 53.3 \\
          \hline
	\end{tabular}
	\caption{Test accuracy 
	(mean\% $\pm$ std\%, computed across three repeats) 
	and memory usage on
    EigenWorms - Part of Table 1 in \cite{FKLMS21}}
	\label{FKLMS21_Table1}
\end{table}

In \cite{FKLLO21} Neural CDEs are combined  
with \textit{Neural Stochastic 
Differential Equations} (Neural SDEs) (see 
\cite{RT19,CDLW20,HHMR20}, for example) to show that
the classical approach to fitting SDEs can be approached
as a special case of (Wasserstein) GANs. This direct
extension makes no reference to pre-specified 
statistics or density functions. As a consequence, in 
the infinite data limit \textit{any} SDE may be learnt.
Full details of this approach and several subsequent 
applications may be found in \cite{FKLLO21}.
Moreover, the follow-up work \cite{FKLL21} introduces
several technical innovations to improve both model 
performance and training speed for Neural SDEs.

Neural RDEs reduce the computational cost of the log-ODE 
method by learning the vector field $\tilde{f}$ 
(cf. \eqref{log-ODE_vf_def} in Section \ref{CDE_log-ODE}) 
as a neural network rather than computing it directly.
The paper \cite{BEN24} introduces 
\textit{Log-Neural Controlled Differential Equations}
(Log-NCDEs)
which are Neural RDEs that do \textit{not} ignore the structure 
of the vector field $\tilde{f}$. 
Indeed Log-NCDEs directly compute the vector field $\tilde{f}$
via the iterated Lie brackets of the original vector field $f$.
Whilst Log-NCDEs are necessarily more costly than 
Neural CDEs/RDEs, it is shown in \cite{BEN24} that 
this exploitation of the underlying 
regularity/structure of the vector fields is demonstrated to 
achieve a higher average test set accuracy than Neural CDEs, 
Neural RDEs, S5, and the linear recurrent unit on a range of 
multivariate time series classification benchmarks; the 
interested reader is directed to
\cite{BEN24} for full precise details.

The 
\href{https://github.com/datasig-ac-uk/signature_applications/tree/master/controlled_neural_differential_equations}{Neural CDE}
notebook
contains an introduction to the technology developed in
both \cite{FKLM20} and \cite{FKLMS21}. 
In particular, the use of the 
\href{https://github.com/patrick-kidger/torchcde}
{\textit{torchcde}} package for 
time series classification and use of the log-ode method 
are demonstrated. By working through the provided 
examples the reader will hopefully gain the experience and 
familiarity required to utilise the Neural CDE approach 
within their own work/projects.

\section{Speech Emotion Recognition}
\label{speech_emot_rec}
Recognising emotions from audio streams has numerous
real-world applications, ranging from the use of
voice-based assistants such as Alexa, Siri and 
Google Home, to being used to assist the detection of 
psychiatric disorders such as bipolar disorder. 
Speech emotion recognition (SER) can use both voice
characteristics and linguistic content to extract emotion.

A key step in SER is determining an effective and 
efficient representation for emotional utterances 
or speech segments.
The complexity of emotional expressions make this 
challenging; the same phrase can indicate completely 
different emotions depending on the tone.

Many SER systems extract frame-level acoustic features 
(e.g. frequency, zero crossing rate, jitter, etc), 
called Low-level Descriptors (LLD) for 
utterances of various lengths, and then apply a set of 
statistical pooling functions (mean, max, variance, etc) 
to obtain fixed-size utterance-level features.
Whilst the global characteristics are captured via such 
pooling functions, temporal variations of speech signals 
are not effectively extracted, diluting important 
regional information \cite{AP17}.

Recently, various types of deep learning models have been 
used to effectively model such temporal information,
including both convolutional and recurrent neural networks 
(CNN and RNN respectively) \cite{NV17,BMZ17,LT18,CHLRSW18}.
The complex network architectures and large number of 
parameters involved mean these models are difficult to
build and tune, time-consuming to train, and often 
require expensive computing resources.

In \cite{LLNNSW19}, the authors focus on the acoustic 
characteristics of the speech signal in order to recognise 
underlying emotions.
Motivated by the inherent sequential structure of emotions 
conveyed by speech, the authors explore the use of 
path signatures for modelling temporal sequences of emotional 
utterances. They demonstrate that this method incorporates 
both the short-term characterisation at the frame-level 
and the long-term aggregation at the utterance-level.
Their approach operates on minimally hand-engineered 
filter-bank energy features to help avoid the overfitting 
issues encountered
by other alternative approaches, see \cite{BMZ17}.

Inspired by \cite{JLNY16}, a hierarchical tree 
structure is used for path signature features and 
tree-based convolutions are adopted
for both the integration of global, regional and 
local information, and for filtering irrelevant and 
redundant information.
The use of dyadic intervals allows a sufficiently fine 
description of both the global and local information 
to be captured, whilst avoiding the exponential growth 
in dimensionality associated with higher order signature 
terms.

\begin{figure}[ht]
    \begin{center}
        \includegraphics[width=\textwidth]
        {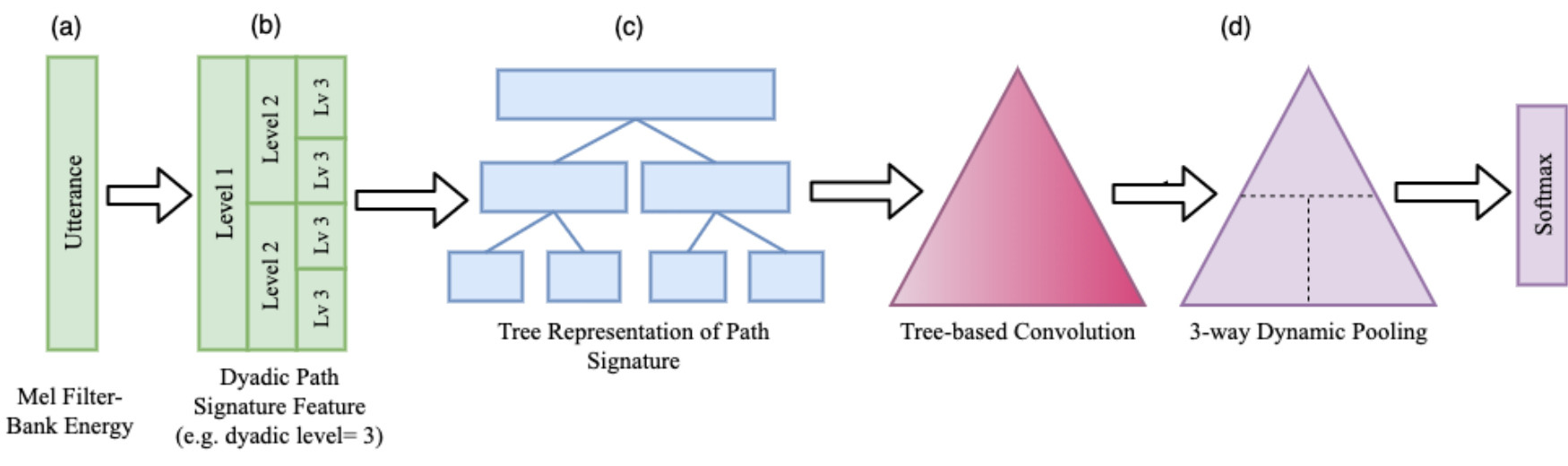}
        \caption{Overview of the proposed SER system
        - Figure 1 in \cite{LLNNSW19}. Reprinted 
        with permission.}
        \label{tree_structure_figure}
    \end{center}
\end{figure}

Figure \ref{tree_structure_figure} illustrates an 
overview of the proposed SER system as explained
below.
\begin{enumerate}[label=(\Alph*)]
    \item A stream of frame-level energy features
    is extracted from an utterance
    \item The entire stream of frames within the
    utterance is segmented to dyadic paths. A truncated
    signature is extracted from each dyadic path
    \item The collection of the signatures of
    each dyadic path is transformed to the dyadic 
    path-tree signature representation
    \item Tree-based convolution and dynamic 
    pooling are applied to learn the underlying 
    structure. An output layer is added for final
    classification.
\end{enumerate}
The signatures are truncated to order 3, and 
a dyadic level of 4 is selected.
The collection of all dyadic pieces forms the 
\emph{dyadic path-tree} from which features at 
different resolutions and structural
information is more easily extracted. Each node 
is given an associated weight matrix depending on 
the nodes relative position, before dynamic pooling
is used to pool the features, again depending on their 
position in the tree. A final softmax layer is added for 
classification, see \cite{LLNNSW19} for full details.
 
The four emotional categories \emph{Angry, Happy, Neutral} 
and \emph{Sad} form the classes for the classification.
Effectiveness is evaluated using data from the Interactive 
Emotional Dyadic Motion Capture (IEMOCAP) database 
\cite{BBCKKLLMN08}.
It comprises approximately 12 hours of audio-visual 
recordings performed by 10 actors. Each recording is 
split in to 5 sessions, and each session is composed of 
two actors, one male and one female. Overall it contains 
10039 (manually segmented) utterances with an 
average duration of 4.5 seconds. The database can be 
further divided into an improvised speech data set and 
a scripted data set. 
In \cite{LLNNSW19}, the improvised data set is chosen 
since the scripted speech exhibits strong correlation 
with the manually labelled emotions leading to bias 
over linguistic content learning.

The openSMILE toolkit \cite{ESW10} is used for extracting 
40-dimensional features from each utterance, with an 
additional dimension added for time.
A \emph{leave-one-speaker-out} scheme is chosen for 
evaluation. This means that for each instance A, 
training is carried out using all instances except A, 
before the learnt model is then evaluated on A itself.
The cross-entropy cost is considered for training, and 
the precise hyper-parameters used are listed in table 1 
of \cite{LLNNSW19}. 
The final SER performance is evaluated using widely 
adopted metrics: weighted accuracy (WA), which is 
the overall classification accuracy; 
and unweighted accuracy (UA), which averages accuracy 
of each emotion category. A range of neural network 
models are considered as baselines, and the results 
are summarised in the Table \ref{SER_results}.

\begin{table}[ht]
	\centering
	\begin{tabular}{|c|c|c|}
        \hline 
		\textbf{Model}& \textbf{UA} & \textbf{WA} \\ 
		\hline	
		COVAREP \cite{GLMS16} & 51.84 & 49.64 \\ 
		DNN-ELM \cite{LT15} & 52.13 & 57.91 \\
		LSTM (Speech) \cite{GLMS16} & 51.85 & 51.94 \\
		LSTM (Glottal) \cite{GLMS16} & 54.56 & 52.82 \\
		Attentive CNN \cite{NV17} & 56.83 & 61.95 \\
		\textbf{PTS-CNN} \cite{LLNNSW19} & 53.03 & 58.90 \\
        \hline
	\end{tabular}
	\caption{Model Performances - Table 2 in \cite{LLNNSW19}}
	\label{SER_results}
\end{table}
\vskip 4pt

The relatively simple Path-Tree-Signature 
based CNN (PTS-CNN) model in \cite{LLNNSW19} 
achieves comparable results 
to networks of complex design. The PTS-CNN
model can deal with utterances of variable length
without preprocessing and uses the openSMILE
toolkit \cite{ESW10} to obtain its features.
This is in stark contrast to the Attentive-CNN 
model, which requires utterances to be processed
to all have the same length and uses the 
tailor-made eGeMAPS feature set \cite{ABDEELNSSST16}. 
With minimal model tuning, or manual engineering,
the PTS-CNN model yields comparable results to 
complex neural network models making use of a 
range of heavily engineered emotion features
(COVAREP, LSTM \cite{GLMS16}).

\section{Health Applications}
\label{health_app}

\subsection{Bipolar and Borderline Personality Disorders}
\label{bipolar}
Historically, the diagnosis of psychiatric disorders 
has been hampered by the inherent inaccuracy of 
retrospective recall of mood states. 
The influx of development in mobile technology (such as 
mobile phones, smart watches, fitbits etc) has allowed 
momentary assessment to obtain more precise 
measures of psychopathology and highlighted the 
shortcomings of current diagnostic categories 
\cite{AGGLS18}. The NIMH Research Domain Criteria 
(RDoC) propose a new data-driven approach bottom up 
approach to diagnosis. The oscillatory nature of 
psychiatric symptomatology poses a significant 
analytic challenge.

In \cite{AGGLS18}, signature-based machine learning 
models are proposed to analyse data obtained from a 
clinical study in \cite{BCDOPSTV16}.
The study explored daily reporting of mood in 
participants with bipolar disorder, borderline 
personality disorder and healthy volunteers. 
The two problems tackled were the classification 
problem of classifying participants on the basis 
of their mood, and the time-series forecasting problems 
of predicting the participants mood the following day.

Mood data was captured from 130 individuals; 48 of whom 
were diagnosed with bipolar disorder (BD), 31 were 
diagnosed with borderline personality 
disorder (BPD) and 51 were healthy. For a minimum of 
two months, the participants rated their mood across 
six different categories (anxiety, elation, sadness, 
anger, irritability and energy) using a 7-point 
Likert scale with values from 1 (not at all) to 
7 (very much). The data was split into streams of 20 
consecutive observations, which did not necessarily 
arise from 20 consecutive days due to failure of 
participants recording. This generated 733 streams of 
data, which were randomly split into 513 training 
instance streams and 220 testing instance streams.
 
\begin{figure}[ht]
\centering
\includegraphics[scale=0.9]{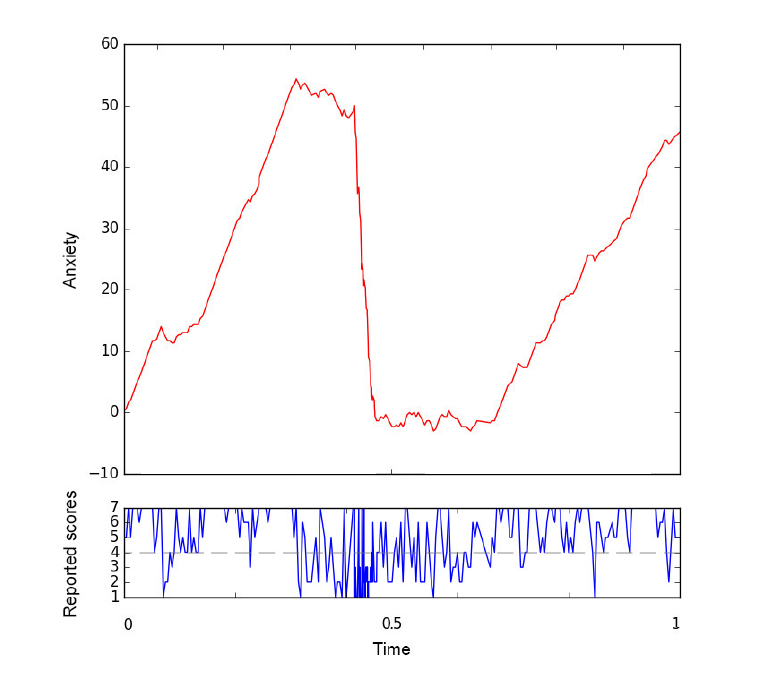}
\caption{Normalised anxiety scores of a participant 
with bipolar disorder (above), which were calculated 
using the reported scores (below). High 
levels of reported scores correspond to upward trends, 
low levels of reported scores correspond to downward 
trends and periods of time of high oscillations in the 
reported scores are represented by oscillations in the 
path --- Figure 1 in \cite{AGGLS18}. Unaltered. 
Licensed under a 
\href{https://creativecommons.org/licenses/by/4.0/}
{Creative Commons Attribution 4.0 International License}.}
\label{example_mood_path}
\end{figure}

The streams were identified, after normalisation, as 
7-dimensional paths (one dimension for time, 6 dimensions 
for the moods) and giving the input-output pairs 
$\big( R_i , Y_i \big)$ where $R_i$ 
is the 7-dimensional path of participant $i$. 
Figure \ref{example_mood_path} shows the normalised 
anxiety scores of a participant with BD and the 
associated two-dimensional path.
Intuitively, if this two-dimensional path has an 
upward trend than the participant scores are greater 
than 4 in that category.
Conversely, a downward trend corresponds to scores 
lower than 4. Drastic mood changes from day to day
will result in a highly oscillating normalised path,
whereas a more stable day to day mood will result
in a more stable normalised path.

\begin{figure}[ht]
\centering
\includegraphics[scale=1.0]
{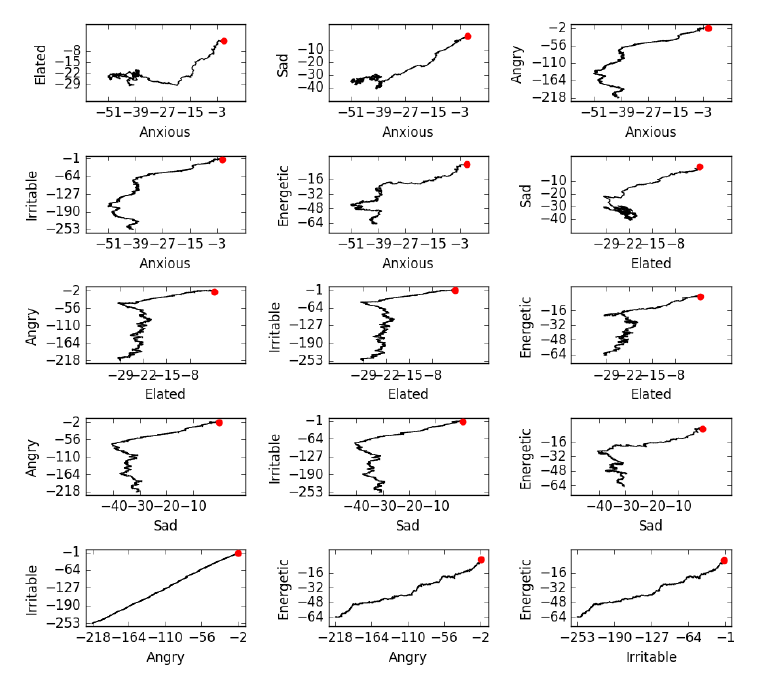}
\caption{Normalised scores of each category plotted 
against all other categories, for a participant with 
bipolar disorder. The red point indicates the starting 
point and each plot has its own scale
- Figure 2 in \cite{AGGLS18}. Unaltered. Licensed 
under a 
\href{https://creativecommons.org/licenses/by/4.0/}
{Creative Commons Attribution 4.0 International License}.}
\label{mood_change_order}
\end{figure}

Figure \ref{mood_change_order} shows the normalised 
scores of each category plotted against all other 
categories, clearly illustrating 
the order in which a participants mood changes. 
For example, consider the Angry vs. Elated plot 
(third row, first column).
The path starts at $(0,0)$ and initially moves left, 
showing that the participant is becoming less elated 
while their anger score remains
almost constant. Suddenly the period of low elation 
stops and the participant begins recording lower 
anger levels, and these
low levels of anger persist for the rest of the path.
On the other hand, the Angry vs. Irritable plot 
(fifth row, first column) is essentially a straight line, 
revealing that the levels of
anger and irritability are highly correlated and roughly 
equal for this participant.

The truncated signatures 
$\Pi_n \big( S( R_i) \big) = S^{(n)}(R_i)$ for 
$n=2,3,4$ are used to provide new input-output pairs. 
These truncated signatures will, in particular, 
capture the order of a participants mood change 
as illustrated in Figure \ref{mood_change_order}.
The strategies for each task are broadly similar.

For the classification, the participant of interest 
was removed from the data before softmax regression 
was used on the remaining pairs
$\Big\{ \big(  S^{(n)} (R_i), Y_i \big) \Big\}$.
Removing the participant of interest from the 
training reduced the risk of overfitting and 
increased the models robustness to new data. 
The model was then tested on 20-observations periods 
from the participant of interest, with the 
proportion of the periods of time for which they were 
classified as each of healthy, bipolar or borderline 
personality recorded.

Earlier work \cite{BCDOPSTV16} already established 
there are differences in mean mood scores between the 
overall groups, hence classifying the streams of 20 
consecutive observations on the basis of comparison 
with the mean score in each mood category was used as 
a baseline model for comparison.

The signature method significantly out-performed the 
naive mean-based model; the mean-based model 
classified $54\%$ of participants correctly, whilst the 
signature model correctly classified $74.85\%$ of 
participants. Incorrect predictions were mainly found 
in the split between bipolar disorder and border 
personality disorder participants, whereas the model 
very clearly distinguished healthy participants from 
the group. Full details may be found in \cite{AGGLS18}.

For the mood prediction, mean absolute error (MAE) 
regression was used on the input-output pairs 
$\Big\{ \big( S^{(n)}( R_i), Y_i \big) \Big\}$.
For testing, a prediction $\hat{Y}_i$ was deemed correct 
if $\big\lvert \hat{Y}_i - Y_i \big\rvert \leq 1$, 
reflecting 
that we are only interested in predicting the correct 
class, rather than capturing the exact score.
The model is benchmarked against the simple model 
of predicting that the next days score will be the 
same as the previous. The results are summarised in 
Table \ref{mood_predict_results}.
More detailed performance results 
for the signature-based
method can be found in Table 3 in \cite{AGGLS18}

\begin{table}[ht]	
\centering
    \begin{tabular}{|c|c|c|c|}
        \hline
        \textbf{Model} & \textbf{Healthy Accuracy $(\%)$}&
        \textbf{Bipolar Accuracy $(\%)$} &
        \textbf{Borderline Accuracy $(\%)$}  \\ 
        \hline		
        Same-as-prior & 61-92 & 46-67 & 44-62 \\
        \textbf{Signature} & \textbf{89-98} &
        \textbf{82-90} & \textbf{70-78} \\
        \hline
    \end{tabular}
\caption{Comparison of mood prediction models ---
Summary of Table 3 in \cite{AGGLS18}}
\label{mood_predict_results}
\end{table}

The subsequent work \cite{LLNSTWW20} extends the use 
of path signature techniques to the classification of 
Bipolar disorder and Borderline Personality disorder 
using multi-modal datasets.
The dataset considered comes from the Automated 
Monitoring of Symptoms Severity (AMoSS) study 
\cite{AGGLS18,BCDOPSTV16}
in which volunteers self-monitored their own daily mood 
via a smartphone app in conjunction with a range of 
wearable devices. Among the 139 participants, 53 had 
been diagnosed with BD, 33 had been diagnosed with BPD, 
and 53 had neither (termed healthy control, HC).
Halfway through the study, 62 participants were 
interviewed by 2 clinicians and 2 psychology graduates 
to gather feedback about the scheme. The interviews 
were either conducted in-person or by telephone.
The interviews were semi-structured with the topics 
remaining within the scope of
\begin{itemize}
	\item Users experience of the smartphone app,
	\item Users experience of the wearable device, 
	\item Benefits of participation in the study, and
	\item Potential improvements.
\end{itemize}
The topics were non-clinical, with no direct effort 
made to establish either a diagnosis or the 
participants mental state at the time of the interview. 

The AMoSS interview (AMoSS-I) dataset considered in 
\cite{LLNSTWW20} consists of 50 randomly sampled 
interviews that were initially transcribed by the same 
interviewers. An alignment procedure was applied to 
the audio recordings and manual transcribed text,
as detailed in \cite{LLNSTWW20}. In total, 67 features 
are extracted from the processed dataset including 28 
linguistic features denoted
as \textbf{LING} (such as mean sentence length, number 
of first person pronouns),
19 semantic content features \textbf{CNT} (such as number of 
words categorised as a certain emotion),
6 dialogue features \textbf{DIAL} (such as number of 
short pauses of less than half a second, number of 
pauses of longer than half a second),
3 features dealing with interruptions and talking 
over others, and 11 averaged features representing 
each pause. Full details of these features their 
extraction may be found in Section 3 of \cite{LLNSTWW20}.
 
A \textit{leave-one-participant-out} evaluation scheme 
is selected and combined with logistic regression with 
the aim of classifying whether a participant was 
BD, BPD or HC based on their interview.
The signature transform truncated to order 3 was 
applied to the path given by each extracted feature. 
The correlation of the features in the training data 
were ranked via a psychological scale (so-called 
International Personality Disorder Examination (IPDE) 
score, see \cite{ABBCCDDFLS94}) and only 
those features with a $p$-value below $0.001$ were 
retained. The selected features are then fed to the 
classifier for 3 separate binary tasks:
\begin{itemize}
	\item BD vs. HC,
	\item BPD vs. HC, and
	\item BD vs. BPD.
\end{itemize}
Three separate experiments are conducted; one in which 
features are only extracted from the speech of the 
participant, one in which features are only extracted 
from the interviewer, and one in which features are 
extracted from the speech of both.

\begin{figure}[ht]
    \begin{center}
        \includegraphics[width=\textwidth]
        {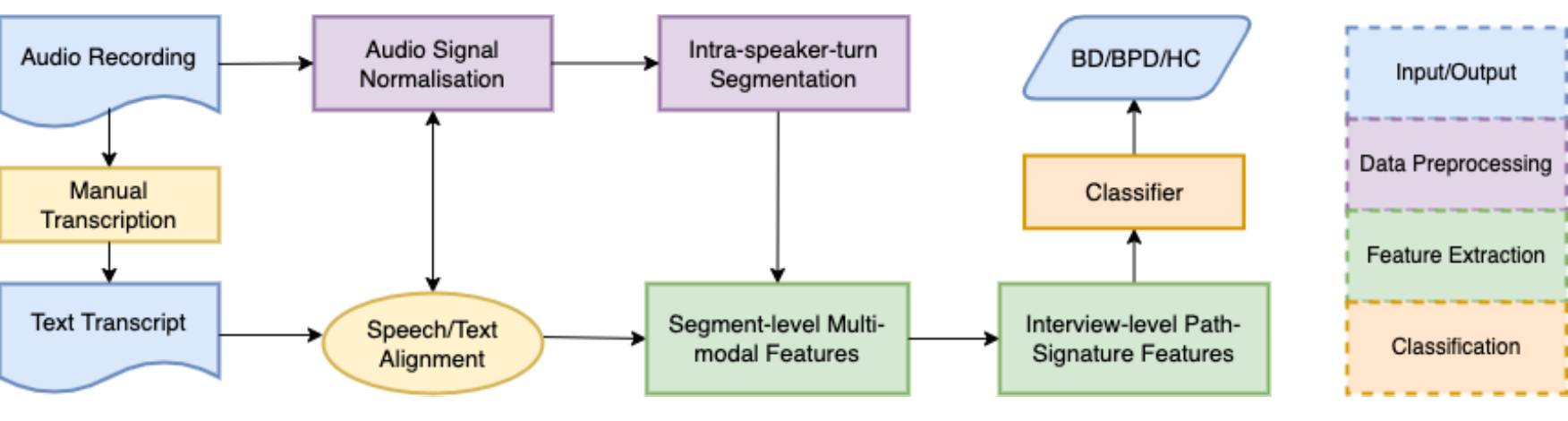}
        \caption{Data pipeline including three stages of 
        data preprocessing (in {\color{purple}purple}), 
        feature extraction (in {\color{green}green})
        and classification (in {\color{orange}orange}).
        Variant of Figure 2 in \cite{LLNSTWW20}.
        Reprinted with permission.}
        \label{fig:data-pipeline}
    \end{center}
\end{figure}

The results for the classification task are summarised 
in Table \ref{interview_auroc_scores}.
The scores presented are the average of the 
AUROCs across all interviews. 
Extracting features from the participant alone 
results in a good classifier for the task.
Given that the interviews were non-clinical and 
not aimed at establishing the mood of a participant, 
it is not surprising that the features extracted 
from the interviewer perform badly.

\begin{table}[ht]	
\centering
    \begin{tabular}{|c|ccc|}
        \hline
        & & \textbf{AUROC} & \\
        \textbf{Subject} & \textbf{BD vs. HC} & 
        \textbf{BPD vs. HC} & \textbf{BD vs. BPD} \\
        \hline	
        Participant & \textbf{0.810} & \textbf{0.733} & 
        \textbf{0.817} \\
        Interviewer & 0.304 & 0.473 & 0.231 \\
        Both & 0.494 & 0.431 & 0.657 \\
        \hline
    \end{tabular}
\caption{Average AUROC for each binary classification ---
Table 3 in \cite{LLNSTWW20}}
\label{interview_auroc_scores}
\end{table}

Ablation experiments were conducted to examine 
the affect of removing each feature type. The 
results are summarised in
Table \ref{ablation_results}.
The \textbf{LING} features are the biggest contributors; 
in fact, when they are removed the $p$-value threshold
has to be made $5$ times larger to allow \textit{any} 
features to be selected. The results illustrate that 
combining different types of features works better 
than relying on any individual type.

\begin{table}[ht]	
\centering
    \begin{tabular}{|cccc|}
        \hline
        \textbf{Features} & \textbf{BD vs. HC} & 
        \textbf{BPD vs. HC} & \textbf{BD vs. BPD} \\
        \hline	
        All & $0.810^{\ast \ast}$ & $0.733^{\ast \ast}$ & 
        $0.817^{\ast \ast}$ \\
        All - \textbf{CNT} & $0.810^{\ast \ast}$ & 
        $0.733^{\ast \ast}$ & $0.787^{\ast \ast}$ \\
        All - \textbf{Dial} & $0.768^{\ast \ast}$ & 
        $0.733^{\ast \ast}$ & $0.811^{\ast \ast}$ \\
        All - \textbf{LING} & $0.625^{\ast}$ & 
        $0.578^{\ast}$ & $0.669^{\ast}$ \\
        All - \textbf{LING} - \textbf{CNT} &
        $0.642^{\ast}$ & $0.703^{\ast}$ & $0.604^{\ast}$ \\
        All - \textbf{LING} - \textbf{Dial} &
        $0.442^{\ast}$ & $0.429^{\ast}$ & $0.550^{\ast}$\\
        All - \textbf{CNT} - \textbf{Dial} &
        $0.768^{\ast \ast}$ & $0.733^{\ast \ast}$ &
        $0.763^{\ast \ast}$ \\
        \hline
    \end{tabular}
\caption{Feature ablation results (AUROC) for each task; 
$p$-value used for feature selection:
$\ast \ast~<0.001$ and $\ast~ <0.005$.
Table 4 in \cite{LLNSTWW20}.}
\label{ablation_results}
\end{table}

The subsequent work \cite{LLSVWW21} investigates the use
of the invisibility reset transform 
(cf. \eqref{inv_reset}) to effectively model the 
conversation dynamics by enhancing the signature
representation of conversation speech.
Further, the subsequent work \cite{GLSW22} uses an 
approach based upon the log signature to overcome the 
common issue of missing responses.
The central idea is to record the omission of data as a 
new channel, and subsequently summarise the resulting 
streams via the log signature.

\subsection{Alzheimer's Disease}
\label{alzh}
Alzheimer's disease (AD) is the most common form 
of dementia in older people, affecting $6\%$ of 
the population aged over 65. 
Memory loss is a mild cognitive impairment (MCI) 
commonly indicating early stage of AD. MCI is 
diagnosed with no 
evidence of dementia, and MCI does not always 
progress to dementia. However, the progressive 
and irreversible loss of 
brain function caused by AD suggests that potential 
drug therapies should be used as early as possible.
Hence there is a demand for reliable early predictions 
of which individuals will develop AD.

Using brain imaging data to derive features for 
predicting a diagnosis of AD has been investigated 
in many works,
see \cite{ABDGKKMMMRRS15,ABILNPS17,KLMMSSSTVWWW11} 
for example.
In \cite{GLM19}, signature-based techniques are 
developed to distinguish between healthy and 
Alzheimer groups from preprocessed MRI data.
Data from the 
\href{https://tadpole.grand-challenge.org/}
{TADPOLE grand challenge competition} in 2019 
was used for the classification tasks considered in 
\cite{GLM19}.
The 1737 participants from the TADPOLE data set are
split into three groups; the 688 who have a diagnosis 
of Alzheimer's disease at some time (labelled $AD$), the 
424 who always have a healthy diagnosis (labelled $NL$), and 
the 484 who always have an MCI diagnosis (labelled $MCI$).
This results in the comparison of participants whose
diagnosis converts to Alzheimer's disease with those
whose diagnosis remains unchanged.

The training data is selected in \cite{GLM19} by selecting
participants with a first diagnosis of AD at 36 months 
from baseline (start of monitoring). 
Each selected participant is further required to have at
least four measurements of all the variables WholeBrain (w), 
Hippocampus (h) and Ventricles (v) 
(see \cite{GLM19} for details of these variables) 
in the 24 months since baseline, with one measurement 
at the 24 month time point.
Finally, each selected participant from the $AD$ set 
is required to have matching counterparts in both the 
$NL$ and $MCI$ sets. Given a member $x \in AD$, an 
individual $y \in NL$ or $y \in MCI$ qualifies as a 
counterpart $x$ if the following requirements are 
satisfied. Firstly, the ages of $y$ and $x$ must differ 
by no more than 5 years. 
Secondly, the diagnosis for $y$ must have remained 
unchanged for the 72 months since their first (baseline)
measurement. 
Finally, $y$ must have at least four measurements of all
the variables w, h, and v up to month 24, again including
a measurement at month 24 itself.

The test data is selected making use of measurements 
starting from 12 months since the first baseline 
measurement. To reflect the real world the test data
imposes no matching/counterparting between individuals 
in $AD$ and those in $NL/MCI$. The measurements used for
analysis of the test data are at 12, 24 and 36 months 
for all three sets ($AD$, $NL$, $MCI$).

Two classification tasks are considered in \cite{GLM19};
AD vs. NL and AD vs. MCI. Binary logistic regression, 
which models the log probabilities of the outputs as 
linear functions of the inputs. This results in the 
model being more easily interpreted than more
sophisticated classifiers such as random forests.
The input features
are selected via Lasso regularisation. The input for LASSO
is a vector formed from the three variables, 
WholeBrain (w), Hippocampus (h) and Ventricles (v), 
and either the signature or the log-signature of the path 
determined by the variables $v$, $h$, and $w$.
The training in \cite{GLM19} uses 10-fold cross-validation.
The LASSO regularisation coefficient $\lambda$ is increased 
to result in a sparse set of variables to act as predictors;
see \cite{GLM19} for full details.

The features selected for each classification task 
are summarised in Table \ref{feat_set_comp}
which is a 
version of Table 2 in \cite{GLM19}.

\begin{table}[ht]	
    \centering
    \begin{tabular}{|c|c|c|}
        \hline
        \textbf{Task} &
        \textbf{Signature Feature Set}& 
        \textbf{Log Signature Feature Set}  \\ 
        \hline	
        ~ & Hippocampus-BL & Hippocampus-BL \\
        ~ & [Incr.Ventricles] & [Incr.Hippocampus] \\
        AD vs. NL & (Hippocampus,Time) & 
        [Incr.Ventricles] \\
        ~ & (Hippocampus,Time) & ~ \\
        ~ & (Hippocampus,WholeBrain) & ~ \\
        \hline  
        ~ & Hippocampus-BL & Hippocampus-BL \\
        ~ & Ventricles-BL & Ventricles-BL \\
        AD vs. MCI & (Hippocampus,Time) & 
        [Incr.Hippocampus] \\
        ~ & (Hippocampus,WholeBrain) & 
        [Incr.Ventricles] \\
        ~ & (Time,Ventricles) & (Time,Ventricles) \\
        ~ & (Hippocampus,Hippocampus) & ~ \\
        \hline
    \end{tabular}
\caption{Comparison of feature sets selected for
each classification task in \cite{GLM19}---Table 2 in
\cite{GLM19}}
\label{feat_set_comp}
\end{table}

The notation used in Table \ref{feat_set_comp} 
is as follows. The baseline value of a variable X is
denoted by X-BL. The increment of a variable
X are denoted [Incr.X] and correspond to depth one 
signature terms. Area terms between variables X and Y 
are denoted by (X,Y) and correspond to particular 
combinations of depth two signature terms. The precise
details may be found in \cite{GLM19}.

The resulting classifiers achieve at least $90\%$ 
accuracy on both classification tasks considered 
on the test data, see Table 3 in \cite{GLM19} for full 
details of the performance. Whilst the features 
provided by the signature method correspond to 
known AD pathology that can be extracted manually, 
the path signature provides a systematic way of
automatically generating these features \textit{without}
requiring any prior knowledge of the pathology.

\subsection{Early Sepsis Detection}
\label{sepsis_detection}
Sepsis is an overaggressive autoimmune response to 
infection that can cause life-threatening damage to 
the body's organs. It was estimated that in 2017 sepsis
affected 50 million people worldwide and caused
11 million deaths \cite{HKLMNS20}.
In America alone it is thought to be responsible for 
one in three hospital deaths \cite{HKLMNS20}, and the 
cost of admission and patient care has been estimated to 
exceed \$41.5 billion \cite{BSS20-I,BSS20-II,BSS20-III}.
The time of detection is critically linked to the mortality
rate of sepsis. In cases of septic shock, it is known that
the risk of death increases roughly 10\% for every hour 
of delay in antibiotic treatment \cite{CFGKKLPRSSTWZ06}. 
Early detection of sepsis is evidently crucial to 
improving sepsis management and mortality rates.

The works \cite{HKLMNS19,HKLMNS20} investigate the use of 
path signature techniques for early prediction of sepsis.
Unlike prior existing machine learning algorithms for 
this task, the Sepsis-3 definition \cite{DSS16} 
is used to denote the onset of sepsis.
The development of the model in \cite{HKLMNS19} was during
the 2019 PhysioNet challenge entitled
``Early Prediction of Sepsis from Clinical Data.”
Challenge participants were invited to submit algorithms 
that were trained on the same set of 
readily attainable ICU data and validated 
under a common performance metric on unseen test data.
The data and performance metrics used are detailed in 
\cite{CJJNRSSW19}. The algorithm proposed in 
\cite{HKLMNS19} was the first placed entry in the official
phase of the challenge \cite{HKLMNS20}.

The aim of this section is to present an overview of the
method developed in \cite{HKLMNS19}. In addition to 
\cite{HKLMNS19}, we will use the follow up work 
\cite{HKLMNS20}. The article \cite{HKLMNS20} is an extension 
of the Computing in Cardiology conference proceedings paper 
\cite{HKLMNS19} which includes further method details, 
additional results, and discussion.

The PhysioNet/Computing in Cardiology Challenge 2019 data 
were sourced from ICU patients in three separate hospital 
systems (which we label Hospital A, Hospital B and Hospital 
C). The data resulting from Hospital A and Hospital B 
were split into a publicly available training set 
and an undisclosed test set, both to be used for model 
development and testing. The test set and the data from 
Hospital C remained private, with 
submitted models being scored against these unseen data 
from all three systems.
The training set comprised of 40 336 patients with 40 
features consisting of demographic, vital signs, and 
laboratory data recorded per patient. The data was indexed
with time at 1-hour increments and predictions were to be 
made sequentially as each hour in a patient's time series
using \textit{only} the information observed up until
that time
(i.e. without making use of any future information).
A custom utility function (detailed in \cite{CJJNRSSW19}) 
was used to score the models.

Loosely speaking, the utility function is designed to 
ensure the following. Firstly, false-positives amongst
patients that never develop sepsis are penalised, whilst 
correct true negatives score zero. Secondly, for patients
that did develop sepsis, early prediction was penalised 
and false negatives were more heavily penalised. In 
addition correct true positives are rewarded. 
The penalisation of early prediction was included to 
promote predictions being made in the desired 6-hour 
window directly proceeding the on-set of sepsis time 
defined using the Sepsis-3 definition \cite{DSS16}.

The datasets are first augmented to include a number of 
additional features thought, based on literature review
and ``expert knowledge", to be useful for discerning 
the onset of sepsis \cite{HKLMNS20}.
The following hand-crafted features derived from the data 
are additionally added (see Table 1 in \cite{HKLMNS19}).

\begin{itemize}
    \item \textbf{Shock Index}: the ratio of the heart 
    rate to the systolic blood pressure.
    \item \textbf{BUN/CR}: the ratio of bilirubin to 
    creatinine.
    \item \textbf{Counter}: variable for the 
    temperature and the laboratory values that records 
    the number of times a given variable is measured over
    a pre-determined look-back window.
    Its inclusion is designed to exploit the idea that
    measurement frequency provides an indication of patient 
    health (i.e. an increasing in sampling rate may indicate
    physician concern about a patient). 
    \item \textbf{Max/Min}: the maximum and minimum of 
    the vital signs over a pre-determined look-back 
    window. 
    \item \textbf{PartialSOFA}: partial 
    construction of the \textit{Sequential Organ Failure 
    Assessment} (SOFA) score \cite{DSS16}.
    The resulting feature is termed PartialSOFA since the 
    dataset did not include all variables that 
    comprise the SOFA, meaning that the PartialSOFA 
    score was calculated only on the required information 
    available in the data. Hence the PartialSOFA score was 
    calculated based on threshold conditions on each of the 
    platelet count, bilirubin, mean arterial pressure (MAP), 
    and creatinine variables.
    \item \textbf{SOFA-Deterioration}: Binary label; 
    $1$ if PartialSOFA has decreased by $2$ in the last
    $24$ hour window. 
\end{itemize}
\noindent
The PartialSOFA and SOFA-Deterioration variables are 
included since deterioration of the SOFA score is a 
requirement of the Sepsis-3 definition. 
The size of the look-back windows chosen for the 
Counter and Max/Min variables are 
treated as hyperparameters and optimised during training.

Once the data has been augmented to include these
hand-crafted 
features, the signature transform is applied to the 
resulting time series. A sliding window approach is used;
signature features are computed for each time point over
a window of pre-determined look-back size. The input 
paths were augmented to include a time dimension, and 
the cumulative sum followed by the lead-lag transformation
were applied as further augmentations prior to truncated
signatures being computed. Both the truncation level and 
the look-back window length are treated as hyper-parameters
to be optimised during training.

The challenge data was pre-labelled with the value 1 at any
location of sepsis occurrence or predefined window around 
sepsis onset and zero otherwise. The method proposed in 
\cite{HKLMNS19} creates an alternative labelling that 
accounts for information about the utility score to enable
the classifier to place greater importance on points that
lead to a larger score if predicted correctly.
If $U_{y}(x,t)$ denotes the utility score of predicting 
$y$ for patient $x$ at time $t$, then the \textit{modified 
utility score} (MUS) is defined as 
$U_M(x,t) := U_1(x,t) - U_0(x,t)$.
It is against this labelling that the regressor is trained.

Stratified five-fold cross-validation, with a uniform 
distribution of time points and sepsis labels in 
each fold, is used for hyper-parameter optimisation. 
Precise details may be found in \cite{HKLMNS20}.
The final values of the parameters can be found 
in Table 2 in \cite{HKLMNS19}. They may be summarised 
as follows. The Counter variables are computed
over a look-back window of size 8, whilst
the Max/Min variables are computed over a 
look-back window of size 6.
The streams given by PartialSOFA, MAP and 
BUN/CR are augmented with a time dimension and a
lead-lag transformation, before signatures truncated
to depth 3 were computed using a look-back window of 
size 7. For the remaining non-stationary streams, 
the cumulative sum augmentation is applied followed
by the lead-lag transformation, before signatures 
truncated to depth 3 are computed using a look-back 
window of size 7.

The performance of the model using the final hyper-parameter
values is recorded via the average utility score over the 
five-folds used in cross-validation.
On the training data and testing data from Hospital A 
this score was 0.442 and 0.433 respectively, whilst on 
the training data and testing data from Hospital B it 
was 0.421 and 0.434 respectively. The similar 
performance on both the training and testing data suggest
the model was not over-fitted when restricted to the same 
hospital system on which it was trained. 

The data from Hospital C was not included in the public 
training data, and was used only for validation. 
The model achieved an average utility score of 
-0.123 on the data from Hospital C. This is significantly 
worse than the performance on the data from Hospital A 
and Hospital B, highlighting the limitations to using the 
method trained on one hospital system to make predictions
on a different one \cite{HKLMNS20}.
The utility scores for the model using the final 
hyper-parameters may be found in Table 1 in \cite{HKLMNS20}.

The impact of the various features on the models 
performance are explored in \cite{HKLMNS20}. 
The cross-validated and averaged utility score 
predictions on the training data for models trained 
using different subsets of features are recorded in 
Table 2 in \cite{HKLMNS20}. Four different subsets of
features are considered; Time only (T), Time + Original 40 
features (TO), Time + Original 40 features + Nonsignature 
features (TON), and Time + Original 40 features + 
Nonsignature features + Signature Features (TONS).

The average utility scores achieved using each subset 
are 0.282 for T, 0.389 for TO, 0.422 for TON and 
0.434 for TONS.
These scores reveal that the time-only feature is the 
single most useful feature. Whilst the inclusion of the
signature features does not result in a dramatic 
improvement (the score increases from 0.422 to 0.434),
it nevertheless illustrates that the representation of 
the information after the signature transformation is 
beneficial to learning.

The model from \cite{HKLMNS19} is designed to optimise 
the predefined utility function.
An extension to allow its use in an in-hospital 
environment to provide clinically actionable information
is proposed in \cite{HKLMNS20}
At each time point, the larger the models output value 
the higher the risk of sepsis. When a specified 
operating point threshold is exceeded, the subject is 
designated as a ``sepsis-risk patient”, indicating that 
closer monitoring or further tests may be warranted.
This operating point threshold can be chosen to achieve 
the most clinically meaningful sensitivity and specificity
\cite{HKLMNS20}. 

Returning to the training set of 40 336 patients, in 
\cite{HKLMNS20} the authors implement this for a range
of operating point thresholds. The results are presented
in a confusion matrix where a true negative represents that 
no call was made and the person did not develop sepsis, 
and a true positive represents that the patient being 
flagged and developing sepsis ``at some point” after this 
call. The choice of operating point threshold so 
that 33\% specificity (the proportion of correct 
positives that were correctly identified) is achieved 
results in the confusion matrix presented in Table
\ref{HKLMNS20_figure_2}.

\begin{table}[ht]	
\centering
    \begin{tabular}{|c|c|c|}
        \hline
        & \textbf{Predicted Sepsis} & 
        \textbf{Predicted No-Sepsis} \\
        \hline 
        \textbf{Actual Sepsis} & 1777 & 1150 \\
        \textbf{Actual No-Sepsis} & 3554 & 33855 \\
        \hline
    \end{tabular}
\caption{Confusion matrix displaying the number of 
people predicted as likely to get sepsis compared 
with those who actually end up with sepsis with 
the threshold tuned to 33\% specificity.
Data originates in Figure 2 in \cite{HKLMNS20}.}
\label{HKLMNS20_figure_2}
\end{table}

The signature-based model presented in 
\cite{HKLMNS19,HKLMNS20} for the early prediction of 
sepsis offers a competitive approach to discerning early
onset sepsis from health data streams.
The \href{https://github.com/datasig-ac-uk/signature_applications/tree/master/sepsis_detection}
{Early Sepsis Detection} notebook provides an introduction 
to implementing the methodology proposed in 
\cite{HKLMNS19,HKLMNS20}.
The data used to train the model in this notebook 
are the sequences of physiological and laboratory-observed 
measurements contained in the 
\href{https://mimic.physionet.org/}{MIMIC-III} dataset.
This consists of electronic health records for 40000 patients 
in intensive care at the at the Beth Israel Deaconess Medical 
Center, Boston, Massachusetts, between 2001 and 2012. 
These data include, for example, patients' heart rates, 
temperatures, and oxygen saturation levels, all recorded 
repeatedly over time for each patient.
The task considered is to use the classifier to predict, 
at time $t$, whether a patient will go on to develop sepsis
by time $t+T$, for some pre-determined $T > 0$, based on 
the patients measurement sequences recorded up until time
$t$. The notebook allows one to work through a version of 
the methodology of \cite{HKLMNS19,HKLMNS20} in a simplified
setting to illustrate the use of signature methods for the 
early detection of sepsis.

\subsection{Information Extraction from Medical Prescriptions}
\label{prescription_work}
Medical prescription notes are a valuable source
of important patient information that may not be 
recorded elsewhere.
Manual extraction and annotation is time-consuming 
and, since it must be done by specialists, expensive.
Natural language processing (NLP) tasks offer an 
approach to automating these tasks. Advances in 
pre-training large-scale contextualised 
language representations such as ELMo \cite{CGILNPZ18} 
and BERT \cite{CDLT18} have improved the performance 
of many NLP tasks. 
Several studies have fine-tuned BERT to clinical text 
(which differs substantially from general text) for 
NLP tasks \cite{ABJMMNWW19, LPY19,RSWX19,AHR20}.
Many tasks can be formulated as a classification or 
regression task, wherein either a simple linear layer 
or sequential models such as LSTM
are added after the BERT encoding of the input text as 
task-specific prediction layer \cite{CGGLLNPTU20}.

BERT's model architecture does not contain any recurrence 
or convolution, meaning additional positional 
encoding is required to model the order of the 
sequence (i.e. word order). In \cite{BLNW20}, the 
authors integrate the signature transform method into 
the Transformer model in order to naturally capture 
sequential ordering information in an effective manner.

The model architecture of BERT is composed of $N$ 
identical layers. Each layer has a multi-head 
self-attention mechanism followed by a position-wise 
fully connected feed forward network. The attention 
function takes three input vectors query Q, key K 
and value V, and generates a weighted sum of the 
value V, with the weights given as the softmax of a 
rescaling of the dot-product $Q K^{T}$
(see Section 2 of \cite{BLNW20} for the precise definition).
Multi-head attention splits Q, K and V into multiple heads 
via linear projection, allowing the model to attend to 
information at different positions from different 
representation subspaces in parallel. Each projected head
first goes through the scaled dot-product attention 
function, then concatenated, and finally projected once
more to output the final values. Positional information 
is not explicitly modelled by the Transformer encoder 
without modifying its inputs to include a representation
of absolute position \cite{BLNW20}.

The main extension to the Transformer architecture 
proposed in \cite{BLNW20} is replacing this attention 
function with a \textit{Sig-Attention} function.
The Attention function itself is first taken, before 
the outputs dimension is reduced by an affine map. 
The truncated signature $S^{(N)}$ for some $N \in \N$ 
is taken to give the final output.
A dropout rate (set to be $0.1$ in \cite{BLNW20}) 
is used to ensure the output dimension is the same 
as the input dimension.
The Sig-Attention mechanism is illustrated in 
Figure \ref{Figure2_BLNW20_a} below.

\begin{figure}[ht]
    \centering
    \includegraphics[scale = 0.5]
    {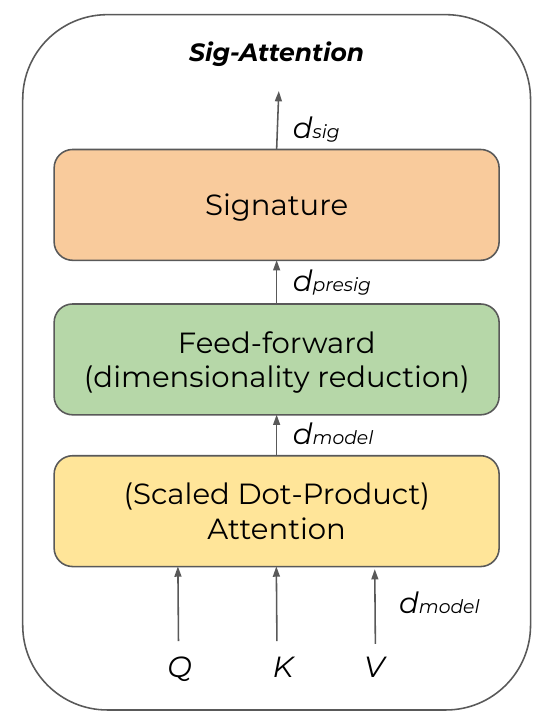}
    \caption{Sig-Attention mechanism---Part of Figure 2 in
    \cite{BLNW20}.
    Reproduced with permission. Colour Alterations.
    Licensed under a 
    \href{https://creativecommons.org/licenses/by/4.0/}
    {Creative Commons Attribution 4.0 International
    License}.}
    \label{Figure2_BLNW20_a}
\end{figure}

\begin{figure}[ht]
\centering
\includegraphics[scale = 0.8]
{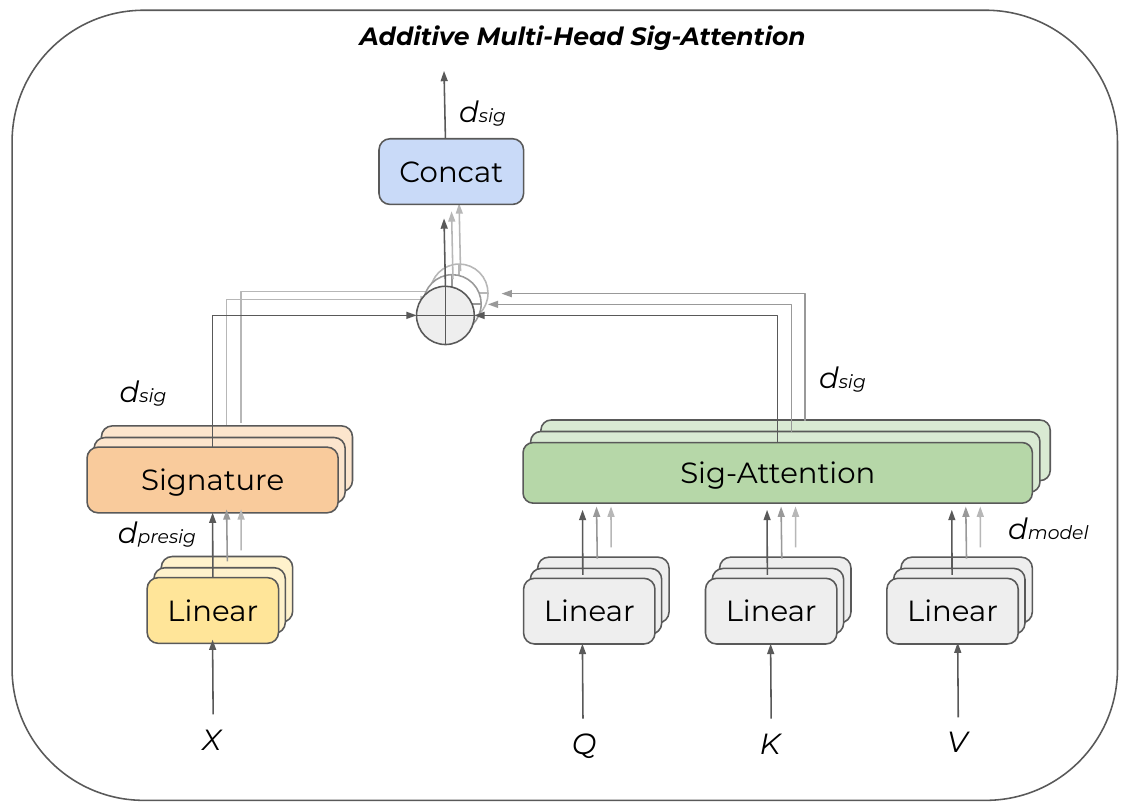}
\caption{Additive Multi-Head Sig-Attention---Part of
Figure 2 in \cite{BLNW20}. 
Reproduced with permission. Colour Alterations. 
Licensed under a 
\href{https://creativecommons.org/licenses/by/4.0/}
{Creative Commons Attribution 4.0 International License}.}
\label{Figure2_BLNW20_b}
\end{figure}

The proposed \textit{Additive Multi-Head Sig-Attention}
model combines the information from the input 
sequence with the output of Sig-Attention in different
representation subspaces; see Figure \ref{Figure2_BLNW20_b} 
below which is part of Figure 2 in \cite{BLNW20}.
The model takes the embedding of an input sequence
as well as the queries, keys and values as input
As shown in Figure \ref{Figure2_BLNW20_b}, the queries, keys
and values as well as the input embeddings are first 
linearly projected to have the required dimensions. 
Sig-Attention is then applied to the query, key and value 
vectors, whilst the signature is taken of the projected 
input embedding.
The two output signatures are combined
to result in one signature vector.
Finally all the signature vectors are concatenated
and passed onto the next sub-layer.

A dataset of 3852 distinct prescriptions for 
beta-blockers, provided by Karolinska University 
Hospital, written in Swedish is used for experimentation. 
Medical practitioners have annotated three labels 
QUANTITY, QUANTITY TAG and INDICATION.
\begin{itemize}
    \item QUANTITY: Total amount of capsules prescribed
    \item QUANTITY TAG: one of 5 classes labelling the 
    quantity prescribed to the patient:
		\begin{enumerate}
		  \item \textit{Not Specified}: the quantity was 
            not specified on the prescription
            \item \textit{Complex}: a range of quantities
            was given and the quantity is an average
            \item \textit{PRN}: prescription to take only
            if needed
            \item \textit{As Per Previous Prescription}:
            refers to guidance in previous prescription
            \item \textit{Standard}: standard prescription
        \end{enumerate}
    \item INDICATION: one of 5 classes covering the purpose 
    of the prescription:
        \begin{enumerate}
            \item \textit{Cardiac}
            \item \textit{Tremors}
            \item \textit{Migraine}
            \item \textit{Others}
            \item \textit{NA (Not Annotated)}
        \end{enumerate}
\end{itemize}
An example annotated prescription is included in Table 
\ref{annotated_prescription}. This is a reproduction 
of Table 1 in \cite{BLNW20}

\begin{table}[ht]	
\centering
\scalebox{0.8}{
    \begin{tabular}{|ccccc|}
        \hline
        \textbf{Swedish} & \textbf{Translated English} & 
        \textbf{Indication} & \textbf{Quantity} & 
        \textbf{Quantity Tag} \\
        \hline 
        $\begin{array}{c}
            \text{1 TABLETT FOREBYGGANDE} \\
            \text{MOT MIGRAN}
        \end{array}
        $
        & 1 tablet prevention against migrain & Migraine &
        1 & Standard 
        \\
        \hline 
        $\begin{array}{c}
            \text{MOT HOGT BLODTRYCK} \\
            \text{OCHHJARTKLAPPNING} \\
            \text{2 TABLETTER KLOCKGAN 08:00} \\
            \text{1 TABLETT KLOCKAN 18:00}
        \end{array}
        $ 
        & 
        $\begin{array}{c}
            \text{Against high blood pressure} \\
            \text{and heart palpitations} \\
            \text{2 tablets at 08:00} \\
            \text{1 tablet at 18:00}
        \end{array}
        $ &
        $\begin{array}{c}
             \\
             \text{Cardiac-} \\
             \text{hypertension} \\
             
        \end{array}
        $ & 3 & Standard \\
        \hline 
        $\begin{array}{c}
            \text{2 TABLETTER KL. 08,} \\
            \text{2 TABLETT KL. 20. DAGLIGEN.} \\
            \text{OBS KVALLSDOSEN HAR} \\
            \text{HOJTS JFRT MED 070427} 
        \end{array}
        $ & 
        $\begin{array}{c}
            \text{2 tablets kl. 08} \\
            \text{2 tablet kl. 20 Daily.} \\
            \text{Note the evening box has} \\
            \text{hojts jfrt with 070427}
        \end{array}
        $ &
        NA & 4 & Standard \\
        \hline 
        1 tablett vid behov mot stress & 1 tablet if 
        needed against stress & Anxiety & 1 & PRN \\	
        \hline 
        $\begin{array}{c}
            \text{FOR BLODTRYCK} \\
            \text{OCH HJARTRYTM-}
        \end{array}
        $ & 
        $\begin{array}{c}
            \text{For blood pressure} \\
            \text{and heart rhythm -} 
        \end{array}
        $ & 
        $\begin{array}{c}
            \text{Cardiac-} \\
            \text{hypertension-} \\
            \text{dysrhythmia}
        \end{array} 
        $ & 0 & Not Specified \\
        \hline 
        $\begin{array}{c}
            \text{1 tabl pA morgonen och} \\
            \text{en halv tabl pA kvAllen} \\
            \text{fOr hjArtrytmen}
        \end{array}
        $ & 
        $ \begin{array}{c}
            \text{1 table in the morning and} \\
            \text{a half table in the evening} \\
            \text{for heart rhythem}
        \end{array}
        $ 
        & 
        $ \begin{array}{c}
            \text{Cardiac-} \\
            \text{dysrhythmia}
        \end{array}
        $ & 1.5 & Standard \\
        \hline 
        1-2 tablett 2 gAnger dagligen & 1-2 tablets 2 
        times daily & NA & 3 & Complex \\
        \hline
    \end{tabular}
}
\caption{Example prescriptions with translations
(via Google translate) and annotations; 
the final three columns are the labels of interest 
that are sought to be extracted automatically. 
Reproduced variant of Table 1 in \cite{BLNW20}.}
\label{annotated_prescription}
\end{table}

\begin{figure}[ht]
\centering
\includegraphics[scale = 0.8]
{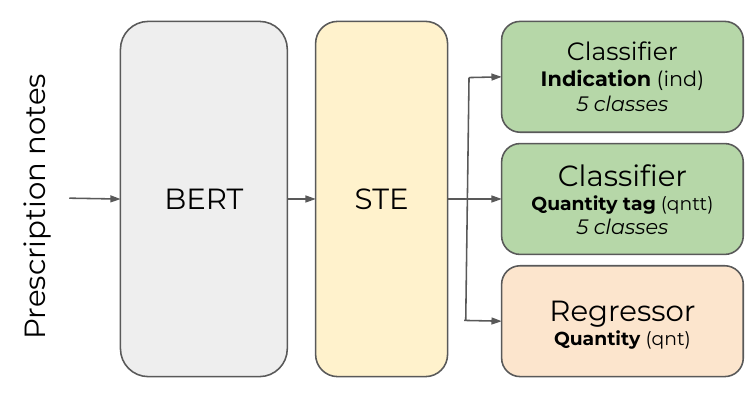}
\caption{Proposed multi-task learning architecture with
Sig-Transformer Encoder (STE). Figure 4 in \cite{BLNW20}.
Reproduced with permission. Colour Alterations.
Licensed under a 
\href{https://creativecommons.org/licenses/by/4.0/}
{Creative Commons Attribution 4.0 International License}.}
\label{Figure4_BLNW20}
\end{figure}

The following two approaches are proposed in 
\cite{BLNW20}.
\begin{itemize}
    \item Encode the Swedish text directly using 
    Multilingual BERT (M-BERT) \cite{CDLT18}, and
    \item Translate the prescriptions and then apply 
    ClinicalBERT \cite{AHR20} that is pre-trained on 
    clinical English.
\end{itemize}
Both methods are applied to the multi-task learning 
problem consisting of a regression problem 
(find the QUANTITY) and two classification
problems (find the QUANTITY TAG and INDICATION). 
The proposed Sig-Transformer Encoder (STE) 
multi-task learning architecture is 
depicted in Figure \ref{Figure4_BLNW20}.

A single linear layer is used as the classifier 
for QUANTITY TAG whilst two-layer networks are 
used for both QUANTITY and INDICATION.
Cross-entropy loss is used for the two classification 
tasks whilst the mean-square error is used for 
the regression. The overall loss function is taken 
to be a weighted sum of the loss functions for all 
three tasks. The STE model is compared against both 
the baseline Base model (i.e. the architecture of 
Figure \ref{Figure4_BLNW20} with the STE step removed) 
and against an LSTM model given by the same
architecture as Figure \ref{Figure4_BLNW20}, but with 
the STE step replaced by an LSTM step. The results 
are summarised in Table \ref{prescription_results_1}.

\begin{table}[ht]	
\centering
    \begin{tabular}{|cccc|}
        \hline 
        \textbf{Model} &
        $
        \begin{array}{c}
            \textbf{QUANTITY} \\ 
            \text{(MSE)}
        \end{array} $ & 
        $ 
        \begin{array}{c}
            \textbf{QUANTITY TAG} \\
            \text{(f1 score)}
        \end{array} $ &
        $
        \begin{array}{c}
             \textbf{INDICATION}  \\
             \text{(f1 score)} 
        \end{array}$ 
        \\
        \hline 
        Base & 0.50 & 0.60 & 0.08 \\
        ClinicalBERT & 0.21 & 0.89 & 0.09 \\
        M-BERT & 0.41 & 0.63 & \textbf{0.10} \\
        \hline 
        Base$+$LSTM & 0.45 & 0.76 & 0.06 \\
        ClinicalBERT$+$LSTM & 0.47 & 0.64 & \textbf{0.10} \\
        M-BERT$+$LSTM & 0.50 & 0.68 & 0.02 \\
        \hline 
        Base$+$STE & 0.23 & 0.78 & 0.03 \\
        ClinicalBERT$+$STE & 0.36 & 0.77 & 0.06 \\
        M-BERT$+$STE & \textbf{0.15} & \textbf{0.92} &
        0.05 \\
        \hline 
    \end{tabular}
\caption{Performance comparisons for information extraction.
Table 3 in \cite{BLNW20}.}
\label{prescription_results_1}
\end{table}

Without using translation (i.e. using M-BERT), 
the addition of LSTM and STE improves performance 
for QUANTITY and QUANTITY TAG.
Overall, the M-BERT$+$STE model proposed in \cite{BLNW20} 
performs best for QUANTITY and QUANTITY TAG.
Whilst the proposed methods gives worse performance 
for INDICATION, the unanimous poor performance by 
all models on this task 
suggests there is an underlying issue with the class 
boundaries. It is likely that the aggregation of the 
original 44 classes to the final 5 classes has left the 
new classes much less distinct since each is now formed 
of a range of topics \cite{BLNW20}.

Further evidence that the INDICATION task is 
unexpectedly more challenging is the strong individual 
class separation achieved by the M-BERT$+$STE model in 
the QUANTITY TAG classification task, illustrated in 
Table \ref{prescription_results_2}.

\begin{table}[ht]	
\centering
    \begin{tabular}{|cccccc|}
        \hline
        ~ & ~ & \textbf{QUANTITY TAG}&
        \textbf{(f1 score)} & ~ & ~ \\
        \textbf{Model} & \textbf{Standard} & \textbf{APPP} & 
        \textbf{PRN} & \textbf{Complex} & \textbf{NS} \\
        \hline 
        Base & 0.71 & 0.50 & 0.10 & 0.76 & 0.95 \\
        ClinicalBERT & 0.83 & 0.97 & 0.89 & 0.99 & 0.79 \\
        M-BERT & 0.94 & 0.41 & 0.04 & 0.81 & 0.98 \\
        \hline 
        Base$+$LSTM & 0.93 & 0.36 & 0.99 & 0.77 & 0.74 \\
        ClinicalBERT$+$LSTM & 0.90 & 0.23 & 0.99 & 0.61 & 
        0.49  \\
        M-BERT$+$LSTM & 0.29 & 0.99 & 0.72 & 0.47 & 0.93 \\
        \hline 
        Base$+$STE & 0.97 & 0.84 & 0.77 & 0.44 & 0.88 \\
        ClinicalBERT$+$STE & 0.99 & 0.70 & 0.85 &
        0.65 & 0.65 \\
        M-BERT$+$STE & 0.86 & 0.97 & 1.00 & 0.89 & 0.86 \\
        \hline 
    \end{tabular}
\caption{Performance for individual classes comparison; 
APPP means ``As Per Previous Prescription" and NS means 
``Not Specified". Table 6 in \cite{BLNW20}.}
\label{prescription_results_2}
\end{table}

The M-BERT$+$STE model performance is consistent across
all classes, and for each class its performance is 
comparable to that of the best model for that particular 
class. In fact, every model other than ClinicalBERT$+$STE 
is considerably out-performed by M-BERT$+$STE on at least
one class. Incorporating signature transform methods 
with the self-attention mechanism has enabled the authors
of \cite{BLNW20} to create a relatively simple model 
whose performance is at least comparable with its 
competitors on the information extraction tasks considered
in \cite{BLNW20}.

\section{Landmark-based Human Action Recognition}
\label{LHAR_app}
Human action recognition (HAR) has a wide range of 
applications such as video surveillance and 
behavioural analysis.
The task is to recognise an action from video footage. 
\textit{Landmark-based Human Action Recognition} (LHAR) regards 
objects as systems of correlated labelled landmarks. 
Recognising actions from the evolution of vectors 
connecting these labelled landmarks is motivated by 
the moving light-spots experiments in \cite{Joh73}.  
They demonstrated that people can detect motion 
patterns and recognise actions from several bright spots 
distributed on a body. The challenge of LHAR is to train 
a computer to recognise actions based on the evolution 
of some set of vectors in a fixed vector space, 
usually $\R^d$ for some $d \in \N$.

LHAR uses anonymous data; only the position of a certain 
number of markers is recorded. The use of de-identified 
data is often more suitable.
For example, monitoring vulnerable people in their homes 
to spot accidents and falls. It is unlikely that many
people would be happy to have full video data constantly 
recorded from within their homes.
However, only having the location of, say, 20 markers
recorded, without any person identifying features, 
is likely a more palatable option.

Two major challenges associated with LHAR are designing 
reliable discriminative features for spatial structural 
representation and modelling the temporal dynamics of motion. 
There are two main approaches to LHAR: joint-based and 
part-based. 
The joint-based approach regard the body as a set of points
and attempt to capture correlation
among the body joints via the pairwise distances or the 
joint orientations.
The parts-based approach focuses on connected segments 
of the set of points. 
Most methods represent spatial poses using predefined 
skeletal structures, only connecting points
via paths following the skeleton of the body, see 
\cite{ACV14} for example.

Whilst these connections seem intuitive, they are not 
guaranteed to be the crucial ones for distinguishing actions.
The connections discarded by imposing a skeletal structure 
may contain valuable non-local information.
For instance, the non-local displacement between two 
hand points is a key feature for the action of clapping.
Not using skeletal structure information ensures that any 
potentially vital non-skeletal connections are not discarded.
A further advantage is that it makes it easier to extend 
the method to the actions of other objects.
If we are not using skeletal structure, then we need only 
specify the landmark locations on the body. 
We can easily consider other objects by experimenting with 
landmark location \textit{without} needing to understand the
underlying structure of this new object.

A limitation of this framework is that the labelled 
data must all be within the same vector space. 
We need to be able to make sense of the difference 
between two joints, with the distance between them 
often being important. This is evidently the case for 
locations on a body, which are naturally vectors in 
$\R^3$ (or $\R^4$ if time is included). 
But if the data consisted of recorded emotions, 
as was the case in Section \ref{bipolar}, then we 
would not be in this framework;
there is no sensible notion of what the difference 
anger - depressed should be, for example.

The evolution of the connected segments is a path, 
and thus using the signature to understand this path is 
natural. In \cite{LNSY17}, the path signature 
feature (PSF) transform is used to provide a 
suitable feature set for spatial and temporal 
representation of LHAR. A pose is localised by 
disintegration into a collection of $m$-node sub-paths, 
with the signatures of these paths encoding non-local 
and non-linear geometric dependencies. This offers a 
resolution to the problematic trade-off between 
hand-designed local descriptors being insufficient 
to capture complex spatio-temporal dependences 
\cite{BGJSZ13,FTZ13}, and the deep RNN, LSTM in 
particular, models learning features that are 
not as easily interpreted.

\subsection{Path Disintegration and Transformations}
The disintegrations considered are 
\textit{Pose Disintegration} and 
\textit{Temporal Disintegration}.
The pose is regarded as an ordered collection 
of points in $\R^d$, with the ordering chosen 
randomly and fixed. The pose disintegration 
localises the pose into all possible subposes 
containing $m$ points. The inherited order converts 
each subpose to a unique $m$-node sub-path that 
visits each node once. These paths are not 
restricted to being parts of the physical body. 
An illustration of the pose disintegration can be seen 
in Figure \ref{Figure2_LNSY17}.
The notation $PSF_m$ is used to refer to computing 
the path signature features of all possible subposes 
containing $m$ points in a single frame.

\begin{figure}[ht]
\centering
\includegraphics[width=1\textwidth]{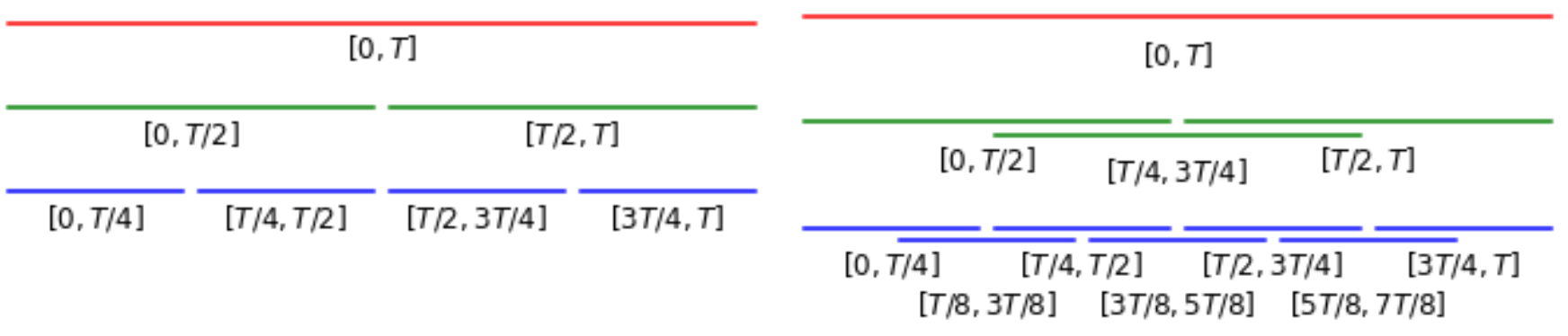}
\caption{On the left is the standard dyadic 
collections $\cd_0^{\dyadic}$, $\cd_1^{\dyadic}$ 
and $\cd_2^{\dyadic}$,
on the right is the collections $\cd_0$, $\cd_1$ 
and $\cd_2$.
Image retrieved from 
\href{https://github.com/datasig-ac-uk/signature_applications/tree/master/human_action_recognition}{Landmark Human Action Recognition Github}.}
\label{dyadic_illus}
\end{figure}

The temporal disintegration splits the interval 
$[0,T]$ into over-lapping dyadic intervals. 
For $j \in \N$, the $j^{\text{th}}$ dyadic level 
of the resulting hierarchical structure is the 
collection of subintervals 
\begin{equation}
	\label{dyadic_split}
		\cd_j := \cd_j^{\dyadic} \cup \cd_j^{\overlap}
\end{equation}
where
\begin{equation}
	\label{Dj_dyadic}
		\cd_j^{\dyadic} := 
		\Bigg\{ \bigg[ \frac{i}{2^{j}}T , 
		\frac{i+1}{2^{j}} T \bigg] \mid 
		i \in \left\{ 1 , \ldots , 2^j - 1 \right\} 
		\Bigg\} 
\end{equation}
and
\begin{equation}
	\label{Dj_overlap}
		\cd_j^{\overlap} := 
		\Bigg\{ \bigg[ \frac{2k+1}{2^{j+1}}T , 
		\frac{2k+3}{2^{j+1}} T \bigg] \mid 
		k \in \left\{ 1 , \ldots , 2^j - 2 \right\} 
		\Bigg\}.
\end{equation}
The inclusion of the overlapping intervals 
$\cd_j^{\overlap}$ prevents the interior points 
$\frac{i}{2^j}T$, for $i\in \{1,\ldots ,2^{j} -1\}$,
being artificially converted to endpoints in 
the collection $\cd_j$. This dyadic decomposition 
of the interval $[0,T]$ is illustrated in 
Figure \ref{dyadic_illus}. 

Signatures are computed over all the smaller intervals, 
reflecting the idea that the low-order signature 
terms over all smaller intervals will be more 
informative than the higher order terms over 
the entire interval $[0,T]$.
The choice of hierarchical level $h \in \N$ is a 
trade-off between improving efficiency and 
introducing local noise over finer intervals.
The notation $\cd_j-PSF_m$ is used to refer 
to computing the path signature features of 
all possible subposes containing $m$
over all the $l^{\text{th}}$ level dyadic 
decomposition $\cd_l$ for $l \leq j$.

Better performance is observed for taking signatures 
of level two over all of $\cd_0$, $\cd_1$ and $\cd_2$ 
than taking signatures of level five over
the whole time interval $[0,T]$ in \cite{LNSY17}. 
Whilst using level two truncations involves simpler 
and quicker computations than using level five 
signatures, the level two signatures on 
$\cd_0$, $\cd_1$ and $\cd_2$ also result in a 
smaller feature set.
Assuming the path is three-dimensional, the level 
five signature over $[0,T]$ involves 
$\frac{3^6 - 1}{3-1} = 364$ terms.
In contrast, the level two signatures over all the dyadic
intervals $\cd_0$, $\cd_1$ and $\cd_2$ involve only  
$11 \times \frac{3^3 - 1}{3-1} = 11 \times 13 = 143$ terms,
which is roughly $40\%$ of the size.

\subsection{Datasets}
The proposed method is trained on four datasets; JHMDB 
\cite{BGJSZ13}, SBU \cite{BCHSY12}, Berkeley MHAD 
\cite{BCKOV13} and NTURGB$+$D \cite{LNSW16}. 
The JHMDB dataset consists of 928 clips, each containing 
between 15 and 40 frames, capturing a single person doing 
one of 21 actions. The 2D joint positions are manually 
annotated, and the loss of information due to 2D projection 
presents an additional challenge.

The SBU Interaction is a 3D Kinect-based dataset. Kinect 
sensors are cost effective depth cameras
that are capable of providing reliable joint locations 
via real-time pose estimation algorithms \cite{BCFFKMSS13}.
SBU contains a total of 282 clips categorised into 8 
classes of two-actor interaction. Self-occlusion (when 
an object is obscured due to overlapping itself) causes 
measurement errors in the joint locations, which again 
presents an additional challenge.

The Berkeley MHAD dataset is $659$ clips captured by a 
marker-based motion capture system. The $3D$ locations 
of $43$ joints are accurately captured using LED markers. 
There are $384$ clips performed by $7$ actors that are used
for training, and $275$ clips performed by $5$ actors that 
are used for testing.

The Kinect-based NTURGB$+$D dataset is one of the 
largest 3D action recognition datasets and contains 
56,000 clips 
categorised into 60 classes. There is no constraint 
on the number of actors appearing
in each clip which, together with the large variation 
in viewpoint, pose a significant challenge for 
analysing this dataset.

\subsection{Method Implementation}
\label{lhar_method}
Gaussian noise is added over joint coordinates to 
simulate the errors caused by estimation. Biometric 
differences are compensated for by 
normalising the coordinate value ranges to $[-1,1]$ 
over the entire clip. Similarly, each feature is 
normalised to the interval $[-1,1]$.
To obtain a fixed length input, $M=10$ clips are 
uniformly sampled from each clip.

\begin{figure}[ht]
\centering
\includegraphics[width=0.8\textwidth]{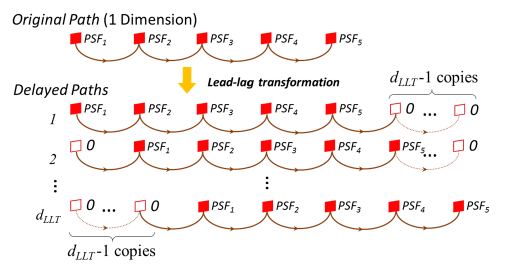}
\caption{Illustration of multi-delayed lead-lag transformation.
The dimension of lead-lag paths is $d_{LLT}$.
The delayed paths are padded with zeros to ensure a fixed
length for each dimension.
Figure 2 in \cite{LNSY17}.
Reproduced under license from Springer Nature
(License Number 5425270874131).}
\label{Figure2_LNSY17}
\end{figure}

The pose disintegrations for $m=2$ and $m=3$ are 
considered. Since the absolute position may be essential 
in some applications (static CCTV monitoring for example) 
the \textit{Invisibility-reset transformation} is used 
to retain this information in the signature. 
Further, multi-delayed no-future-pause Lead-Lag 
transformations with no-future-pause are considered,
with those with smaller delays encoding short-term 
dependencies, and those with longer delays encoding 
long-term dependencies.
See Section \ref{dataset_to_path} for definitions of 
these transformations and Figure \ref{Figure2_LNSY17}
for an illustration of the multi-delayed Lead-Lag 
transformations.

The spatial features extracted by $PSF_2$ with 
truncation level $n_{SP}$ of the paths corresponding 
to the pairs of joints in each frame (i.e. $m=2$) 
are labelled \textbf{S-P-PSF}. The spatial features 
extracted by the $PSF_3$ with truncation level $n_{ST}$
of the paths corresponding to the triples of joints in 
each frame (i.e. $m=3$) are labelled \textbf{S-T-PSF}.
The truncation levels $n_{SP} := 2$ 
and $n_{ST} := 3$ were found to be optimal in 
Section 6 of \cite{LNSY17}.

\begin{figure}[ht]
\centering
\includegraphics[width=0.8\textwidth]{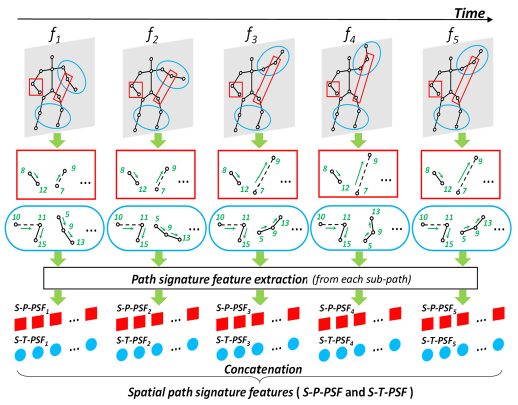}
\caption{The illustration of spatial feature 
(\textbf{S-P-PSF} and \textbf{S-T-PSF}) extraction 
(N= 15 in this figure). The red 
quadrangles denote the feature extraction 
of joint pairs, while the blue ellipses denote 
that of joint triples. All possible pairs and 
triples of joints are considered. 
Figure 3 in \cite{LNSY17}. Reproduced under license
from Springer Nature (License Number 5425270874131).}
\label{Figure3_LNSY17}
\end{figure}

The temporal features extracted by $\cd_{h_{TJ}}-PSF_1$ 
with truncation level $n_{TJ}$ of the paths corresponding 
to the evolution of the joint locations (i.e. $m=1$) 
are labelled \textbf{T-J-PSF}. 
Each dimension of the features \textbf{S-P-PSF} and 
\textbf{S-T-PSF} characterises one particular spatial 
constraint of a pose. The temporal features extracted 
by $\cd_{h_{TS}}-PSF_1$ with truncation level $n_{TS}$ 
of multi-delay lead-lag transformations, with 
no-future-pause, of the one-dimensional paths 
corresponding to the evolution of each particular 
spatial constraint of a pose are labelled \textbf{T-S-PSF}. 
The choices of truncation levels $n_{TJ} := 5$ and 
$n_{TS} := 2$ and the choices of the hierarchical levels
$h_{TJ} := 3$ and $h_{TS} := 3$ for the dyadic 
decomposition level were found to be optimal in 
Section 6 of \cite{LNSY17}.

\begin{figure}[ht]
\centering
\includegraphics[width=\textwidth]{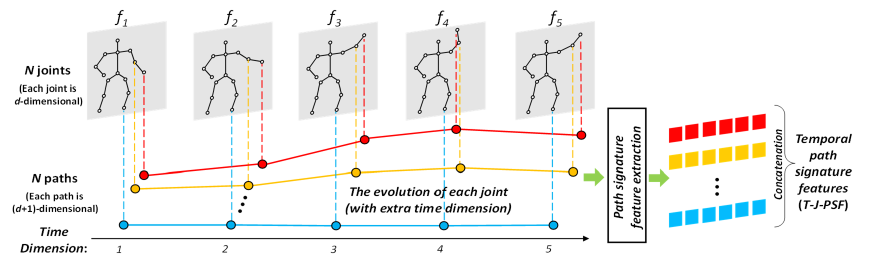}
\caption{Illustration of temporal features 
extracted from the evolution of each 
corresponding joint (\textbf{T-J-PSF}).
Figure 4 in \cite{LNSY17}. Reproduced under license
from Springer Nature (License Number 5425270874131).}
\label{Figure4_LNSY17}
\end{figure}

\begin{figure}[ht]
\centering
\includegraphics[width=\textwidth]{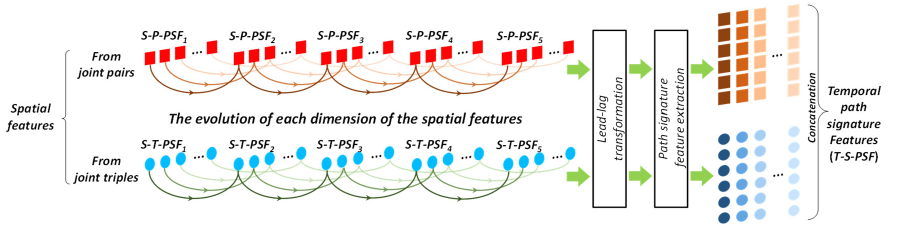}
\caption{Illustration of temporal features 
extracted from the evolution of spatial context 
(\textbf{T-S-PSF}). Each dimension of the spatial 
features is treated equally and individually. 
Figure 5 in \cite{LNSY17}. Reproduced under license
from Springer Nature (License Number 5425270874131).}
\label{Figure5_LNSY17}
\end{figure}

Illustrations of the extraction of the temporal 
features \textbf{T-J-PSF} and \textbf{T-S-PSF} are provided 
in Figures \ref{Figure4_LNSY17} and 
\ref{Figure5_LNSY17} respectively.

Finally, the $d$-dimensional coordinates of the 
joint locations at each frame are recorded and 
labelled \textbf{S-J}. The features considered in \cite{LNSY17}
are summarised below, starting with the 
\textbf{Spatial Structural Features} in each frame.
\begin{itemize}
    \item \textbf{S-J}: The $d$-dimensional coordinates of 
    each of the joints
    \item \textbf{S-P-PSF}: The PSF over each pair of joints 
    up to signature level $n_{SP}$
    \item \textbf{S-T-PSF}: The PSF over each triple of joints
    up to signature level $n_{ST}$
\end{itemize}
The \textbf{Temporal Dynamical Features} in each frame are
as follows.
\begin{itemize}
    \item \textbf{T-J-PSF}: The temporal PSF over the evolution
    of each joint up to signature level $n_{TJ}$
    \item \textbf{T-S-PSF}: The evolution of each dimension of 
    spatial PSF is treated as a path over which the temporal
    PSF up to signature level $n_{TS}$ is extracted
\end{itemize}
\noindent
Having extracted the features, we now describe the proposed 
models architecture.
A single-hidden-layer neural network is chosen for the 
classifier, with the hidden layer consisting of 64 neurons. 
The input dimension is determined by the signature transform 
(i.e. by \textbf{S-P-PSF}, \textbf{S-T-PSF},
\textbf{T-J-PSF} and \textbf{T-S-SPF}) 
and the output is a probability distribution given by 
a softmax layer over all the class labels in the dataset. 
Training uses stochastic gradient descent, with early 
stopping implemented to allow a maximum of 300 epochs, 
on the cross-entropy cost function with a learning rate 
that decays exponentially in the number of epochs run.

A random proportion of the connections between the input 
and the single hidden layer are omitted using Dropconnect,
see \cite{FLWZZ13}, for the purposes of regularisation. The 
high dimension of the data means that a high ratio of 
Dropconnect is required to prevent over-fitting, and the 
rate is set at 0.95.

\subsection{Comparison with Specifically-Tailored Methods}
The models performance on the four datasets MHAD, SBU, 
JHMDB and NTURGB$+$D is compared the performance of several 
models developed via other non-signature based techniques 
in \cite{LNSY17}. The results obtained for each of these 
four datasets are summarised below.

For the JHMDB dataset the off-the-shelf pose estimation 
called Alphapose (with Poseflow) \cite{FLTX17} was used 
to obtain 17 estimated joints from the RGB videos. 
The signature method proposed in \cite{LNSY17} is 
compared with the skeleton-based methods 
DT-FV \cite{SW13}, HLPF \cite{BGJSZ13}, Novel HLPF 
\cite{FTZ13}, and P-CNN \cite{CLS15}, all
trained using the evolution of these 17 joints, 
in \cite{LNSY17}. As recorded in \cite{LNSY17}, the 
accuracy $\mathbf{80.4\%}$ achieved by the Path Signature 
method \cite{LNSY17} is better than the accuracy 
achieved by any of the other methods 
( $65.9\%$ for DT-FV \cite{SW13},
$74.6\%$ for P-CNN \cite{CLS15}, 
$76.0\%$ for HLPF \cite{BGJSZ13}, and 
$79.6\%$ for Novel HLPF \cite{FTZ13}).

For the SBU dataset the two human bodies were regarded as 
a single united system with a total of 30 joints in 3D.
The signature method proposed in \cite{LNSY17} is 
compared with the skeleton-based methods 
Yun et al. \cite{BCHSY12}, Ji et al. \cite{CJY14}, 
CHARM \cite{CLLW15}, HBRNN \cite{DWW15},
Deep HBRNN \cite{LLSXXZZ16}, Co-occurrence \cite{LLSXXZZ16}, 
STA-LSTM \cite{LLSXZ17}, ST-LSTM-Trust Gate 
\cite{LSWX16,CLSWX17}, SkeletonNet \cite{ABBKS17}, and 
GC-Attention-LSTM \cite{DHKLW17}, all trained using the
evolution of the 30 joints, in \cite{LNSY17}.
As recored in Table 6 in \cite{LNSY17}, the accuracy of 
$\mathbf{96.8\%}$ for the Path Signature method \cite{LNSY17}
is better than the accuracy achieved by any of the other 
methods
($80.3\%$ for Yun et al. \cite{BCHSY12}, 
$86.8\%$ for Ji et al. \cite{CJY14}, 
$83.9\%$ for CHARM \cite{CLLW15}, 
$80.4\%$ for HBRNN \cite{DWW15},
$86.0\%$ for Deep HBRNN \cite{LLSXXZZ16}, 
$90.4\%$ for Co-occurrence \cite{LLSXXZZ16}, 
$91.5\%$ for STA-LSTM \cite{LLSXZ17}, 
$93.3\%$ for ST-LSTM-Trust Gate \cite{LSWX16,CLSWX17}, 
$93.5\%$ for SkeletonNet \cite{ABBKS17}, and  
$94.1\%$ for GC-Attention-LSTM \cite{DHKLW17}). 

The joint locations are recorded precisely enough in the 
Berkely MHAD dataset to enable methods to achieve 
perfect accuracy.
The signature method proposed in \cite{LNSY17} is 
compared with the skeleton-based methods 
Vantigodi et al. \cite{BV13}, 
Ofli et al. \cite{WWY13},
Vantigodi et al. \cite{RV14},
Kapsouras et al. \cite{KN14},
HBRNN \cite{DWW15}, and 
ST-LSTM-Trust Gate \cite{LSWX16,CLSWX17} 
on the Berkeley MHAD dataset in \cite{LNSY17}.
As recorded in Table 7 in \cite{LNSY17}, the perfect 
accuracy of $\mathbf{100\%}$ achieved by the Path Signature
method \cite{LNSY17} matches the perfect accuracy of
$100\%$ achieved by both the HBRNN \cite{DWW15} and
SL-LSTM-Trust Gate \cite{LSWX16,CLSWX17} methods, 
and is better than the accuracy achieved by the remaining
methods 
($96.1\%$ for Vantigodi et al. \cite{BV13}, 
$95.4\%$ for Ofli et al. \cite{WWY13},
$97.6\%$ for Vantigodi et al. \cite{RV14}, and
$98.2\%$ for Kapsouras et al. \cite{KN14}).

For the NTURGB$+$D dataset, experiments were conducted 
on recognising both the cross-subject (i.e. actions 
performed by different people) and the cross-view 
(i.e. actions recorded from different view points) tasks.
This involved an initial classification step to sort 
the data into 1-body and 2-body actions, before extracting 
the joints to be used.
For both experiments the signature method proposed in
\cite{LNSY17} is compared with the skeleton-based methods
Dynamic Skeletons \cite{HLZZ15},
HBRNN \cite{DWW15},
Part-aware LSTM \cite{LNSW16},
ST-LSTM-Trust Gate \cite{LSWX16,CLSWX17},
STA-LSTM \cite{LLSXZ17},
SkeletonNet \cite{ABBKS17},
Joint Distance Maps \cite{HLLW17}, and
GC-Attention-LSTM \cite{DHKLW17} in \cite{LNSY17}.
As recorded in Table 9 in \cite{LNSY17},
for the cross-subject task recognition, the accuracy 
of $\mathbf{78.3\%}$ achieved by the Path Signature method
\cite{LNSY17} is better than the accuracy achieved by any 
of the other methods 
($60.2\%$ for Dynamic Skeletons \cite{HLZZ15},
$59.1\%$ for HBRNN \cite{DWW15},
$62.9\%$ for Part-aware LSTM \cite{LNSW16},
$69.2\%$ for ST-LSTM-Trust Gate \cite{LSWX16,CLSWX17},
$73.4\%$ for STA-LSTM \cite{LLSXZ17},
$75.9\%$ for SkeletonNet \cite{ABBKS17},
$76.2\%$ for Joint Distance Maps \cite{HLLW17}, and
$74.4\%$ for GC-Attention-LSTM \cite{DHKLW17}).
Additionally, as recorded in Table 9 in \cite{LNSY17},
for the cross-view task recognition, the accuracy 
of $\mathbf{86.3\%}$ achieved by the Path Signature method
\cite{LNSY17} is better than the accuracy achieved by any 
of the other methods 
($65.2\%$ for Dynamic Skeletons \cite{HLZZ15},
$64.0\%$ for HBRNN \cite{DWW15},
$70.3\%$ for Part-aware LSTM \cite{LNSW16},
$77.7\%$ for ST-LSTM-Trust Gate \cite{LSWX16,CLSWX17},
$81.2\%$ for STA-LSTM \cite{LLSXZ17},
$81.2\%$ for SkeletonNet \cite{ABBKS17},
$82.3\%$ for Joint Distance Maps \cite{HLLW17}, and
$82.8\%$ for GC-Attention-LSTM \cite{DHKLW17}).

The experiments conducted in \cite{LNSY17} illustrate that,
within the class of recognising actions via the evolution of 
connected segments between landmark locations, the Path 
Signature method proposed in \cite{LNSY17}, a relatively
simple linear shallow fully-connected neural network, 
achieves comparable results to other more complex models.

\subsection{Demo Notebook}
\label{demo_notebook}
The 
\href{https://github.com/datasig-ac-uk/signature_applications/tree/master/human_action_recognition}
{Landmark Human Action Recognition} python notebook, 
created by Peter Foster and Kevin Schlegel, is based on 
\cite{LNSY17}. It provides an introduction to the methodology 
of \cite{LNSY17}, presenting python code that 
uses the \href{http://jhmdb.is.tue.mpg.de/}{JHMDB} dataset 
to generate a feature set as described in Section 
\ref{lhar_method}, before training a simple classifier 
action classifier using this feature set (again following 
the process discussed in Section \ref{lhar_method}).

The learning uses the PyTorch python package. The correct 
version of PyTorch to install depends on the particular 
hardware and operating system used. Consequently
it is highly recommended to install PyTorch manually 
following the official instructions.
After installing PyTorch, following the notebook is 
straightforward and once complete, the learnt classifier 
achieves an accuracy of $67.54\%$. 
This is lower than the $80.4\%$ reported in \cite{LNSY17} 
(see Table 5 in \cite{LNSY17}) since the notebook chose path 
transformations for the purpose achieving 
a high speed of computations (with the feature set generated 
within seconds and the network training completing within a 
few minutes) rather than for maximal accuracy.

By changing the path transformations and combining feature 
vectors of different sets of transformations the accuracy 
can be increased.
This notebook highlights the simplicity of the signature 
techniques; within 20 minutes it is possible to use the 
method proposed in \cite{LNSY17} on a CPU to train a 
classifier that achieves comparable accuracy with a 
specialised tailor-made models, see Table 5 in \cite{LNSY17}.

\section{Distribution Regression via the Expected Signature}
\label{dist_reg_expect_sig} 
Many real-world tasks fall within the 
\textit{Distribution Regression} framework, 
where the aim is to learn the functional relationship 
between multiple time-series and a single output.
Examples range from determining the temperature of a gas 
from the trajectories of particles \cite{Hil86,Rei99,Sch89}, 
to using high-resolution climatic data to predict overall 
end-of-year crop yields \cite{APP10,DR14,ELLLY17}, and 
estimating mean-reversion parameters from observed 
financial market dynamics \cite{LP11,GJR18,BGW00}. 

In Section \ref{expect_sig}, an approach to distribution 
regression via the expected signature was outlined. 
This approach is implemented in \cite{BDLLS20}, and we 
will provide a more detailed summary of their methodology 
within this section. 
Consider $M$ input-output pairs 
$\Big\{ \big\{ \bx^{i,j} \big\}_{j=1}^{N_i} , 
y^i \Big\}_{i=1}^M$
where each pair $i$ is given by a target $y^i \in \R$ 
and a collection of $N_i$ time-series
$\bx^{i,j} = \Big\{ \big( t_1 , x_1^{i,j} \big) , 
\ldots , \big( t_{l_{i,j}} , x_{l_{i,j}}^{i,j} 
\big) \Big\}$
of lengths $l_{i,j} \in \Z_{\geq 1}$, time stamps 
$t_1 \leq \ldots \leq t_{l_{i,j}}$ and values in 
a $b$-dimensional Banach space $W$. 
By using the method of transforming a stream of values 
into a stream of increments outlined in Subsection 
\ref{subsec:math_framework}, we update each stream 
$\bx^{i,j}$ to now be a stream of increments taking its
values in a $d$-dimensional Banach space $V$.
In particular, we must have that $d = 2(b+1) \in \Z_{\geq 2}$
and that, as a set, $V = W \times \R^{b+2}$.

Fix a compact interval $[a,b] \subset \R$.
Following the notation used in 
\cite{BDLLS20}, we let $\cc ( [a,b] , V)$ 
denote the set of paths resulting
from the removal of all the tree-like (see \cite{HL10}) 
paths from $\cv^1 ( [a,b], V)$. This has no 
practical impact on the following strategies \cite{BDLLS20}.
Consider, for each $i \in \{1, \ldots , M\}$ and 
$j \in \{1, \ldots , N_i\}$ the contour $\Gamma_{\bx^{i,j}}$
resulting from the concatenation of the entries of 
$\bx^{i,j}$.
Choose a parameterisation $x^{i,j} : [a,b] \to V$
of $\Gamma_{\bx^{i,j}}$ such that 
$x^{i,j} \in \cc([a,b],V)$.

After doing so, 
we have $M$ groups of input-output pairs of the form
\begin{equation}
	\label{in_out_pairs}
		\Big( \big\{ x^{1,j} : [a,b] \to V 
		\big\}_{j=1}^{N_1} , y^1 \in \R \Big), 
		\ldots ,
		\Big( \big\{ x^{M,j} : [a,b] \to V 
		\big\}_{j=1}^{N_M} , y^M \in \R \Big).
\end{equation}
The collection of trajectories in group $i$ is summarised 
by considering the measure
$\de^i := \frac{1}{N_i} \sum_{j=1}^{N_i} \de_{x^{i,j}} 
\in \p \cc ( [a,b],V )$
where $\de_{x^{i,j}}$ is the Dirac measure centred at 
$x^{i,j}$.
The input-output pairs in \eqref{in_out_pairs} can be 
represented as
\begin{equation}
	\label{in_out_meas}
		\Big( \de^1 \in \p \big(\ck\big) , y^1 
		\in \R \Big), \ldots ,
		\Big( \de^M \in \p \big(\ck\big) , y^M 
		\in \R \Big)
\end{equation}
and we aim to learn a function 
$F : \p \cc ([a,b], V) \to \R$ 
from the pairs in \eqref{in_out_meas}.

The expected signature can characterise 
probability measures on paths (cf. Theorem 
\ref{dist_uni_nonlin}) and two methods for 
distribution learning using the expected signature
are proposed in \cite{BDLLS20}. 
The first is the approach outlined 
at the end of Section \ref{expect_sig}.
To provide the details of this approach, we first
recall that the pathwise expected signature 
$\Phi_{\Path} : \p \cc ( [a,b],V ) \to 
\cc ( [a,b] , T((V)) )$ is defined by
\begin{equation}
    \label{pathwise_expected_sig_def_2}
        \Phi_{\Path} (\mu)_t := 
        \E_{\mu} \big[ S_{a,t}(X) \big].
\end{equation} 
A consequence of the universality of the
pathwise expected signature 
(cf. Theorem \ref{dist_uni_nonlin}) is that 
for any compact subset 
$\ck \subset \cc ( [a,b], V)$, if $F$ 
denotes our target function we want to learn, then 
\begin{equation}
    \label{good_span}
        F \rvert_{\p (\ck)} \in \Span
        \Big\{  S_{a,b} \big( 
        \Phi_{\Path} (\mu) \big)^{
		\bJ 
        } \mid
		\mu \in \p (\ck) , 
		\bJ = (j_1 , \ldots , j_k ) \in 
		\{ 1 , \ldots , k \}^d , 
		k \in \N_0 \Big\}
		=: \ca (\ck).
\end{equation}
The \textit{SES} method of \cite{BDLLS20} uses a 
truncation of $\ca(Z)$, where 
$Z := \big\{ \de^i \mid i \in \{1, \ldots ,M \} \big\}$,
as the feature functions before applying linear regression
to find the optimal combination with respect to the mean 
squared error (MSE). The truncation is in terms of the 
length of the multi-indices allowed,
i.e. in \eqref{good_span} we restrict to $k \leq K_0$ 
for some $K_0 \in \Z_{\geq 1}$ so that only the coordinate 
iterated integrals of order up to $K_0$ are used.

The second method introduced in \cite{BDLLS20} combines 
the expected signature with a Gaussian kernel.
Recall that the expected signature 
$\bS :\cc ( [a,b], V) \to T((V)) $ is defined 
by
\begin{equation}
    \label{expect_sig_def_2}
        \bS (\mu) := 
        \E_{\mu} \big[ S_{a,b}(X) \big] 
        =  \prod_{n=0}^{\infty} 
        \E_{\mu} \big[ S^n_{a,b}(X) \big].
\end{equation}
The \textit{KES} model proposed in \cite{BDLLS20} 
considers the kernel 
$k : \p \cc ( [a,b],V ) \times \p \cc ([a,b],V) \to \R$ 
defined by
\begin{equation}
    \label{exp_sig_ker}
        k(\mu , \nu) := 
        \exp \Big( 
        -\sigma^2 \big\lvert \big\lvert \bS(\mu) - 
        \bS(\nu) \big\rvert \big\rvert^2_{T((V))} \Big)
\end{equation}
for $\sigma > 0$. It is established that the kernel 
$k$ is universal in the sense that if 
$\ck \subset \cc ( [a,b] , V )$ is compact, then
the associated \textit{Reproducing Kernel Hilbert Space} 
(RKHS) is 
dense in $C^0 \big( \p ( \ck) ; \R \big)$ 
(see Theorem 3.3 in \cite{BDLLS20}).

To evaluate $k ( \de^i , \de^j )$ for 
$i,j \in \{1, \ldots , M \}$ we first compute that
\begin{align*}
    \big\lvert\big\lvert \bS \big( \de^i \big) - 
    \bS \big( \de^j \big) \big\rvert\big\rvert^2_{T((V))} 
    &= 
    \Bigg\lvert \Bigg\lvert 
    \frac{1}{N_i} \sum_{k=1}^{N_i} 
    \E_{\de_{x^{i,k}}} \big[ S_{a,b}(X) \big] - 
    \frac{1}{N_j} \sum_{l=1}^{N_j} \E_{\de_{x^{j,l}}} 
    \big[ S_{a,b}(X) \big] \Bigg\rvert\Bigg\rvert^2_{T((V))} \\
    &=\Bigg\lvert\Bigg\lvert
    \frac{1}{N_i} \sum_{k=1}^{N_i}  
    S_{a,b} \big( x^{i,k} \big)  - \frac{1}{N_j} 
    \sum_{l=1}^{N_j} S_{a,b} \big( x^{j,l} \big) 
    \Bigg\rvert\Bigg\rvert^2_{T((V))} \\
	&= E_{ii} + E_{jj} - 2 E_{ij}
\end{align*}
where 
\begin{equation}
	\label{Eij}
		E_{ij} := \frac{1}{N_i N_j} 
		\sum_{k=1}^{N_i} \sum_{l=1}^{N_j} 
		\Big< S_{a,b} \big( x^{i,k} \big) , 
		S_{a,b} \big( x^{j,l} \big) \Big>_{T((V))}
		=\frac{1}{N_i N_j} \sum_{k=1}^{N_i} 
		\sum_{l=1}^{N_j} K_{x^{i,k} , x^{j,l}} (a,b).
\end{equation}
In \eqref{Eij}, $K_{\cdot , \cdot} ( \cdot , \cdot)$ 
denotes the signature kernel 
discussed in Section \ref{full_sig_kernel_section}.
For the readers convenience, we recall that if 
$x \in \cv^1 ( [a,b] , V )$ and 
$y \in \cv^1 ( [c,d] , V )$ then
$K_{x,y} (s,t) := 
\big< S_{a,s}(x) , S_{c,t}(y) \big>_{T((V))}$ for 
$s \in [a,b]$ and $t \in [c,d]$ (cf. \eqref{sig_kernel}).
The relatively simple numerical scheme developed in 
\cite{CFLSY20} (see Section \ref{full_sig_kernel_section})
allows the inner products in \eqref{Eij} to be computed 
by a simple call to any numerical PDE solver of choice 
\cite{BDLLS20}.

The KES model uses the maps 
$\mu \mapsto  k( \mu , \de^i )$ for 
$i \in \{1, \ldots , M\}$ for feature functions 
before applying linear regression to optimise the 
linear combination with respect to the mean squared 
error (MSE).

In Section 5 of \cite{BDLLS20}, the performance of 
the KES and SES models are compared with the performance 
of DeepSet method \cite{KPRSSZ17} and other existing 
kernel-based methods for a variety of tasks.
The kernel-based techniques considered all correspond 
to the same general framework.
Distributions $\mu$ are first mapped to a RKHS $H_1$ 
via the map $\mu \mapsto \int_{X} k_1 (\cdot, x) \mu(dx)$
where $k_1 : X \times X \to H_1$ is the reproducing
kernel for $H_1$.
A second kernel $k_2$ is then used for the regression 
step to approximate a function $F : H_1 \to \R$ via 
the minimisation of a loss function.
A more detailed summary may be found in Section 4 of 
\cite{BDLLS20}, with the full details appearing in 
\cite{GSSS07}, \cite{DFMS12} \cite{Fla15}, \cite{GPSS16}, 
\cite{FLSS17}.
The notation DR-$k_1$ refers to the model produced by 
choosing $k_2$ to be a Gaussian kernel.

The particular kernel-based models considered are 
DR-RBF, DR-Matern32 and DR-GA, where GA refers to 
the Global Alignment kernel for time-series from 
\cite{BCMV07}, and the definition of all three may 
be found in Appendix B of \cite{BDLLS20}.
Kernel Ridge Regression is used to train KES and 
DR-$k_1$ for all choices of $k_1$ whilst Lasso 
Regression is used for SES.
All models are run 5 times on each task with the mean 
and standard deviation of the predictive MSE recorded.
The hyper-parameters of KES and DR-$k_1$ were selected 
by cross-validation on the training set of each run. 
Full details of the implementation can be found in 
appendix B of \cite{BDLLS20}.

The first experiment considered determining the 
temperature $T$ of an ideal gas. The simulation 
modelled 50 different gases, each consisting of 20 
particles, by randomly initialising all velocities 
and letting the particles evolve at constant speed. 
The task is to learn $T$ 
(sampled uniformly at random from [1, 1000]) from the 
set of trajectories traced by the particles in the gas. 
The complexity of the large-scale dynamics depends on 
both $T$ and the radius of the particles. The results 
of two experiments, one where particles
have a small radius (few collisions) and another where 
they have a bigger radius (many collisions), are presented
in Table \ref{GAS_results}.

\begin{table}[ht]	
\centering
\begin{tabular}{|c|c|c|}
    \hline
    \textbf{Model}& \textbf{Predictive MSE - Few Collisions} &
    \textbf{Predictive MSE - Many Collisions}  \\ 
    \hline		
    DeepSets  & $8.69 \pm 3.74 $& $5.61 \pm 0.91$ \\ 
    DR-RBF & $3.08 \pm 0.39$ & $4.36 \pm 0.64$ \\
    DR-Matern32 & $3.54 \pm 0.48$ & $4.12 \pm 0.39$ \\
    DR-GA & $2.85 \pm 0.43$ & $3.69 \pm 0.36$ \\
    \textbf{KES} & $\mathbf{1.31 \pm 0.34}$ & 
    $\mathbf{0.08 \pm 0.02}$ \\
    \textbf{SES} & $\mathbf{1.26 \pm 0.23}$ & 
    $\mathbf{0.09 \pm 0.03}$ \\
    \hline
\end{tabular}
\caption{Inferring the temperature of an ideal gas.
Predictive MSE have been scaled by a factor of 100; if X
is a predictive MSE value appearing in this table,
then the actual predictive MSE is X/100.
Table 1 in \cite{BDLLS20}.}
\label{GAS_results}
\end{table}

The performances are comparable in the simpler setting, 
with KES and SES being slightly better.
In the more collisions setting the KES and SES models 
significantly out-perform all other methods.

The \textit{fractional Ornstein--Uhlenbeck} (fOU) process
$\sigma_t = \exp ( P_t)$, with 
$dP_t = -a(P_t - m) dt + v dB_t^H$ for $a , m ,v \geq 0$
and $B^H_t$ denoting a fractional Brownian motion of 
Hurst parameter $H \in (0,1)$,
is used as a model for volatility \cite{PSS20}.
Due to findings in \cite{GJR18}, $H$ is chosen to be $0.2$
The authors consider the task of estimating the 
mean-reversion parameter $a$ from simulated sample 
paths of $\sigma_t$ in \cite{BDLLS20}.
For this purpose, 50 mean-reversion values
$\{ a_i \}_{i=1}^{50} \subset \big[ 10^{-6} , 1\big]$
are chosen uniformly, before each $a_i$ is regressed
on a collection of $N \in \{ 20,50,100 \}$ (time-augmented) 
trajectories $\Big\{ \sigma_t^{i,j} \Big\}_{j=1}^{N}$ 
of length 200.
The results, presented in Table \ref{Finance_results},
illustrate that the models KES and SES out-perform 
the others, and that
the performance of both KES and SES progressively 
with the number of samples $N$.

\begin{table}[ht]	
\centering
    \begin{tabular}{|c|ccc|}
        \hline
        \textbf{Model} & & \textbf{Predictive MSE} & \\ 
        & \textbf{N=20} & \textbf{N=50} & \textbf{N=100} \\ 
        \hline		
        Deepsets &$74.43 \pm 47.57$ & $74.07 
        \pm 49.15$ & $74.03 \pm 47.12$ \\
        DR-RBF &$52.25 \pm 11.20$ & $58.17 
        \pm 19.05$ & $44.30 \pm 7.12$ \\ 
        DR-Matern32 & $48.62 \pm 10.30$ & $54.91 
        \pm 12.02$ & $32.99 \pm 5.08$\\
        DR-GA & $3.17 \pm 1.59$ & $2.45 
        \pm 2.73$ & $0.70 \pm 0.42$ \\
        \textbf{KES} & $\mathbf{1.41 \pm 0.40}$ & 
        $\mathbf{0.30 \pm 0.07}$ & $\mathbf{0.16 \pm 0.03}$ \\
        \textbf{SES} & $\mathbf{1.49 \pm 0.39}$ & 
        $\mathbf{0.33 \pm 0.12}$ & $\mathbf{0.21 \pm 0.05}$ \\
        \hline
    \end{tabular}
\caption{Estimating mean-reversion parameters. 
Predictive MSE have been scaled by a factor of 1000;
if X is a predictive MSE value appearing in this 
table, then the actual predictive MSE is X/1000. 
Table 2 in \cite{BDLLS20}.}
\label{Finance_results}
\end{table}

The final task considered in \cite{BDLLS20} is a crop yield
prediction task.
The goal is to predict the yield of 
wheat crops over a region from the longitudinal
measurements of climatic variables recorded across
different locations of the region. The  
\href{https://ec.europa.eu/eurostat/data/database}
{Eurostat}
dataset containing the total annual regional yield
of wheat crops in mainland France - divided in 22
administrative regions - from 2015 to 2017 is used.
Climatic measurements (temperature, soil humidity
and precipitation), recorded at a frequency of
every 6 hours, were extracted for each region.
The total number of recordings varies across regions,
and half of each regions recordings were randomly
discarded as a further subsampling step.
In addition to the MSE, the 
\textit{mean absolute percentage error} (MAPE) 
was recorded. The baseline method predicts the 
average yield calculated from the training set.
The results are presented in Table \ref{Crop_results}.
Both the KES model and the SES model achieve larger 
improvements over the baseline than the other methods, 
with the SES model providing the best performance.

\begin{table}[ht]	
\centering
\begin{tabular}{|c|c|c|}
    \hline
    \textbf{Model}& \textbf{MSE} & \textbf{MAPE}  \\ 
    \hline		
    Baseline & $2.38 \pm 0.60$ & $23.31 \pm 4.42$	\\
    DeepSets & $2.67 \pm 1.02$ & $22.88 \pm 4.99$ \\
    DR-RBF & $0.82 \pm 0.22$ & $13.18 \pm 2.52$ \\
    DR-Matern32 & $0.82 \pm 0.23$ & $13.18 \pm 2.53$ \\
    DR-GA & $0.72 \pm 0.19$ & $12.55 \pm 1.74$ \\
    \textbf{KES} & $\mathbf{0.65 \pm 0.18}$ & 
    $\mathbf{12.34 \pm 2.32}$ \\
    \textbf{SES} & $\mathbf{0.62 \pm 0.10}$ & 
    $\mathbf{10.98 \pm 1.12}$ \\
    \hline
\end{tabular}
\caption{Predicting wheat crop yield.
Table 3 in \cite{BDLLS20}.}
\label{Crop_results}
\end{table}

\section{Conformance and SigMahaKNN Method for
Anomaly Detection}
\label{anom_detec}
Anomaly detection is the task of determining whether
a given observation is unusual compared to a corpus
of observations that have been deemed to be usual. 
It arises in numerous fields such as medicine
\cite{BCCHVV13}, financial fraud \cite{CLNTZ16} and
cybersecurity \cite{JS00}. 
A natural approach is to use a distance metric and
view an event as an anomaly if it is at least some
distance from the set of observations.
Approaches to unsupervised anomaly detection for
multivariate data include density-based approaches
\cite{BKNS00}, clustering \cite{DHX03}, random forests
\cite{LTZ12}, support vector machines \cite{AAG13}, 
neural networks \cite{CC19} and nearest neighbour-based 
approaches \cite{FHK04,HW07,KS20}.
The article \cite{CCPT14} surveys a wide-range
of anomaly detection techniques.

Throughout the remainder of this section we consider the
following generic problem framework.
We assume that we have a corpus $\cc$ of observations that
are all deemed to be normal.
The aim is to use this corpus $\cc$ to determine whether a
new observation \textit{outside} of $\cc$ is an anomaly.
In this setup we assume that the corpus $\cc$ is \textit{not}
polluted; no point within this given corpus $\cc$ should be
considered anomalous.
In particular, we do not consider the separate challenge
of dealing with a polluted corpus where it is possible for
points in the given corpus to be anomalous.

It is important to fix a sensible notion of what determines
an outlier. It may initially seem tempting to
determine an outlier by its difference to \textit{most} 
other instances; i.e. by comparison to some
averaged quantity determined by the entire
corpus $\cc$. 
However, this viewpoint runs the risk of incorrectly deeming
data points to be outliers. 

For an example of incorrectly deeming a data point to be
an outlier,
suppose we are interested in the time taken by a person to run
100m.
%
Assume we have an initial large corpus
$\cc$ of people that contains at least one Olympic level sprinter.
Let $\mu_{\cc}$ denote the average time taken to run 100m
taken over the corpus $\cc$.
If a new persons time is compared to $\mu_{\cc}$,
it is likely that an Olympic
sprinter would be considered an outlier.
But of course they are not an
outlier since there is at least one \textit{other}
Olympic sprinter
exhibiting similar performance amongst the corpus $\cc$.

For an example of incorrectly deeming a data point to not be
an outlier, assume we have a collection of points 
$\cc \subset S^1$ where 
$S^1 := \{ x \in \R^2 \mid \lvert x \rvert = 1 \} \subset \R^2$ 
denotes the unit-circle in $\R^2$ with respect to
the Euclidean distance on $\R^2$. 
Assume that every point $(x,y) \in \R^2$ satisfies 
that the angle $\tan^{-1}(y/x)$ is rational.
Provided the cardinality of $\cc$ is sufficiently large, 
the mean of $\cc$ will be the origin $(0,0) \in \R^2$.
Then suppose $(z,w) \in S^1$ with 
$\tan^{-1}(w/z) \in \R \setminus \Q$.
The distance of $(z,w)$ to the mean $(0,0)$ is the same 
as the distance of \textit{any} point $(x,y) \in \cc$ to 
$(0,0)$ (i.e. the distance is $1$) and hence any 
distance-to-the-mean based anomaly detection procedure
will not deem $(z,w)$ an outlier.
But $(z,w)$ making an irrational angle to the positive
x-axis means it differs from \textit{every} point in 
the corpus $\cc$, and therefore it should be an outlier.

With this in mind, a more useful point-of-view is to 
determine outliers by \textit{not} being similar to 
\textit{any} other instance within the dataset.
This avoids the issue of incorrectly deeming a rare
observation as anomalous.
But, from this point-of-view, it is important to allow the
corpus of observations that are deemed to be ``normal"
to be updated over time; after observing a rare event
that is nevertheless \textit{not} an anomaly,
we want to add it to the ``normal" corpus to ensure
that future observations of this event will not
be classified as an anomaly.

The \textit{k-nearest neighbour} (k-NN) approach uses
this idea by comparing an instance to its k
nearest neighbours within the dataset. It is based 
on the assumption that usual data points have close 
neighbours in the training set, while unusual points are
located far from their neighbours \cite{FHK04}.
Loosely, a point is declared an outlier if it is located
far from its k-nearest neighbours. 
The choice of k can be treated as a hyper-parameter to be
optimised.

The k-NN approach requires one to make a choice of 
distance metric to use for the determination of a points 
k-nearest neighbours. The \textit{Mahalanobis distance} 
\cite{Mah36}, providing a notion of distance between a 
point and a distribution, is a popular choice considered 
in the works \cite{CSSQZ20,KS20}, for example.
The recent work \cite{CCFLS20} introduces the
\textit{variance norm} as a data-driven metric that offers
a mathematically rigorous extension of the Mahalanobis
distance.
In particular, the variance norm coincides with the
Mahalanobis distance when the sample covariance matrix is
finite-dimensional and has full column rank \cite{CCFLS20}. 
However, unlike the Mahalanobis distance, the variance 
norm remains well-defined when these assumptions are
\textit{not} satisfied.
As an example, multi-collinearity may often lead to data 
exhibiting rank-deficiency \cite{CCFLS20}.

We turn our attention to the anomaly detection method
proposed in \cite{CCFLS20} of using a notion of
\textit{conformance} to
determine whether an entire stream is an anomalous
object compared to a corpus of normal streams.
A version of this work first appeared in 2020 \cite{ACFL20}.
Our presentation in this section focuses on the majorly
updated version of the article \cite{CCFLS20} that
appeared in December 2023.

Unlike earlier tasks considered in this article, 
anomaly detection is often \textit{semi-supervised}.
The data for training consists of a collection of 
streams of increments, with the collection denoted by 
$\Omega_{\s(V)}$ to match our notation, that are normal objects.
Only the normal objects are available during training, 
unlike the availability of two types of objects in 
binary classification problems. 
In fact, the anomaly detected might not even belong to 
the same class as the normal objects; their only 
defining characteristic is that they are different 
from all the objects in the corpus.
A brief outline of the approach proposed in \cite{CCFLS20} 
is the following.

\begin{enumerate}
    \item Transform each stream $\bx \in \Omega_{\s(V)}$
    to its path signature in $T((V))$.
    \item Fit the data-driven variance norm to the 
    resulting corpus of path signatures in $T((V))$.
    \item Use the nearest neighbour algorithm in the 
    tensor algebra $T((V))$ as a downstream metric-based 
    anomaly detector.
\end{enumerate}
\vskip 4pt
\noindent
Whilst the variance norm is considered on the tensor algebra
$T((V))$ (or possibly on truncations of the tensor algebra),
it actually makes sense in a general Banach space setting.
Thus, to both match the framework assumed in \cite{CCFLS20}
and avoid clashing with the notation we have adopted 
throughout this article, we fix a (possibly infinite 
dimensional) Banach space $W$ and explain how the variance
norm is defined with respect to a finite collection of elements 
$\cc = \{ w_1  , \ldots , w_n \} \subset W$ 
for some integer $n \in \Z_{\geq 1}$.

When $\dim(W) < \infty$ one can consider the Mahalanobis
distance in this setting. 
To be more precise, we recall that if
$D$ is a distribution with
mean $\mu$ and positive-definite covariance matrix $S$,
then the Mahalanobis distance between a point $y$ and
the distribution $D$ is defined to be 
\begin{equation}
    \label{eq:Mahalanobis_distance_def}
        d_M(y,D) := \sqrt{(y-\mu)^T S^{-1} (y - \mu)}
\end{equation}
where the superscript $T$ denotes taking the transpose.
If we take $D := \frac{1}{n} \sum_{i=1}^n \de_{w_i}$ 
to be the empirical distribution summarising the corpus
$\cc$ then $\mu = \mu_{\cc} := \frac{1}{n} \sum_{i=1}^n w_i$
and $S = (X - \mathbf{1}_n\mu_{\cc})^T 
(X - \mathbf{1}_n\mu_{\cc})$ where 
$X$ is the $(n \times d)$ matrix whose rows are the 
elements $w_1 , \ldots , w_n$ with respect to some basis 
of $W$, and $\mathbf{1}_n := (1, \ldots , 1) \in \R^n$.
Then the Mahalanobis distance of $y \in W$ to the 
empirical distribution summarising the corpus $\cc$ is 
\begin{equation}
    \label{eq:Mahalanobis_distance_empirical_dist}
        d_M(y,X) := \sqrt{(y - \mu_{\cc})^T 
        \left[(X - \mathbf{1}_n \mu_{\cc})^T 
        (X - \mathbf{1}_n \mu_{\cc}) \right]^{-1}
        (y - \mu_{\cc})}.
\end{equation}
Two limitations of
\eqref{eq:Mahalanobis_distance_empirical_dist}
observed in \cite{CCFLS20} are:

\begin{itemize}
    \item Computing a distance to the mean for anomaly 
    detection may not be suitable for certain situations.
    The example of having the corpus $\cc$ uniformly 
    distributed in the Euclidean 
    unit circle $S^1 \subset \R^2$ discussed at the start 
    of this section illustrates this point. 
    To provide a second example, let $S^2 \subset \R^3$
    denote the Euclidean unit sphere in $\R^3$, i.e.
    $S^2 := \big\{ (x,y,z) \in \R^3 \mid x^2 + y^2 + z^2 = 1 
    \big\}$.
    Then suppose the corpus $\cc$ is uniformly distributed
    in $S^2_{z=0} := \big\{ (x,y,z) \in S^2 \mid z=0 \big\}$.
    Provided the cardinality of the corpus $\cc$ is 
    large enough, the mean will be the origin
    $(0,0,0) \in \R^3$.
    Hence in terms of the distance to the mean, the point
    $(0,0,1) \in S^2$ is not anomalous compared to the corpus
    $\cc$. However, it evidently should be classified as
    an anomaly since it differs from every point in the 
    corpus $\cc$.
    \item It is unclear how to proceed when the 
    empirical distribution covariance matrix 
    $(X - \mathbf{1}_n\mu_{\cc})^T 
    (X - \mathbf{1}_n \mu_{\cc})$ is singular.
    This is often the case when, for example, the 
    streamed data exhibits multi-colinearity 
    between the channels.
\end{itemize} 
\vskip 4pt
\noindent
These limitations are addressed by the variance norm 
introduced in \cite{CCFLS20}.
We now define this notion in the setting of the 
corpus $\cc = \{ w_1 , \ldots , w_n \} \subset W$, 
and stress that there is no assumption made on the 
dimension of $W$. 
Let $\mu_{\cc} := \frac{1}{n} \sum_{i=1}^n w_i$
be the mean of the corpus $\cc$.
We first define a bilinear form 
$\mathbf{q} : W^{\ast} \times W^{\ast} \to \R$ 
by setting, for $\sigma , \vph \in W^{\ast}$,
\begin{equation}
    \label{eq:covariance_bilin_form_def}
        \mathbf{q}(\sigma,\vph) := \sum_{i=1}^n 
        \sigma(w_i - \mu_{\cc}) \vph(w_i - \mu_{\cc}).
\end{equation} 
Subsequently define a map 
$\mathbf{p} : W \times W \to \R$ by setting, 
for $w,u \in W$,
\begin{equation}
    \label{eq:variance_bilin_form_def}
        \mathbf{p}(w,u) := \sup \Big\{ 
        \sigma(w - \mu_{\cc})\sigma(u - \mu_{\cc}) 
        \mid \sigma \in W^{\ast} \text{ with }
        \mathbf{q}(\sigma,\sigma) \leq 1 \Big\}.
\end{equation}
Then the \textit{variance norm} of $w \in W$ is defined 
to be 
\begin{equation}
    \label{eq:variance_norm_def}
        \lvert\lvert w\rvert\rvert_{\var} 
        := \sqrt{\mathbf{p}(w,w)}.
\end{equation}
When $W$ is finite dimensional, 
the following key properties of
the variance norm $\lvert\lvert \cdot\rvert\rvert_{\var}$
are established in
Theorem 3.1 in \cite{CCFLS20}.

\begin{theorem}[Properties of $||\cdot||_{\var}$ when 
$\dim(W) < \infty$; 
Theorem 3.1 in \cite{CCFLS20}]
\label{theorem:p_props_finite_dim_W}
Assume $W$ is a finite dimensional real Banach space, 
$n \in \Z_{\geq 1}$ and
$\cc = \{w_1 , \ldots , w_n \} \subset W$. 
Fix a basis for $W$ and let $X$ be the matrix whose rows
are the elements $w_1 , \ldots , w_n$ expressed with 
respect to this basis.
Let $\mathbf{1}_n := (1, \ldots , 1) \in \R^n$
and $\mu_{\cc} := \frac{1}{n} \sum_{i=1}^n w_i$
be the mean of the corpus $\cc$.
Then the following properties are true.
\begin{enumerate}[label=(\Alph*)]
    \item If $w \in \Span( \cc )$ then
    $\lvert \lvert w \rvert \rvert_{\var}^2 =
    \mathbf{p}(w,w) = 
    (w - \mu_{\cc})^T \big[(X-\mathbf{1}_n \mu_{\cc})^T 
    (X - \mathbf{1}_n \mu_{\cc}) \big]^{\dag} 
    (w - \mu_{\cc})$
    where the superscript $\dag$ is used to denote the 
    Moore-Penrose pseudo-inverse. 
    \item If $w \notin \Span(\cc)$ then 
    $\lvert \lvert w \rvert \rvert_{\var}^2
    = \mathbf{p}(w,w) = \infty$.
\end{enumerate}
\end{theorem}
\vskip 4pt
\noindent 
We stress that the choice of a basis of $W$ in Theorem 
\ref{theorem:p_props_finite_dim_W} is purely for the 
purposes of expressing the matrix $X$. 
The result itself is \textit{not} dependent on the choice of 
basis; one could select any other basis and obtain the 
same result, only now with the matrix $X$ expressed with 
respect to the new basis of $W$ instead.

Using the variance norm we define the
\textit{conformance distance} of an 
element $w \in W$ to the corpus $\cc$ is defined to be 
(see Definition 3.2 in \cite{CCFLS20})
\begin{equation}
    \label{eq:conformance_dist_def}
        d_{\cc}(w, \cc) := 
        \min_{i \in \{1, \ldots ,n\}} 
        \big\lvert\big\lvert w - w_i 
        \big\rvert\big\rvert_{\var}^2. 
\end{equation}
Thus in the case that $W$ is finite dimensional we have, 
via Theorem \ref{theorem:p_props_finite_dim_W},
for every $w \in W$ that either 
$d_{\cc}(w,\cc) = \infty$ if $w \notin \Span(\cc)$, 
or 
\begin{equation}
    \label{eq:conformance_dist_finite_dim_in_span}
        d_{\cc}(w,\cc) = 
        (w - w_i - \mu_{\cc})^T 
        \big[(X-\mathbf{1}_n \mu_{\cc})^T 
        (X - \mathbf{1}_n \mu_{\cc}) \big]^{\dag} 
        (w - w_i - \mu_{\cc})
\end{equation} 
for some $i \in \{1, \ldots ,n\}$ if $w \in \Span(\cc)$.
It follows from \eqref{eq:conformance_dist_finite_dim_in_span}
that if the matrix $X$ has full column rank and 
$w \in \Span(\cc)$ then the conformance distance is equal to 
the nearest neighbour Mahalanobis distance \cite{CCFLS20}.

Allowing the conformance distance to be infinity in 
some cases is a distinction from most Mahalanobis distance 
based anomaly detection methods \cite{CCFLS20}.
But this provides the benefit of avoiding an insensitivity 
introduced by the use of the pseudo-inverse of the matrix 
$(X-\mathbf{1}_n \mu_{\cc})^T (X - \mathbf{1}_n \mu_{\cc})$.
As explained in section 3.2.2 of \cite{CCFLS20}, 
the quantity 
$(w - w_i - \mu_{\cc})^T 
\big[(X-\mathbf{1}_n \mu_{\cc})^T 
(X - \mathbf{1}_n \mu_{\cc}) \big]^{\dag} 
(w - w_i - \mu_{\cc})$
remains invariant under translation of $w$ by a 
particular class of elements $z \in W$. Loosely, 
the class of elements is determined via the SVD of 
the matrix $X$; see section 3.2.2 in \cite{CCFLS20} 
for full details.

This is problematic from an anomaly detection perspective
since the class of elements $z \in W$ for which this
quantity is invariant under $w \mapsto w + z$ enables one 
to map $w \in \Span(\cc)$ to an element
$w + z \notin \Span(\cc)$.
Being outside $\Span(\cc)$ ought to make $w+z$ an anomaly; 
but the invariance means it will not be marked as an outlier 
by a method based solely on the pseudo-inverse.
However, the conformance distance does mark it as an 
exceptional point by returning the value $\infty$; 
despite such elements determining a different type of 
anomaly, the conformance distance remains able to detect them
\cite{CCFLS20}.

We now illustrate how the variance norm and conformance 
distance are used in \cite{CCFLS20} to define the 
SigMahaKNN method for detecting anomalies in streamed data.
In the notation used in this article, the SigMahaKNN method
is developed in the setting that $V := \R^d$ for some 
integer $d \in \Z_{\geq 1}$.
That is, for an integer $M \in \Z_{\geq 1}$, we have 
a collection of streams of increments 
$\Omega_{\s(\R^d)} = 
\big\{ \bx_1 , \ldots , \bx_M \big\} \subset \s\big(\R^d\big)$.
The SigMahaKNN method is designed to classify whether 
a stream
$\bx \in \s\big(\R^d\big) \setminus
\Omega_{\s\big(\R^d\big)}$ 
is an outlier compared to the collection
$\Omega_{\s\big(\R^d\big)}$.

We first pick an integer $N \in \Z_{\geq 1}$ and consider 
the corpus $\cc := \big\{ S^{(N)}(\bx) \mid 
\bx \in \Omega_{\s(\R^d)} \big\} \subset
T^{(N)} \big( \R^d \big)$
of elements in the truncated tensor algebra
$T^{(N)} \big( \R^d \big)$
obtained by taking the truncation to depth $N$ of the 
signature of each stream
$\bx \in \Omega_{\s \big( \R^d \big)}$ 
(cf. Section \ref{good_props}). 
Recalling Subsection \ref{ten_alg}, we can view the 
corpus $\cc$ as a collection of $M$ points in the Euclidean 
space $\R^{m(d,N)}$ for 
\begin{equation}
    \label{eq:dim_m_def}
        m = m(d,N) := \sum_{j=0}^N d^j = 
        \Bigg\{ 
        \begin{array}{ccc}
             N+1 & \text{if} & d = 1 \\
             \frac{d^{N+1} -1}{d - 1} & \text{if} & d \geq 2.
        \end{array}
\end{equation}
For notational convenience, we let 
$\cc = \{ z_1 , \ldots , z_M \} \subset \R^M$.
Define an empirical measure 
$\de_{\cc} := \frac{1}{M} \sum_{j=1}^M \de_{z_j}$
summarising $\cc$ (cf. Section \ref{dist_reg_expect_sig}), 
and subsequently let 
$\mu_{\cc} := \frac{1}{M} \sum_{j=1}^M z_j$
denote its mean.
Finally, to match the notation adopted in \cite{CCFLS20},
we let $\lvert \lvert \cdot \rvert\rvert_{\Sig^N(\cc)}$
denote the 
conformance distance $d_{\cc}(\cdot , \cc)$ defined 
by \eqref{eq:conformance_dist_def} for the corpus
$\cc \subset \R^m$.

It is $\lvert \lvert \cdot \rvert\rvert_{\Sig^N(\cc)}$
that determines 
the anomaly score for the SigMahaKNN algorithm;
streams with conformance distance higher than a 
user-determined threshold are anomalies.
That is, suppose we determine an appropriate threshold
$a > 0$.
Then let $\bx \in \s \big(\ R^d \big)$ be a stream that is not
in the original collection $\Omega_{\s \big( \R^d \big)}$.
The SigMahaKNN method determines whether $\bx$ is an 
anomaly compared to the collection
$\Omega_{\s \big( \R^d \big)}$ as follows.

\begin{enumerate}[label=(\Alph*)]
    \item Compute
    $S^{(N)}(\bx) \in T^{(N)} \big(\R^d \big) \cong \R^m$, 
    the truncated to depth $N$ signature of
    the stream $\bx$.
    \item Compute
    $\big\lvert \big\lvert S^{(N)}(\bx)
    \big\rvert \big\rvert_{\Sig^N(\cc)}$,
    the conformance distance of $S^{(N)}(\bx)$ relative
    to the corpus $\cc$.
    \item If 
    $\big\lvert \big\lvert S^{(N)}(\bx)
    \big\rvert \big\rvert_{\Sig^N(\cc)} > a$ then
    the stream $\bx$ is classified as an anomaly; 
    otherwise the stream $\bx$ is classified as normal.
\end{enumerate}
\noindent 
An obvious question is how the threshold value $a > 0$
should be determined. 
The authors propose using the nearest neighbour distance
to set an appropriate threshold in \cite{CCFLS20}.
This is done as follows.

\begin{enumerate}[label=(\Alph*)]
    \item Randomly split $\cc = \cc_1 \sqcup \cc_2$ 
    into two equal-sized parts $\cc_1$ and $\cc_2$.
    \item Compute the empirical cdf of the nearest 
    neighbour distance for $\cc_1$ using $\cc_2$ as 
    the corpus.
    \item Choose an appropriate tail quantile in the 
    empirical cdf to set the threshold value $a > 0$.
\end{enumerate}
\noindent
The choice of the depth $N \in \Z_{\geq 1}$ to which
the signature is truncated determines a parameter that
may be varied.
Intuitively, the order $N$ is a measure of the 
resolution at which the streams are viewed.
For small $N$, only general features of the streams are 
considered, with more details of the streams considered as 
$N$ increases.

The following theoretical properties of the conformance
distance $\lvert \lvert \cdot \rvert \rvert_{\Sig^N(\cc)}$
as the truncation level
$N \in \Z_{\geq 1}$ varies are established in \cite{CCFLS20}.
The first establishes that for a fixed stream
$\bx \in \s(\R^d)$, the mapping
$N \mapsto \big\lvert \big\lvert S^{(N)}(\bx)
\big\rvert \big\rvert_{\Sig^N(\cc)}$ 
is monotonically increasing 
(see Proposition 4.1 in \cite{CCFLS20}).
Hence the extent to which a stream $\bx$ is considered an 
outlier compared to the original collection 
$\Omega_{\s \big( \R^d \big)}$
increases as the resolution $N$ at 
which we examine the streams increases.

The second establishes that if
$\bx \notin \Omega_{\s \big( \R^d \big)}$
then there exists an integer $N_{\ast} \in \Z_{\geq 1}$
such that for every integer $N \in \Z$ with $N \geq N_{\ast}$
the conformance distance of $S^{(N)}(\bx)$ with respect 
to the corpus $\cc$ is $\infty$
(see Proposition 4.2 in \cite{CCFLS20}).
Thus if a stream $\bx$ is \textit{not} in the
original collection
$\Omega_{\s \big( \R^d \big)}$ of streams,
we are guaranteed that it
will be classified as an outlier provided we examine the 
streams at a sufficiently high resolution.

Four major benefits of the SigMahaKNN method proposed in 
\cite{CCFLS20} are:
\begin{itemize}
    \item It is \textit{dimensionless} in the sense that it 
    is independent of the unit of measurement of the streams.
    Moreover it is invariant to time-reparameterisation and
    concatenation of streams.
    These are direct consequence of the invariance of the 
    signature covered in Theorem \ref{tree_like_unique} 
    in this article. Particular theoretical results 
    establishing these property for the SigMahaKNN method 
    may be found in Theorems 4.1, 4.2, and 4.3 in 
    \cite{CCFLS20}.
    \item It is a data-driven extension of Mahalanobis 
    distance based models that continues to make sense 
    even when the sample covariance matrix is singular. 
    In particular, it can deal with data exhibiting 
    multi-collinearity between channels.
    \item No distribution assumptions are required for 
    the original collection of streams 
    $\Omega_{\s \big( \R^d \big)} \subset \s \big( \R^d \big)$. 
    The method can be applied to any finite collection 
    of streams in $\s \big( \R^d \big)$.
    \item Different types of anomalies that are missed 
    by Mahalanobis distance based models are detected 
    by the SigMahaKNN model.
\end{itemize}
\noindent
For the remainder of this section we turn our attention
to the numerical performance of the SigMahaKNN method 
as reported in section 5 of \cite{CCFLS20}.
A detailed algorithm for the implementation of the 
SigMahaKNN method is provided in algorithm 1 in 
\cite{CCFLS20}.
In particular, it includes consideration of numerical 
rank deficiency issues involved in the computation of 
the Moore-Penrose pseudo-inverse via SVD factorisation 
that we have omitted.

For the two example problems we present here, 
the SigMahaKNN method is compared with 
\textit{isolation forest} \cite{LTZ08} 
and the \textit{local outlier factor method} \cite{BKNS00}.
In order to use these well-known anomaly detection methods 
on streamed data, the authors of \cite{CCFLS20} 
consider them with either the moment features (mean and 
covariance of different channels of the stream), 
of the signature features.
This leads to four baselines: \textbf{IF-M} (Isolation 
Forest with Moment features), \textbf{IF-S} (Isolation 
Forest with Signature features), \textbf{LOF-M} 
(Local Outlier Factor with Moment features), and 
\textbf{LOF-S} (Local Outlier Factor with Signature features).

The SigMahaKNN method is evaluated using the 
PenDigitis-orig data set \cite{DG19}, consisting
of 10992 instances of handwritten digits captured from 
44 subjects, in section 5 of \cite{CCFLS20}.
Each instance is represented as a 
two-dimensional stream, based on sampling of the 
pen position. 

Given a digit $k$, a set of normal data $\ci_{\Normal}$
is defined to be the set of instances representing the 
digit $k$.
The subset of $\ci_{\Normal}$ labelled as `training' by 
the annotators is taken for the collection
$\Omega_{\s \big( \R^2 \big)}$
of normal streams.
Further $\cY$ is taken to be the set of instances labelled
as `testing' by the annotators, which results in 
$\#(\cY) = 3498$.
Finally, $\ci_{\Anomaly}$ is the subset of $\cY$ not 
representing digit $k$. 
On average, taken over the consideration of all possible 
digits $k$, one has the cardinalities 
$\# \Big(\Omega_{\s \big( \R^2 \big)}\Big) = 749.4$ and 
$\# \big(\ci_{\Anomaly}\big) = 3148.2$.
Min-Max normalisation is applied to each individual stream
since each digit class is invariant to translation and 
scaling.

Truncated signatures of orders $N \in \{1 , \ldots , 5\}$ are
considered without any augmentation. 
The results, based on aggregating conformance values 
across the set of possible digits, are summarised in Table 
\ref{Handwritten_Anom}.
Once the truncation level is at least $2$ the SigMahaKNN 
out-performs all the baseline comparison models.

\begin{table}[ht]	
    \centering
    \begin{tabular}{|c|c|c|c|c|c|}
        \hline
        & $N=1$ & $N=2$ & $N=3$ & $N=4$ & $N=5$ \\ 
        \hline 
        \textbf{SigMahaKNN} & 0.870 & \textbf{0.942} & 
        \textbf{0.948} & \textbf{0.954} & \textbf{0.956} \\
        \textbf{IF-M} & - & - & - & - & 0.618 \\
        \textbf{IF-S} & \textbf{0.888} & 0.931 & 0.916 
        & 0.875 & 0.834 \\
        \textbf{LOF-M} & - & - & - & - & 0.514 \\
        \textbf{LOF-s} & 0.563 & 0.584 & 0.582 
        & 0.582 & 0.582 \\
        \hline
    \end{tabular}
    \caption{Handwritten digits data: performance quantified 
    using ROC AUC in response to signature
    order $N$. Bootstrapped standard errors based on 
    $10^4$ samples are around $0.003$. 
    Table 1 in \cite{CCFLS20}}
    \label{Handwritten_Anom}
\end{table}

The second numerical example from \cite{CCFLS20} that 
we cover considers a sample of marine 
\href{https://coast.noaa.gov/htdata/CMSP/AISDataHandler/2017/index.html}{vessel traffic data} 
based on the the automatic identification
system (AIS) which reports a ship's geographical position 
alongside other vessel information.
This dataset, collected by the US Coast Guard in January 2017, 
contains a total of 31 884 021 geographical positions 
recorded for 6282 distinct vessel identifiers.
The authors of \cite{CCFLS20} 
consider the stream of timestamped latitude/longitude 
position data associated with each vessel.

Vessel data with invalid identifiers of invalid length 
information were discarded. Streams are compressed to 
only contain points that are at least 10m away from the 
previous position, and, to help ensure that streams are 
faithful representations of ship movement, 
vessels whose starting and 
finishing points are within 5km are discarded. To 
evaluate the effect of the stream length, the streams are
disintegrated so that the length between initial and final 
points in each sub-stream remains constant with
$D \in \{ 4\km , 8\km, 16\km, 32\km \}$. Only sub-streams 
whose maximum distance between successive points
is less than 1km are subsequently retained. A sub-stream 
is deemed normal if it belongs to a vessel of length
at least 100m, and deemed anomalous if it corresponds to 
a vessel of length no greater than 50m.

A finite collection of streams is obtained from 607 vessels 
whose sub-streams total between 10111 $(D=32\km)$
and 104369 $(D=4\km)$. 
A subset of normal instances used for testing is obtained 
from 607 vessels whose sub-streams total between 11254 
$(D=32\km)$ and 114071 $(D=4\km)$.
Finally, a subset of anomalous instances is obtained from 997
vessels whose sub-streams total between 8890 $(D=32\km)$ 
and 123237 $(D=4\km)$.

Imbalance in the number of sub-streams is accounted for 
by using, for each of the aforementioned three subsets, 
a weighted sample of 5000 instances. 
After computing the sub-streams and transforming them 
as outlined above, Min-Max normalisation is applied.
The difference between successive timestamps is 
augmented into the streams as a new channel to account 
for velocity (cf. Time Augmentation in Subsection 
\ref{dataset_to_path}).
Signatures truncated to depth $N=3$ are considered. 
For baseline comparison models, each sub-stream is 
also summarised by estimating its component-wise mean
and covariance, retaining the upper triangular part of 
the covariance matrix. 
The resulting set of features has dimensionality  
$\frac{1}{2}(n^2 +3n)$, and these features are provided as 
input to an isolation forest \cite{LTZ08}. The 
isolation forest is trained using 100 trees and for 
each tree in the ensemble using 256 samples represented 
by a single random feature. 

Table \ref{vessels_Anom} displays results for the proposed 
signature approach in comparison to the baseline, for 
combinations of stream transformations and values of 
the sub-stream length D. 
Signature conformance yields higher ROC-AUC scores than 
the baseline models for the majority of the parameter 
combinations. The maximum ROC-AUC score
(highlighted in bold in Table \ref{vessels_Anom}) 
is achieved by the signature
conformance for $D = 32\km$ using a combination of 
lead-lag, time differences and invisibility reset 
transformations. The score of 0.891 is a 10 
percentage points gain compared to the best-performing 
baselines parameter combination for $D=32\km$.
This best-performing baseline parameter combination for 
$D=32\km$ is achieved by the IF-M model using a combination 
of time differences and invisibility reset
transformations, achieving a ROC-AUC score of 0.828.

\begin{table}[H]	
	\centering
        \scalebox{0.9}{
	\begin{tabular}{|c|c|c|c|c|c|c|c|}
        \hline
		\textbf{Model} & \textbf{Lead-Lag} & 
		\textbf{Time-Diff} & \textbf{Time-inv}& 
		$\mathbf{D=4\km}$ & $\mathbf{D=8\km}$ & 
		$\mathbf{D=16\km}$ & $\mathbf{D=32\km}$  \\ 
		\hline	
		\textbf{SigMahaKNN}  & No & No & No & 
		0.723 & 0.706 & 0.705 & 0.740 \\
        \textbf{IF-M} & No & No & No & 
		0.714 & 0.712 & 0.727 & 0.727 \\
        \textbf{IF-S} & No & No & No & 
		0.627 & 0.623 & 0.645 & 0.660 \\
        \textbf{LOF-M} & No & No & No & 
		0.543 & 0.542 & 0.522 & 0.513 \\
        \textbf{LOF-S} & No & No & No & 
		0.484 & 0.500 & 0.491 & 0.492 \\
        \hline	
		\textbf{SigMahaKNN}  & No & No & Yes & 
		0.776 & 0.789 & 0.785 & 0.805 \\
        \textbf{IF-M} & No & No & Yes & 
		0.781 & 0.785 & 0.776 & 0.790 \\
        \textbf{IF-S} & No & No & Yes & 
		0.686 & 0.701 & 0.715 & 0.718 \\
        \textbf{LOF-M} & No & No & Yes & 
		0.555 & 0.585 & 0.564 & 0.535 \\
        \textbf{LOF-S} & No & No & Yes & 
		0.565 & 0.572 & 0.562 & 0.520 \\
        \hline	
		\textbf{SigMahaKNN}  & No & Yes & No & 
		0.810 & 0.813 & 0.818 & 0.848 \\
        \textbf{IF-M} & No & Yes & No & 
		0.767 & 0.772 & 0.786 & 0.804 \\
        \textbf{IF-S} & No & Yes & No & 
		0.731 & 0.714 & 0.737 & 0.771 \\
        \textbf{LOF-M} & No & Yes & No & 
		0.547 & 0.543 & 0.526 & 0.520 \\
        \textbf{LOF-S} & No & Yes & No & 
		0.511 & 0.505 & 0.493 & 0.494 \\
        \hline	
		\textbf{SigMahaKNN}  & No & Yes & Yes & 
		0.839 & 0.860 & 0.863 & 0.879 \\
        \textbf{IF-M} & No & Yes & Yes & 
		0.830 & 0.823 & 0.831 & 0.828 \\
        \textbf{IF-S} & No & Yes & Yes & 
		0.777 & 0.784 & 0.789 & 0.808 \\
        \textbf{LOF-M} & No & Yes & Yes & 
		0.559 & 0.589 & 0.572 & 0.545 \\
        \textbf{LOF-S} & No & Yes & Yes & 
		0.565 & 0.572 & 0.561 & 0.520 \\
        \hline	
		\textbf{SigMahaKNN}  & Yes & No & No & 
		0.811 & 0.835 & 0.824 & 0.837 \\
        \textbf{IF-M} & Yes & No & No & 
		0.696 & 0.704 & 0.711 & 0.724 \\
        \textbf{IF-S} & Yes & No & No & 
		0.617 & 0.596 & 0.634 & 0.668 \\
        \textbf{LOF-M} & Yes & No & No & 
		0.543 & 0.541 & 0.522 & 0.513 \\
        \textbf{LOF-S} & Yes & No & No & 
		0.484 & 0.500 & 0.491 & 0.493 \\
        \hline	
		\textbf{SigMahaKNN}  & Yes & No & Yes & 
		0.812 & 0.835 & 0.833 & 0.855 \\
        \textbf{IF-M} & Yes & No & Yes & 
		0.758 & 0.759 & 0.767 & 0.773 \\
        \textbf{IF-S} & Yes & No & Yes & 
		0.692 & 0.701 & 0.691 & 0.713 \\
        \textbf{LOF-M} & Yes & No & Yes & 
		0.566 & 0.556 & 0.542 & 0.533 \\
        \textbf{LOF-S} & Yes & No & Yes & 
		0.564 & 0.569 & 0.558 & 0.518 \\
        \hline	
		\textbf{SigMahaKNN}  & Yes & Yes & No & 
		0.845 & 0.861 & 0.862 & 0.877 \\
        \textbf{IF-M} & Yes & Yes & No & 
		0.747 & 0.763 & 0.780 & 0.785 \\
        \textbf{IF-S} & Yes & Yes & No & 
		0.725 & 0.692 & 0.716 & 0.757 \\
        \textbf{LOF-M} & Yes & Yes & No & 
		0.547 & 0.543 & 0.526 & 0.520 \\
        \textbf{LOF-S} & Yes & Yes & No & 
		0.530 & 0.533 & 0.553 & 0.600 \\
        \hline	
		\textbf{SigMahaKNN}  & Yes & Yes & Yes & 
		\textbf{0.848} & \textbf{0.863} & \textbf{0.870} 
        & \textbf{0.891} \\
        \textbf{IF-M} & Yes & Yes & Yes & 
		0.811 & 0.813 & 0.809 & 0.823 \\
        \textbf{IF-S} & Yes & Yes & Yes & 
		0.779 & 0.782 & 0.801 & 0.823 \\
        \textbf{LOF-M} & Yes & Yes & Yes & 
		0.572 & 0.563 & 0.554 & 0.544 \\
        \textbf{LOF-S} & Yes & Yes & Yes & 
		0.574 & 0.588 & 0.589 & 0.584 \\
        \hline
	\end{tabular}
        }
        \caption{SigMahaKNN, IF-M, IF-S, LOF-M, and 
        LOF-S on Marine vessel traffic data: performance 
        quantified using ROC AUC. 
        Best across all transformations and models 
        are in bold.
        Combination of 
        Tables 2, 3, 4, 5, and 6 in \cite{CCFLS20}}
	\label{vessels_Anom}
\end{table}

Subsequent to the appearance of \cite{CCFLS20}, 
the outlier detection framework of the SigMahaKNN method 
is considered for anomaly detection on radio astronomy
data in \cite{Arr24}.
In particular, the authors of \cite{Arr24} develop the 
SigNova method, inspired by the SigMahaKNN framework,
that can detect much fainter 
\textit{radio frequency interference} (RFI) than the 
two main frameworks SSINS 
\cite{BBHMRW19} and 
\href{https://gitlab.com/aroffringa/aoflagger}{AOFlagger} 
\cite{GOR12,HKOW15} are capable of detecting.
The SigNova method implements all the steps of the SigMahaKNN 
framework with an additional application of the 
\href{https://github.com/datasig-ac-uk/pysegments}{pysegments} 
segmentation algorithm \cite{Arr24}.

In \cite{CGZ24} the authors propose a novel framework for
Mahalanobis-type anomaly detection on any Banach space
(including, in particular, any Hilbert space) that is
based upon the generalised notion of variance norm
introduced in \cite{CCFLS20} and ideas from Cameron-Martin 
spaces (see e.g. \cite{Lif12,Bog15}). 
In the Hilbert space setting, it is established that the
variance norm depends solely on the given inner product,
and hence that the kernelised Mahalanobis distance can be
recovered by working on 
\textit{Reproducing Kernel Hilbert Spaces} (RKHS) \cite{CGZ24}.
This underpins the authors introduction of the
\textit{Kernelised Nearest-Neighbour Mahalanobis} distance for
semi-supervised anomaly detection in \cite{CGZ24}.
Both theoretical justifications and empirical validations of
this approach can be found in \cite{CGZ24}.

The paper \cite{AGTZ22} considers the use of path signature 
techniques for anomaly detection with the aim of detecting 
market manipulation attempts from financial data.
There are two main reasons we delay discussing this approach 
until Section \ref{sec:randomised_signature}.
The first is that the authors of \cite{AGTZ22} make use of
``off-the-shelf" anomaly detection algorithms without 
alteration; this is quite distinct in spirit from the 
approach in \cite{CCFLS20} of introducing a novel measure 
of similarity related to a reformulation of what 
being an anomaly \textit{should} mean.
The second is that a major novelty of the approach proposed
in \cite{AGTZ22} is the use of \textit{randomised signatures}, 
which are noteworthy enough to warrant their own dedicated 
section.

\section{Randomised Signature}
\label{sec:randomised_signature}
In this section we discuss
\textit{randomised signatures}  
as presented in, for example, \cite{AGTZ22}.
Loosely speaking, randomised signatures exploit the notion of
\textit{reservoir computing} for modelling input-output systems
to provide more concrete
realisations (or, algebraically speaking, representations) 
of the abstract Tensor algebra (cf. Subsection \ref{Ten_Alg}).
The close link between path signatures and CDEs discussed in
Section \ref{CDE_log-ODE} is central to this approach.
We begin our presentation with a brief introduction to 
reservoir computing.

Fix non-negative integers $n,k,e \in \Z_{\geq 1}$.
\textit{Reservoir Computers} (RCs) are state-space 
transformations
given by recurrent neural networks (RNNs) determined by two
maps; namely, a \textit{reservoir} (or \textit{state}) map
$F: \R^n \times \R^e \to \R^n$, 
and a \textit{readout} map $h : \R^n \to \R^k$.
The integer $n \in \Z_{\geq 1}$ is the number
of virtual neurons of the system.
RCs transform (or filter) a discrete-time input 
$z = (z_m)_{m \in \Z} \subset \R^e$ into an output signal
$y = (y_m)_{m \in \Z} \subset \R^k$ using the state-space 
transformation given, for $t \in \Z$, by 
\begin{equation}
    \label{eq:RC_state-space_transform}
        y_t := h \big( x_t \big) \in \R^k
        \qquad \text{for} \qquad 
        x_t := F \big( x_{t-1} , z_t \big) \in \R^n.
\end{equation} 
The RC satisfies the \textit{echo state property} (ESP) if
any input signal $(z_m)_{m \in \Z} \subset \R^e$
results in a unique output under the transformation
detailed in \eqref{eq:RC_state-space_transform}.
Typically, the reservoir map $F$ does \textit{not} depend on 
the specific dynamics, and instead provides a collection (or
reservoir) of features.
The readout map $h$ is the object that is actually trained
to use the features provided by the reservoir map $F$ to
match the observed collection of outputs sufficiently well.

The RC approach enables one to translate infinite dimensional
problems regarding filters to analogous
questions related to the reservoir and readout maps that
generate the filter and are defined on
finite dimensional spaces. 
An in-depth discussion of reservoir computing covering
numerous important theoretical results can be found in 
\cite{GO18}, \cite{CGGOT20}, and the references there in.
A particularly noteworthy result is the universality property
established for a particular class of RCs in \cite{GO18}.
In \cite{GO18} the authors establish that the class of RCs 
for which the reservoir map $F : \R^n \times \R^e \to \R^n$ 
is given by 
$F(x,y) := \sigma ( \bA x + \bB y + \xi )$, 
for real matrices 
$\bA \in \R^{n \times n}$, $\bB \in \R^{n \times e}$, a vector
$\xi \in \R^n$, and a sigmoid function $\sigma : \R \to \R$ 
applied component wise, and the readout map 
$h : \R^n \to \R^k$ is given by 
$h(x) := \bW x$, for a real matrix $\bW \in \R^{k\times n}$,
are universal uniform approximants for a large class of 
discrete-time filters; see Theorem 4.1 in \cite{GO18} for 
the full detailed statement.

Reservoir computing offers an approach to the
task of describing the solution trajectories of
CDEs without having access to the vector fields. 
To be more precise, fix non-negative integers 
$d,l \in \Z_{\geq 1}$, fix $T > 0$, and (for simplicity) 
suppose that
$X : [0,T] \to \R^d$ is a continuous path of bounded variation, 
i.e. $X \in \cv^1([0,T];\R^d)$.
Further suppose that
$f: \R^l \to \bLL \big( \R^d ,\R^l \big)$
is continuous. Then, for a given $x_0 \in \R^l$, 
we may consider trying to find a path 
$z : [0,T] \to \R^l$ solving the following CDE driven by $X$: 
\begin{equation}
    \label{eq:RC_CDE_eqn_A}
        z_0 = x_0 \qquad \text{and} \qquad
        dz_t = f(z_t) dX_t
\end{equation} 
for every $t \in (0,T]$.
In Section \ref{CDE_log-ODE} we observed that if the CDE 
\eqref{eq:RC_CDE_eqn_A} is linear in the sense that
$f(z_t)dX_t = B(dX_t)z_t$ 
for a bounded linear map 
$B : \R^d \to \bLL \big(\R^l,\R^l\big)$, 
then the solution to \eqref{eq:RC_CDE_eqn_A} is given by
a linear combination of the terms of the signature of the
path $X$.
To be more precise, in this case the solution
to \eqref{eq:RC_CDE_eqn_A} is given by 
(cf. \eqref{lin_CDE_sol_sig_involve})
\begin{equation}
    \label{eq:RC_lin_CDE_sig_solution}
        z_t = \Bigg( \sum_{i=0}^{\infty} B^{\otimes i} 
        \Big( S^i_{0,t}(X) \Big) \Bigg)x_0
\end{equation}
Evidently, for linear CDEs, 
one approach is to take the (truncated) 
signature as the reservoir map, and subsequently take
the readout map to be a trainable linear map.
This strategy is completely analogous to the generic
strategy followed throughout this survey article.

However, as noted in \cite{CGGOT21}, 
the difficulty in calculating the precise values
of the reservoir given by the signature of a control path
$X \in \cv^1([0,T];\R^d)$ is in sharp contrast to the
usual setting for reservoir computing in which reservoirs
are easy to evaluate.
In \cite{CGGOT21} the authors overcome this issue by
introducing the \textit{randomised signature}.
The information of the signature is compressed via a random
projection to a lower dimensional Euclidean space. 
A key aspect is that the image of this projection can be
well-approximated by a dynamical system on the
lower dimensional
Euclidean space with random characteristics. 
In order to provide a more detailed explanation
of the randomised
signature we additionally make use of the presentation
provided in \cite{AGTZ22}.

Coarsely, the idea of randomised signatures is to provide a 
more concrete realisation of the tensor algebra
$T\big(\big(\R^{d+1}\big)\big)$ 
via mapping the canonical basis
$e_0 , \ldots ,e_d$ of $\R^{d+1}$ 
to a collection of informative vector fields
$\R^k \to \R^k$ for some moderately sized
integer $k \in \Z_{\geq 1}$.
The switch from $d$ to $d+1$ reflects that path $X$ is
replaced by its time-augmented variant where the running time
is recorded as a new additional zeroth component.
The choice of the form of vector fields is partly motivated by
both the aim to model a random projection of the signature,
and the universality results established in \cite{GO18} and
\cite{CGGOT20}.

To specify the class of vector fields $\R^k \to \R^k$ 
considered, fix a choice of a
bounded real analytic activation function $\sigma : \R \to \R$,
and, for each
$j \in \{0 , \ldots , d\}$, a choice of $k \times k$ matrix
$\bA_j \in \R^{k \times k}$ and vector $b_j \in \R^k$.
Then one may consider the map from $\R^{d+1}$ to the set of
vector fields on $\R^k$ given by mapping, for each
$j \in \{0, \ldots , d\}$, the basis element $e_j$ to
the vector field $f_j : \R^k \to \R^k$ defined, for 
$x \in \R^k$, by 
\begin{equation}
    \label{eq:vec_field_fj}
        f_j(x) := \sigma \big( \bA_j x + b_j \big).
\end{equation}
In \eqref{eq:vec_field_fj}, the activation function
$\sigma$ is applied component-wise.
Following the notation used in Definition 2.12 in 
\cite{AGTZ22}, the solution 
$\RS : [0,T] \to \R^k$ to the linear CDE
\begin{equation}
    \label{eq:random_sig_cde}
        \RS_0 \in \R^k
        \qquad \text{with} \qquad
        d \RS_t = \sum_{j=0}^d f_j(\RS_t) dX^j_t
        = 
        \sum_{j=0}^d 
        \sigma \big( \bA_j \RS_t + b_j \big) dX^j_t
\end{equation}
for every $t \in (0,T]$ is called \textit{randomised signature}.

Theoretical results in both \cite{CGGOT21} and \cite{AGTZ22} 
establish that, under certain assumptions,
randomised signatures
are universal approximants in the space of
continuous functions.
The approach in \cite{CGGOT21} is inspired
by reservoir computing
techniques, culminating in Theorem 3.3 which makes use of the
\textit{Johnson--Lindenstrauss} lemma to establish that
randomised signatures are as expressive as the signature
itself in a quantified sense.
In a similar spirit, Theorem 2.13 in \cite{AGTZ22} gives a more
direct argument to verify that randomised signatures inherit
the universality of the full signature provided the
representation from the free algebra generated by
$e_0 , \ldots , e_d$ to the algebra of differential operators
on $\R^k$ defined in \eqref{eq:vec_field_fj} is injective.
It is additionally required in Theorem 2.13 in \cite{AGTZ22} 
that $b_0 , \ldots , b_d , \bA_0 , \ldots , \bA_d$ are 
independent samples of a probability law absolutely continuous
with respect to the Lebesgue measure.

We end this section with an illustration
of the use of randomised
signature for anomaly detection presented in \cite{AGTZ22}.
In particular, the task of detecting market fraud by
identifying so-called \textit{Pump-and-Dump} (PD) 
attempts within various cryptocurrencies is considered in
\cite{AGTZ22}.
The lack of regulation of cryptocurrency brokers has resulted
in them being the target of numerous
market manipulation attempts.
A PD scheme is a particular type of
market manipulation strategy that is highly risky
for participants \cite{AGTZ22}.
Typically, a PD scheme involves the following general
steps happening over the course of 2--3 minutes.
\begin{itemize}
    \item \textbf{Pre-Pump}: 
    Organisers discretely advertise to their members
    which cryptocurrency to buy, which cryptocurrency to use
    to make the purchases, and the exact time to do so.
    \item \textbf{Pump}: 
    Members are urged to buy the target coin and hold it
    to cause a short-term price inflation.
    \item \textbf{Dump}: 
    After the price increases dramatically, some 
    members consolidate
    their gains by selling. This leads to a cascade as more
    and more participants choose to sell, often leading to 
    the price returning close to its pre-pump level.
\end{itemize}
A more thorough discussion of PD schemes may be found in 
subsections 3.3--3.5 in \cite{AGTZ22}. 
Further details may also be found in
\cite{LX19} or \cite{MMSS21}, for example.

The authors of \cite{AGTZ22} consider data obtained from the 
database created by \cite{MMSS21}.
The database consists of a list of PD attempts. 
Every entry of the list contains
\begin{enumerate}[label=(\arabic*)]
    \item The acronym for the name of the coin, 
    e.g. BTC for Bitcoin, 
    called \textit{symbol} of the coin;
    \item The name of the Telegram/Discord group where this 
    information was retrieved;
    \item The date (day) of the PD attempt;
    \item The time (hour and minute) of the PD attempt;
    \item The exchange on which the PD attempt took place.
\end{enumerate}
Using this information and the Python-library 
\href{https://docs.ccxt.com/#/}{ccxt}, the authors of 
\cite{AGTZ22} download trades, that is all orders that were
placed and also realised by selling or buying cryptocurrency
in a neighbourhood of the PD attempt.
Every trade is characterised by the following information:
\begin{enumerate}[label=(\arabic*)]
    \item Trading pair of symbol and currency, e.g. 
    BTC/USD means Bitcoin is traded using US Dollars;
    \item Timestamp, an integer, denoting the UNIX time in 
    milliseconds;
    \item Datetime, i.e. the ISO8601 datetime with milliseconds;
    \item Side, string, either ``buy" or ``sell" to distinguish 
    the operation kind: for the trading pair BTC/USD ``buy" 
    means buying BTC using USD, while ``sell" means receiving 
    USD for BTC;
    \item Price, float, indicating the price at which the pair 
    was traded;
    \item Amount, float, denoting the quantity (in base 
    currency, i.e. USD in the example BTC/USD) which was 
    traded.
\end{enumerate}
Data was retrieved for windows of length
$3$, $6$, and $14$ days centred on the moment
of the PD attempt.
The underlying idea is that the activity corresponding
to the PD attempt will be anomalous compared to the
normal activity contained within the window.

The raw data was preprocessed as follows.
Whenever trades have the same timestamp, side, and price then
the trade volumes are summed accumulated since one can
interpret these trades as one larger trade without any loss of
information.
The variable side is translated into a numerical format by
encoding all ``buy" signals as $0.5$ and all ``sell" signals as
$-0.5$.
Volume is calculated as the product of price and amount; 
prices are used to compute the simple returns.
Thus the final time series considered consists of the channels 
trade time, returns, volume, and trade side, with each channel 
normalised to live within the interval $[0,1]$.

Let $D_c \subset \R^{n + 4}$ denote the vector of time series 
composed of trades' timestamp, side, price, and volume for a 
coin $c$. Here $n$ denotes the number of trades during the 
pre-specified time interval around the PD attempt.
Split $D_c$ as into $\lceil n/o \rceil$ subsets of $w$ trades, 
where $o$ denotes the offset of how many trades to move 
forward in time.
To be more precise, identify the trade-window specified
by $w$ trades and then move on shifting among the listed trades.
Denote this splitting by 
\begin{equation}    
    \label{eq:D_c_split}
        D_c = \bigcup_{i=1}^{\lceil n/o \rceil} D^i_c.
\end{equation}
Transform each subset $D_c^i$ into a feature set by either 
calculating the exact signature or the randomised signature.
Let $R^i_c$ and $\tilde{R}_c^i$ denote the exact and randomised 
signatures respectively.
The final feature set is then given by
\begin{equation}
    \label{eq:feat_sets_final}
        \bR := 
        \bigcup_{c} \bigcup_{i=1}^{\lceil n/o \rceil} 
        R^i_c 
        \qquad \text{and} \qquad
        \tilde{\bR} :=
        \bigcup_{c} \bigcup_{i=1}^{\lceil n/o \rceil} 
        \tilde{R}^i_c. 
\end{equation} 
The offset $o$ and the window size $w$ can be viewed as 
hyper-parameters which can be optimised. 
The choice of $w=100$ and $o=5$ was found to give a good
balance between run time and precision in \cite{AGTZ22}.

The remaining hyper-parameters required to compute the
randomised signatures were chosen as follows.
The function $\sigma$ is taken as $\tanh$, the reservoir 
dimension $k$ was chosen as $50$, the mean and the variance 
of the matrices $\bA$ were $0.05$ and $0.1$ 
respectively, and the mean and the variance of the vectors
$b$ were $0$ and $1$ respectively 
(cf. Table 1 in \cite{AGTZ22}).

Once the feature sets $\bR$ and $\tilde{\bR}$ are computed, 
the authors of \cite{AGTZ22} use the 
\href{https://scikit-learn.org/stable/}{Sklearn Python Package}
implementations of the robust covariance and isolation forest
anomaly detection algorithms.
In contrast to the signature-based approach to 
anomaly detection 
proposed in \cite{CCFLS20} (which is covered in Section 
\ref{anom_detec} of this article), these ``off-the-shelf" 
anomaly detection algorithms are used \textit{without} 
alteration in \cite{AGTZ22}.
Consequently, any drawbacks of these anomaly
detection algorithms are inherited. 
The method using the robust covariance anomaly
detection algorithm requires the number of samples to exceed
the square of the number of features, which leads to a
restriction on the truncation level that may be considered.
The method using the isolation forest anomaly
detection algorithm 
requires the use of sub-sampling techniques
due to the improved performance of this algorithm on 
smaller data sets.

More detailed overviews of the robust covariance and isolation
forest anomaly detection algorithms are provided in subsections
3.8.1 and 3.8.2 of \cite{AGTZ22} respectively.
The reader seeking extensive coverage is directed to \cite{DR99}
for the robust covariance anomaly detection algorithm
and to \cite{LTZ08} for the isolation forest anomaly detection
algorithm.

The predictions of the resulting methods are compared with
manually labelled PD attempts. 
The unsupervised anomaly detection method proposed in 
\cite{KK18} is used as a benchmark.
Performance is additionally compared against the 
method proposed in \cite{MMSS21} that is based upon supervised
learning and requires labelled training data.
The signature and randomised signature methods are themselves
unsupervised techniques; no labelled training data is required.
The precision, recall, and F1 scores are recorded for 
comparison (the F1 score is the harmonic mean of the precision
and the recall).
The results are summarised in Table 
\ref{table:random_sig_anom_detect_results}
(cf. Table 3 in \cite{AGTZ22}).

Excluding the Exact Truncated Signature (Robust Covariance)
method, all the signature and randomised signature
based methods 
proposed in \cite{AGTZ22} out-perform the
benchmark unsupervised 
model from \cite{KK18} for all time windows.
Moreover, the best classifier from \cite{AGTZ22}, which is 
an unsupervised method, comes 
relatively close to matching the performance of the classifier
in \cite{MMSS21} which is a supervised method.
Finally, the similar levels of performance between the 
exact truncated signature methods and the randomised signature 
methods provides an empirical illustration of the theoretical 
property that randomised
signatures contain the same information as the signature itself.

\begin{table}[H]	
	\centering
        \scalebox{0.9}{
	\begin{tabular}{|c|c|c|c|c|}
            \hline
            \textbf{Classifier} & \textbf{Days} 
            & \textbf{Precision (\%)} 
            & \textbf{Recall (\%)} & \textbf{F1 (\%)} \\
            \hline
            Kamps et al. & 3 & 65 & 86 & 74 \\
             & 6 & 48 & 86 & 62 \\
             & 14 & 29 & 81 & 43 \\
             \hline
            La Morgia et al. & 3 & 98 & 91 & 95 \\
             & 6 & - & - & - \\
             & 14 & - & - & - \\
             \hline
            Randomised Signature (Isolation Forest) 
             & 3 & 93 & 83 & 88 \\
             & 6 & 82 & 83 & 82 \\
             & 14 & 75 & 75 & 75 \\
             \hline
            Randomised Signature (Robust Covariance) 
             & 3 & 80 & 94 & 86 \\
             & 6 & 80 & 84 & 82 \\
             & 14 & 75 & 75 & 75 \\
             \hline
            Exact Truncated Signature (Isolation Forest) 
             & 3 & 83 & 92 & 87 \\
             & 6 & 81 & 82 & 81 \\
             & 14 & 71 & 74 & 72 \\
             \hline
            Exact Truncated Signature (Robust Covariance) 
             & 3 & 61 & 61 & 60 \\
             & 6 & 61 & 47 & 54 \\
             & 14 & 58 & 32 & 44 \\
             \hline
        \end{tabular}
        }
        \caption{Kamps et al. denotes the method proposed 
        in \cite{KK18}.
        La Morgia et al. denotes the method proposed in 
        \cite{MMSS21}; the results for this method are reported 
        directly from \cite{MMSS21} which does not consider
        windows of 6 or 14 days.
        Randomised Signature (Isolation Forest) denotes the 
        method proposed in \cite{AGTZ22}
        of using randomised signatures as the features 
        for an isolation forest.
        Randomised Signature (Robust Covariance) denotes the 
        method proposed in \cite{AGTZ22} 
        of using randomised signatures as the features 
        for the robust covariance algorithm.
        Exact Truncated Signature (Isolation Forest)
        denotes the 
        method proposed in \cite{AGTZ22}
        of using truncated signatures as the features 
        for an isolation forest.
        Exact Truncated Signature (Robust Covariance)
        denotes the 
        method proposed in \cite{AGTZ22}
        of using truncated signatures as the features 
        for the robust covariance algorithm.
        Table 3 in \cite{AGTZ22}.}
	\label{table:random_sig_anom_detect_results}
\end{table}

\bibliographystyle{apalike}
\bibliography{SMIML_EMSS_References}
\vskip 4pt 
\noindent
{\sc University of Oxford, Radcliffe Observatory,
Andrew Wiles Building, Woodstock Rd, Oxford, 
OX2 6GG, UK.}
\vskip 4pt
\noindent
TL: tlyons@maths.ox.ac.uk \\
\url{https://www.maths.ox.ac.uk/people/terry.lyons}
\vskip 4pt
\noindent
AM: andrew.mcleod@maths.ox.ac.uk \\
\url{https://www.maths.ox.ac.uk/people/andrew.mcleod}
\end{document}